%% file: main.tex
\documentclass[sts]{imsart}
\usepackage{lmodern}
\usepackage{booktabs,multirow}
\usepackage{enumitem}
\usepackage{amsthm,amsmath,bm,bbm,amsfonts,amssymb,mathtools}
\usepackage{xcolor}
\usepackage[numbers,sort&compress]{natbib}
\usepackage[colorlinks,citecolor=blue,urlcolor=blue]{hyperref}
\usepackage{graphicx,subfig}
\input{mathcommands}
\allowdisplaybreaks
\hypersetup{
  pdftitle={TabPFN: One Model to Rule Them All?},
  pdfauthor={Qiong Zhang, Yan Shuo Tan, Qinglong Tian, Pengfei Li}
}

\begin{document}

\begin{frontmatter}
\title{TabPFN: One Model to Rule Them All?}
\runtitle{TabPFN: One Model to Rule Them All?}

\begin{aug}
\author[A]{\fnms{Qiong}~\snm{Zhang}\thanksref{t1}\ead[label=e1]{qiong.zhang@mbzuai.ac.ae}},
\author[B]{\fnms{Yan Shuo}~\snm{Tan}\thanksref{t1}\ead[label=e2]{yanshuo@nus.edu.sg}},
\author[C]{\fnms{Qinglong}~\snm{Tian}\thanksref{t1}\ead[label=e3]{qinglong.tian@uwaterloo.ca}},
\and
\author[D]{\fnms{Pengfei}~\snm{Li}\ead[label=e4]{pengfei.li@uwaterloo.ca}}
\thankstext{t1}{Co-first author.}

\address[A]{Division of Computing \& Mathematical Sciences, MBZUAI, United Arab Emirates\printead[presep={\ }]{e1}.}
\address[B]{Department of Statistics \& Data Science, National University of Singapore, Singapore\printead[presep={\ }]{e2}.}
\address[C]{Department of Statistics \& Actuarial Science, University of Waterloo, Canada\printead[presep={\ }]{e3}.}
\address[D]{Department of Statistics \& Actuarial Science, University of Waterloo, Canada\printead[presep={\ }]{e4}.}
\end{aug}

\begin{abstract}
Hollmann et al. (\emph{Nature} \textbf{637} (2025) 319–326) recently introduced TabPFN, a transformer-based deep learning model for regression and classification on tabular data, which they claim ``outperforms all previous methods on datasets with up to 10,000 samples by a wide margin, using substantially less training time.''
Furthermore, they have called TabPFN a ``foundation model'' for tabular data, as it can support ``data generation, density estimation, learning reusable embeddings and fine-tuning''.
In this paper, we provide a tailored explanation of how TabPFN works for a statistics audience, by emphasizing its interpretation as approximate Bayesian inference.
We then explore the significance of TabPFN to the field of statistics: We show that an out-of-the-box application of TabPFN can sometimes outperform specialized state-of-the-art methods for semi-supervised parameter estimation, prediction under covariate shift, and heterogeneous treatment effect estimation.
As a partial explanation for the predictive effectiveness of TabPFN, we show that it can simultaneously adapt to both nonparametric structure and parametric structure, for instance, sometimes outperforming LASSO even when assumptions are correctly specified.
Reproduction code is available at \url{https://github.com/qinglong-tian/tabpfn_study}.
\end{abstract}

\begin{keyword}
\kwd{Foundation models}
\kwd{in-context learning}
\kwd{meta-learning}
\kwd{tabular data}
\kwd{transformers}
\end{keyword}
\end{frontmatter}

\input{sections/intro}
\input{sections/tabpfn}
\section{Case studies}
\label{sec:case_studies}
\input{sections/semisupervised}
\input{sections/causal}
\input{sections/covariateshift}
\input{sections/analysis}
\input{sections/noisy_label}
\input{sections/conclusion}


\begin{acks}
This research was enabled in part by support provided by Cedar (\url{cedar.alliancecan.ca}) and the Digital Research Alliance of Canada (\url{alliancecan.ca}). 
The authors would like to thank Krishnakumar Balasubramanian, Arik Reuter, Chandan Singh, and Yuqian Zhang for useful discussions.
This work was conducted while Qiong Zhang was at the Institute of Statistics and Big Data, Renmin University of China.
Qiong Zhang is now affiliated with Mohamed bin Zayed University of Artificial Intelligence.
\end{acks}

\begin{funding}
Qiong Zhang is supported by the National Key R\&D Program of China Grant 2024YFA1015800 and the National Natural Science Foundation of China Grant 12301391.
Yan Shuo Tan is supported by NUS Startup Grant A-8000448-00-00 and the Singapore Ministry of Education (MOE) AcRF Tier 1 Grants A-8002498-00-00 and A-8004458-00-00.
Qinglong Tian is supported by the Natural Sciences and Engineering Research Council of Canada (NSERC) under Grant RGPIN-2023-03479.
Pengfei Li is supported by NSERC under Grant RGPIN-2026-04406 and by the Faculty of Mathematics Research Chair at the University of Waterloo.
\end{funding}

\bibliographystyle{imsart-nameyear}
\bibliography{biblio}

\begin{appendix}
\input{sections/appendix}
\end{appendix}

\end{document}

%% file: mathcommands.tex
\usepackage{amsmath,amsfonts,bm, bbm}

\newcommand{\eg}{{\em e.g.,~}}
\newcommand{\ie}{{\em i.e.,~}}

\DeclareMathAlphabet{\mathsfit}{\encodingdefault}{\sfdefault}{m}{sl}
\SetMathAlphabet{\mathsfit}{bold}{\encodingdefault}{\sfdefault}{bx}{n}

\def\gD{{\mathcal{D}}}

\def\gF{{\mathcal{F}}}

\def\gN{{\mathcal{N}}}

\def\sE{{\mathbb{E}}}

\def\sP{{\mathbb{P}}}

\def\sR{{\mathbb{R}}}

\newcommand{\indep}{\perp\!\!\!\perp}

\DeclareMathOperator*{\argmin}{arg\,min}

\newcommand{\trans}{^{\intercal}}

\newcommand{\bea}{\begin{eqnarray*}}
\newcommand{\eea}{\end{eqnarray*}}
\newcommand{\ba}{\begin{eqnarray*}}
\newcommand{\ea}{\end{eqnarray*}}
\newcommand{\be}{\begin{equation}}
\newcommand{\ee}{\end{equation}}
\newcommand{\bi}{\begin{itemize}}
\newcommand{\ei}{\end{itemize}}

\makeatletter
\newcommand*\rel@kern[1]{\kern#1\dimexpr\macc@kerna}
\newcommand*\widebar[1]{%
  \begingroup
  \def\mathaccent##1##2{%
    \rel@kern{0.8}%
    \overline{\rel@kern{-0.8}\macc@nucleus\rel@kern{0.2}}%
    \rel@kern{-0.2}%
  }%
  \macc@depth\@ne
  \let\math@bgroup\@empty \let\math@egroup\macc@set@skewchar
  \mathsurround\z@ \frozen@everymath{\mathgroup\macc@group\relax}%
  \macc@set@skewchar\relax
  \let\mathaccentV\macc@nested@a
  \macc@nested@a\relax111{#1}%
  \endgroup
}
\makeatother

%% file: sections/intro.tex
\section{Introduction}
\label{sec:intro}

Deep learning has achieved remarkable success in modeling unstructured data types such as text and images.
However, for tabular data—arguably the most prevalent and consequential data type across science—ensemble tree methods have remained the gold standard predictive modeling strategy, consistently outperforming even deep learning on benchmark datasets~\citep{caruana2008empirical,fernandez2014we,olson2018data,grinsztajn2022tree}.
This long-held wisdom has now been challenged by \citet{hollmann2025accurate}, who introduced a deep learning method for tabular data, called TabPFN, which they claimed ``outperforms all previous methods on datasets with up to 10,000 samples by a wide margin, using substantially less training time.''
As evidence, they showed that TabPFN significantly outperforms random forests, XGBoost, LightGBM, and CatBoost on $50$+ benchmark classification and regression datasets.

Beyond its impressive predictive accuracy, TabPFN is claimed to be robust to outliers and missing data, and automatically estimates predictive distributions, thereby providing uncertainty quantification.
Furthermore, it is claimed to support ``data generation, density estimation, learning reusable embeddings and fine-tuning''.
Because of this wide range of capabilities, \citet{hollmann2025accurate} call TabPFN a ``tabular foundation model''~\citep{pmlr-v235-van-breugel24a}, appropriating a term that has been used to describe the transformative role played by large language models (LLMs)~\citep{bommasani2021opportunities} in natural language processing.
TabPFN shares other similarities with LLMs: It makes use of the same type of deep learning architecture (transformers)~\citep{vaswani2017attention} and the same learning paradigm (in-context learning)~\citep{brown2020language}, which differs fundamentally from traditional statistical and machine learning approaches.

Our goal in this paper is not to further test the claims of \citet{hollmann2025accurate}, which have now been further validated on other benchmarks and by other sources (see for instance \citealt{erickson2025tabarena}).
Instead, we aim to provide an explanation of how TabPFN works that is tailored to a statistics audience (Section \ref{sec:tabpfn}) and, more importantly, to explore the significance of TabPFN to the field of statistics.

Towards this goal, we first demonstrate the utility of TabPFN in handling three currently researched estimation problems, namely semi-supervised parameter estimation (Section \ref{subsec:semi-supervised}), heterogeneous treatment effects estimation (Section \ref{sec:causal}), and prediction under covariate shift (Section \ref{sec:covariate_shift}).
For each problem, we compare a simple out-of-the-box application of TabPFN to a ``state-of-the-art'' method proposed by a selected paper recently published at a top statistics journal.
Using the same simulation set-up described by the paper, augmented with further comparisons on real datasets, we show that \emph{TabPFN often achieves better results than the comparator method}\footnote{The version of the TabPFN model used in our study is the one introduced by \citet{hollmann2025accurate}, referred to there as TabPFNv2.
While our paper was undergoing revision, the developers released updated models, TabPFN-2.5~\citep{grinsztajn2025tabpfn}, TabPFN-3~\citep{priorlabs2026tabpfn3}, and TabPFN-3.5~\citep{jaeger2026tabpfn35technicalreport}.
For consistency, all experiments and discussion in this paper are based on TabPFNv2.%
}.

In addition, we investigate how the statistical principle of adaptivity \citep{donoho1995adapting} can provide a partial explanation for the predictive effectiveness of TabPFN.
In classical nonparametric statistics, adaptivity refers to a procedure’s ability to automatically adjust to unknown aspects of the data-generating process, such as smoothness, sparsity, or latent structure, achieving optimal rates without relying on these properties being specified in advance.
In Section \ref{sec:inductive_biases}, we perform numerical experiments suggesting that TabPFN, without any hyperparameter tuning, can simultaneously adapt to both parametric and nonparametric structure: It is competitive with and sometimes even outperforms parametric methods (e.g., LASSO and linear discriminant analysis) even in settings where the parametric assumptions are correct.
At the same time, it outperforms local averaging and smoothing methods in misspecified settings—for example, classification problems with substantial label noise.
Such adaptivity properties are beyond the reach of ensemble tree methods and other previous machine learning or nonparametric methods.

While TabPFN is not a perfect model, its capabilities indicate that ``in-context learning'' with a ``foundation model'' could become a powerful alternative to conventional statistical methodologies across a variety of tasks.
This emerging paradigm appears sufficiently promising to merit serious consideration alongside more established approaches.
We define these terms and elaborate on their implications in Section \ref{sec:discussion}, where we also discuss avenues for statisticians to guide and influence the field’s development.

%% file: sections/tabpfn.tex
\section{An overview of TabPFN}
\label{sec:tabpfn}
TabPFN makes use of \emph{in-context learning} (ICL), which is a new learning paradigm that differs from the standard statistical regression and classification and has gathered significant interest in the machine learning community.
The meaning of ICL has changed over time.
In TabPFN, ICL takes the form of approximate Bayesian inference; we elaborate on this perspective below and defer a discussion on the historical development of ICL to Section~\ref{sec:discussion}.

The \textbf{first key idea} of TabPFN is to approximate posterior predictive distributions (PPD) using a transformer model, the same deep learning architecture behind LLMs.
Specifically, let $\Pi$ denote the space of joint distributions on covariate-label pairs $(X,Y)$ and let $n$ be a fixed sample size.
Let $p$ be a prior on $\Pi$.
Suppose the observed training data $\mathcal{D} = \{(X_i,Y_i)\}_{i=1}^{n}$ are drawn IID from a fixed distribution $\pi^* \in \Pi$, and let $p(\pi|\mathcal{D})$ denote the posterior on $\Pi$ given $\mathcal{D}$.
With slight abuse of notation, the posterior predictive distribution (PPD) for $Y_{n+1}$ given $X_{n+1}=x$ is defined as $p(y|x,\mathcal{D}) = \int_\Pi \pi(y|x)p(\pi |\mathcal{D})d\pi$.
The TabPFN transformer $M$ is trained to approximate the mapping $(\mathcal{D},x) \mapsto p(y|x,\mathcal{D})$.
\citet{muller2022transformers}, who developed this approach, called the resulting model a \emph{prior-fitted network} (PFN).

Approximating the PPD of a Bayesian regression or classification model is far from a new idea.
Indeed, the popular Bayesian Additive Regression Trees (BART) algorithm does the same job via Markov chain Monte Carlo (MCMC) \citep{chipman2010bart}.
The novelty of the PFN approach is that while previous approaches (MCMC, variational inference, likelihood-free inference, etc.) attempt to approximate the map $x \mapsto p(y|x,\mathcal{D})$ for a fixed training dataset $\mathcal{D}$, PFN tries to approximate the map jointly over $x$ and $\mathcal{D}$.
It has the computational benefit of amortizing computation, which is the basis of the ``substantially less training time'' claim by \citet{hollmann2025accurate} regarding TabPFN.%
\footnote{Similar ideas have been explored in the variational inference community under the name of \emph{Neural Processes} (NP) \citep{garnelo2018neural}.
Both the NP and PFN approaches use a ``meta-learning'' perspective to approximate the PPD, but they differ in their implementation choices.
NPs employ an encoder-decoder architecture with a parametric distribution over the latent variable $z$, naturally lending itself to variational inference.
PFNs, on the other hand, adopt a transformer architecture that avoids explicit distributional assumptions in its internal representations.
As a result, their training objectives differ slightly: NPs explicitly optimize a variational lower bound, while PFNs frame learning as a direct optimization problem.}
Such an approximation may also yield statistical benefits via implicit regularization that are not yet fully understood.

The \textbf{second key idea} of TabPFN is its choice of prior on the space of joint distributions $\Pi$.
The large number of parameters of a transformer model ($\approx$ 7M in the case of TabPFN) means that it could potentially approximate PPDs defined in terms of complicated priors.
TabPFN chooses to define its prior in terms of structural causal models (SCMs) \citep{pearl2009causality}.
An SCM defines a directed acyclic graph $\mathcal{G}$ whose nodes represent variables $Z_1,\ldots,Z_p$ and whose edges represent causal relationships.
Specifically, each variable $Z_k$ is defined via the equation $Z_k = f_k(Z_{\operatorname{PA}_{\mathcal{G}}(k)},\varepsilon_k)$, where $f_k$ is a deterministic function, $\operatorname{PA}_{\mathcal{G}}(k)$ denotes the parents of $Z_k$, and $\varepsilon_1,\ldots,\varepsilon_p$ are independent ``exogenous'' random variables.
SCMs have long been used to model causal inference from observational data, in which case $f_1,\ldots,f_p$ are usually defined to be linear functions.
Instead, TabPFN allows them to be neural networks or decision trees, thereby modeling nonlinear and even discontinuous structure.

These two key ideas cohere to provide an efficient training strategy for TabPFN.
To fit the model's parameters $\theta$, a collection $\mathfrak{D}$ of approximately 130M synthetic datasets were generated.
Each dataset $\mathcal{D} \in \mathfrak{D}$ contains IID examples from a joint distribution $\pi$, which is itself drawn from the SCM prior.
Note that while $\pi$ may have a complex density, the SCM structure gives a simple way to sample from it.
Next, $\mathcal{D}$ is partitioned into a ``pseudo''-training set $\mathcal{D}_{\text{train}}$ and ``pseudo''-test set $\mathcal{D}_{\text{test}}$.
For each $(X_{\operatorname{test}}, Y_{\operatorname{test}}) \in \mathcal{D}_{\text{test}}$, the transformer output $M_\theta(X_{\text{test}},\mathcal{D}_{\text{train}})$ approximates the PPD $p(\cdot|X_{\operatorname{test}},\mathcal{D}_{\text{train}})$.
The loss
\begin{multline*}
-\frac{1}{|\mathfrak{D}|}\sum_{(X_{\operatorname{test}}, Y_{\operatorname{test}}) \in \mathcal{D}_{\text{test}}}  \log M_\theta(X_{\operatorname{test}},\mathcal{D}_{\text{train}})(Y_{\operatorname{test}})
\end{multline*}
is minimized by stochastic gradient descent.
Its expectation equals the expected KL divergence up to a $\theta$-independent constant, making the two minimization objectives equivalent.
See \citet{hollmann2025accurate} for details on the prior and model training.

%% file: sections/semisupervised.tex
\subsection{Study I: parameter estimation in a semi-supervised setting}
\label{subsec:semi-supervised}

\subsubsection{Problem and motivation}
In regression and classification problems, labeled data are often scarce or costly to obtain, while unlabeled data are typically more abundant.
This abundance arises because unlabeled data (\eg medical images, sensor readings, or customer transactions) are often generated organically during routine operations, whereas the corresponding labels (\eg diagnoses, system states, or purchase intentions) require expert annotation or time-consuming measurement.
Semi-supervised learning methods (SSL)~\citep{Zhu2010} attempt to improve the efficiency of statistical analyses on the labeled data by leveraging information contained in unlabeled data, usually under the assumption that both sets of data share the same data-generating distribution.

We consider the problem of parameter estimation in an SSL setting.
To formalize this, let $X \in \sR^{p}$ denote the feature vector and $Y \in \sR$ the response variable.
We observe:
\begin{itemize}[leftmargin=*]
    \item A labeled dataset $\gD_{L} = \{(X_i, Y_i)\}_{i=1}^n$ with $n$ IID observations from the joint $P_{X,Y}$;
    \item An unlabeled dataset $\gD_{U} = \{X_i\}_{i=n+1}^{n+m}$ with $m$ IID observations from the marginal $P_X$.
\end{itemize}
The goal is to estimate a parameter $\theta$ of $P_{X,Y}$, such as the mean of $P_Y$ or the regression coefficients of $Y$ on $X$.
Generalizing earlier work, \citet{Song02042024} and \citet{angelopoulos2023prediction} both considered an M-estimation framework, in which the parameter $\theta \in \sR^q$ is defined as the minimizer of the expected loss:
\be
\label{eq:working_model}
\theta^* := \argmin_{\theta \in \Theta} \mathbb{E}_{P_{X,Y}}\{l(\theta; X, Y)\},
\ee
where $l: \Theta \times \sR^p \times \sR \to \sR$ is a pre-specified loss function.
Examples of such estimands are shown in Table \ref{tab:simulation_settings}.
Note that these working models do not have to be correctly specified in order for the estimands to be meaningful.

\begin{table}[htbp]
\centering
\caption{Parameters of interest in the three simulation settings.
  Here we define $\Vec{X}=[1,X\trans]\trans$ and $\tau\in(0,1)$ is a given value.}
\label{tab:simulation_settings}
\resizebox{\columnwidth}{!}{
\begin{tabular}{ll}
\toprule
Setting &Parameter of Interest: $\theta$ \\
\midrule
1 (linear reg.)& $\left\{\sE(\Vec{X}\Vec{X}^\top)\right\}^{-1} \sE(\Vec{X}Y)$ \\
2 (logistic reg.)& $\argmin_{\theta} \sE \left\{\log\left[1+\exp(\Vec{X}^\top\theta)\right] - Y\Vec{X}^\top\theta\right\}$ \\
3 (quantile reg.)& $\argmin_{\theta} \sE \left\{\left[I(Y\leq\Vec{X}^\top\theta)-\tau\right](Y-\Vec{X}^\top\theta)\right\}$ \\
\bottomrule
\end{tabular}}
\end{table}

\subsubsection{State-of-the-art}
We describe two recently proposed methodologies.
To do so, we first define the empirical risk minimizer with respect to a labeled dataset $\gD$ as:
\be
\label{eq:ss_erm}
\widehat{\theta}_{\text{ERM}}(\gD) = \argmin_{\theta \in \Theta} \sum_{(X,Y)\in \gD} l(\theta; X, Y).
\ee
\begin{itemize}[leftmargin=*]
    \item \citet{angelopoulos2023prediction} proposed a simple imputation and debiasing framework that makes use of an existing black-box regression or classification model $M$: (a) Use $M$ to impute labels $\hat Y_i$ on both the labeled and unlabeled datasets.
    Denote these imputed datasets as $\widehat{\gD}_{L} = \{(X_i, \widehat{Y}_i)\}_{i=1}^{n}$ and $\widehat{\gD}_{U} = \{(X_i, \widehat{Y}_i)\}_{i=n+1}^{n+m}$ respectively; (b) Estimate the imputation bias via: $\Delta = \widehat{\theta}_{\text{ERM}}(\widehat{\gD}_{L}) - \widehat{\theta}_{\text{ERM}}(\gD_{L})$;
(c) Return the estimate $\widehat{\theta}_{\text{ERM}}(\widehat{\gD}_{U}) - \Delta$.
    \item \citet{Song02042024} (\textbf{SLZ}) proposed a method that makes a projection-based correction to the ERM loss \eqref{eq:ss_erm}.
    They proved that their method has optimal asymptotic efficiency under certain assumptions and further showed superior performance over other methods~\citep{kawakita2013semi, Chakrabortty2018efficient, azriel2022semi} via numerical experiments.
    On the other hand, their method requires hyperparameter tuning such as a choice of the projection space.
\end{itemize}

\subsubsection{Experimental design}
We repeat \citet{Song02042024}'s experiments, but add two comparators based on TabPFN:
\begin{itemize}
\item \textbf{TabPFN-Debiased (TabPFN-D)}: Following \citet{angelopoulos2023prediction}'s approach, we impute each label using the posterior predictive mean (for continuous labels) or mode (for discrete labels), with PPD approximated by TabPFN supplied with $\gD_{L}$ as a training set.%
\footnote{\citet{angelopoulos2023prediction} do not specify how to fit the imputation model.}
We then return the estimate $\widehat{\theta}_{\text{ERM}}(\widehat{\gD}_{L} \cup \widehat{\gD}_{U}) - \Delta$ (i.e. solve ERM on the union of imputed versions of both the labeled and unlabeled datasets).
Since TabPFN is fitted on the same labeled sample used to estimate $\Delta$, this is a heuristic debiasing variant lacking the prediction-powered inference guarantees.
\item \textbf{TabPFN-Imputed (TabPFN-I)}: Follow the same approach as in TabPFN-D, but return the ERM solution without debiasing: $\widehat{\theta}_{\text{ERM}}(\widehat{\gD}_{L} \cup \widehat{\gD}_{U})$.
\end{itemize}
We use the original authors' implementation for the SLZ method.
As a further baseline, we consider the estimator $\widehat{\theta}_{\text{ERM}}(\gD_{L})$ that only makes use of labeled data (\textbf{Vanilla}).

We consider three types of estimands, corresponding to parameters from linear, logistic, and quantile regression working models (Table~\ref{tab:simulation_settings}).
These are called ``working models'' because the true data-generating process differs from the estimation models.
We vary:
\begin{itemize}
\item Feature dimensions: $p \in \{4,\dots,9\}$
\item Sample sizes: $n \in \{300,500\}$ (labeled), $m \in \{500,1000,2000\}$ (unlabeled)
\item Quantile levels (Setting 3 only): $\tau \in \{0.25, 0.5\}$
\end{itemize}
yielding at least $6 \times 2 \times 3 = 36$ configurations.
For each configuration, we approximate the true $\theta^*$ via Monte Carlo integration using $10^7$ samples for the linear-regression setting and $10^6$ samples for the logistic- and quantile-regression settings.
Using $500$ replicates for each configuration, we compute estimates for $\theta^*$, and use them to estimate the bias and mean squared error (MSE).
Further experimental details are given in Appendix~\ref{app:semisup}.

\subsubsection{Results}
\begin{figure}[ht]
\centering
\includegraphics[width=0.45\linewidth]{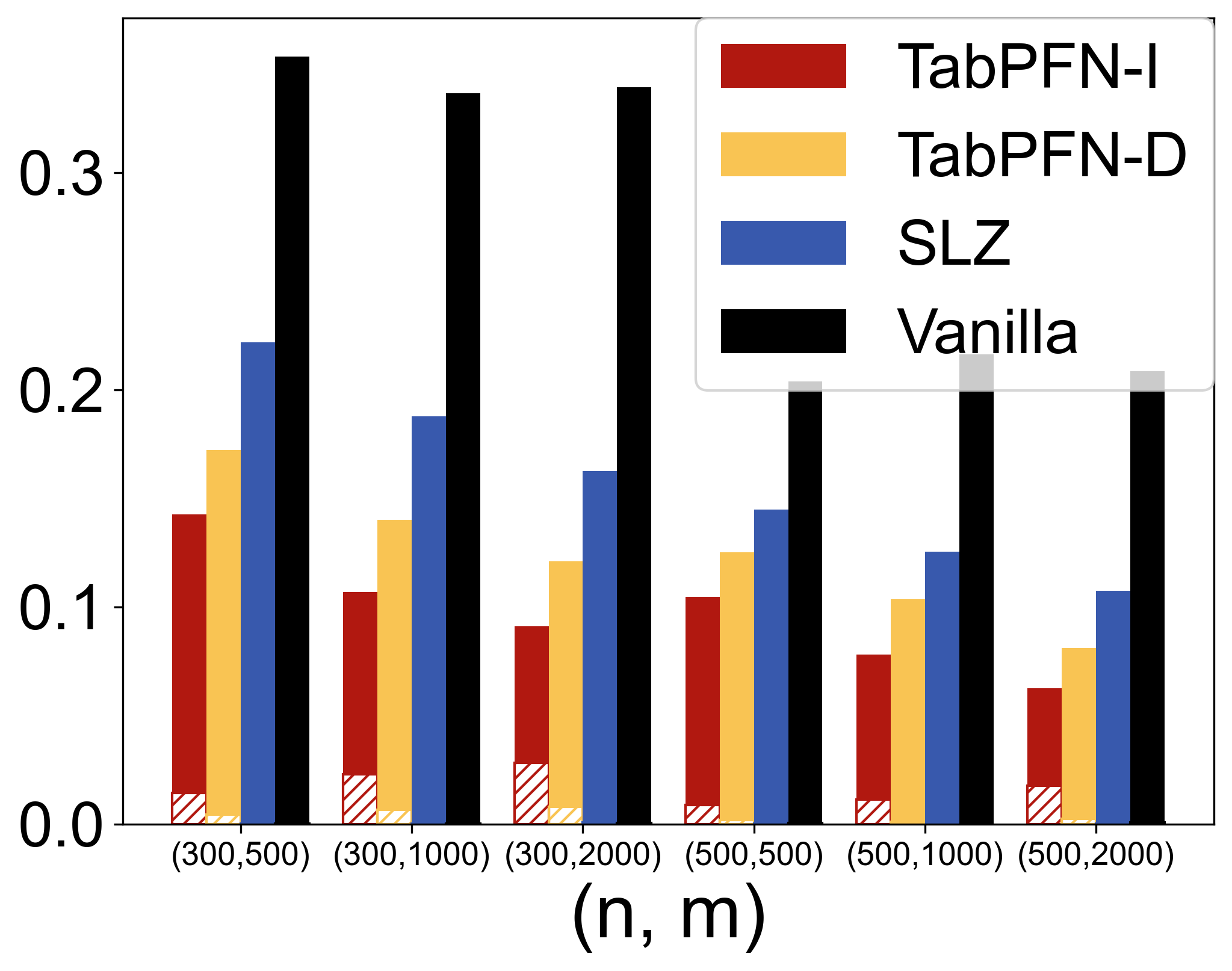}
\includegraphics[width=0.45\linewidth]{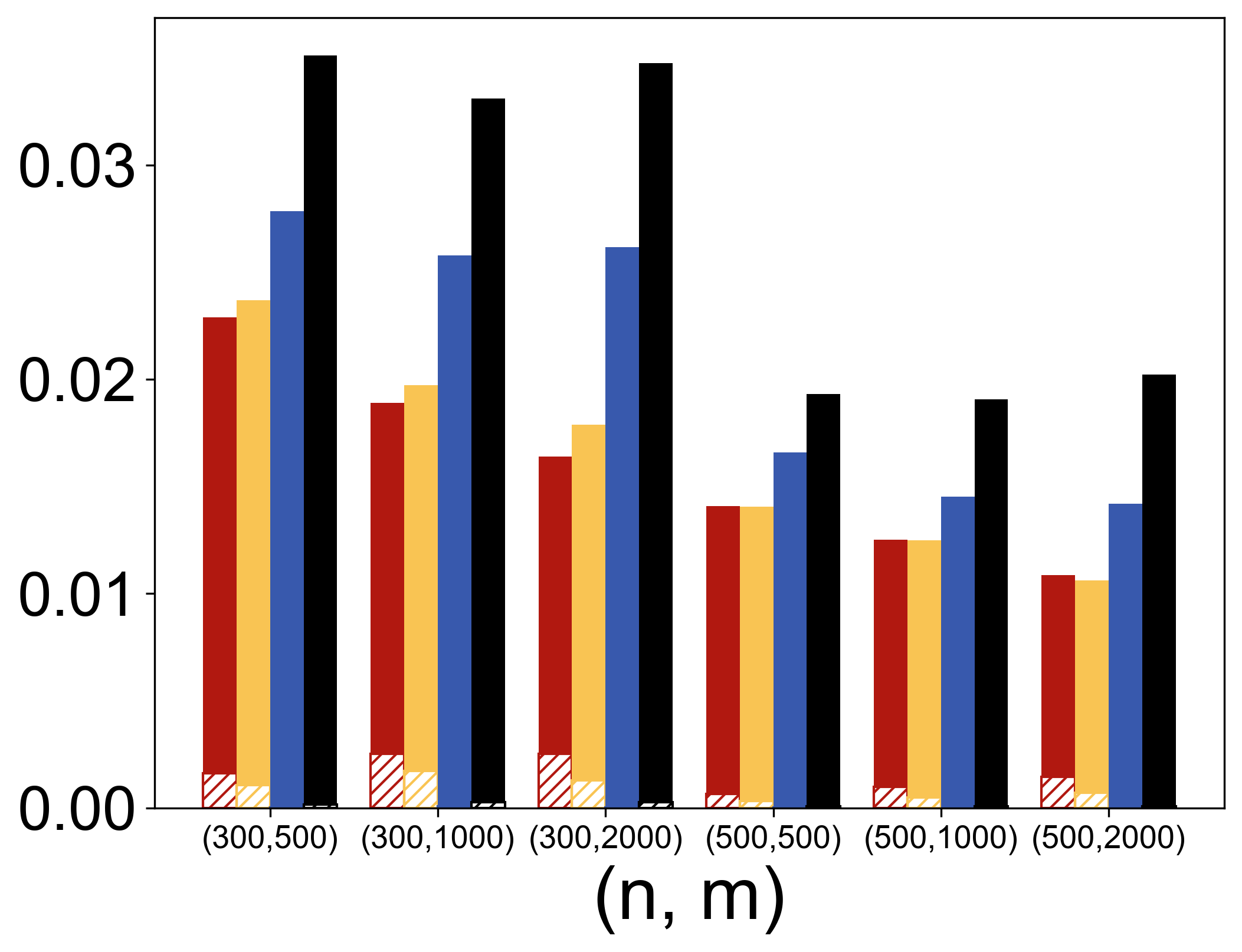}\\
\includegraphics[width=0.45\linewidth]{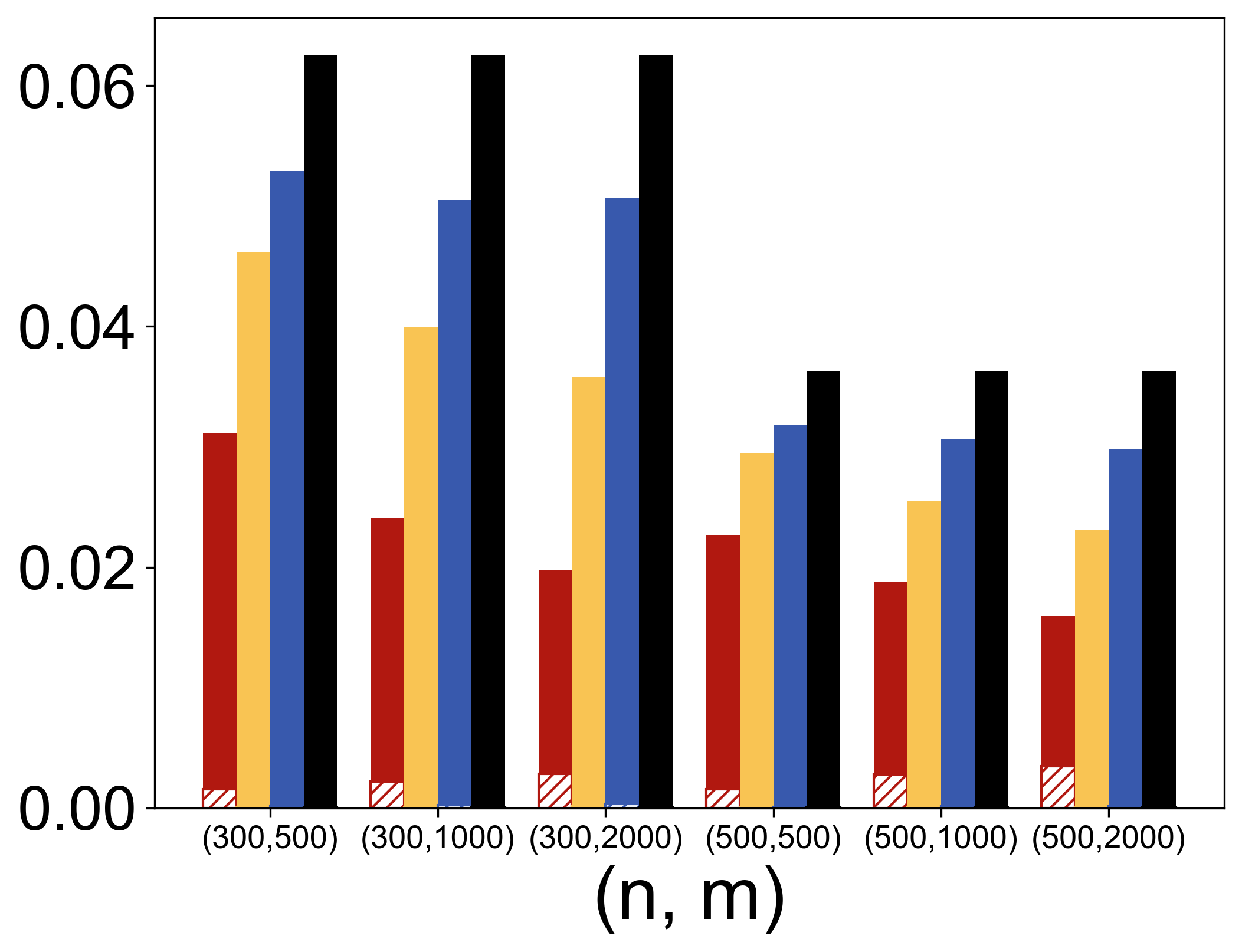}
\includegraphics[width=0.45\linewidth]{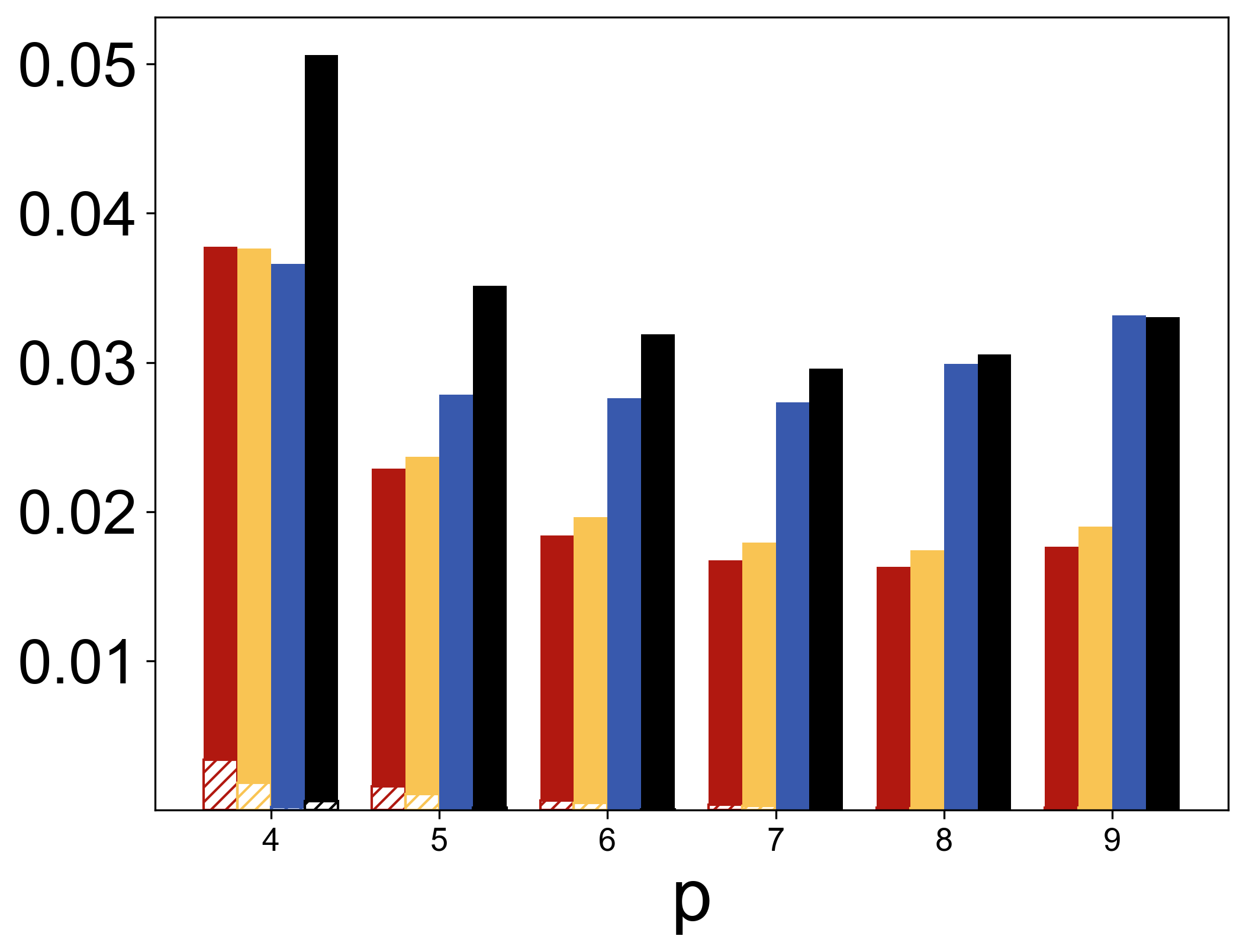}
\caption{MSE results for linear, logistic, and quantile ($\tau=0.25$) regressions when $p=5$, and logistic regression with various $p$ when $(n,m)=(300,500)$.}
\label{fig:semisup-logistic}
\end{figure}

Representative results of our experiments are shown in Figure~\ref{fig:semisup-logistic}, with the complete set of results deferred to Appendix~\ref{app:semisup}.
The first three panels (from the left) show MSE with respect to linear, logistic, and quantile regression respectively when $p=5$, with various combinations of $(n,m)$.
In the last panel, we show how the results for logistic regression vary with $p$, observing that SLZ is only able to outperform the TabPFN approaches when $p=4$.
In the settings examined, TabPFN-I and TabPFN-D generally have lower MSE than SLZ.
We also decompose MSE into squared bias (shaded bar) and variance (solid bar) for the four methods and observe that while TabPFN-I has larger bias than TabPFN-D (as well as the other two methods), it has smaller variance, which offsets the larger bias and leads to smaller MSE.
These comparisons are finite-sample bias--variance results; the asymptotic interpretation of TabPFN-I is discussed in Appendix~\ref{subsec-supp:task1-interpretation}.

\subsubsection{Real data comparison.}
\label{sec:real_data_semi}

To evaluate performance in real-world scenarios, we test the methods on ten OpenML datasets \citep{feurer2021openml}, comprising five for regression and five for binary classification.
For each dataset, we use the linear-model coefficient (for regression) or logistic-regression coefficient (for classification) fitted on the full dataset as a reference target.
We then create labeled--unlabeled splits by randomly selecting $n$ observations as labeled data, where $n$ is the smaller of 5,000 and 10\% of the full dataset for regression or 20\% for classification, and treating the remainder as unlabeled (using only their features).
Because the reference target is itself estimated from the full dataset, the reported errors should be interpreted as discrepancies from the full-data fit rather than errors relative to a known population parameter.
From these splits, we compute the Vanilla, TabPFN-I, TabPFN-D, and SLZ estimators.
In the logistic regression setting, we additionally explore imputation by the posterior predictive mean for TabPFN-I and TabPFN-D, a procedure that has also been called fractional imputation~\citep{rubin1987multiple}.%
\footnote{Fractional imputation replaces each observation with a missing binary response by the two pseudo-observations with responses $1$ and $0$ and sample weights $P(Y=1\mid X=x)$ and $P(Y=0\mid X=x)$.
This yields the same contribution to the score function as imputing the missing response with the single mean value $\sum_{y\in\{0,1\}} y\,P(Y=y\mid X=x)=P(Y=1\mid X=x)$.}
We repeat this process 50 times and report the average MSE of the estimated coefficients across all repetitions.
The data-selection criteria and dataset details are provided in Appendix~\ref{app:ssl_real_data}.
\begin{table}[!tp]
\centering
\caption{Real data SSL: Linear regression.
  Entries are the coefficient MSE multiplied by $10^{4}$.}
\label{tab:ssl_real_data_reg}
\resizebox{0.9\linewidth}{!}{
\begin{tabular}{@{}ccccc@{}}
\toprule
ID & Vanilla & TabPFN-I & TabPFN-D & SLZ\\
\midrule
189&6.3621&2.3105&1.8860&6.5636\\
218&15.2800&4.3926&2.5916&3.1214$\times10^{10}$\\
225&7.0790&1.5684&3.0303&7.0924\\
572&1.4113&0.4698&0.4589&0.9370\\
1200&1.9185&0.8884&0.7787&1.7529\\
\bottomrule
\end{tabular}}
\end{table}

\begin{table}
\centering
\caption{Real data SSL: Logistic regression.}
\label{tab:ssl_real_data_clf}
\resizebox{\linewidth}{!}{
\begin{tabular}{@{}ccccccc@{}}
\toprule
ID & Vanilla & TabPFN-I & TabPFN-D & TabPFN-I & TabPFN-D& SLZ\\
\cmidrule(l{1em}r{1em}){3-4}
\cmidrule(l{1em}r{1em}){5-6}
&&\multicolumn{2}{c}{mode imputation}&\multicolumn{2}{c}{mean imputation}&\\
\midrule
251&0.0195&2.6087&0.2679&0.0149&0.0181&0.0200\\
807&0.0036&0.0116&0.0065&0.0020&0.0017&0.0037\\
816&0.0060&0.1081&0.01550&0.0026&0.0044&0.0060\\
1046&0.0025&0.0030&0.0022&0.0013&0.0011&0.0025\\
40983&5.5566&2.3351&2.5356&1.2346&1.9884&5.9513\\
\bottomrule
\end{tabular}}
\end{table}

The results for linear and logistic regression are presented in Table~\ref{tab:ssl_real_data_reg} and Table~\ref{tab:ssl_real_data_clf}, respectively.
For linear regression, both TabPFN-I and TabPFN-D typically outperform the Vanilla and SLZ methods.
TabPFN-D's slight advantage over TabPFN-I suggests a benefit from debiasing in these presumably noisier and more complex settings.
For logistic regression, Vanilla outperforms SLZ and TabPFN with mode imputation, whereas a TabPFN estimator with mean imputation has the lowest reported MSE on each of the five datasets.
We attribute the weaker performance of mode imputation to its inherent bias: it deterministically assigns a class label even when predictions are uncertain, whereas mean imputation preserves uncertainty by weighting labels according to predicted probabilities, improving stability and accuracy~\citep{rubin1987multiple}.

%% file: sections/causal.tex
\newcommand{\flamlImplementationNote}{%
AutoML uses cross-validation (CV) and a smart exploration strategy to select from a collection of the most popular machine learning models and to tune their hyperparameters.%
}

\subsection{Study II: heterogeneous treatment effects estimation}
\label{sec:causal}

\subsubsection{Problem and motivation}
While researchers were previously satisfied with estimating the average effect of a treatment --- such as a drug, social intervention, or economic policy --- across a population of interest, there has been rising interest in estimating heterogeneous treatment effects (HTE), that is, how different groups of individuals might benefit differentially from the same treatment.
Accurate HTE estimation enables personalized treatments, which may lead to better outcomes.

HTE is usually quantified via the Neyman–Rubin potential outcomes framework~\citep{rubin1974estimating}.
We consider $n$ IID samples $(Y_i(0), Y_i(1), T_i, X_i)$ from a population, where $X_i \in \mathbb{R}^d$ is a $d$-dimensional feature vector, $T_i \in \{0,1\}$ is a binary treatment assignment indicator, and $Y_i(0)$ and $Y_i(1) \in \mathbb{R}$ are potential outcomes.
The observed outcome is $Y_i^{\text{obs}} = Y_i(T_i)$.
Define the response functions under control and treatment respectively as
\[
\mu_{j}(x) = \mathbb{E}[Y(j) \mid X = x], \quad j = 0,1,
\]
and the conditional average treatment effect (CATE) function as $\tau(x) = \mu_{1}(x) - \mu_{0}(x)$.
The propensity score is defined as $e(x) = \mathbb{E}(T|X=x)$.
Unlike the average treatment effect (ATE) $\tau = \mathbb{E}[\tau(X)]$, the CATE captures HTE and is hence our estimand of interest.

\subsubsection{State-of-the-art (SOTA)}
Many methods have recently been proposed to estimate the CATE under the assumption of ignorability ($(Y(1),Y(0))\indep T |X $).
Most but not all of these methods can be subsumed under \citet{kunzel2019metalearners}'s framework for ``meta-learners'', which they defined as ``the result of combining supervised learning or regression estimators (i.e., base learners) in a specific manner while allowing the base learners to take any form''.
These include:
\begin{itemize}[leftmargin=*]
\item \textbf{S-Learner}:
(a) Form an estimate $\hat\mu(t,x)$ for the function $ (t,x) \mapsto \mu_t(x)$ by jointly regressing $Y^{\text{obs}}_i$ on $T_i$ and $X_i$;
(b) Set $\hat{\tau}_S(x) = \hat{\mu}(1,x) - \hat{\mu}(0,x)$.

\item \textbf{T-Learner}:
(a) Form estimates $\hat{\mu}_0(x), \hat{\mu}_1(x)$ for $\mu_0(x)$ and $\mu_1(x)$ by regressing $Y^{\text{obs}}_i$ on $X_i$ within the control and treatment groups respectively;
(b) Set $\hat{\tau}_T(x) = \hat{\mu}_1(x) - \hat{\mu}_0(x)$.

\item \textbf{R-Learner}~\citep{nie2020quasi}:
(a) Form estimates $\hat{e}(x)$ for $e(x)$ and $\hat{m}(x)$ for $m(x) = \mathbb{E}(Y^{\text{obs}}|X=x)$;
(b) Compute residuals $\widetilde{Y}_i = Y_i^{\text{obs}} - \hat{m}(X_i)$, $\widetilde{T}_i = T_i - \hat{e}(X_i)$;
(c) Solve
\be
\label{eq:r_learner}
\hat{\tau}_{R}(\cdot) = \argmin_{\tau} \frac{1}{n} \sum_{i=1}^n \big(\widetilde{Y}_i - \tau(X_i)\widetilde{T}_i\big)^2.
\ee

\item \textbf{DR-Learner}~\citep{kennedy2023towards}:
(a) Form estimates $\hat\mu_0(x)$ for $\mu_0(x)$, $\hat\mu_1(x)$ for $\mu_1(x)$, and $\hat{e}(x)$ for $e(x)$;
(b) Compute doubly robust pseudo-outcomes $\widetilde{Y}_i$ using these models\footnote{$\tilde{Y}_i = \frac{T_i(Y_i - \hat{\mu}_1(X_i))}{\hat{e}(X_i)} - \frac{(1-T_i)(Y_i - \hat{\mu}_0(X_i))}{1-\hat{e}(X_i)} + \hat{\mu}_1(X_i) - \hat{\mu}_0(X_i)$};
(c) Form an estimate $\hat\tau_{\text{DR}}(x)$ for $\tau(x)$ by regressing $\widetilde{Y}_i$ on $X_i$.

\item \textbf{X-Learner}~\citep{kunzel2019metalearners}:
(a) Form estimates $\hat\mu_0(x)$ for $\mu_0(x)$ and $\hat\mu_1(x)$ for $\mu_1(x)$;
(b) Impute treatment effects: $\widetilde{D}_i^1 = Y_i^{\text{obs}} - \hat{\mu}_0(X_i)$ (treatment group) and $\widetilde{D}_i^0 = \hat{\mu}_1(X_i) - Y_i^{\text{obs}}$ (control group);
(c) Form preliminary estimates $\hat{\tau}_j(x)$ for $\tau(x)$ by regressing $\widetilde{D}_i^j$ on $X_i$ for $j=0,1$;
(d) Combine the two estimates via $\hat{\tau}_X(x) = g(x)\hat{\tau}_0(x) + (1-g(x))\hat{\tau}_1(x)$ (typically $g(x) = \hat{e}(x)$).

\end{itemize}

\citet{foster2023orthogonal} advocated for either the R-Learner or DR-Learner as, unlike the S-, T- and X-Learners, they satisfy Neyman orthogonality, which results in theoretically faster estimation rates so long as the base learners used to estimate the nuisance parameters (e.g. $e(x)$ and $m(x)$) satisfy certain convergence properties.
They used \citet{wang2021flaml}'s AutoML algorithm\footnote{\flamlImplementationNote} as the base learner in the meta-learner strategies detailed above, and showed that the R-Learner or DR-Learner indeed achieved better performance.

\subsubsection{Experimental design}
We adapt the six simulation setups of \citet{foster2023orthogonal} and compare TabPFN with AutoML as the base learner.
We fit the DR-, S-, T-, and X-Learners with each base learner.
The R-Learner optimization problem~\eqref{eq:r_learner} is solved by weighted least squares, which TabPFN does not currently support, so we fit the R-Learner only with AutoML.
We also fit an R-Learner with the true nuisance functions $m(x)$ and $e(x)$ and an AutoML final stage, giving 10 estimators in total.
This true-nuisance estimator is labeled ``Oracle'' to match the original implementation, but it is a diagnostic rather than a lower bound on achievable error and is not used to support our comparative conclusions.
Each estimator is named by its method followed by the base learner in parentheses (e.g., `S-Learner (TabPFN)').

Within each repetition, we draw $X_i \sim \operatorname{Unif}([-.5,.5]^6)$, $T_i |X_i\sim\operatorname{Bernoulli}\{e(X_i)\}$, and $Y_i=b(X_i)+(T_i-.5)\tau(X_i)+\epsilon_i$, where $\epsilon_i \sim \mathcal{N}(0,\sigma^2)$.
We use $6$ setups for the functions $e(x)$, $\tau(x)$, and $b(x) = (\mu_0(x) + \mu_1(x))/2$, but show $2$ representative setups here:
\begin{itemize}
\item \textbf{Setup A}: complicated $e(x)$ and $b(x)$, but a relatively simple $\tau(x)$.
\item \textbf{Setup E}: large $b(x)$ resulting in substantial confounding, discontinuous $\tau(x)$.
\end{itemize}
The specifications of the functional forms are given in Appendix~\ref{app:causal}.
We use training-set sizes $n\in\{500,1000,2000\}$ and noise variances $\sigma^2\in\{0.5,1,2\}$.
For each of the resulting 54 configurations, we generate 100 training datasets.
We evaluate all estimators on the same test set of size $10{,}000$ and report the median test MSE over the 100 repetitions.
\begin{figure}[htbp]
\centering
\includegraphics[width=0.45\linewidth]{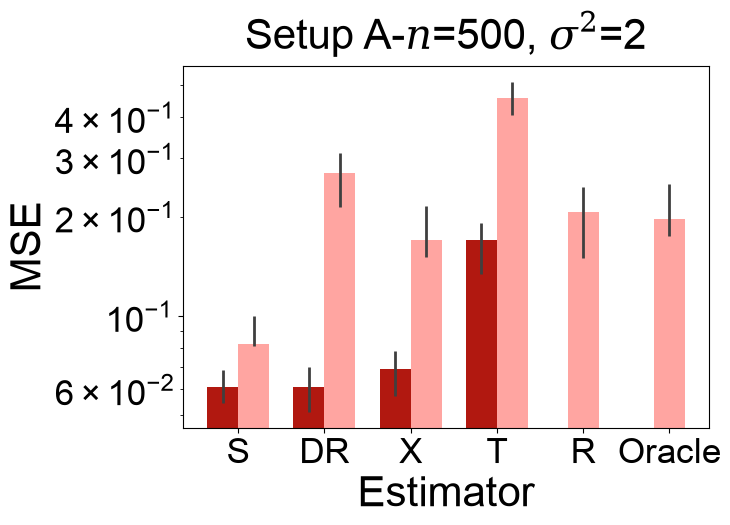}
\includegraphics[width=0.45\linewidth]{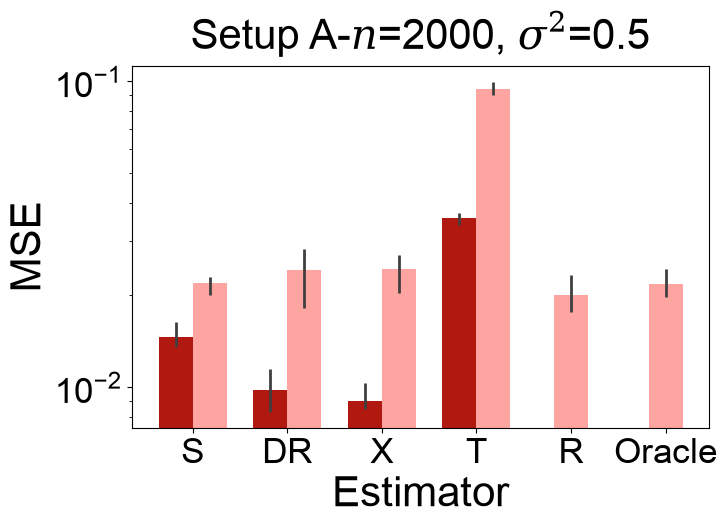}\\
\includegraphics[width=0.45\linewidth]{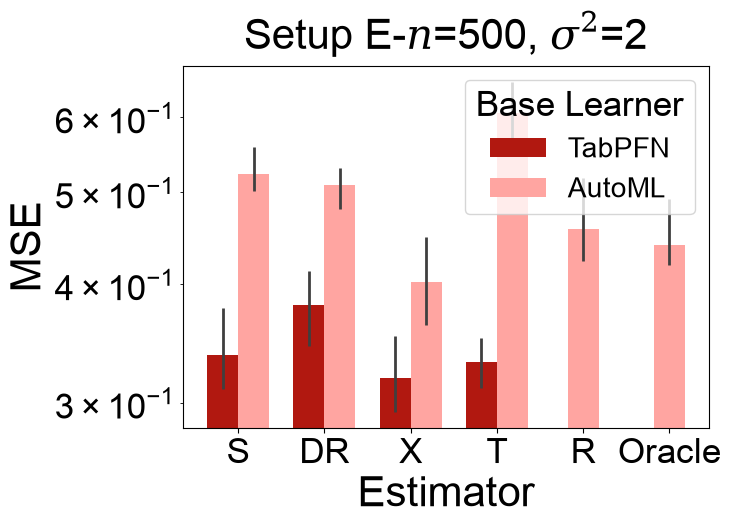}
\includegraphics[width=0.45\linewidth]{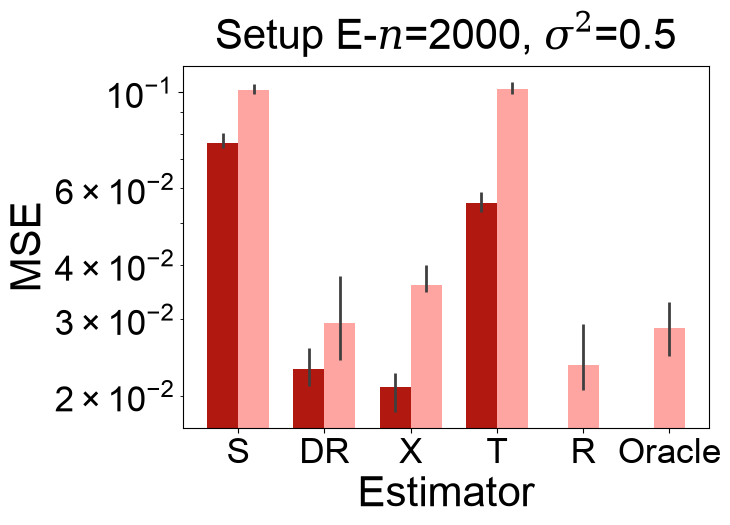}
\caption{Median test MSE over 100 repetitions for four representative configurations under Setups A and E. 
Vertical lines are $95\%$ nonparametric bootstrap intervals for the median, and the vertical axis is logarithmic.}
\label{fig:causal}
\end{figure}

\subsubsection{Results}
We display the results for Setups A and E in Figure~\ref{fig:causal} and defer the complete set of experimental results to Tables~\ref{tab:CATE_full}--\ref{tab:CATE_full_D-F} in Appendix~\ref{app:causal}.
Across the matched comparisons, using TabPFN as the base learner yields a lower median test MSE than its AutoML counterpart in 209 of 216 cells.
The seven exceptions all occur for the S-Learner in Setup D.
Among all 10 estimators, a TabPFN-based estimator has the smallest median test MSE in every one of the 54 simulation settings.
Among the TabPFN-based estimators, the X-Learner has the smallest median test MSE in 24 settings, followed by the DR-Learner in 13 settings, the T-Learner in nine settings, and the S-Learner in eight settings.
The X-Learner (TabPFN) also ranks among the top two estimators in 50 of the 54 settings.
Thus, no single meta-learner dominates uniformly, although the X-Learner (TabPFN) is the most consistently competitive in these simulations.
These finite-sample results do not contradict the asymptotic motivation for orthogonal learners in \citet{foster2023orthogonal}; rather, they show that the choice of base learner materially affects CATE estimation error at the sample sizes considered here.

\subsubsection{Real data comparison}

We evaluate the TabPFN-based meta-learners on the semi-synthetic ACIC 2017 competition benchmark based on the IHDP covariates~\citep{brooks1992effects,dorie2019automated,hahn2019atlantic}.
The benchmark provides 6,000 datasets generated from real covariates under a controlled set of data-generating processes (DGPs), varying (i) the error structure (IID, non-additive, and group-correlated), (ii) the selection strength $\kappa$ (i.e. variability of the true propensity score), (iii) the treatment effect size, and (iv) the noise level.
Each scenario includes ground-truth CATEs, allowing performance assessment by RMSE.
We compare the four TabPFN-based meta-learners (S-, T-, X-, and DR-Learner) against the 16 methods described in \citet{hahn2019atlantic} that are reported with complete benchmark results.
These methods notably include Causal Forests (GRF, \citealt{athey2019generalized}), Targeted Learning (TL, \citealt{coyle2020targeting}), various versions of BART, and Bayesian Causal Forests (BCF, \citealt{hahn2020bayesian}), all of which are considered in the literature to be strong baselines.
We do not compare with the AutoML-based meta-learners due to the much higher computational overhead required in comparison with using TabPFN.

\begin{figure*}[!tp]
    \centering
    \includegraphics[height=0.2\linewidth]{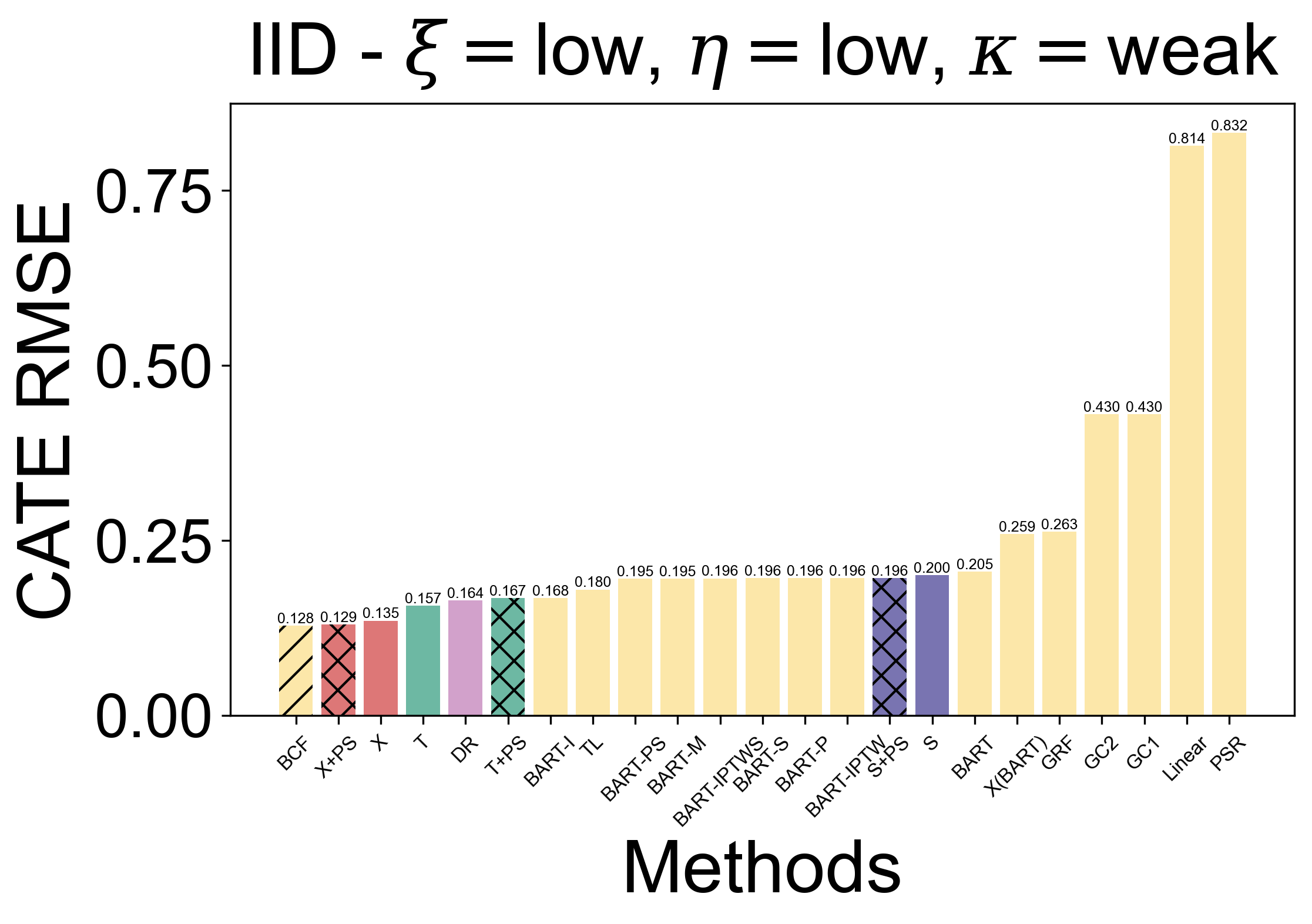}
    \includegraphics[height=0.2\linewidth]{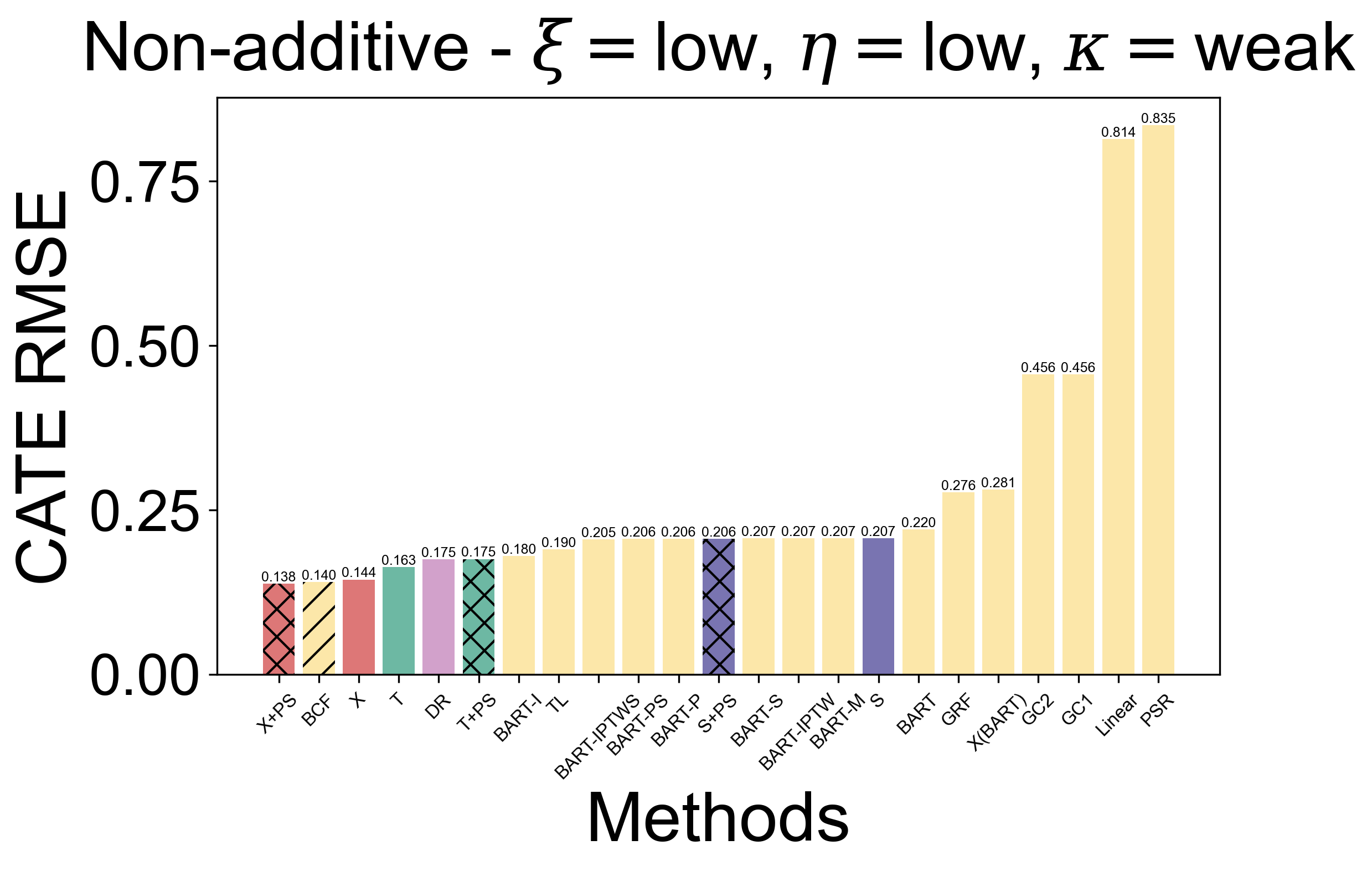}
    \includegraphics[height=0.2\linewidth]{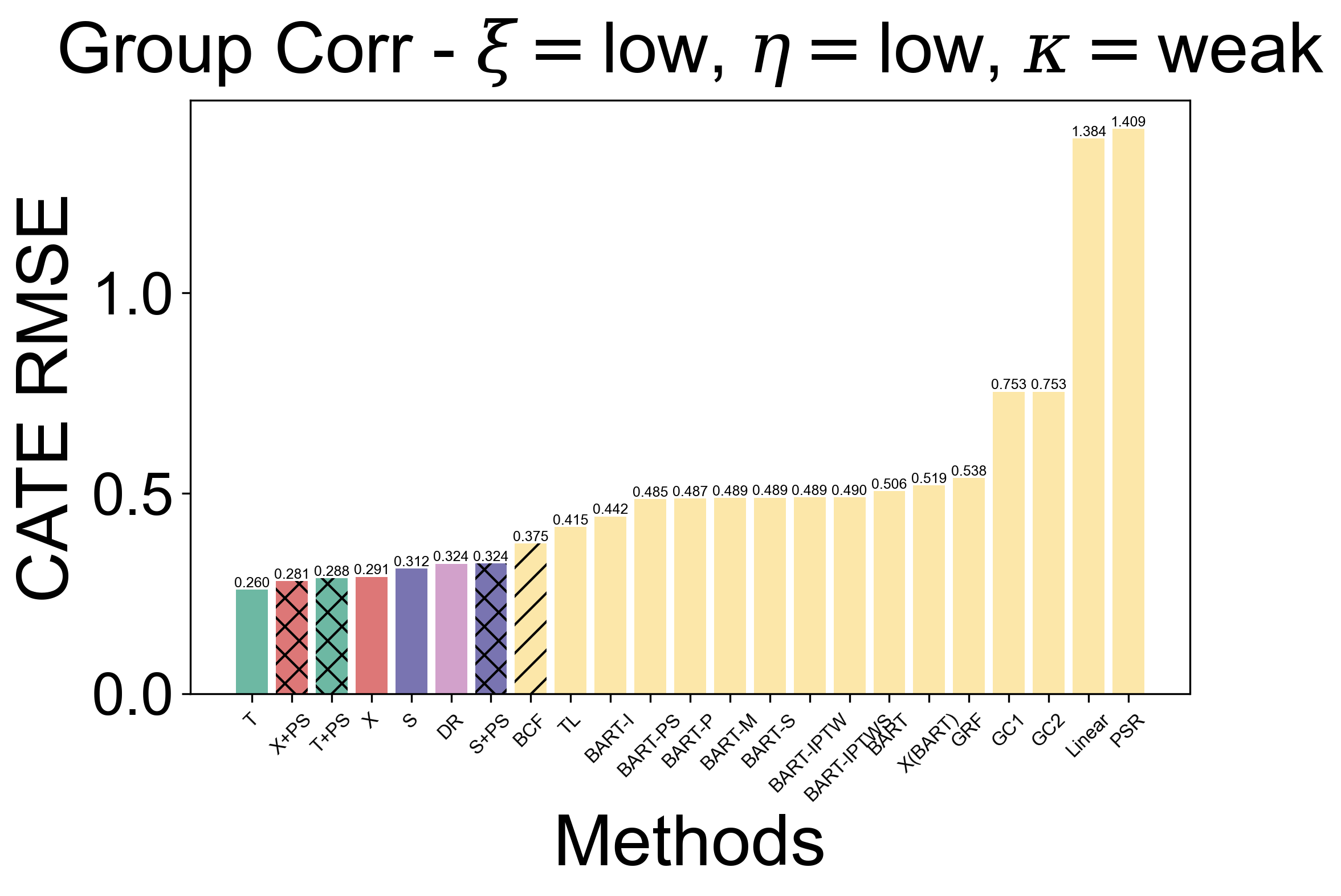}\\
    \includegraphics[height=0.2\linewidth]{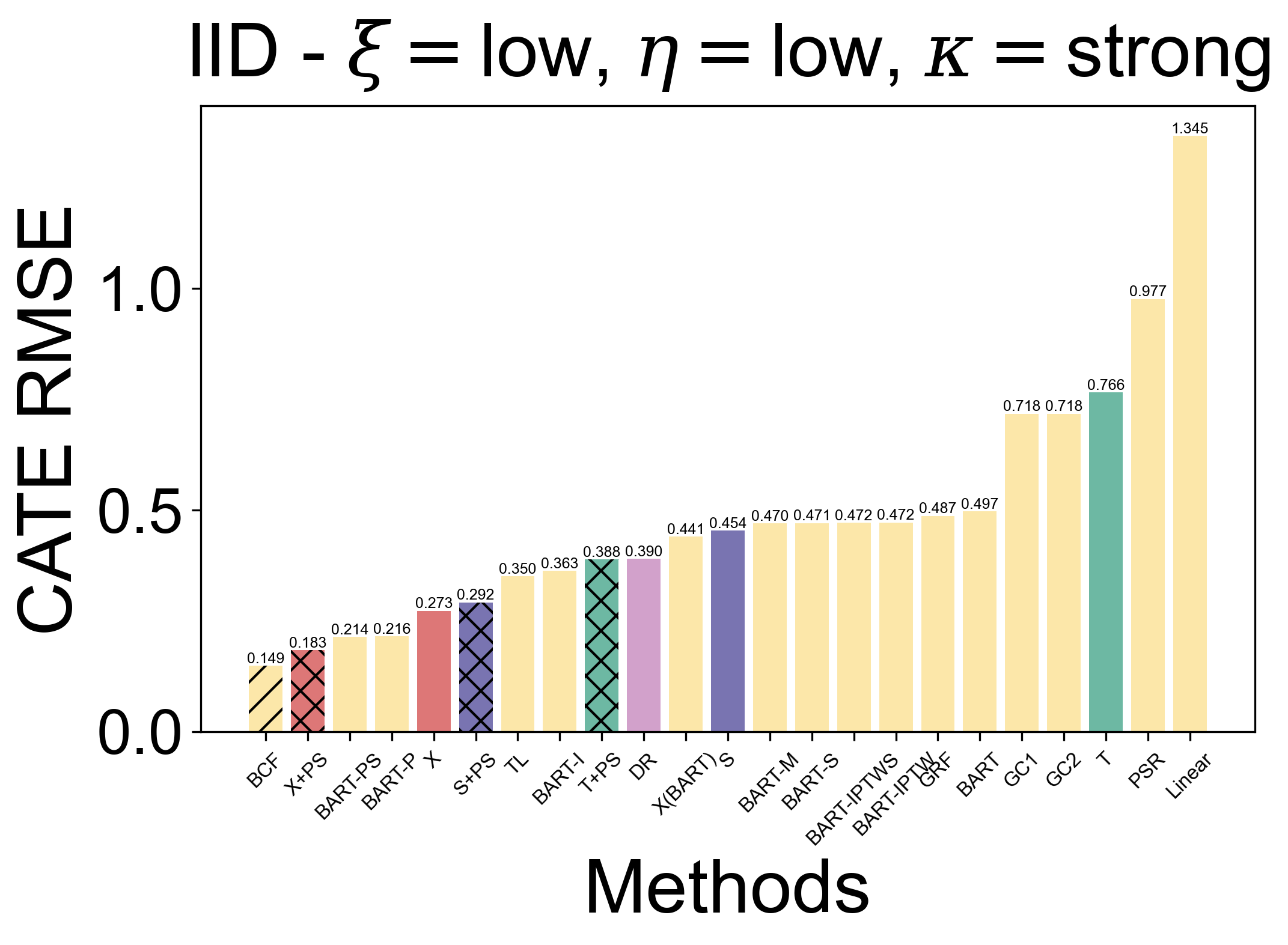}
    \includegraphics[height=0.2\linewidth]{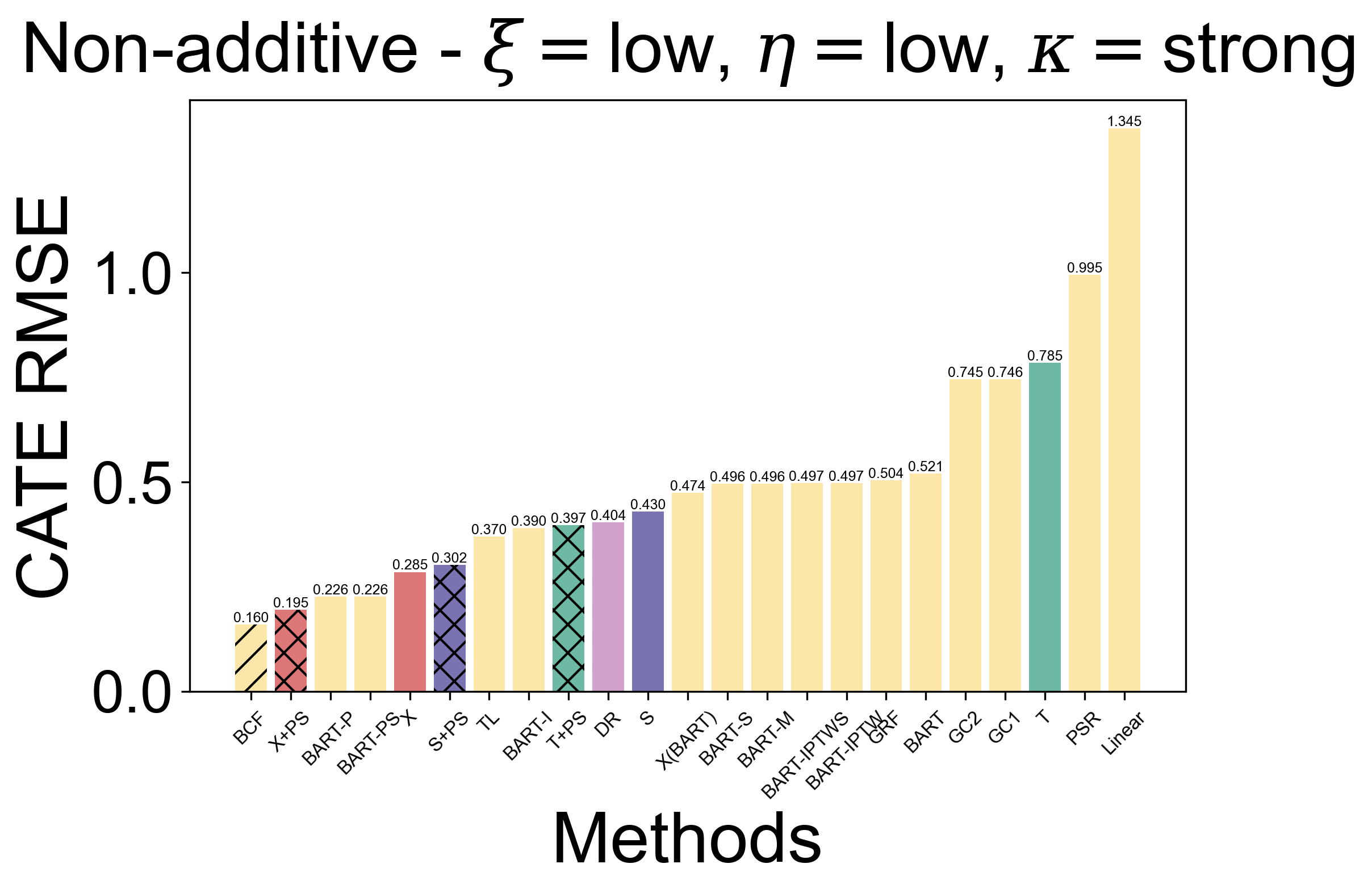}
    \includegraphics[height=0.2\linewidth]{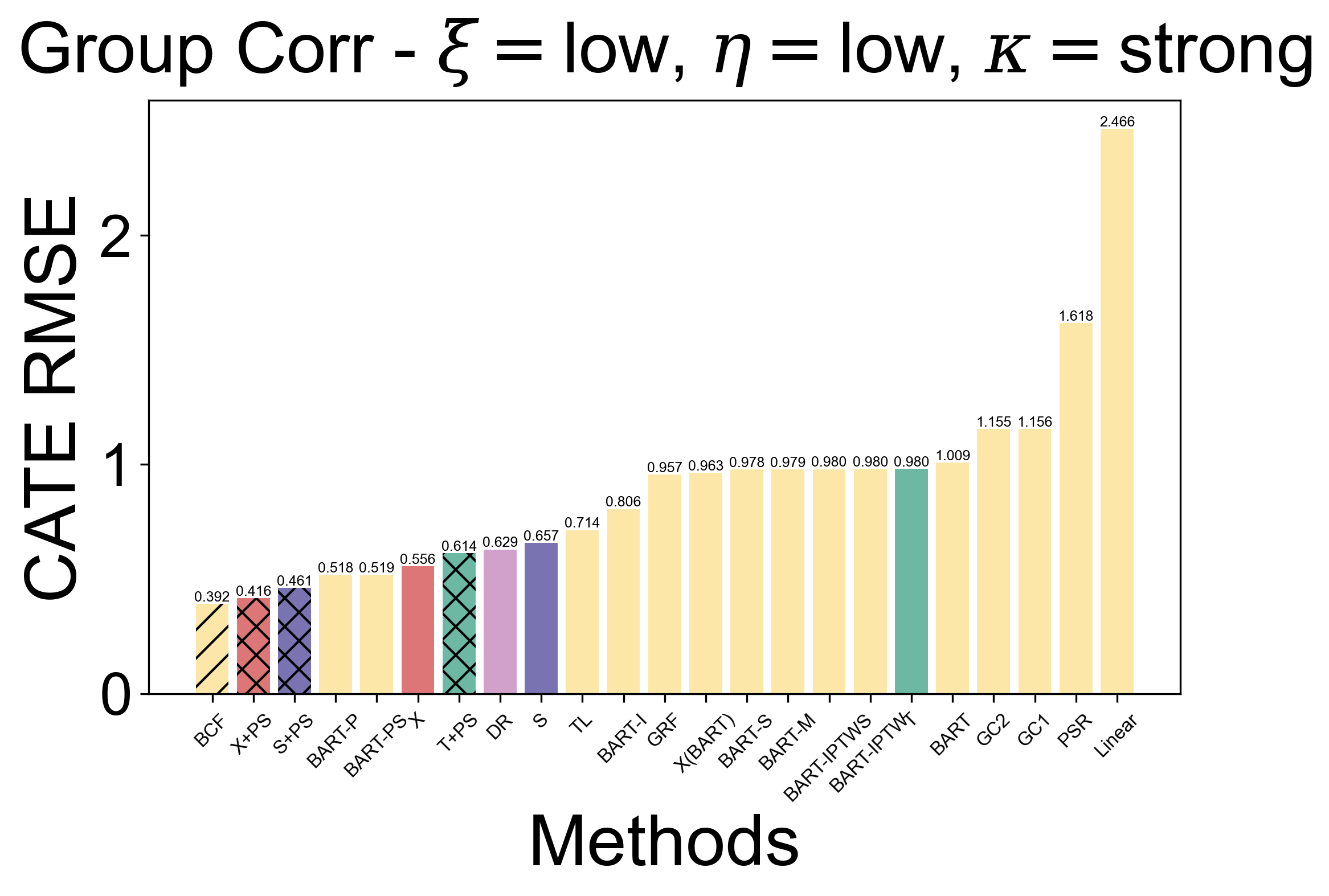}
    \caption{RMSE of TabPFN-based CATE estimators against state-of-the-art methods on the ACIC 2017 benchmark.
      Results are shown across panels organized by error type (columns) and selection strength $\kappa$ (rows).
      Colored bars represent meta-learners built on TabPFN: \textcolor[HTML]{403990}{S-Learner}, \textcolor[HTML]{2F9B7D}{T-Learner}, \textcolor[HTML]{CF3D3E}{X-Learner}, \textcolor[HTML]{BF7AB5}{DR-Learner}.
      The S-, T-, and X-learners enhanced with a propensity score covariate are indicated with the same colors but decorated with a shaded pattern.
      \textcolor[HTML]{FBDD85}{Bayesian Causal Forests} is shown as the shaded yellow bar.}
    \label{fig:cate_real_data}
\end{figure*}
Figure~\ref{fig:cate_real_data} reports the RMSE of each method averaged within each error type and selection strength level.
Under weak selection strength, all TabPFN-based estimators match or outperform the baselines.
Under strong selection strength, their performance degrades somewhat, which we believe is due to the phenomenon of regularization-induced confounding described in \citet{hahn2020bayesian}---an issue that BCF was specifically designed to fix.
Borrowing an idea from BCF, we enhanced the S-, T-, X-Learners by first fitting a propensity score model using a TabPFN classifier and then using the fitted score values as an additional covariate when fitting each base learner model.%
\footnote{The variant is not used for the DR-learner since it already makes use of the propensity score to impute pseudo-outcomes.}
Under strong selection strength, the resulting methods, especially the X-Learner, all enjoy improved performance that is competitive with the state-of-the-art.
Finally, we note that the DR-learner, despite its semi-parametric optimality guarantees, again does not have the best performance in any setting.
Full experimental details and per-scenario results are provided in Appendix~\ref{app:causal_acic}.

%% file: sections/covariateshift.tex
\subsection{Study III: prediction under covariate shift}
\label{sec:covariate_shift}

\subsubsection{Problem and motivation}
In regression and classification problems, covariate shift refers to the scenario where the marginal covariate distribution $p(x)$ differs between training and test datasets, while the predictive distribution $p(y|x)$ remains the same.
This distributional mismatch frequently arises in real-world applications—for instance, due to outdated training data or sampling biases between datasets.
Since covariate shift can substantially degrade model performance, it has become an active area of research~\citep{qiu2023prediction,yang2024doubly,qin2024distribution}.

More precisely, we observe a training dataset $\gD_s$ comprising $n$ IID observations from a source distribution $p_s(x)p(y|x)$ and wish to predict the responses on a test dataset $\gD_t$ comprising $m$ IID observations from the target distribution $p_{t}(x)p(y|x)$, with accuracy evaluated via prediction MSE.
Furthermore, $p_s(x)$ and $p_t(x)$ are usually unknown.

\subsubsection{State-of-the-art}
Classical approaches to address covariate shift make use of importance weighting techniques.
More recently, the problem has been studied under the assumption that the regression function $f(x) = \sE(Y|X = x)$ has bounded norm in a reproducing kernel Hilbert space, in which case kernel ridge regression (KRR) with a carefully tuned regularization parameter $\lambda$ was shown to achieve optimal rates, whereas a na{\"i}ve, unadjusted estimator is suboptimal~\citep{ma2023optimally}.
\citet{wang2024pseudo} introduced a method based on ``pseudo-labeling'' (\textbf{PL}) for selecting $\lambda$, claiming that it improves upon \citet{ma2023optimally}'s work by relaxing certain distributional assumptions.
Specifically, PL fits KRR on the first half of $\gD_s$ and uses the fitted model to impute labels for an auxiliary unlabeled dataset $\widetilde{\gD}_{t}$ drawn from $p_t(x)$.
It then fits a collection of KRR models with different $\lambda$ parameters on the second half of $\gD_s$ and uses $\widetilde{\gD}_{t}$ with the imputed labels to select the best-performing model.

\subsubsection{Experimental design}
We repeat \citet{wang2024pseudo}'s experiments, but add a simple comparator approach which directly fits a model on $\gD_s$ using \textbf{TabPFN} and does not make any adjustment.
In addition to PL, the following methods were used as comparators:
\begin{itemize}[leftmargin=*]
\item \textbf{Naive}: Fit a gradient boosting regression model on $\gD_{s}$, tuned using five-fold CV.

\item \textbf{Importance-Weighted (IW)}: Fit a gradient boosting model on $\gD_{s}$ similarly to the naive approach, but with the loss function weighted by the true density ratio $p_t(x)/p_s(x)$.
\item \textbf{Wang-Oracle}: A variant of PL where the true conditional mean $\mathbb{E}(Y|X=x)$ is used for $\widetilde{\gD}_{t}$ instead of imputed values.
This serves as an oracle benchmark within \citet{wang2024pseudo}.
\end{itemize}
We consider univariate covariate distributions that are mixtures of uniform distributions.
The train and test distributions have the same components but different mixing weights:
\[
\begin{aligned}
p_s &= (5/6)\mathrm{Unif}(0,0.5)+(1/6)\mathrm{Unif}(0.5, 1),\\
p_t &= (1/6)\mathrm{Unif}(0, 0.5)+(5/6)\mathrm{Unif}(0.5, 1).
\end{aligned}
\]
The response variable follows $Y|X=x \sim \gN(f(x), 1)$, where we investigate five distinct mean functions: (i) $f(x)=\cos(2\pi x)-1$; (ii) $f(x)=\sin(2\pi x)$; (iii) $f(x)=|x-1/2|-1/2$; (iv) $f(x)=f_1(x)+f_2(x)-2$, $f_1(x)=\min\{1,|4x-1|\}$ and $f_2(x)=\min\{1,|4x-3|\}$; and (v) $f(x)=x\sin(4\pi x)$.
In each experiment, we generate training and test datasets as described above, with the size of the test set fixed at $m=10^4$, while we vary the training set size $n \in \{500,600,\dots,1,100\}$.
The results are averaged over 500 replicates.

\subsubsection{Results}
\begin{figure}[!ht]
\centering
\includegraphics[width=0.45\columnwidth]{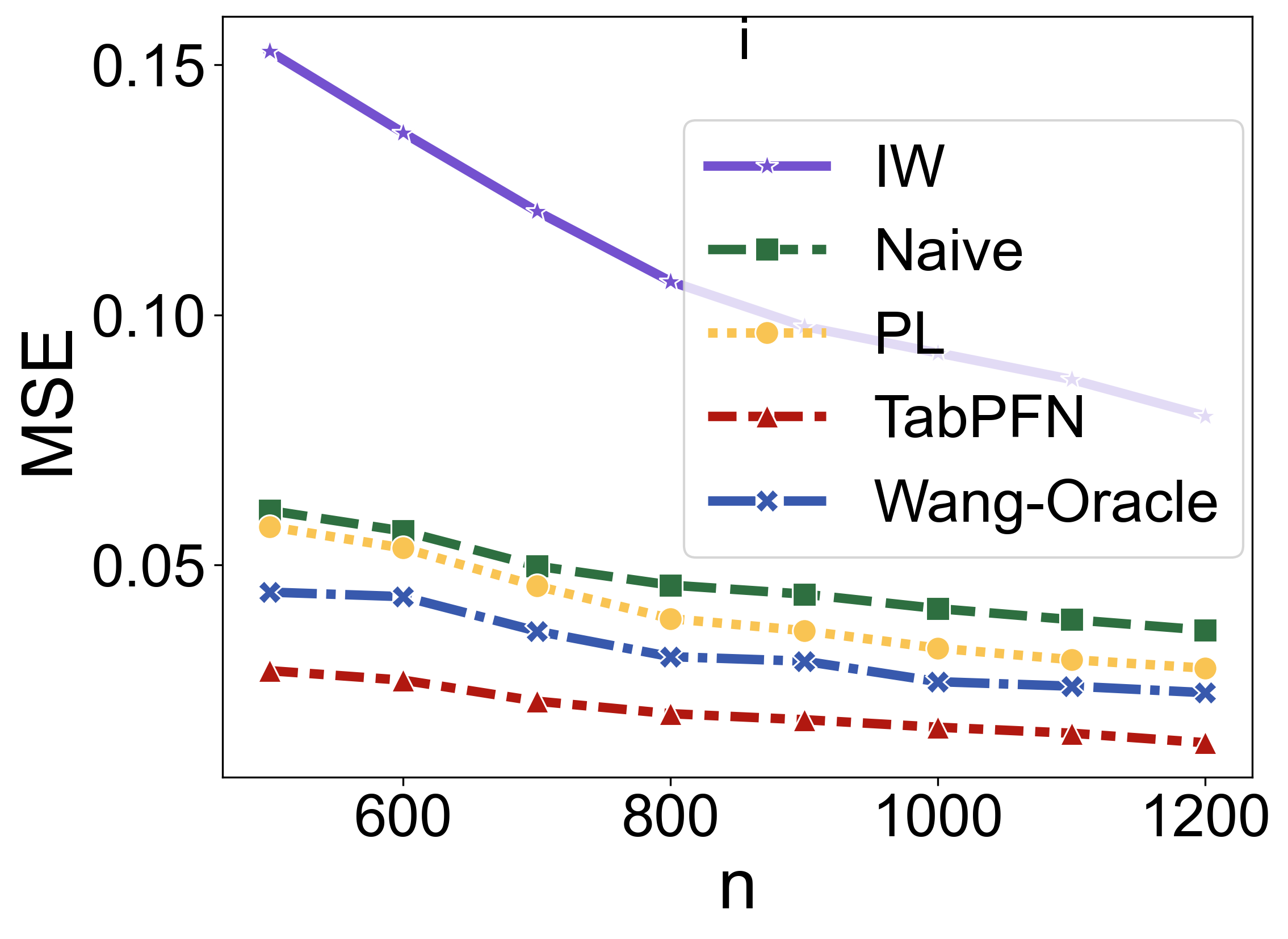}
\includegraphics[width=0.45\columnwidth]{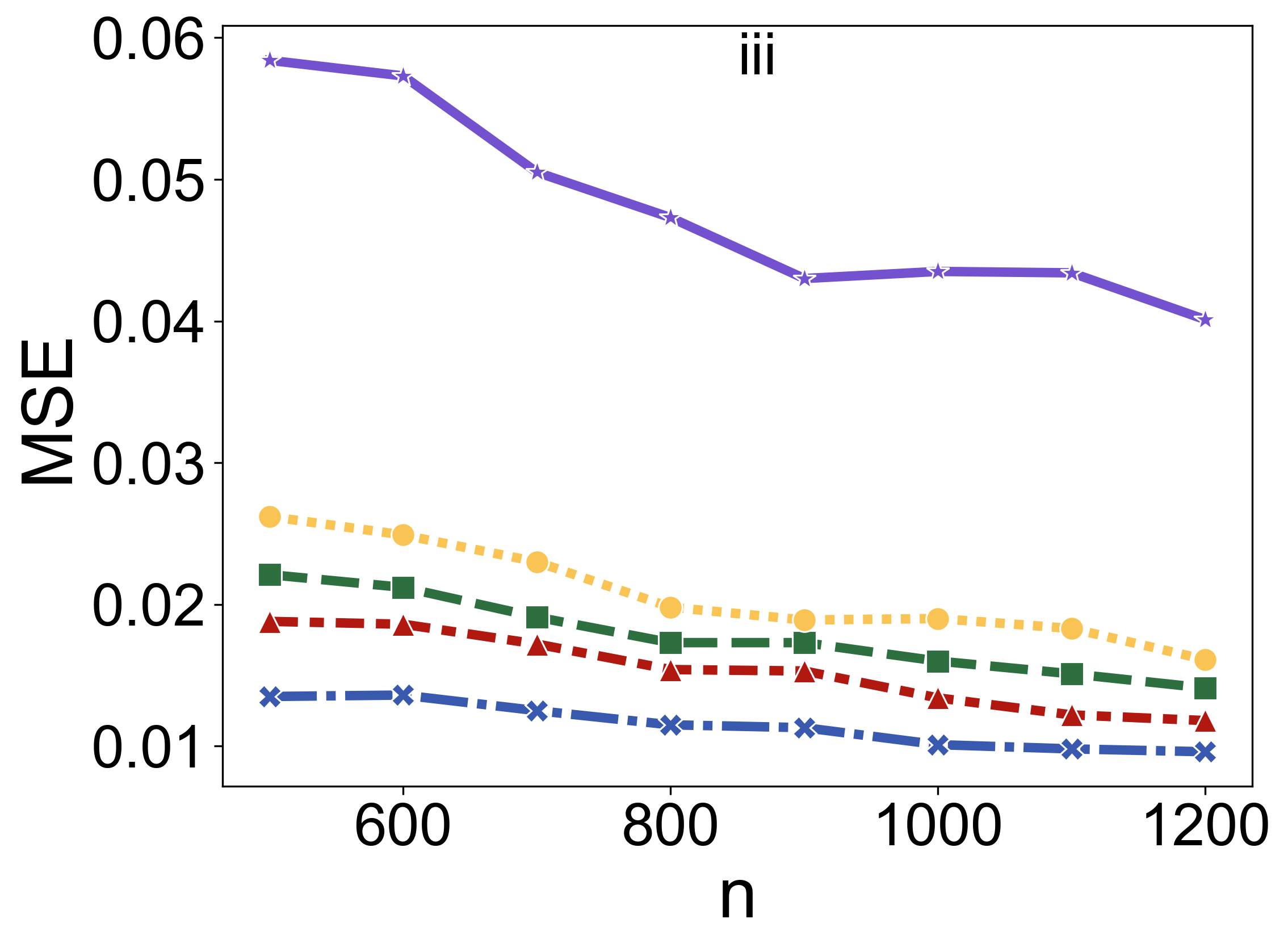}\\
\includegraphics[width=0.45\columnwidth]{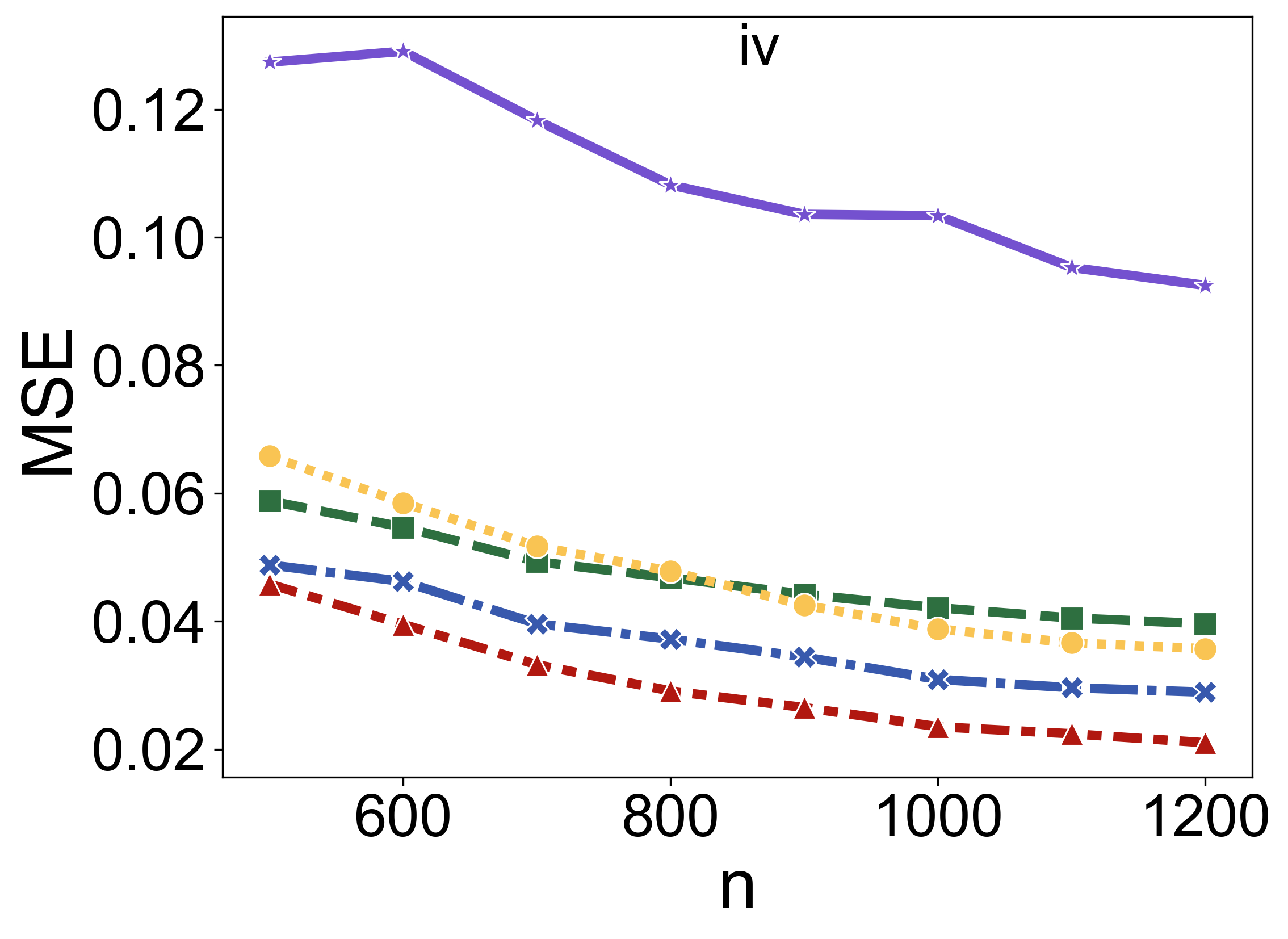}
\includegraphics[width=0.45\columnwidth]{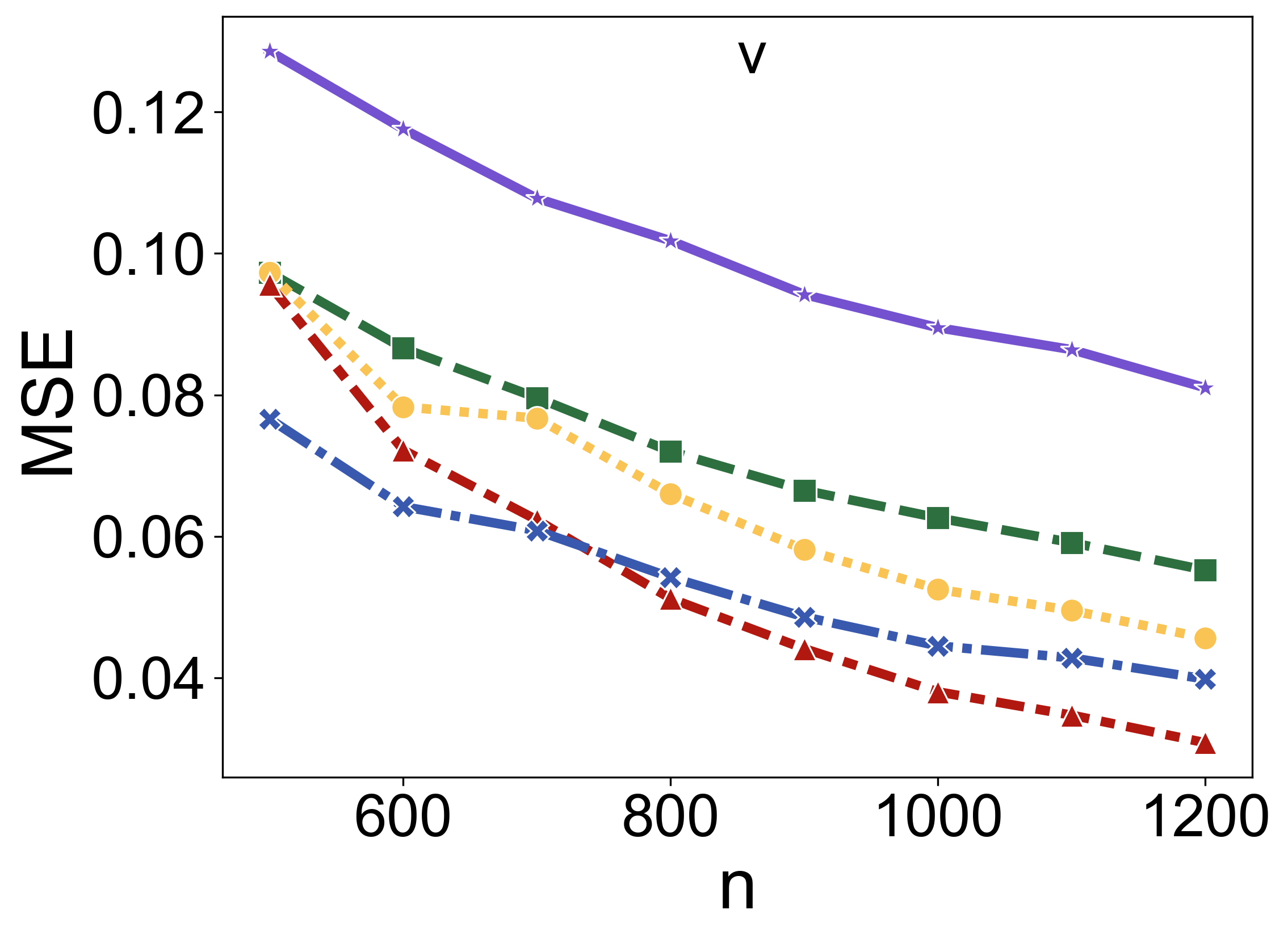}
\caption{Comparison of prediction MSE for different covariate shift methods across varying sample sizes ($n$) and scenarios.}
\label{fig:covariate-shift}
\end{figure}

Figure~\ref{fig:covariate-shift} shows the prediction MSE as a function of $n$ for all five methods, with respect to the mean functions (i), (iii), (iv), and (v) (results for setting (ii), which shows similar patterns to (i), are provided in Appendix~\ref{app:covariate-shift}).
We observe that TabPFN consistently outperforms the PL method, even matching or exceeding the oracle method's performance at larger sample sizes ($n \geq 800$), in all settings apart from (iii).
This is despite TabPFN not making use of the auxiliary information in $\widetilde{\gD}_t$ required by PL and the oracle method.

\subsubsection{Real data comparison}
To assess performance beyond the univariate synthetic setting of \citet{wang2024pseudo}, we evaluate the methods on two multivariate UCI regression datasets: \textit{Airfoil Self-Noise} ($n=1503$, 5 covariates) and \textit{Concrete Compressive Strength} ($n=1030$, 8 covariates).
For each dataset, we perform a 70/30 train--test split and resample the test set with replacement using weights $\omega(x)=\exp(x^\top\boldsymbol{\beta})$, following the procedure of~\citet{tibshirani2019conformal}.
Two shift vectors are considered per dataset, and results are averaged over 10 repetitions.

Table~\ref{tab:covariate-shift-real} reports the mean prediction MSE and standard error across repetitions.
Across all datasets and shift settings, TabPFN achieves the lowest prediction MSE, often by a substantial margin.
Full experimental details are provided in Appendix~\ref{app:realdata_covshift}.

\begin{table}[tp]
\centering
\caption{Prediction MSE (mean $\pm$ standard error) over 10 repetitions under covariate shift.
  Each dataset is evaluated under two shift vectors.}
\label{tab:covariate-shift-real}
\resizebox{\linewidth}{!}{%
\begin{tabular}{llccccc}
\toprule
Dataset & Setting & \shortstack{KRR test-label\\oracle}
  & PL & Naive XGBoost & IW XGBoost & TabPFN \\
\midrule
Airfoil & $\boldsymbol{\beta}_{1}$
  & $12.60 \pm 1.51$ & $14.12 \pm 1.66$
  & $11.64 \pm 3.93$ & $11.74 \pm 3.90$
  & $3.11 \pm 0.76$ \\
 & $\boldsymbol{\beta}_{2}$
  & $8.47 \pm 0.55$ & $8.99 \pm 0.62$
  & $3.57 \pm 0.44$ & $3.60 \pm 0.19$
  & $1.65 \pm 0.15$ \\
\midrule
Concrete & $\boldsymbol{\beta}_{1}$
  & $47.54 \pm 13.16$ & $97.23 \pm 20.44$
  & $5.73 \pm 1.40$ & $5.28 \pm 0.71$
  & $2.56 \pm 0.45$ \\
 & $\boldsymbol{\beta}_{2}$
  & $61.12 \pm 22.44$ & $120.60 \pm 31.10$
  & $7.80 \pm 1.05$ & $6.99 \pm 0.85$
  & $4.95 \pm 0.67$ \\
\bottomrule
\end{tabular}%
}
\end{table}

%% file: sections/analysis.tex
\section{TabPFN adaptivity}
\label{sec:inductive_biases}

In this section, we illustrate the adaptivity of TabPFN, which we believe provides a compelling explanation for its strong predictive performance across a wide range of tasks, as demonstrated in the previous sections.
By adaptivity, we refer to the ability of a procedure to automatically adjust to unknown properties of the underlying data-generating process while still achieving near-optimal performance.
This notion has deep roots in nonparametric statistics, where it was classically studied in the context of regression.
Foundational work by \citet{lepskii1991problem}, and later by \citet{donoho1995adapting}, established estimators that could adapt to unknown smoothness levels of a target function, achieving optimal rates without prior specification of regularity.

However, the form of adaptivity exhibited by TabPFN appears both different and broader.
Unlike traditional methods that adapt within a single nonparametric framework—typically modulating local smoothness or bandwidth—TabPFN seems capable of flexibly behaving like a highly nonparametric method when the signal is irregular, yet also exploiting parametric structure when present.
In other words, rather than committing to either a parametric or nonparametric regime, TabPFN dynamically adjusts across these regimes, a property that is rare among classical estimators.
We will investigate this property in the context of both regression and classification.

\subsection{Adaptivity in regression}

To investigate adaptivity in regression, we consider the problem of sparse linear regression and compare the performance of TabPFN against that of specialized methods, namely LASSO, SCAD, and MCP regression, across a range of data-generating settings.
We find that TabPFN can outperform these methods and can even be used to estimate regression parameters with smaller bias than LASSO.
To complement this analysis, we also perform a visual comparison of TabPFN with Gaussian process regression in Section~\ref{app:extrapolation}, observing that the two methods perform similarly, but with subtle differences in their inductive biases.
Section~\ref{app:extrapolation} also evaluates predictive uncertainty.
Across five real regression benchmarks, TabPFN's mean $95\%$ coverage is $95.2\%$, compared with $93.8\%$ for Gaussian process regression, and its mean CRPS is lower ($0.212$ versus $0.257$); the corresponding reliability and PIT diagnostics are reported in Figure~\ref{fig:realdata-uq} and Table~\ref{tab:realdata-uq}.

\subsubsection{Experimental design}
We follow the experimental design of~\citet{hastie2020best} and generate $Y = X^{\top} \beta^* + \epsilon$, with two types of $\beta^* \in \sR^{p}$ having sparsity $s$ (\ie $s$ entries are $1$ and the rest are $0$):
\begin{itemize}
    \item Beta-type I: $\beta^*$ has $s$ components equal to $1$, spaced approximately evenly across the $p$ indices, and the rest equal to $0$;
    \item Beta-type II: $\beta^*$ has its first $s$ components equal to $1$, and the rest equal to $0$.
\end{itemize}
Features corresponding to non-zero (zero) coefficients are termed relevant (irrelevant).
We fix $p=100$ and vary $s\in \{1,5,10,20,30\}$.
The features $\{x_i:i\in [n]\}$ are drawn i.i.d.
from $\gN_p(0,\Sigma)$, where $\gN_{p}$ denotes a $p$-dimensional Gaussian and $\Sigma$ is either the identity matrix or a matrix with $(i, j)$-th entry equal to $0.35^{|i-j|}$ (banded correlation).
Since beta-type I and II are equivalent when $\Sigma$ is identity, this gives $3$ distinct experimental scenarios.
The response $y_i$ is generated with additive Gaussian noise, whose variance $\sigma^2$ is chosen to meet the desired signal to noise ratio (SNR), denoted by $\nu$.
Specifically, we set $\sigma^2 = (\beta^*)^{\top}\Sigma\beta^*/\nu$.
We consider SNR values $\nu \in \{0.05, 0.25, 1.22, 6\}$, spanning low to high noise regimes.
Finally, we select $n \in \{50, 500\}$.
This yields $5~(\text{sparsity})\times 3~(\text{beta and design matrix}) \times 4~(\text{SNR}) \times 2~\text{(sample size)}=120$ experimental settings.
For each setting, we evaluate performance on an independent test set of size $1000$, with the results averaged over $100$ experimental replicates.
Regularization parameters for the comparators are selected using prediction error on an independent validation set of size $1{,}000$.

\subsubsection{Prediction performance comparison}
\begin{figure*}[!ht]
\centering
\includegraphics[width=0.24\linewidth]{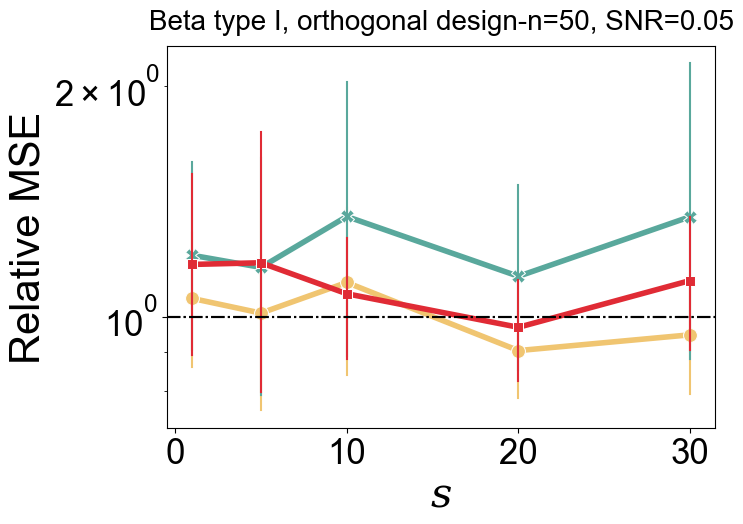}
\includegraphics[width=0.24\linewidth]{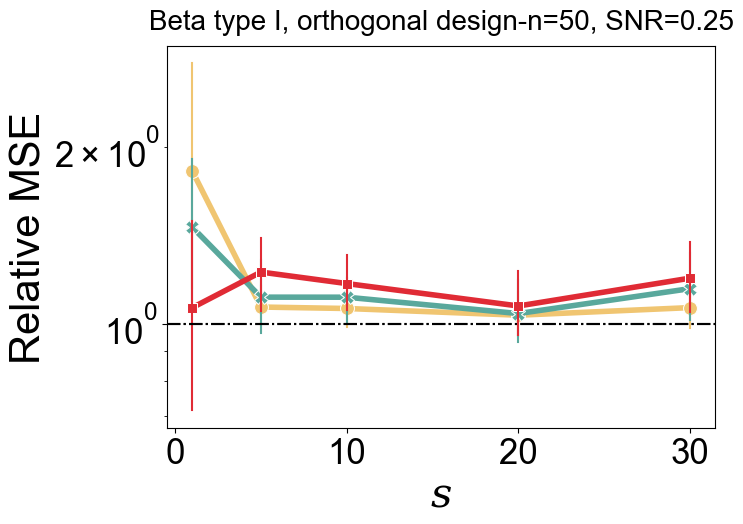}
\includegraphics[width=0.24\linewidth]{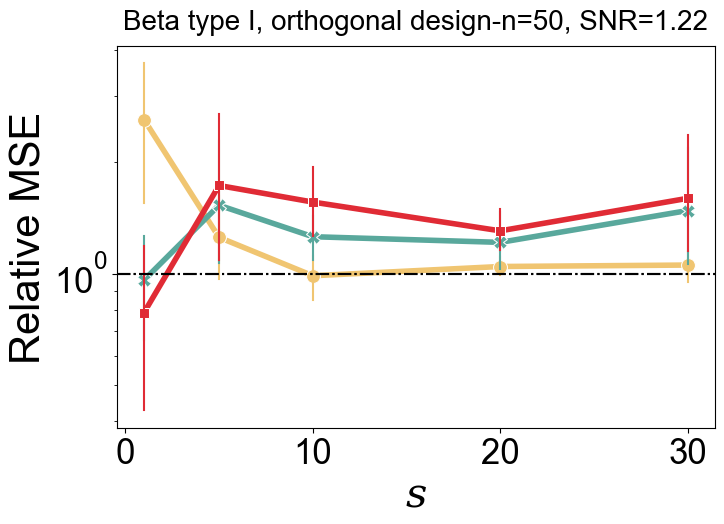}
\includegraphics[width=0.24\linewidth]{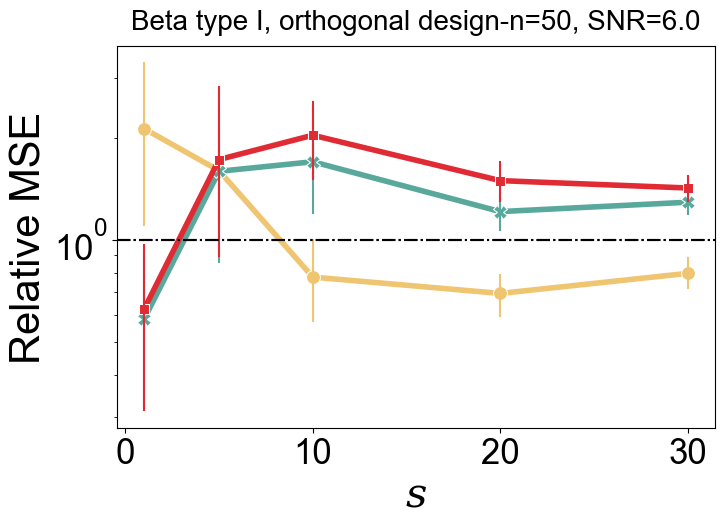}\\
\includegraphics[width=0.24\linewidth]{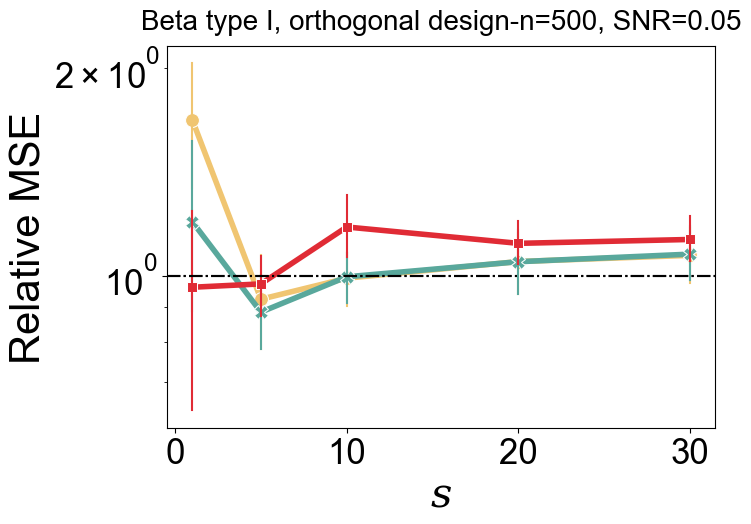}
\includegraphics[width=0.24\linewidth]{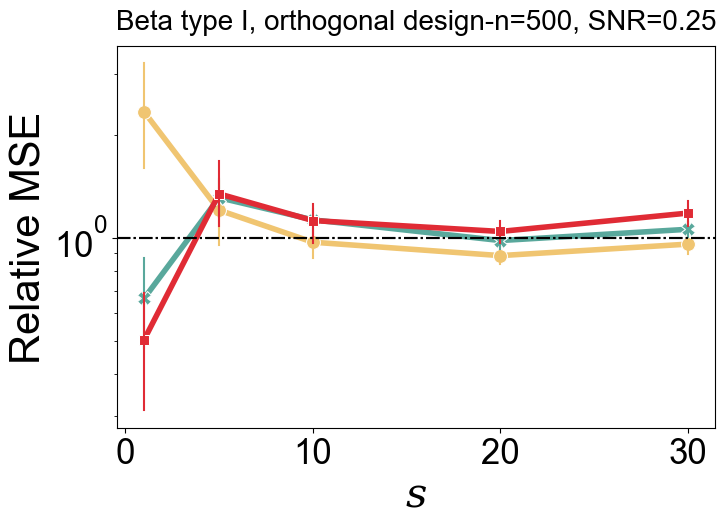}
\includegraphics[width=0.24\linewidth]{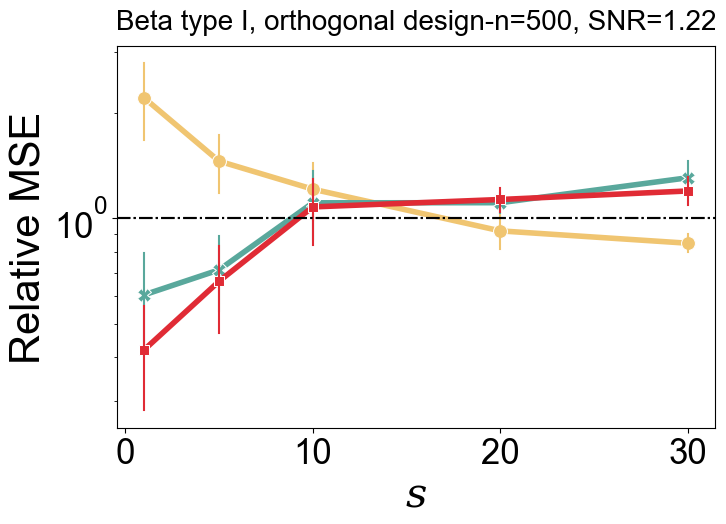}
\includegraphics[width=0.24\linewidth]{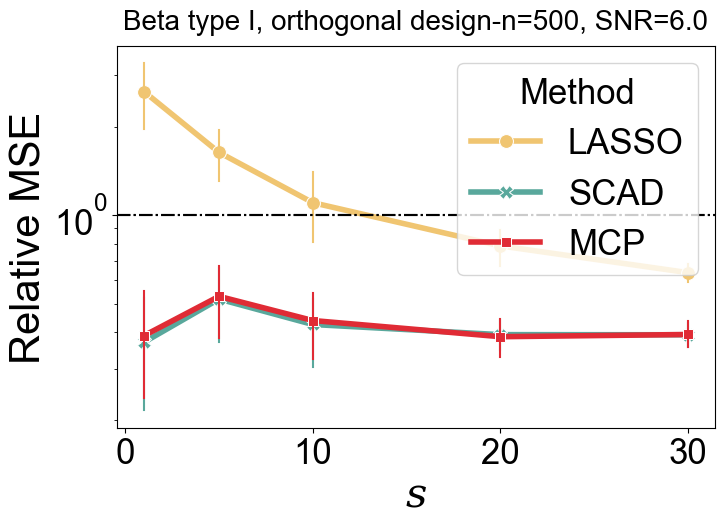}
\caption{Relative test MSE compared to TabPFN, using a beta-type I orthogonal design.
  Rows show results for $n=50$ (top) and $n=500$ (bottom); columns show increasing SNR.}
\label{fig:test_MSE}
\end{figure*}

In Figure~\ref{fig:test_MSE}, we plot the relative test MSE (averaged over $100$ repetitions) of the comparator methods to that of TabPFN.
We focus on the orthogonal feature case, with the results for correlated features deferred to Appendix~\ref{app:linear_regression_more}.
Compared to LASSO, we observe that TabPFN tends to perform better for $s \leq 10$, with the performance gap increasing for higher sample size and SNR.
Conversely, compared to SCAD and MCP, TabPFN tends to perform better for $s \geq 5$ and for lower sample size and SNR.
In other words, TabPFN seems to achieve a type of middle-ground between the two classes of methods.

\begin{table*}[!tp]
\caption{$n=500$, beta-type I, orthogonal design (simple linear regression)}
\label{tab:linear}
\centering
\resizebox{\textwidth}{!}{
\begin{tabular}{@{}cc@{\hspace{0.5em}}cc@{\hspace{1.5em}}cc|@{\hspace{0.5em}}cc@{\hspace{1.5em}}cc|@{\hspace{0.5em}}cc@{\hspace{1.5em}}cc|@{\hspace{0.5em}}cccc@{}}
\toprule
& & \multicolumn{4}{c}{TabPFN} & \multicolumn{4}{c}{LASSO} & \multicolumn{4}{c}{SCAD} & \multicolumn{4}{c}{MCP} \\
\cmidrule(lr){3-6} \cmidrule(lr){7-10} \cmidrule(lr){11-14} \cmidrule(lr){15-18}
SNR & $s$ & \multicolumn{2}{c}{Relevant} & \multicolumn{2}{c}{Irrelevant} & \multicolumn{2}{c}{Relevant} & \multicolumn{2}{c}{Irrelevant} & \multicolumn{2}{c}{Relevant} & \multicolumn{2}{c}{Irrelevant} & \multicolumn{2}{c}{Relevant} & \multicolumn{2}{c}{Irrelevant} \\
& & Bias$^2$ & Var & Bias$^2$ & Var & Bias$^2$ & Var & Bias$^2$ & Var & Bias$^2$ & Var & Bias$^2$ & Var & Bias$^2$ & Var & Bias$^2$ & Var \\
\midrule
\multirow{5}{*}{1.22}
& 1 & 0.0000 & 0.0012 & 0.0000 & 0.0000 & 0.0058 & 0.0016 & 0.0000 & 0.0000 & 0.0000 & 0.0012 & 0.0000 & 0.0000 & 0.0000 & 0.0012 & 0.0000 & 0.0000 \\
& 5 & 0.0004 & 0.0081 & 0.0000 & 0.0004 & 0.0172 & 0.0085 & 0.0000 & 0.0006 & 0.0001 & 0.0077 & 0.0000 & 0.0001 & 0.0001 & 0.0077 & 0.0000 & 0.0002 \\
& 10 & 0.0028 & 0.0196 & 0.0000 & 0.0025 & 0.0244 & 0.0174 & 0.0000 & 0.0020 & 0.0002 & 0.0180 & 0.0000 & 0.0007 & 0.0002 & 0.0179 & 0.0000 & 0.0007 \\
& 20 & 0.0175 & 0.0500 & 0.0004 & 0.0099 & 0.0303 & 0.0360 & 0.0001 & 0.0079 & 0.0005 & 0.0400 & 0.0001 & 0.0051 & 0.0005 & 0.0407 & 0.0001 & 0.0049 \\
& 30 & 0.0387 & 0.0783 & 0.0018 & 0.0177 & 0.0301 & 0.0579 & 0.0002 & 0.0166 & 0.0011 & 0.0707 & 0.0002 & 0.0146 & 0.0011 & 0.0719 & 0.0002 & 0.0142 \\
\midrule
\multirow{5}{*}{6}
& 1 & 0.0000 & 0.0003 & 0.0000 & 0.0000 & 0.0022 & 0.0002 & 0.0000 & 0.0000 & 0.0000 & 0.0002 & 0.0000 & 0.0000 & 0.0000 & 0.0002 & 0.0000 & 0.0000 \\
& 5 & 0.0000 & 0.0016 & 0.0000 & 0.0001 & 0.0038 & 0.0017 & 0.0000 & 0.0001 & 0.0000 & 0.0015 & 0.0000 & 0.0000 & 0.0000 & 0.0016 & 0.0000 & 0.0000 \\
& 10 & 0.0004 & 0.0037 & 0.0000 & 0.0006 & 0.0053 & 0.0035 & 0.0000 & 0.0004 & 0.0000 & 0.0033 & 0.0000 & 0.0001 & 0.0000 & 0.0033 & 0.0000 & 0.0001 \\
& 20 & 0.0022 & 0.0089 & 0.0001 & 0.0028 & 0.0070 & 0.0073 & 0.0000 & 0.0014 & 0.0001 & 0.0067 & 0.0000 & 0.0002 & 0.0001 & 0.0067 & 0.0000 & 0.0002 \\
& 30 & 0.0055 & 0.0168 & 0.0003 & 0.0061 & 0.0079 & 0.0116 & 0.0000 & 0.0027 & 0.0001 & 0.0110 & 0.0000 & 0.0004 & 0.0001 & 0.0110 & 0.0000 & 0.0004 \\
\bottomrule
\end{tabular}}
\end{table*}

\subsubsection{Descriptive accuracy}

Beyond predictive accuracy, we also consider the descriptive accuracy of the TabPFN model---how well it reflects the sparse linear generating model and how it may be used to estimate the true regression coefficients.
We can visually inspect the fitted TabPFN model using Accumulated Local Effects (ALE) plots, which summarize the dependence of the model on each individual feature \citep{apley2020visualizing}.%
\footnote{ALE plots avoid both the confounding associated with conditional dependence plots and the out-of-sample instability associated with marginal dependence plots.}
Figure~\ref{fig:ALE} in the Appendix exhibits the results when the features are orthogonal.
We observe that for moderate to high SNR ($\geq 1.22$), almost all ALE plots of relevant features are roughly linear.
In addition, when $n=500$, the ALE plots for the irrelevant features display much less variation, which suggests that the TabPFN model is indeed well-approximated by a low-dimensional linear surface.
For smaller SNR, the ALE plots are generally nonlinear and are not shown.

The linearity of the ALE plots suggests a two-step pipeline for using TabPFN to estimate the true regression coefficients.
We first fit a TabPFN model to the data and then fit an ordinary least squares model to the TabPFN fitted values.
In Table~\ref{tab:linear}, we compare the accuracy of this procedure to that of LASSO, SCAD, and MCP, further breaking down the error into squared bias and variance.
For smaller values of $s$, we observe that TabPFN performs similarly to SCAD and MCP.
Its coefficients have much smaller bias than LASSO while having comparable variance, thus avoiding the well-known tendency of LASSO to bias its estimates via shrinkage.

%% file: sections/noisy_label.tex
\subsection{Adaptivity in classification}
\label{sec:label_noise}
We examine the classification performance of TabPFN in the presence of label noise, that is, when the observed labels $\tilde{Y}_1,\tilde{Y}_2,\ldots,\tilde{Y}_n$ in our training set are corrupted from their true values $Y_1,Y_2,\ldots,Y_n$, such as by coding errors.
For simplicity, we consider homogeneous noise in a binary class setting: $\mathbb{P}(\widetilde{Y} \neq Y | X=x, Y=y) = \rho, y \in \{0,1\},$ where $\rho$ is the corruption fraction.
The goal is to train a classifier to predict the \emph{original} label, namely, we want to estimate a mapping $C: \mathbb{R}^{d} \to \{0,1\}$ minimizing the risk $R(C) = \mathbb{E}[C(X) \neq Y].$

This problem is especially interesting when the conditional distribution of the true label, $p(Y|X=x)$, has a simple parametric form.
For example, if $X|Y=y$ is Gaussian, the noise-free classification problem can be solved efficiently via linear discriminant analysis (LDA).
On the other hand, the conditional distribution of the corrupted label, $p(\tilde{Y}|X=x)$, does not have the correct functional form, which makes LDA inconsistent.
In comparison, \citet{cannings2020classification} proved that $k$-nearest neighbors ($k$NN) and kernelized support vector machines (SVM) can be robust to label noise.
However, these methods, which are based on local averaging, pay a price in terms of being less efficient when the labels are noise free.

We show that TabPFN is able to break this robustness-efficiency trade-off.
At the sample sizes studied here, it appears robust to label noise.
On the other hand, it is more efficient than $k$NN, SVM and even LDA across all simulation settings, including noiseless settings.

\subsubsection{Experimental design}
Following~\citet{cannings2020classification}, we generate data from two models:
\begin{itemize}
    \item[\textbf{M1}] $\mathbb{P}(Y=1)=0.9$ with $X|Y = r \sim \gN_{5}(\mu_r, I)$, where $\mu_1 = (3/2, 0, \ldots, 0)^{\top} = -\mu_0$.
    \item[\textbf{M2}] $X \sim U([0, 1]^5)$ with $\mathbb{P}(Y = 1 | X = x) = \min\{4(x_1 - 1/2)^2 + 4(x_2 - 1/2)^2, 1\}$.
\end{itemize}
For each model, we generate an IID sample of size $n+10^4$, reserving the first $n$ observations for training and the remaining $10^4$ for testing.
For the training set, we set $\rho \in \{0.1, \ldots, 0.4\}$.
The optimal Bayes classifier minimizes this risk: $C^{\text{Bayes}}(x) = \mathbbm{1}(\eta(x) > 0.5),$ where $\eta(x) = \mathbb{P}(Y=1|X=x)$.
For any classifier $C'$, we define its excess risk as $R(C') - R(C^{\text{Bayes}})$.

\begin{figure}[!htbp]
\centering
\includegraphics[width=0.45\linewidth]{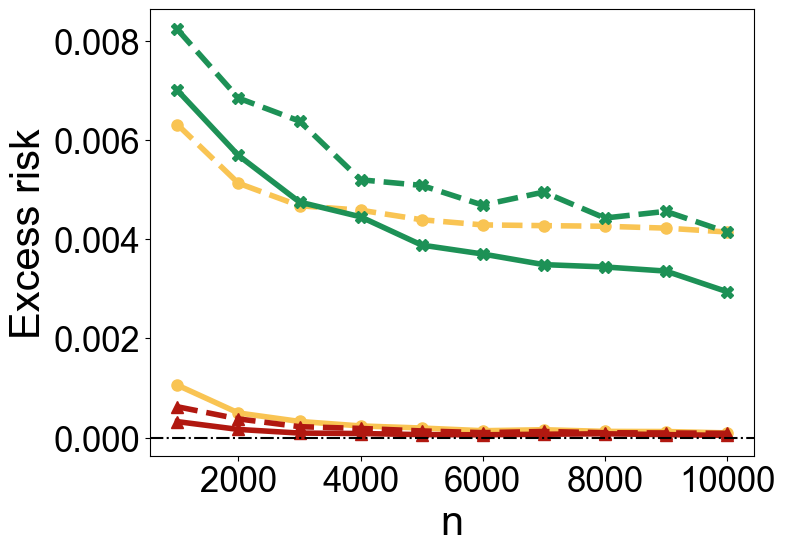}
\includegraphics[width=0.45\linewidth]{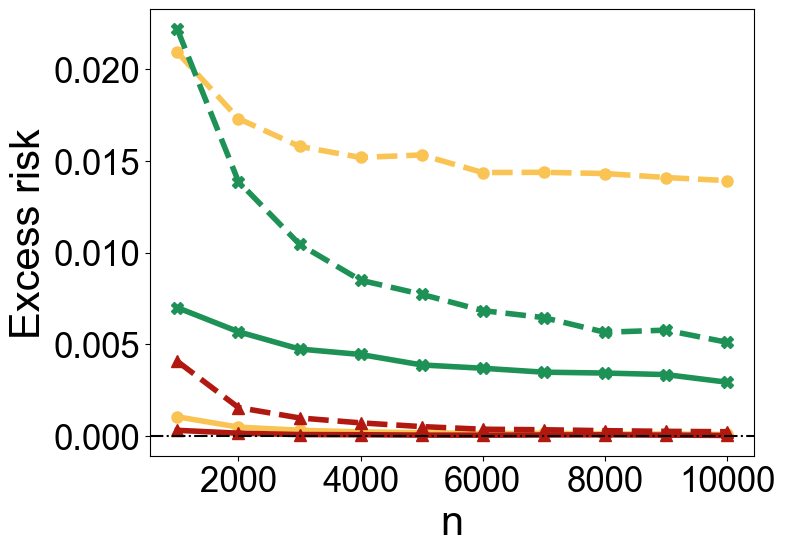}\\
\includegraphics[width=0.45\linewidth]{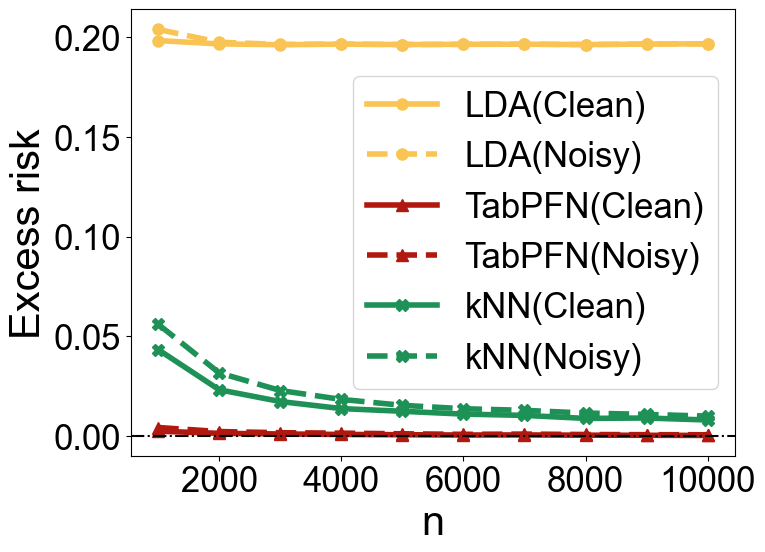}
\includegraphics[width=0.45\linewidth]{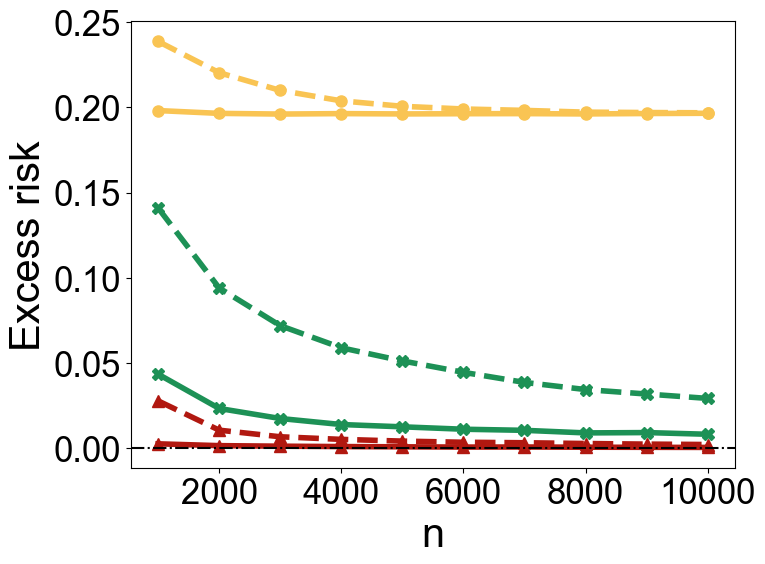}
\caption{Excess risks under Model M1 (left) and M2 (right).
  Left panels: $\rho=0.1$; Right panels: $\rho=0.3$.}
\label{fig:label_noise_performance}
\end{figure}

We compare three fitted classifiers ($k$-NN, LDA, and TabPFN) with the Bayes rule.
For each training dataset, each classifier is trained on two versions of the data: a clean training set and a noise-corrupted version, yielding six fitted classification models in total.
Details such as hyper-parameter selection are deferred to Appendix~\ref{app:noisy_label}.
This process is repeated 1000 times, with Figure~\ref{fig:label_noise_performance} showing the average excess risk across repetitions.

\subsubsection{Prediction performance comparison.}
\begin{figure}[htbp]
\centering
\includegraphics[width=0.45\linewidth]{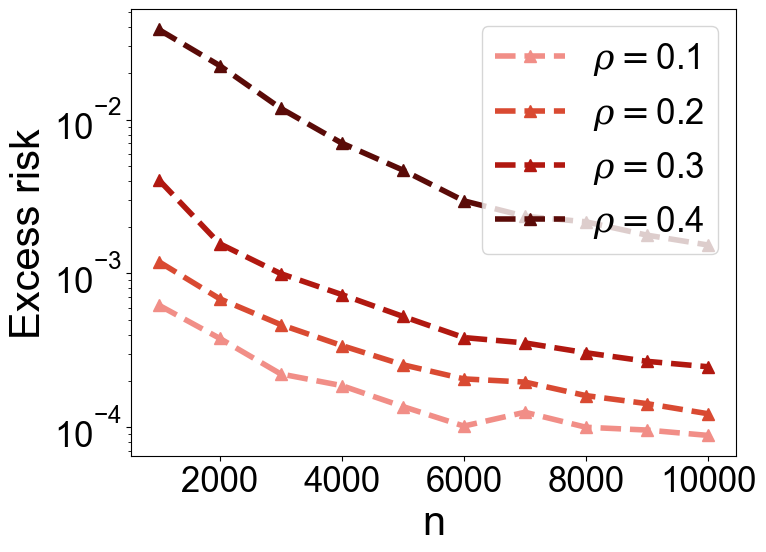}
\includegraphics[width=0.45\linewidth]{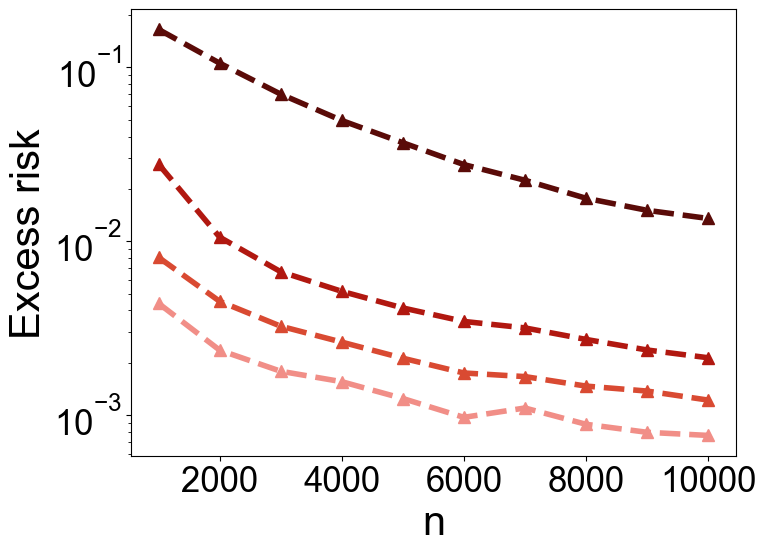}
\caption{TabPFN (noisy) excess risk for varying $\rho$ values under M1 (left) and M2 (right).}
\label{fig:tabpfn_noisy_varying_rho}
\end{figure}

Under both generative models, TabPFN (clean) performs as well as the optimal Bayes classifier when the training set size exceeds $2{,}000$.
TabPFN (noisy) shows slightly worse performance than its clean-data counterpart but remains robust to label noise across the evaluated settings.
Even with noise levels as high as $\rho=0.3$, TabPFN (noisy) nearly matches the optimal classifier given a sufficiently large training set ($\geq 5{,}000$ samples).
LDA exhibits no robustness under M1 and performs poorly under M2 due to the nonlinear Bayes decision boundary.
While the $k$-NN classifier shows some consistency—particularly under M2 with low noise levels—it is less robust than TabPFN in other scenarios.
Moreover, Figure~\ref{fig:tabpfn_noisy_varying_rho} demonstrates that TabPFN remains robust even at $\rho=0.4$.
Interestingly, we notice that there is a significant performance gap between $\rho=0.3$ and $\rho=0.4$ for small training set sizes.

%% file: sections/conclusion.tex
\section{Discussion}
\label{sec:discussion}

\subsection{Foundation models}

In 2021, \citet{bommasani2021opportunities} coined the term ``foundation model'' to refer to ``any model that is trained on broad data (generally using self-supervision at scale) that can be adapted (e.g., fine-tuned) to a wide range of downstream tasks''.
They also noted a paradigm shift in artificial intelligence which de-prioritized the development of task-specific models in favor of building these foundation models, with LLMs being the most prominent examples.
Until recently, the modality of tabular data has resisted this trend, but the tide may be turning.
\citet{hollmann2025accurate} provided a proof-of-concept that, in addition to regression and classification, TabPFN can support ``data generation, density estimation, learning reusable embeddings and fine-tuning'', consequently claiming that TabPFN is a tabular foundation model.%
\footnote{These capabilities were not fully investigated by \citet{hollmann2025accurate} and some are not supported by their publicly available code at the time of writing.}
Our investigations support this claim by showing that simple applications of TabPFN can surpass state-of-the-art approaches to some statistical problems and suggest that tabular foundation models have the potential to reshape how a wide range of statistical tasks are approached.
In addition, \citet{pmlr-v235-van-breugel24a} have a bold vision of tabular foundational models being able to perform automatic meta-analyses, out-of-domain simulation, and other tasks for which adequate methodology does not yet exist.
This line of thinking---that there may one day be \emph{one model to rule them all}---is of course contrary to mainstream statistical thinking, which has traditionally emphasized bespoke models for different problems.
Just as AI scientists wrestle with the academic and societal consequences of foundation models, statisticians may also be forced to confront profound changes in our field.

These studies are of course preliminary, and we do not suggest that TabPFN is a perfect model---it cannot handle $\geq 500$ features and has MSE curves that seem to plateau for $n \geq 10{,}000$ samples \citep{nagler2023statistical}.%
\footnote{These drawbacks have been substantially addressed in TabPFN-2.5 \citep{grinsztajn2025tabpfn} and TabPFN-3 \citep{priorlabs2026tabpfn3}, which scale to datasets with up to 1M rows and 2,000 features.}
Furthermore, an important caveat for the ``foundation model'' framing concerns \emph{pretraining prior mismatch}.
Since TabPFN approximates Bayesian inference under its SCM prior, performance may degrade systematically in settings that are poorly represented in the pretraining distribution.
Importantly, this degradation may persist even as the sample size increases, representing a qualitatively different failure mode from ordinary estimation error.
For example, we found that, in a dense linear regression setting with 100 covariates and all true coefficients equal to $1$, TabPFN performs substantially worse than ordinary least squares, and its performance does not improve with larger sample sizes.
We hypothesize that such dense linear data-generating processes are underrepresented in the SCM pretraining prior.
Direct verification is not currently possible, as the $\sim$130M synthetic pretraining datasets are proprietary and not publicly available.
TabPFN's strong performance on the real datasets in our evaluation and in large-scale external benchmarks~\citep{erickson2025tabarena} is consistent with adequate prior support for the specific benchmark settings evaluated here, but it does not establish that practical settings are generally covered by the proprietary pretraining prior.

\subsection{In-context learning}

Another way in which TabPFN challenges current statistical thinking is via its illustration of the power of ``in-context learning'', whose historical development we now discuss.
In a landmark paper, \citet{brown2020language} studied the LLM GPT-3 and discovered that it could learn how to perform tasks (e.g., language translation) that it was not explicitly trained to do and without changing any of the model's parameters, simply by providing examples of the task in a prompt (text input) to the model.
They called this ability ``in-context learning'' (ICL).
\citet{garg2022can} later framed ICL as a statistical learning problem.
Under their framework, a prompt $P$ is a random sequence $(X_1,f(X_1),\ldots,X_n,f(X_n),X_{n+1})$, where $X_1,\ldots,X_{n+1}$ are drawn IID from a covariate distribution $D_X$ and $f$ is drawn from a distribution $D_f$ over a function class $\gF$.
A model $M$ is a fixed mapping from prompts to the (real-valued) label space and is said to have learned $\gF$ in-context (with respect to $(D_X,D_f)$) if $\sE_P[l(M(P),f(X_{n+1}))] \leq \epsilon$ for some tolerance $\epsilon$ and loss function $l$.
\citet{garg2022can} was able to show via numerical experiments that a transformer, trained from scratch on IID prompts generated as described above, could successfully learn simple function classes (e.g. linear functions) in-context, inspiring a wave of theoretical interest in the problem (see for instance \citet{zhang2024trained,bai2023transformers,kim2024transformers}).

On the other hand, it seems that \citet{brown2020language} and \citet{garg2022can}'s definitions of ICL are respectively too broad and too narrow to capture the scope of the learning strategy employed by TabPFN and other recent work claiming to implement ICL to solve statistical estimation problems.
These include causal discovery~\citep{dhir2025meta}, posterior inference for generalized linear models, factor models, and Gaussian mixture models~\citep{reuter2025can}, RNA folding time density estimation~\citep{scheuer2025kinpfn}, and Poisson mean estimation under empirical Bayes~\citep{teh2025solvingempiricalbayestransformers}.
None of these works provides a definition of ICL, but we find that the common denominator is that they \emph{train a transformer model using a large number of synthetic datasets to learn a mapping on the space of datasets}.
The target of the mapping can be a causal graph, a posterior distribution on a parameter vector, or a Poisson mean vector estimate.
In each problem, the ICL approach achieves higher accuracy than several state-of-the-art methods, such as variational inference for posterior approximation or non-parametric maximum-likelihood estimation in the case of empirical Bayes.

\subsection{Future directions}

These advances present both opportunities and challenges for the statistical community.
TabPFN's success stems from a synthesis of ideas across disciplines, coupled with non-trivial engineering efforts.
The current version~\citep{hollmann2025accurate} builds on its predecessor~\citep{hollmann2022tabpfn}, with key improvements in (1) prior data generation and (2) transformer architecture---both empirically shown to enhance performance.
As statisticians, we can contribute to this evolving field by addressing critical questions:
\begin{itemize}[leftmargin=*]
\item \textbf{Interpretability \& theoretical foundations}: Transformers, with millions of parameters, often lack interpretability.
Understanding when TabPFN succeeds or fails—along with its statistical guarantees—remains an open problem.

\item \textbf{Reliability in practice}: While TabPFN and related methods show promise as off-the-shelf tools for biomedical and other applications, their real-world reliability—particularly for inference tasks—requires rigorous evaluation.

\item \textbf{Scalability \& limitations}: Current implementations struggle with dense/high-dimensional regression and $>10$K samples.
Addressing this bottleneck is essential for broader applicability.
Recent work ~\citep{qu2025tabicl,grinsztajn2025tabpfn} offers preliminary solutions, but fundamental constraints remain.
\end{itemize}

%% file: sections/appendix.tex
\section{Experiment details for study I}
\label{app:semisup}
We adopt the same experimental setup as in~\citet{Song02042024}.
For completeness, we detail the simulation settings in Section~\ref{subsec-supp:task1-setting} and present additional results in Section~\ref{subsec-supp:task1-results}.

\subsection{Simulation settings}
\label{subsec-supp:task1-setting}
For each of the three settings in Table~\ref{tab:simulation_settings} of the main text, the data-generating model is as follows.
\begin{itemize}
\item \textbf{Linear regression working model}.
In this setting, the data is generated from
\[
Y=\alpha_0+\alpha_1^{\top} X+\alpha_2^{\top}\left[X^3-X^2+\exp(X)\right]+\epsilon,
\]
where $X\sim\gN_{p}(0,I)$ and $\epsilon\sim\gN(0,4)$.
The power and exponential operations on the vector $X$ are performed element-wise.
We let $\alpha_0=1$, with $\alpha_1$ and $\alpha_2$ both being all-ones vectors.

\item \textbf{Logistic regression working model}.
In this setting, the data is generated from
\[
\sP(Y=1\mid X)=[1+\exp(-\alpha_0-\alpha_1^{\top} X-\alpha_2^{\top} X^2)]^{-1},
\]
where $ X\sim 0.5\gN_{p}(1, \Sigma)+0.5\gN_{p}(-1,\Sigma), $ and $\Sigma$ is a matrix with $(i,j)$-th entry equal to $0.5^{\vert i-j\vert}$.
We set $\alpha_0=11$, $\alpha_1$ as a vector of all $1$s, and $\alpha_2$ as a vector of all $-1$s.

\item \textbf{Linear quantile regression working model}.
In this setting, the data is generated from
\[
Y=\alpha_0+\alpha_1^{\top} X+\alpha_2\sum_{j=1}^p x_j^2+(1+\alpha_3^{\top} X)\epsilon,
\]
where $X\sim\gN_{p}(0,I)$, and $\epsilon\sim \gN(0,1)$.
We set $\alpha_0 = \alpha_2 = 1$, $\alpha_1 = 0.5\mathbf{1}_p$, and $\alpha_3 = (0.5\mathbf{1}_{r_p}, \mathbf{0}_{p-r_p})$, where $r_p=\operatorname{round}(p/2)$ is computed using R's ties-to-even convention.
where $\mathbf{1}_d$ and $\mathbf{0}_d$ denote $d$-dimensional all-ones and all-zeros vectors, respectively.
\end{itemize}

\subsection{Interpretation of the imputation estimators}
\label{subsec-supp:task1-interpretation}
TabPFN-I is a plug-in estimator proposed for $\theta^*$, obtained by replacing the responses with TabPFN predictions and minimizing the corresponding empirical loss.
Although this estimator is well defined, its probability limit need not equal $\theta^*$ because imputation can change the corresponding population risk.
Imputation-based ERM retains the original population target when the estimating score is affine in the response and the imputed value is the conditional mean, as in linear regression and logistic regression with posterior-mean imputation, but this condition does not generally hold for quantile regression or logistic regression with mode imputation.
Consequently, we interpret the comparisons in this study as finite-sample bias--variance results rather than claims of asymptotic efficiency or consistency.
At the sample sizes considered, the variance reduction obtained from the imputed observations can outweigh the resulting imputation bias and yield a smaller MSE.

\subsection{Additional simulation results}
\label{subsec-supp:task1-results}
\begin{figure*}[htbp]
    \centering
    \subfloat[\scriptsize{Linear, $p=4$}]{
        \includegraphics[width=0.24\textwidth]{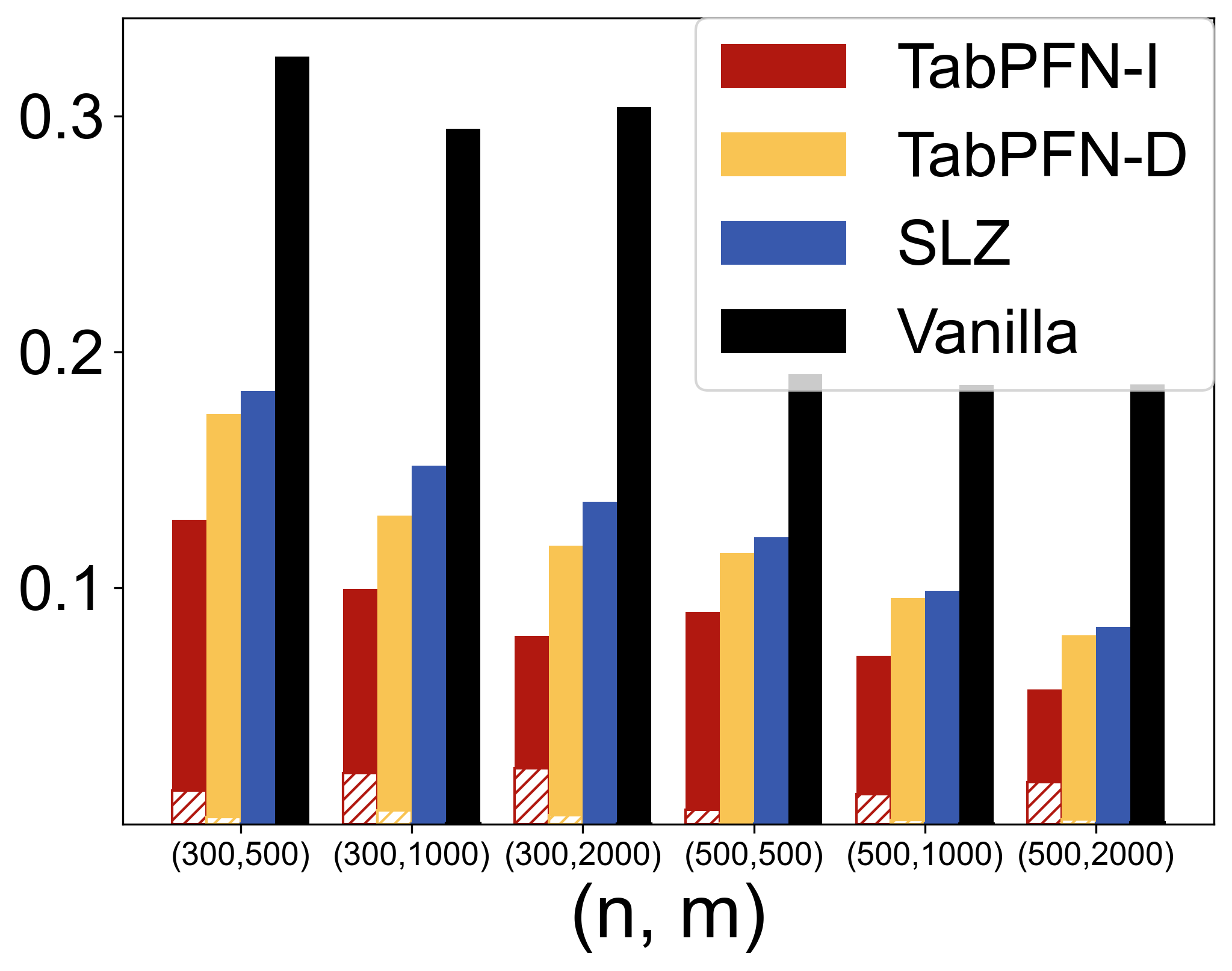}
    }
    \subfloat[\scriptsize{Logistic, $p=4$}]{
        \includegraphics[width=0.24\textwidth]{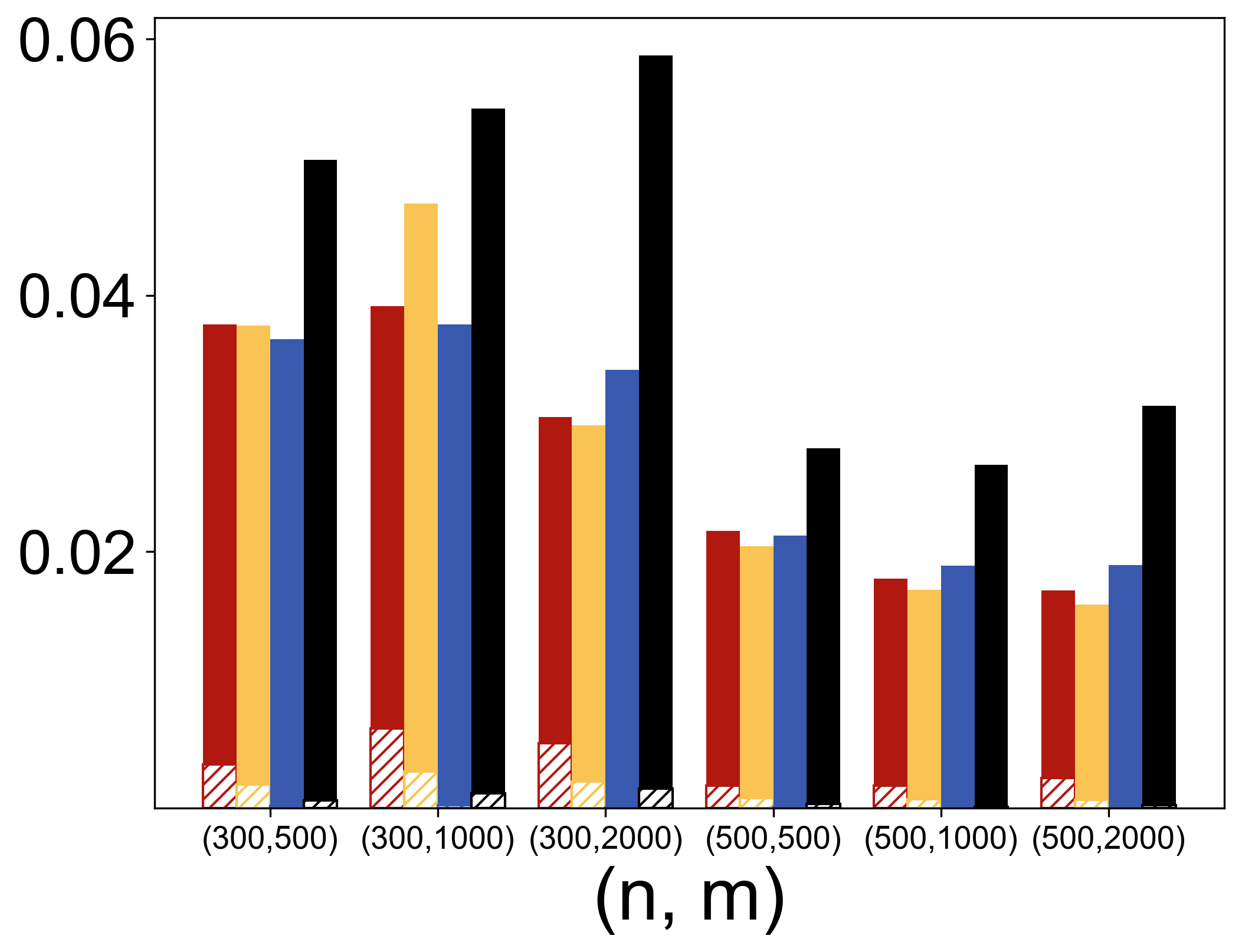}
    }
    \subfloat[\scriptsize{Quantile, $p=4, \tau=0.5$}]{
        \includegraphics[width=0.24\textwidth]{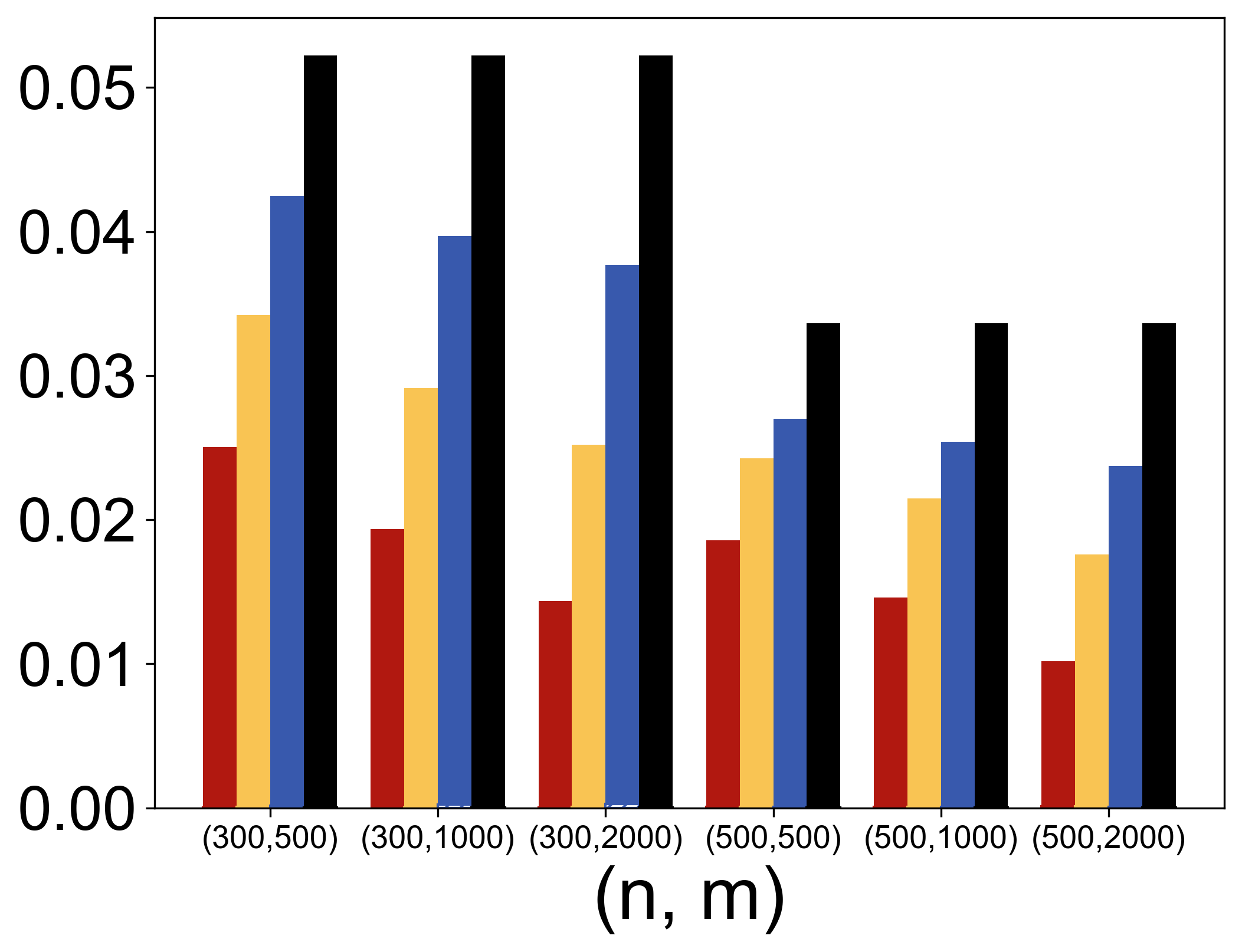}
    }
    \subfloat[\scriptsize{Quantile, $p=4, \tau=0.25$}]{
        \includegraphics[width=0.24\textwidth]{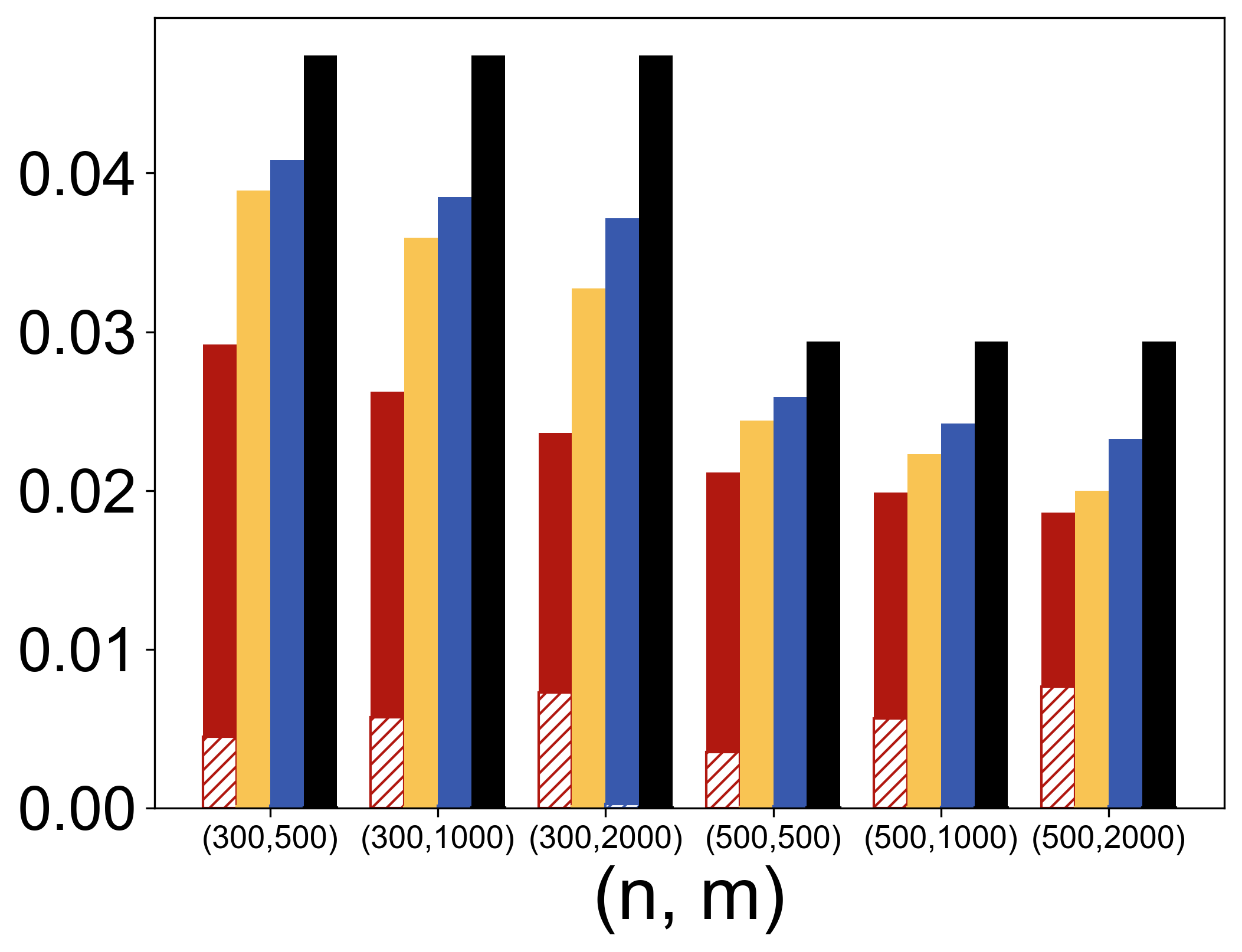}
    }
    \caption{MSE and bias results for $p=4$ for the semi-supervised settings.}
    \label{fig:p4_combined_h}
\end{figure*}

\begin{figure*}[htbp]
    \centering
    \subfloat[\scriptsize{Linear, $p=5$}]{
        \includegraphics[width=0.24\textwidth]{figure/semisupervised/linear_d_5.png}
    }
    \subfloat[\scriptsize{Logistic, $p=5$}]{
        \includegraphics[width=0.24\textwidth]{figure/semisupervised/logistic_d_5.png}
    }
    \subfloat[\scriptsize{Quantile, $p=5, \tau=0.5$}]{
        \includegraphics[width=0.24\textwidth]{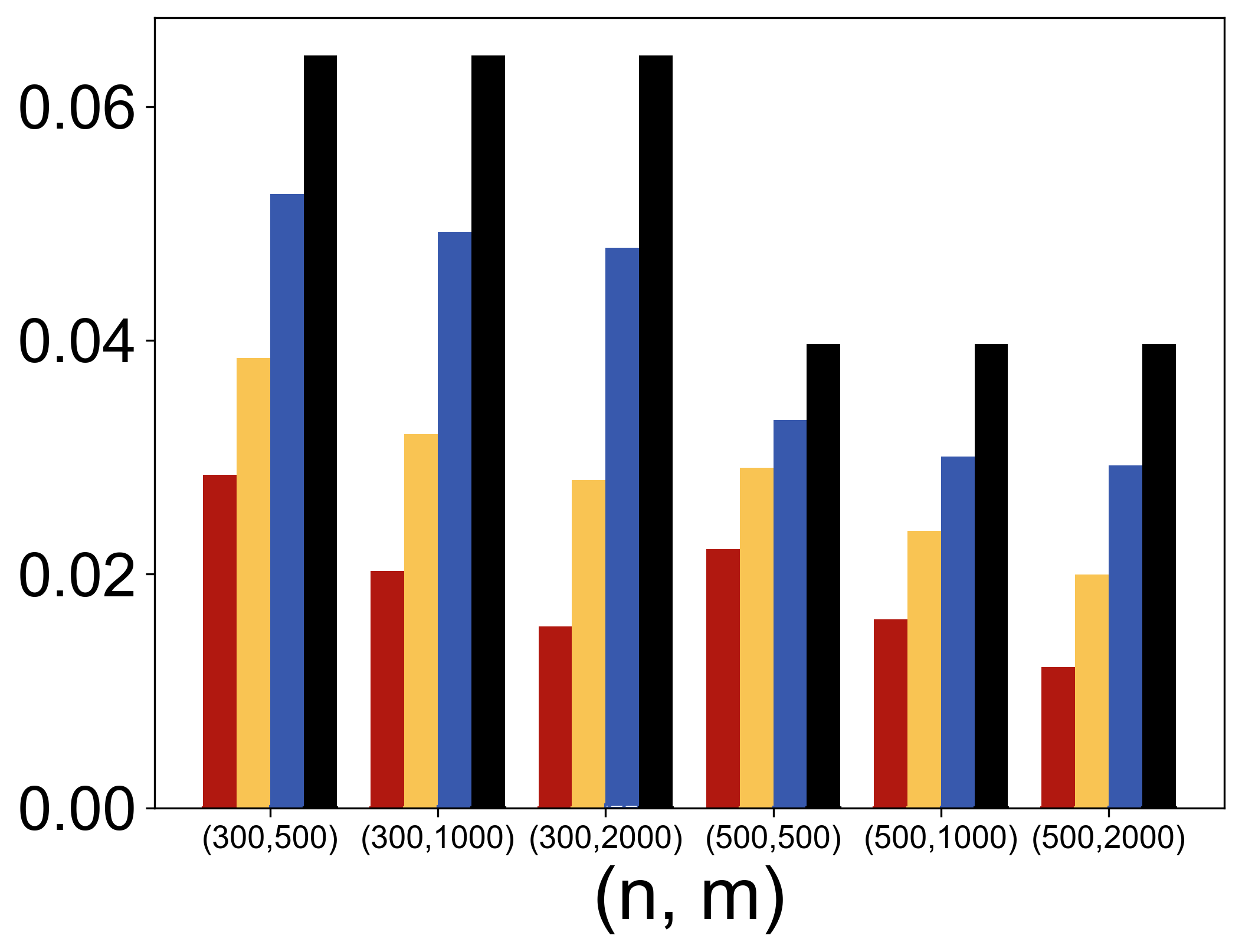}
    }
    \subfloat[\scriptsize{Quantile, $p=5, \tau=0.25$}]{
        \includegraphics[width=0.24\textwidth]{figure/semisupervised/quantile_tau_0.25_d_5.png}
    }
    \caption{MSE and bias results for $p=5$ for the semi-supervised settings.}
    \label{fig:p5_combined_h}
\end{figure*}

\begin{figure*}[htbp]
    \centering
    \subfloat[\scriptsize{Linear, $p=6$}]{
        \includegraphics[width=0.24\textwidth]{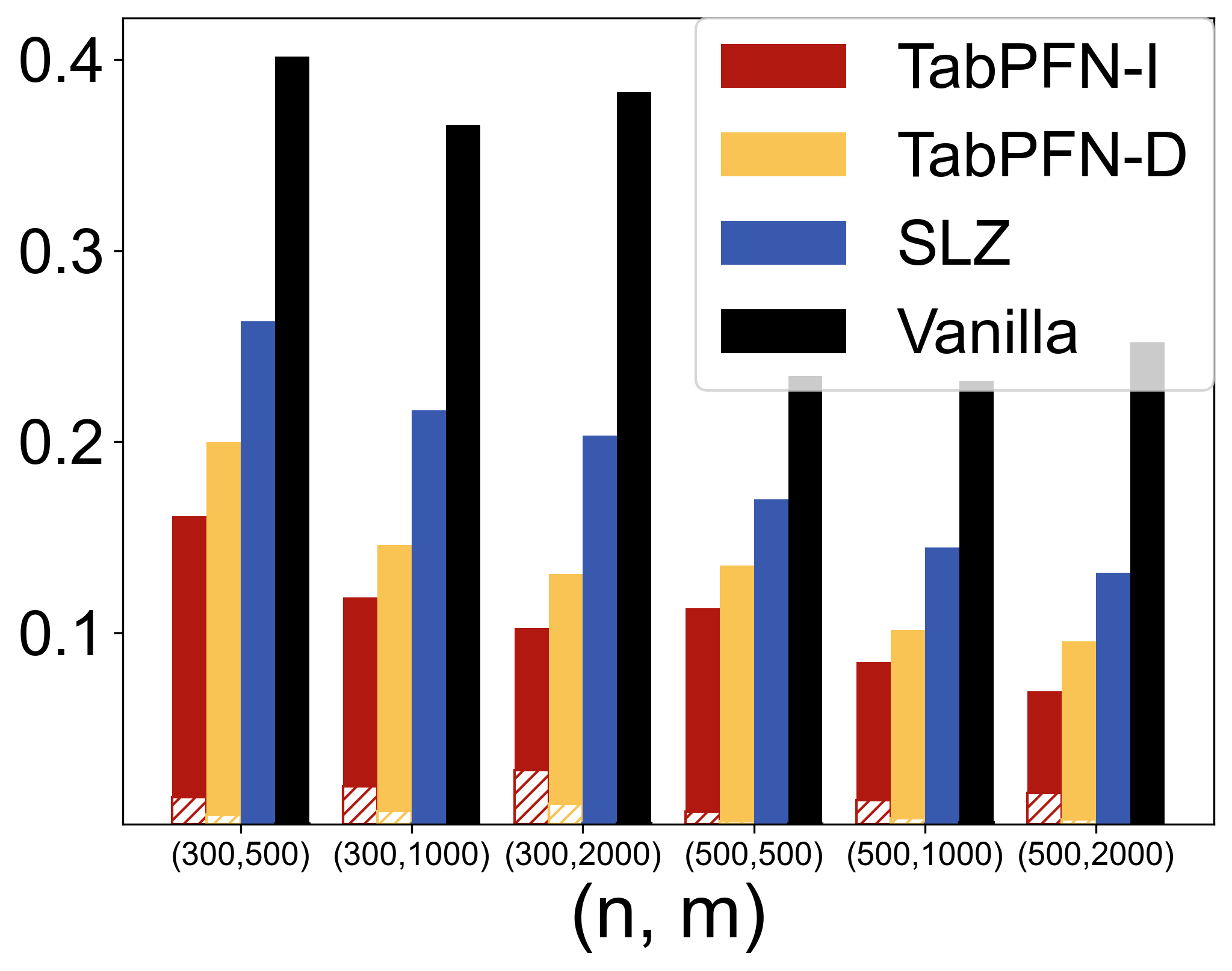}
    }
    \subfloat[\scriptsize{Logistic, $p=6$}]{
        \includegraphics[width=0.24\textwidth]{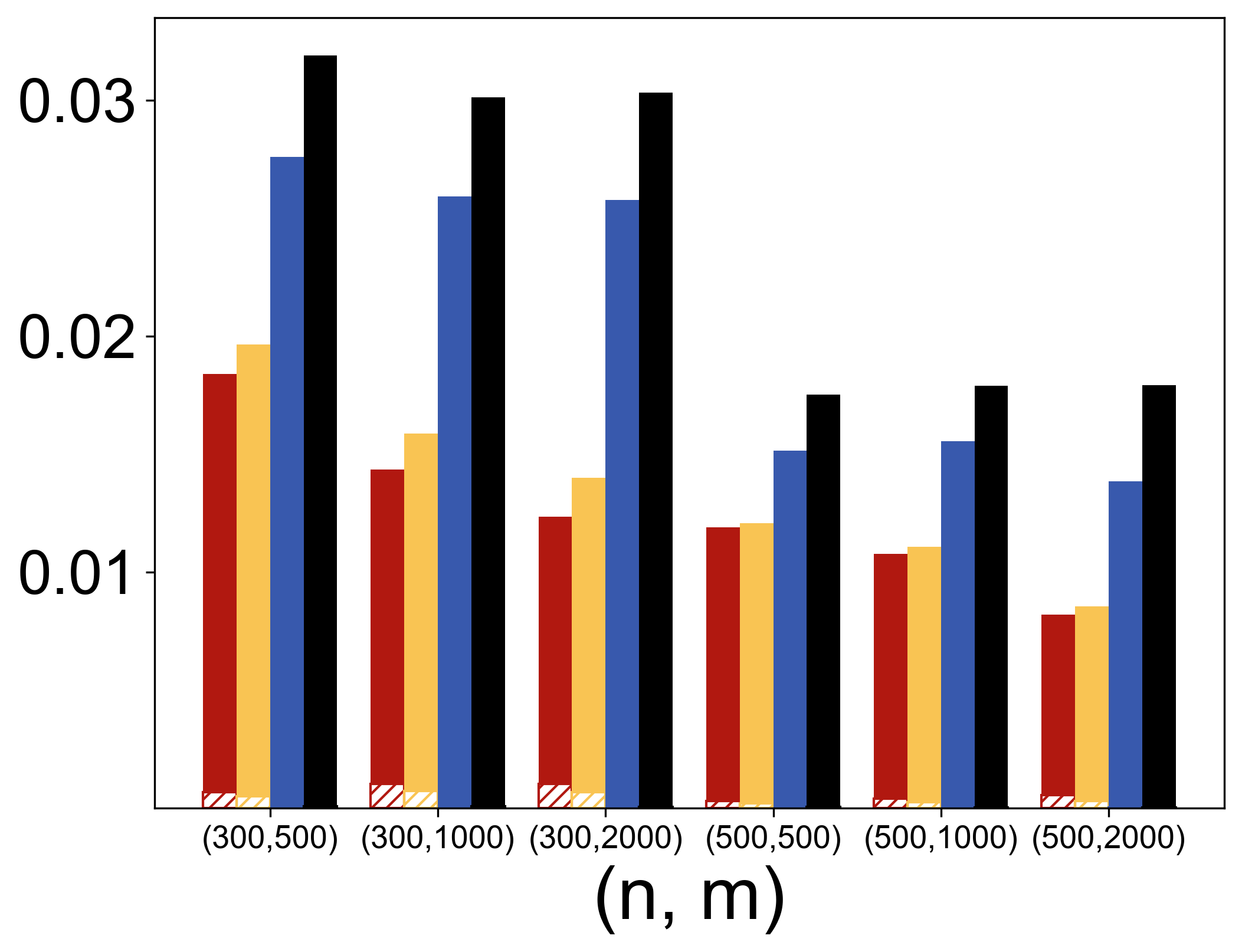}
    }
    \subfloat[\scriptsize{Quantile, $p=6, \tau=0.5$}]{
        \includegraphics[width=0.24\textwidth]{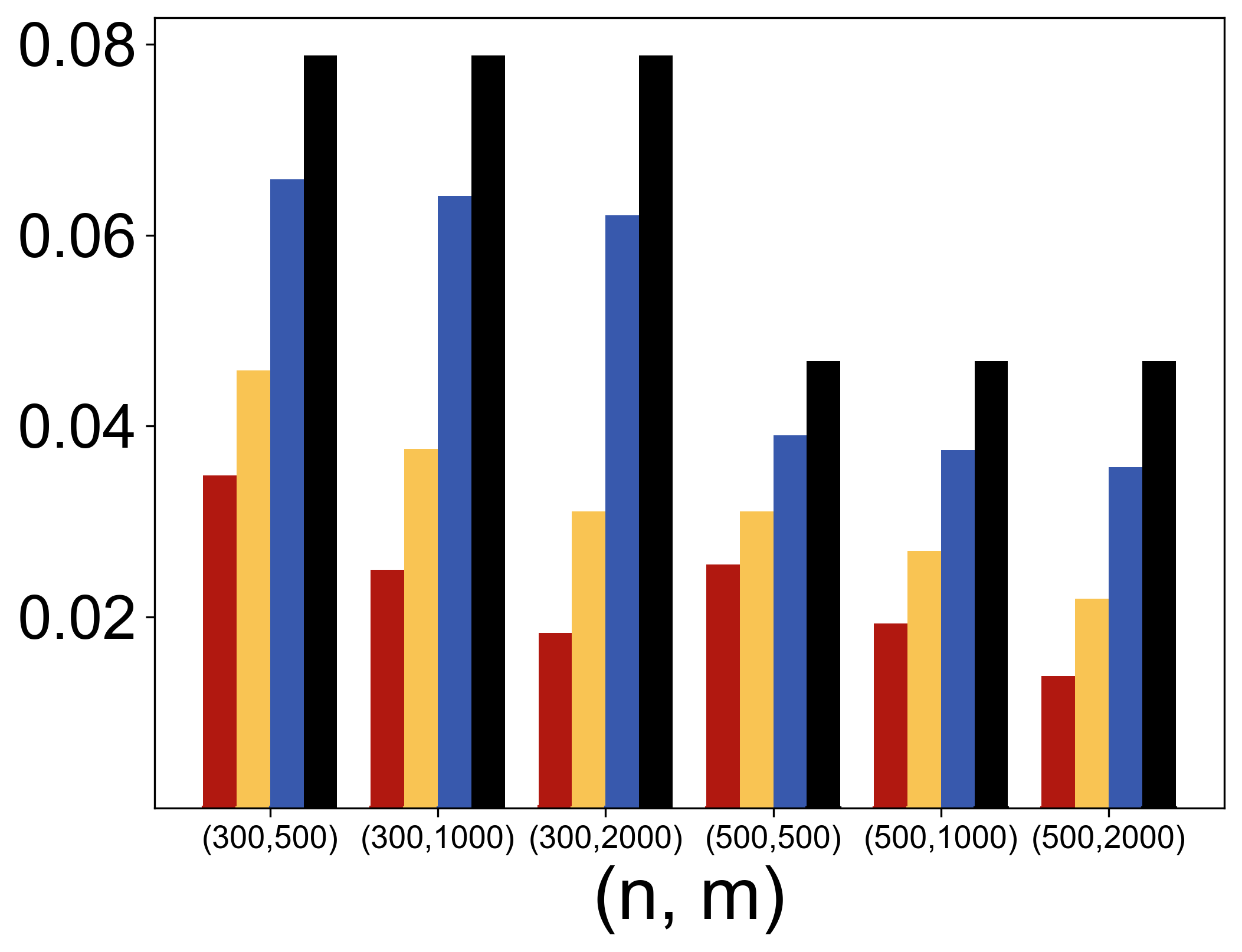}
    }
    \subfloat[\scriptsize{Quantile, $p=6, \tau=0.25$}]{
        \includegraphics[width=0.24\textwidth]{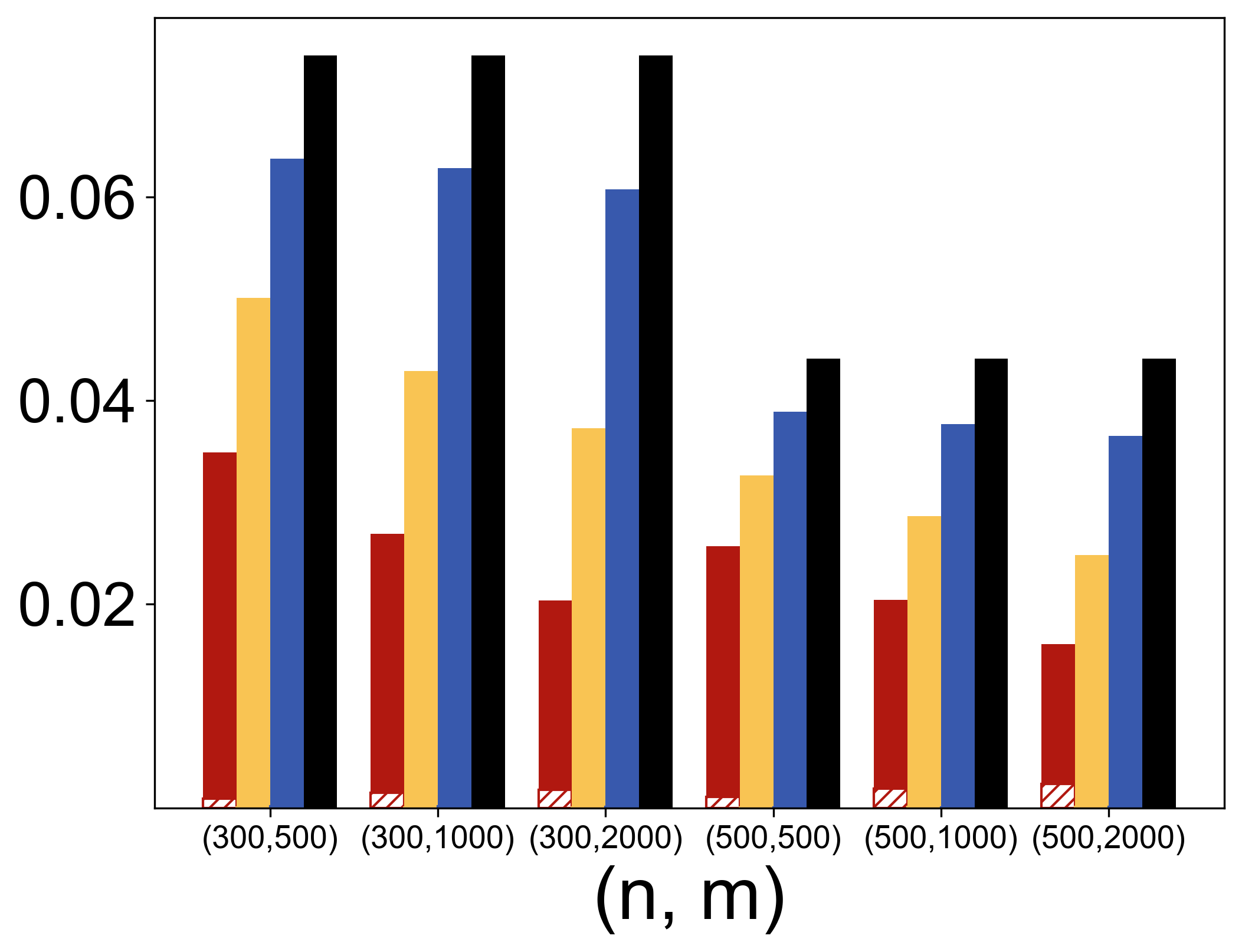}
    }
    \caption{MSE and bias results for $p=6$ for the semi-supervised settings.}
    \label{fig:p6_combined_h}
\end{figure*}

\begin{figure*}[htbp]
    \centering
    \subfloat[\scriptsize{Linear, $p=7$}]{
        \includegraphics[width=0.24\textwidth]{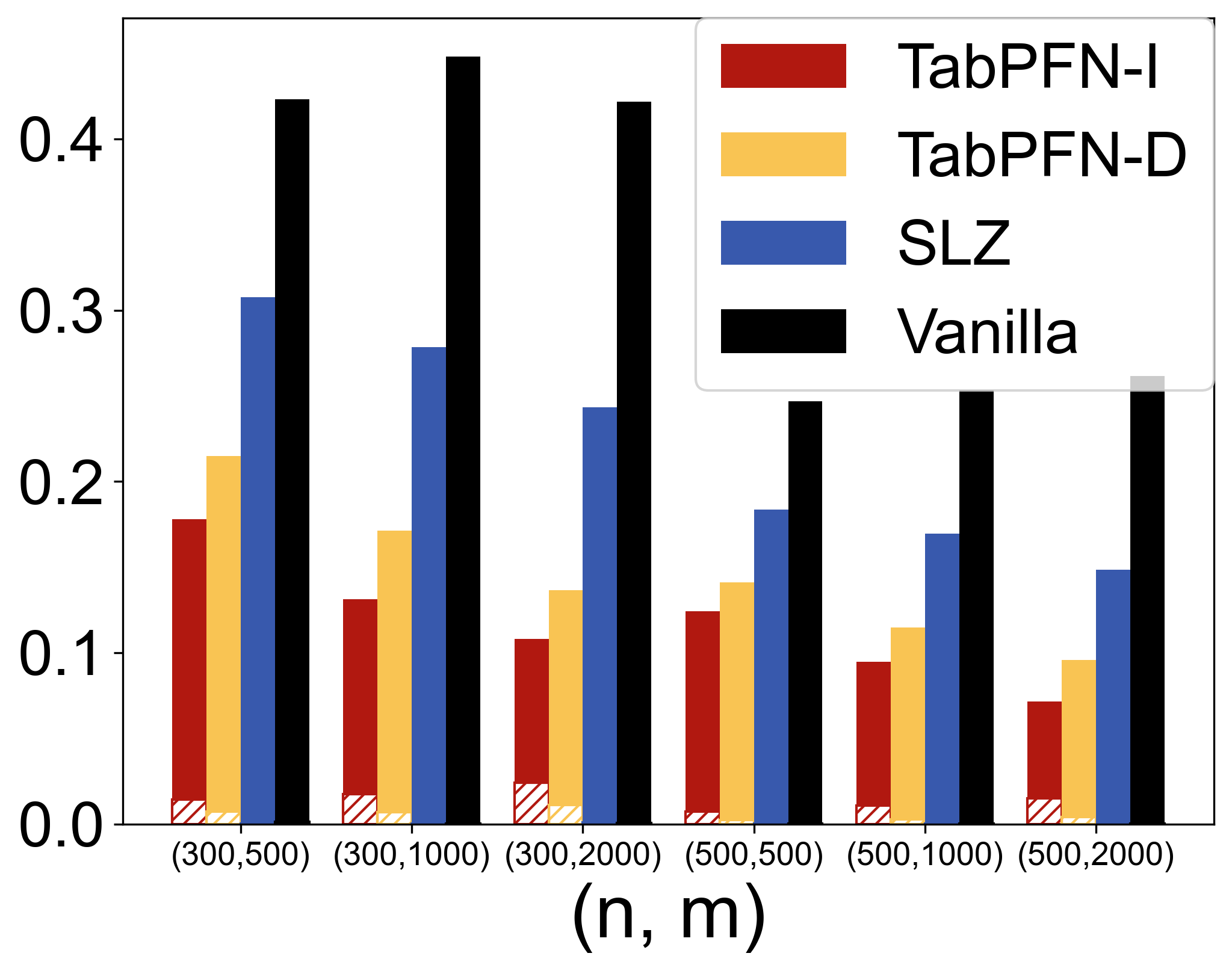}
    }
    \subfloat[\scriptsize{Logistic, $p=7$}]{
        \includegraphics[width=0.24\textwidth]{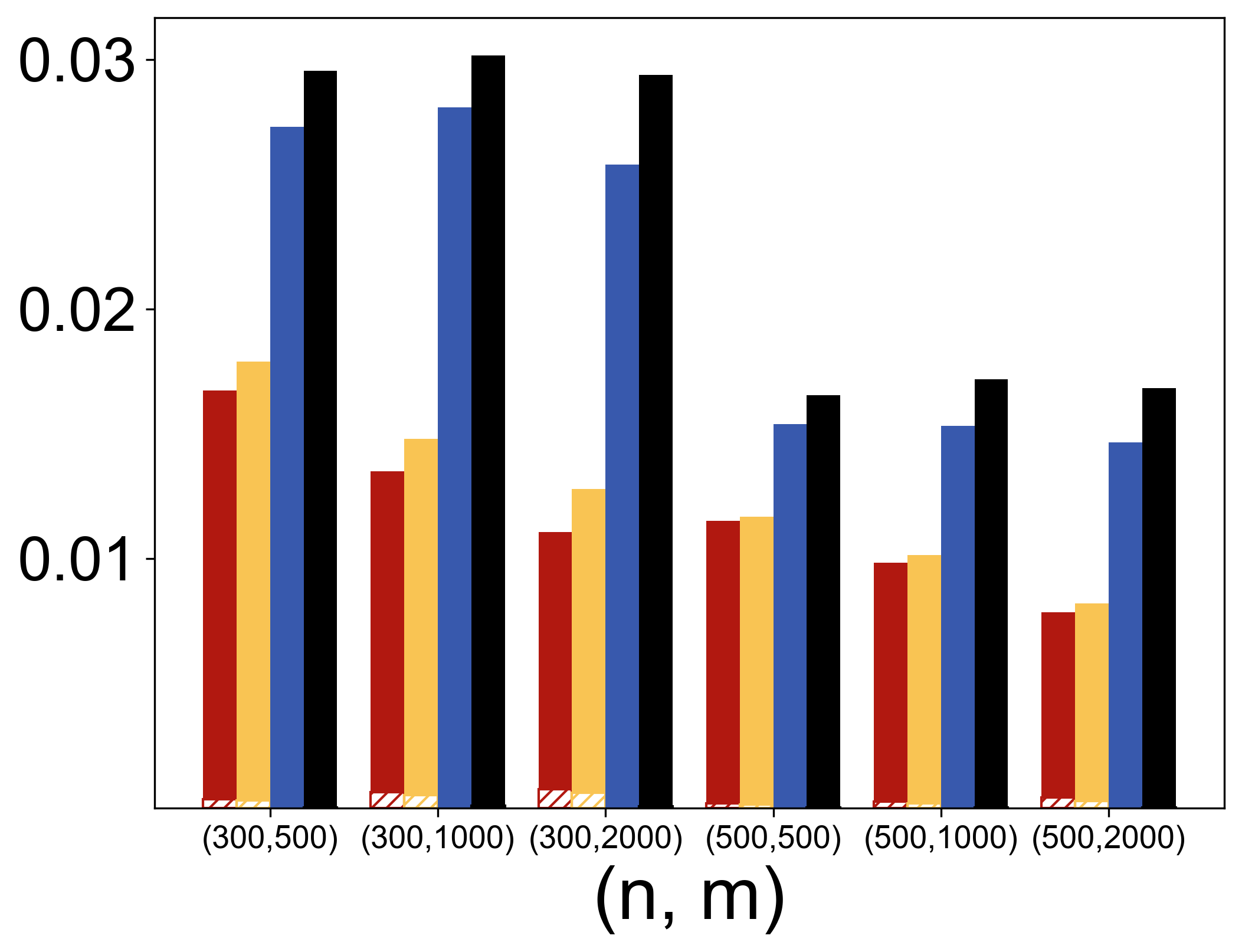}
    }
    \subfloat[\scriptsize{Quantile, $p=7, \tau=0.5$}]{
        \includegraphics[width=0.24\textwidth]{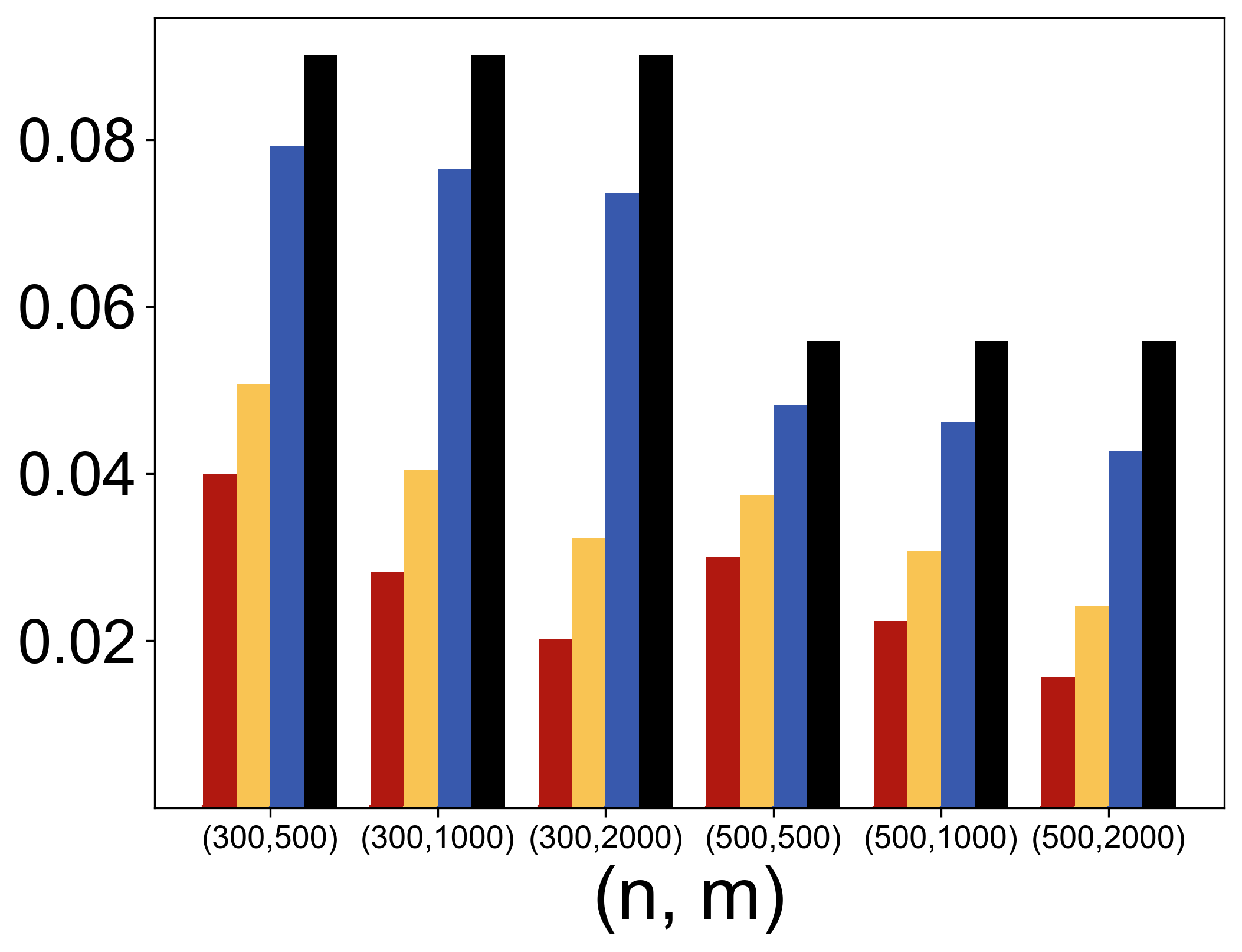}
    }
    \subfloat[\scriptsize{Quantile, $p=7, \tau=0.25$}]{
        \includegraphics[width=0.24\textwidth]{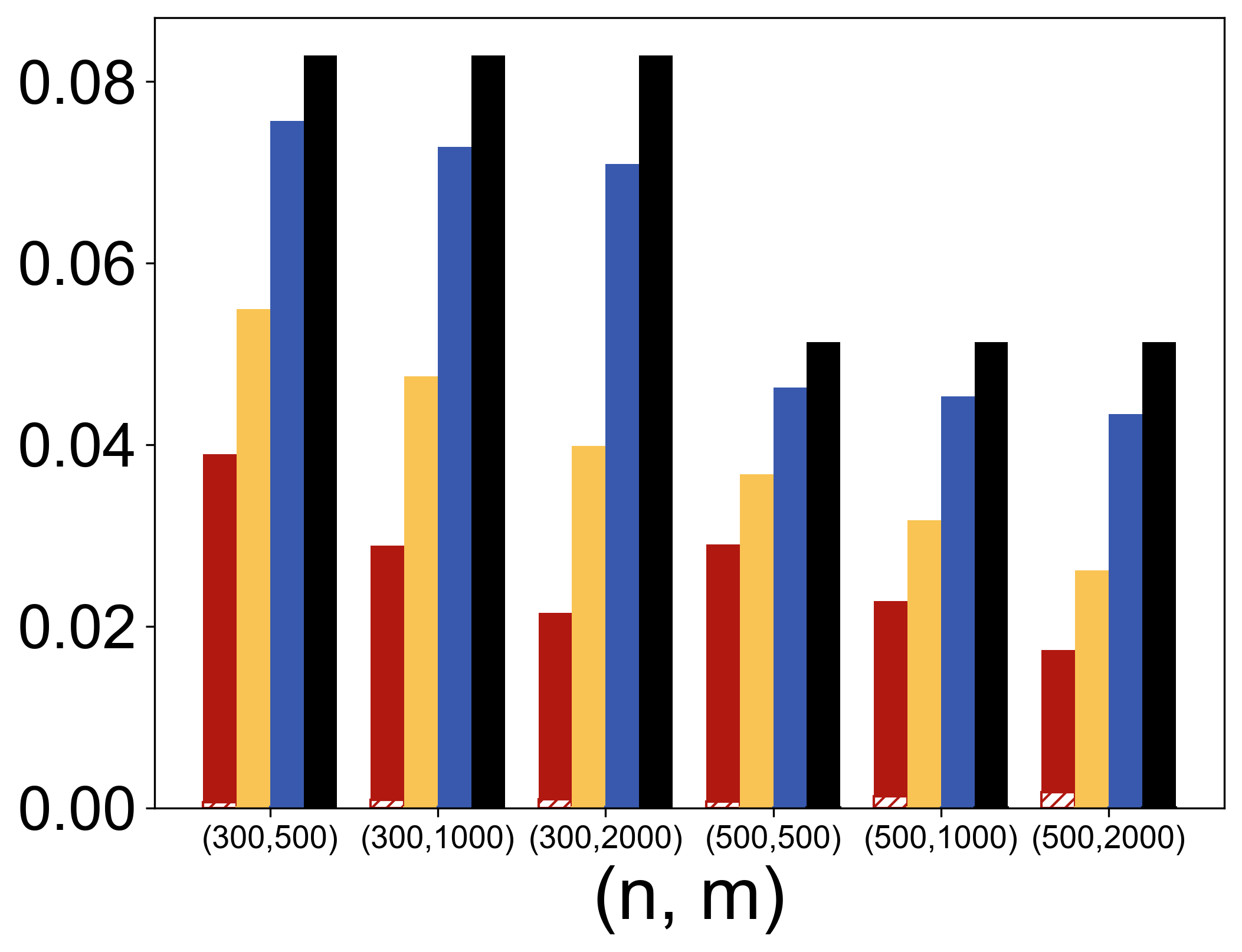}
    }
    \caption{MSE and bias results for $p=7$ for the semi-supervised settings.}
    \label{fig:p7_combined_h}
\end{figure*}

\begin{figure*}[htbp]
    \centering
    \subfloat[\scriptsize{Linear, $p=8$}]{
        \includegraphics[width=0.24\textwidth]{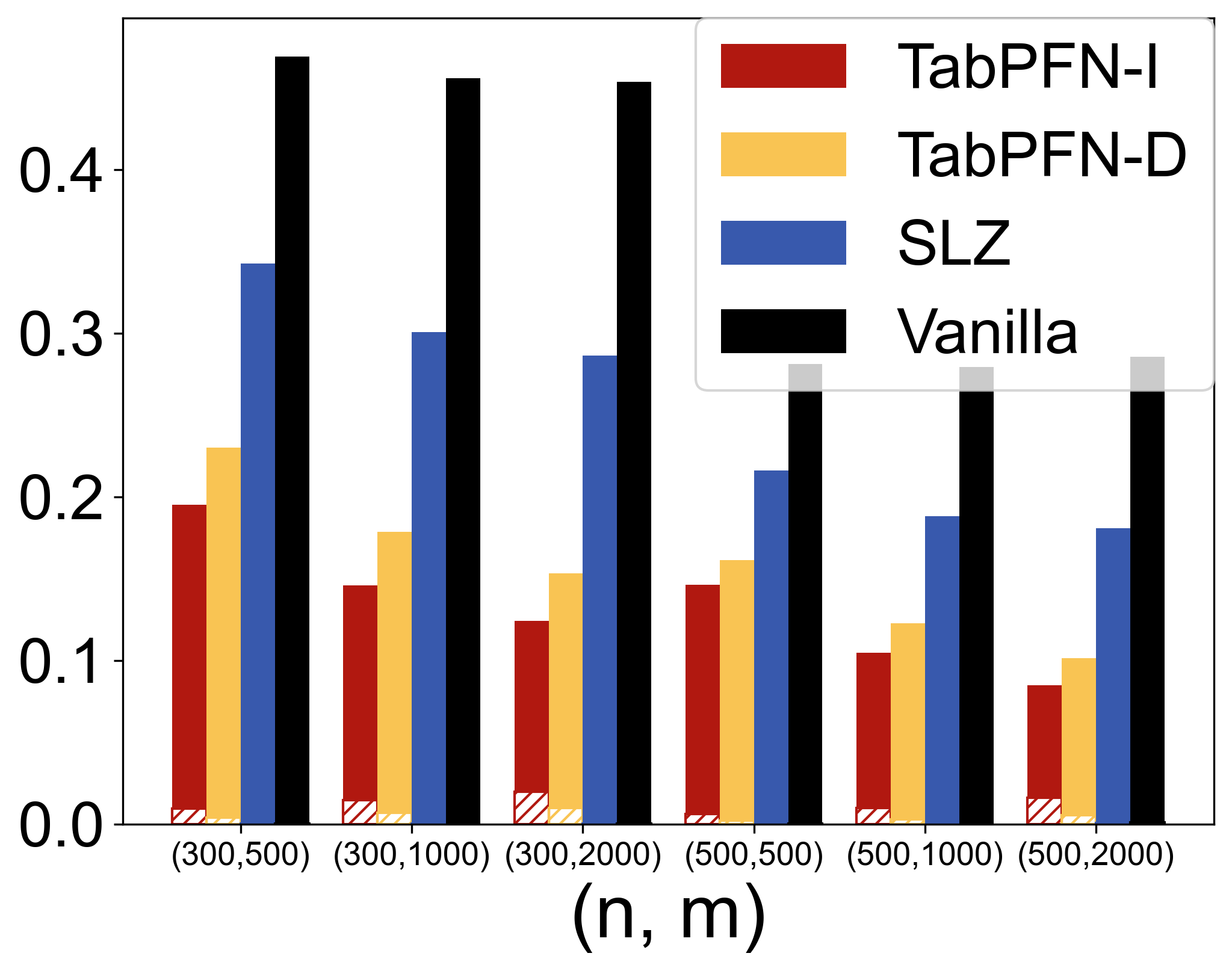}
    }
    \subfloat[\scriptsize{Logistic, $p=8$}]{
        \includegraphics[width=0.24\textwidth]{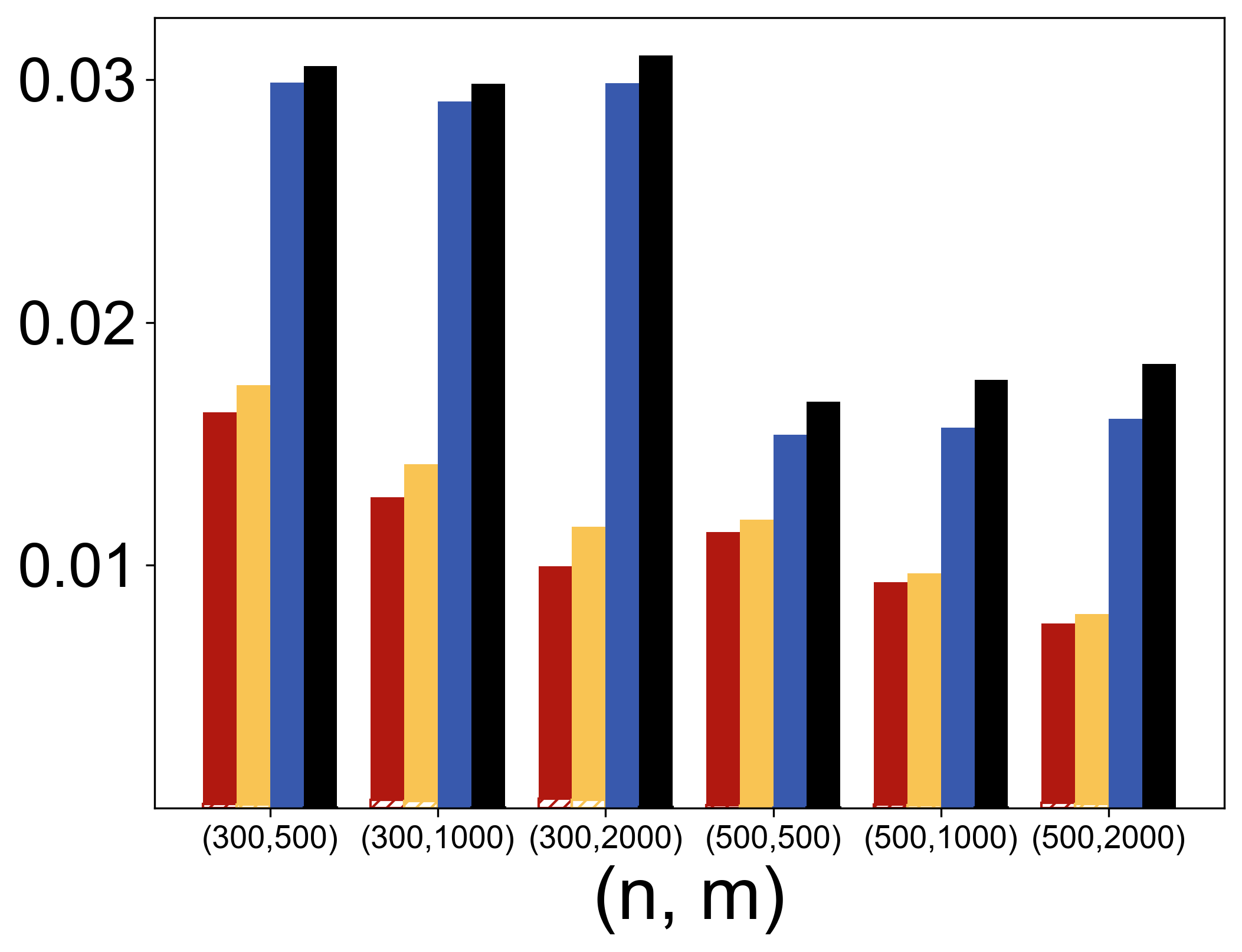}
    }
    \subfloat[\scriptsize{Quantile, $p=8, \tau=0.5$}]{
        \includegraphics[width=0.24\textwidth]{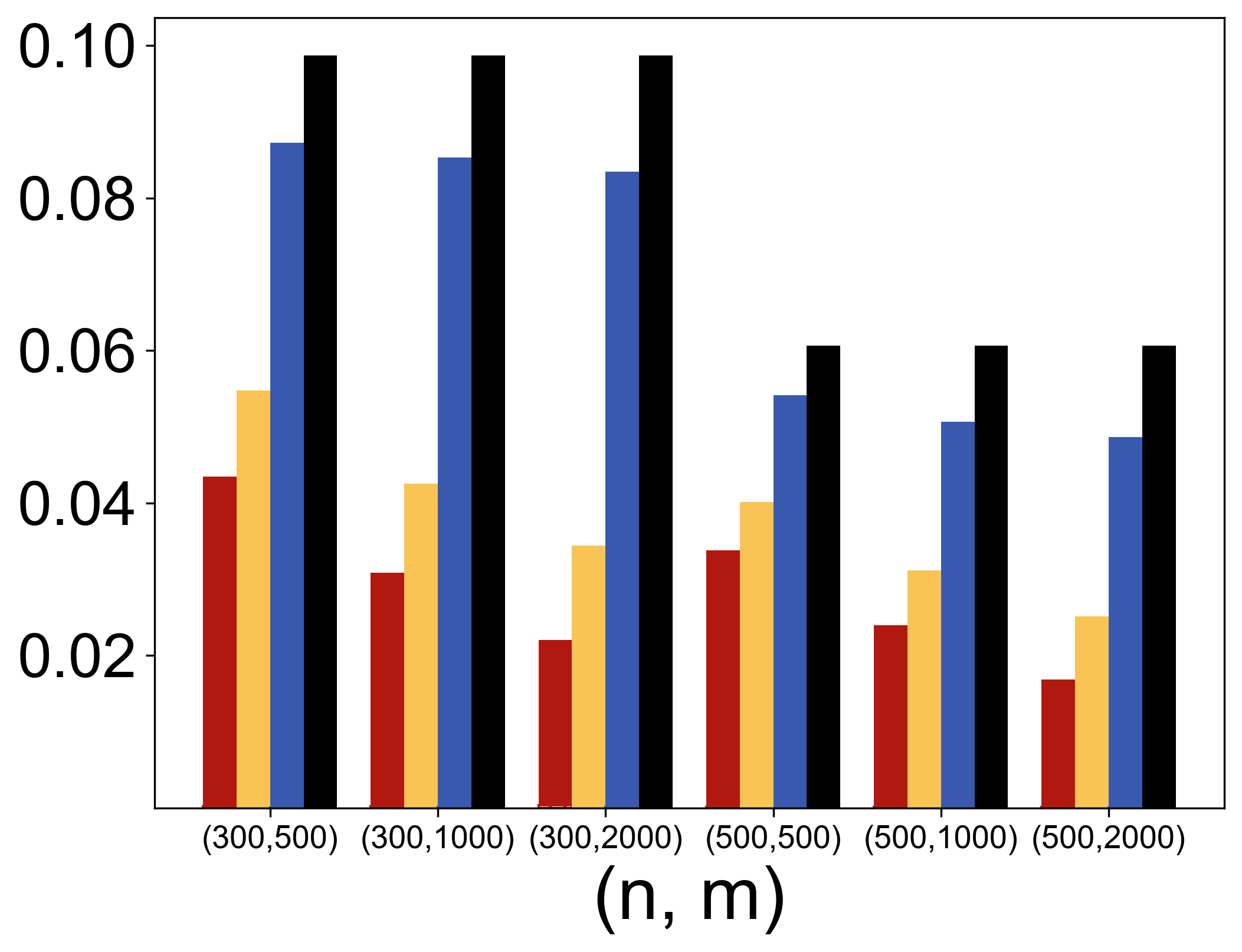}
    }
    \subfloat[\scriptsize{Quantile, $p=8, \tau=0.25$}]{
        \includegraphics[width=0.24\textwidth]{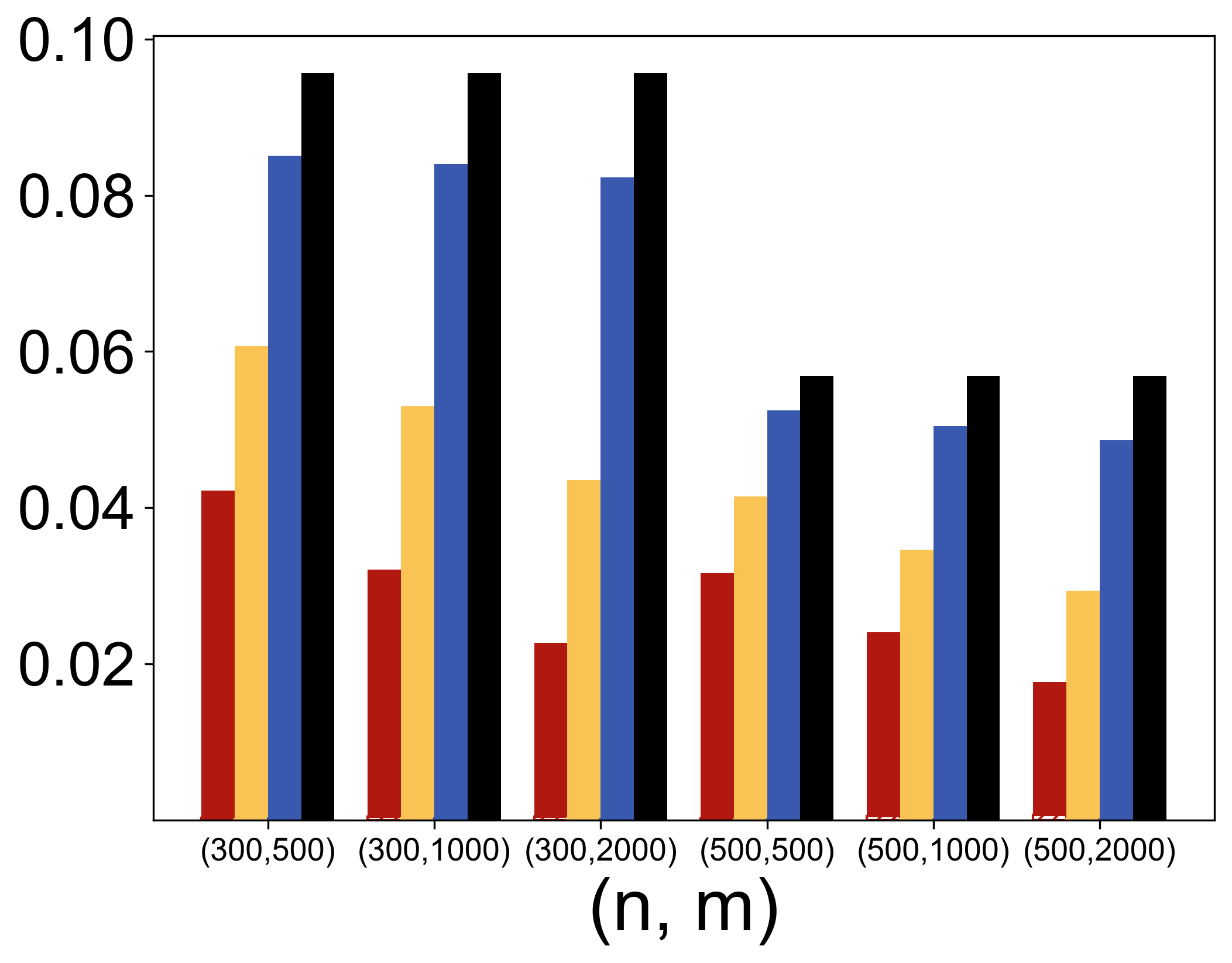}
    }
    \caption{MSE and bias results for $p=8$ for the semi-supervised settings.}
    \label{fig:p8_combined_h}
\end{figure*}

\begin{figure*}[htbp]
    \centering
    \subfloat[\scriptsize{Linear, $p=9$}]{
        \includegraphics[width=0.24\textwidth]{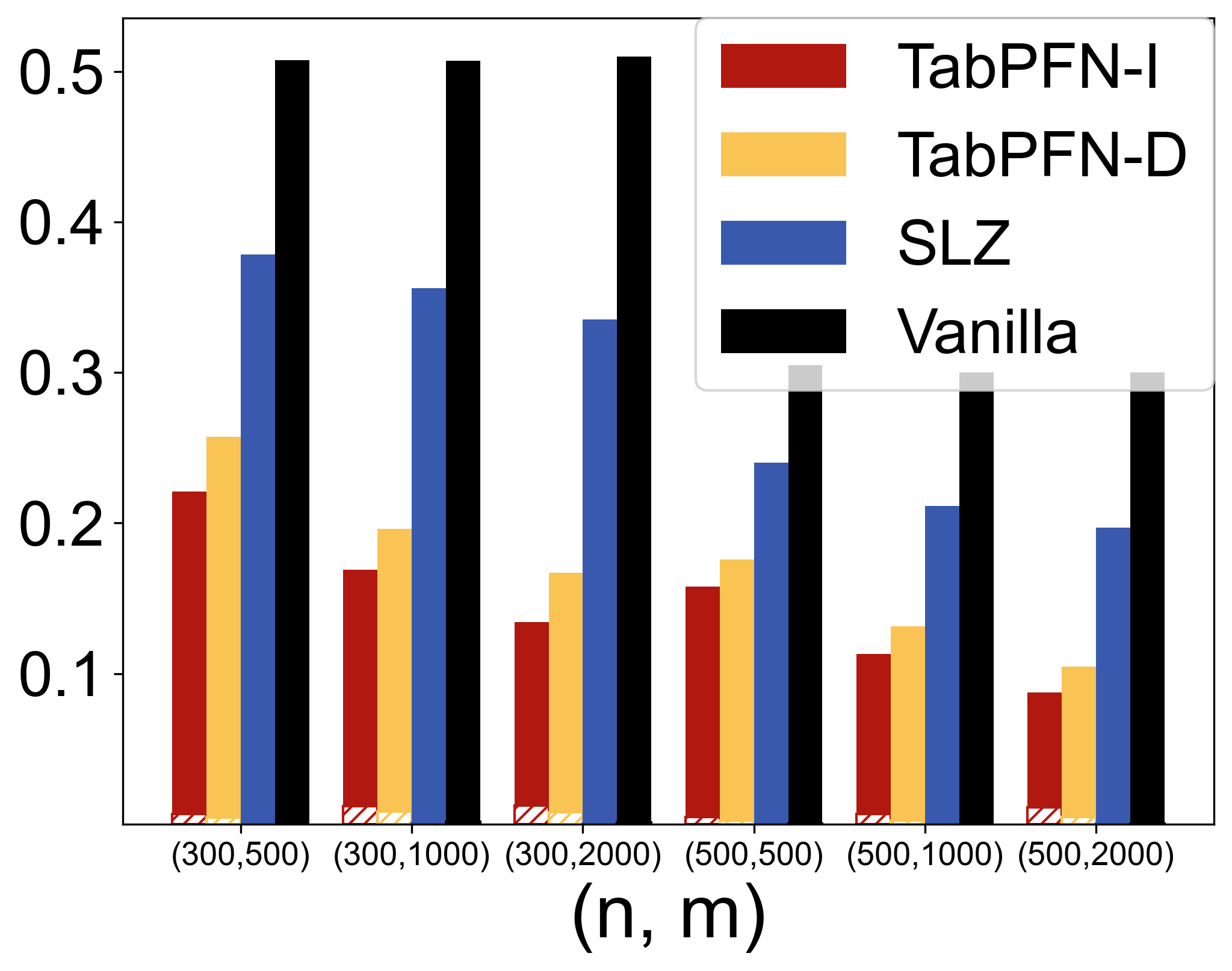}
    }
    \subfloat[\scriptsize{Logistic, $p=9$}]{
        \includegraphics[width=0.24\textwidth]{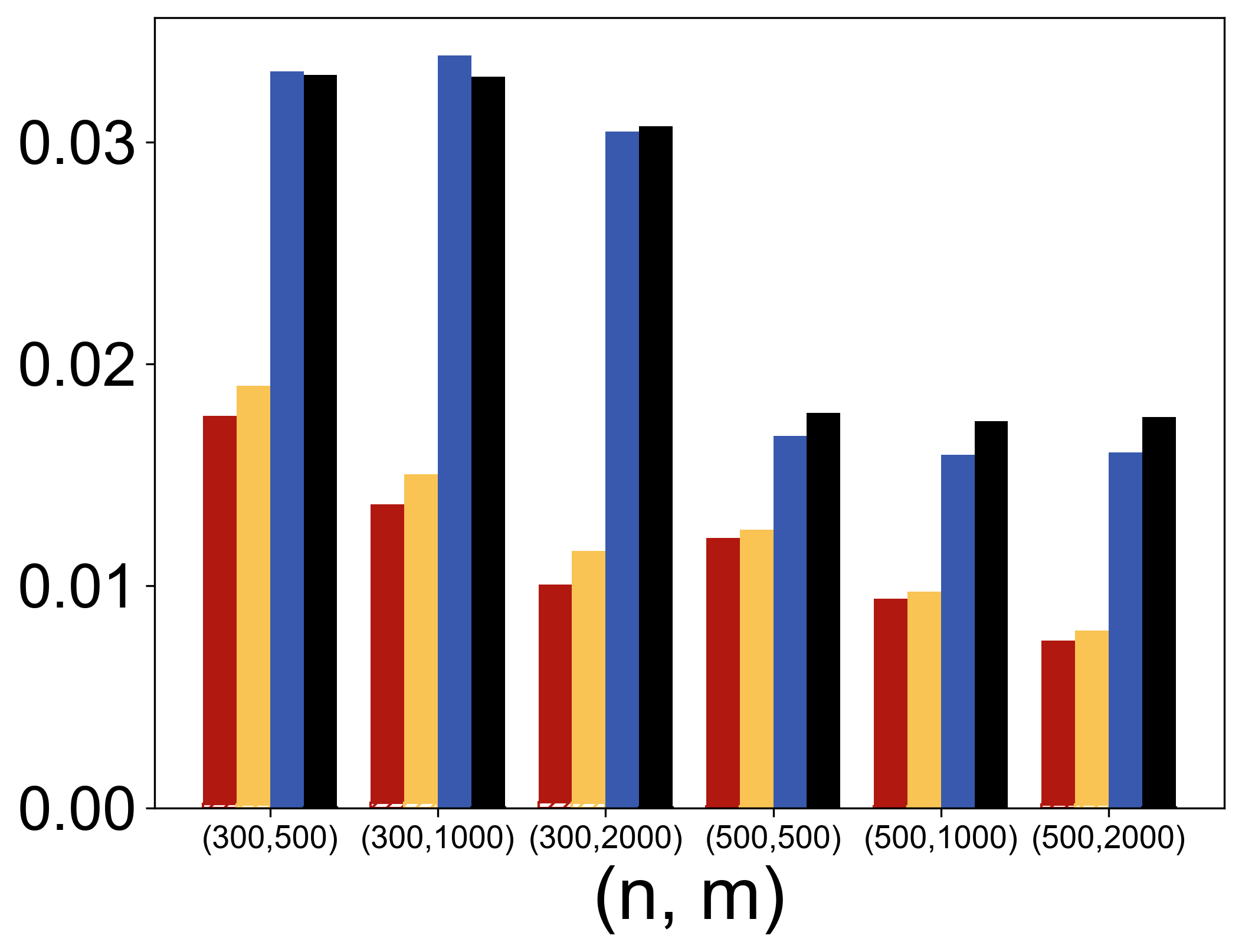}
    }
    \subfloat[\scriptsize{Quantile, $p=9, \tau=0.5$}]{
        \includegraphics[width=0.24\textwidth]{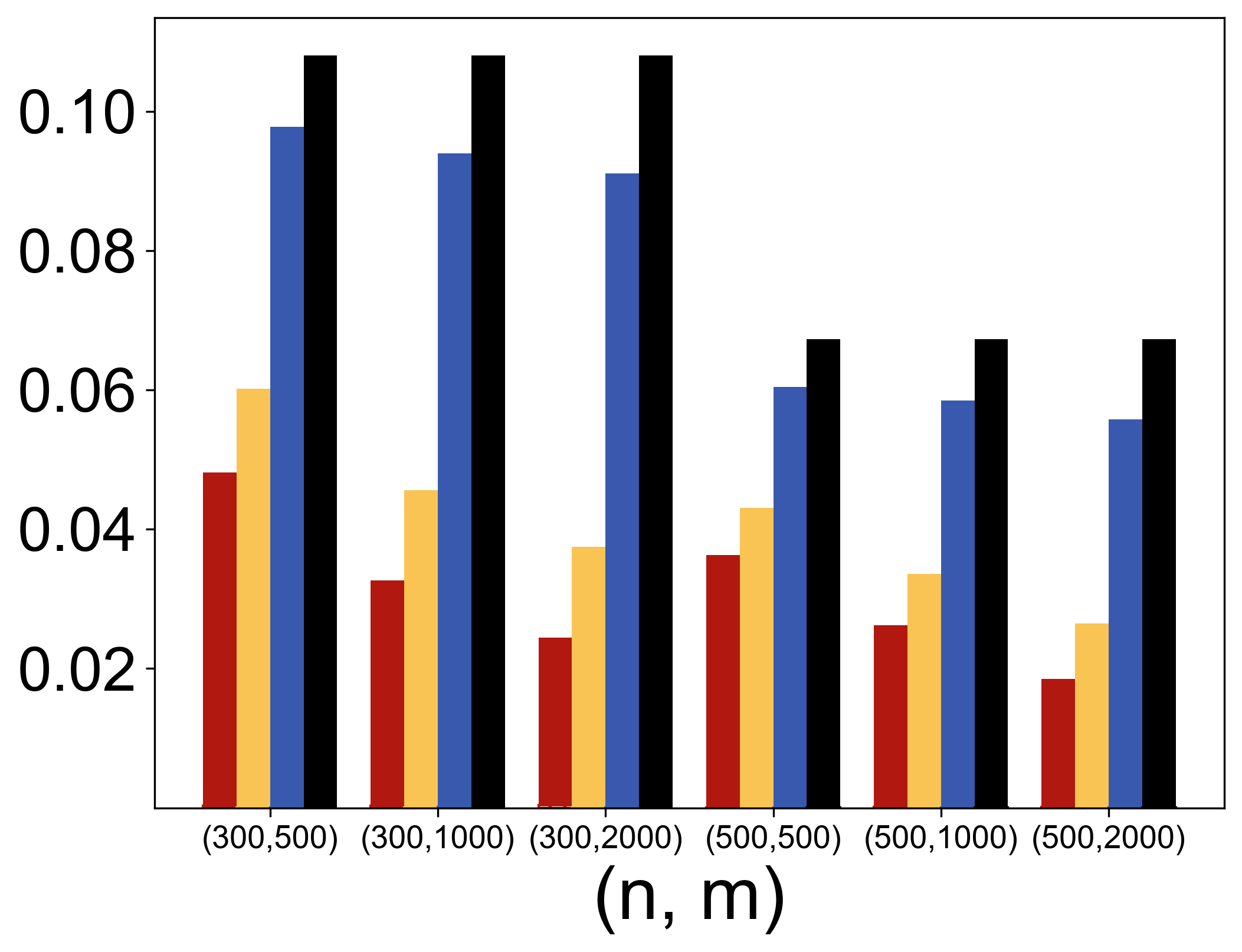}
    }
    \subfloat[\scriptsize{Quantile, $p=9, \tau=0.25$}]{
        \includegraphics[width=0.24\textwidth]{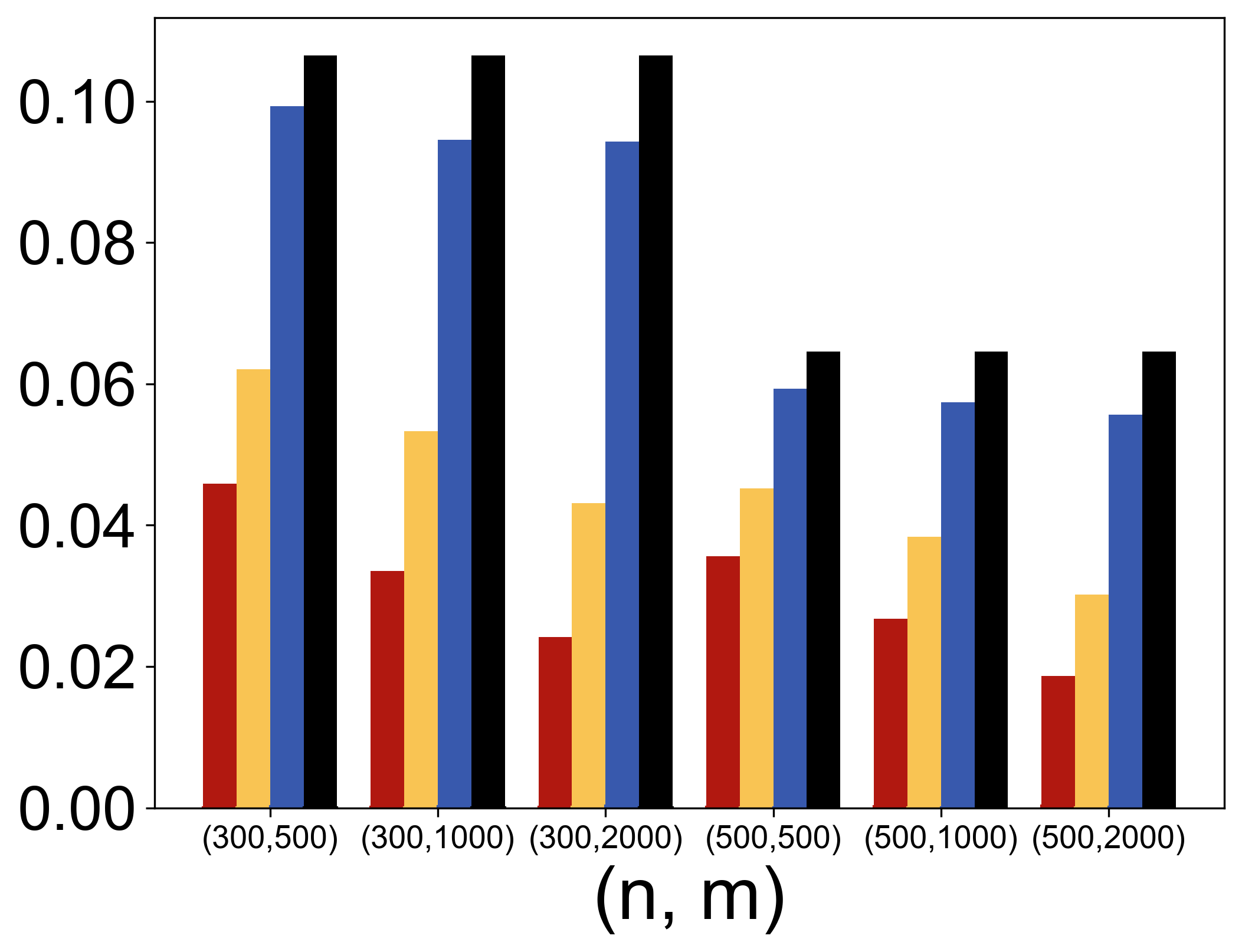}
    }
    \caption{MSE and bias results for $p=9$ for the semi-supervised settings.}
    \label{fig:p9_combined_h}
\end{figure*}

We include all the simulation results in Figures~\ref{fig:p4_combined_h} to \ref{fig:p9_combined_h}.
With the sole exception of the $p=4$ case under logistic regression, TabPFN-I has lower empirical MSE than all competing methods in the settings examined.

\subsection{Real data comparison}
\label{app:ssl_real_data}
The datasets are obtained from OpenML.
We used the dataset search interface at \url{www.openml.org/}, selecting datasets with fewer than 10 features and more than 1{,}000 instances, and verified that all features are numerical.
We then downloaded datasets from the top of this filtered list.
The restriction to fewer than $10$ features is imposed because the SLZ implementation can exhibit numerical instability when the dimensionality is large.
The requirement of a sufficiently large number of instances ensures that the oracle estimator provides a reasonable approximation to the true parameter values.
To maintain computational efficiency, we further restricted datasets to have fewer than 50{,}000 instances.
Even after this filtering, we occasionally encountered matrix-inversion issues in the SLZ implementation; when such cases occurred, we skipped the problematic dataset and selected the next one on the list.
This leads to the following datasets for regression and classification respectively:
\begin{itemize}[leftmargin=*]
\item \textbf{Regression}: kin8nm (ID:189), house\_8L (ID: 218), puma8NH (ID: 225), BNG (stock) (ID: 1200), bank8FM (ID: 572)
\item \textbf{Classification}: BNG(breast-w) (ID:251), kin8nm (ID:807), puma8NH (ID: 816), mozilla4 (ID: 1046), wilt (ID: 40983)
\end{itemize}

\section{Experiment details for study II}
\label{app:causal}
We provide the details for the causal inference experiment and the full experiment result in this section.

\subsection{Simulation settings and additional results}
The specification of the feature distribution, propensity function, base function, and treatment effect function under all setups are given in Table~\ref{tab:setups}.
\begin{table*}[tp]
\centering
\caption{Experimental setups with distributional and functional specifications}
\label{tab:setups}
\resizebox{\textwidth}{!}{
\begin{tabular}{l l l l}
\toprule
\textbf{Setup} & \textbf{Propensity $e(X)$} & \textbf{Base $b(X)$} & \textbf{Effect $\tau(X)$}\\
\midrule
$\mathbf{A}$ &
$\min\{0.8,~\max\{\sin(\pi X_1 X_2), 0.2\}\}$ &
$\sin(\pi X_1 X_2) + 2(X_3 - .5)^2 + X_4 + .5 X_5$ &
$.2 + (X_1 + X_2)/2$ \\
\addlinespace

$\mathbf{B}$ &
$.5$ &
$\max\{0, X_1 + X_2, X_3\} + \max\{0, X_4+X_5\}$ &
$X_1 + \log[1 + \exp(X_2)]$ \\
$\mathbf{C}$ &
$[1 + \exp(X_2 + X_3)]^{-1}$ &
$2 \log[1 + \exp(X_1 + X_2 + X_3)]$ &
$1$\\
$\mathbf{D}$ &
$[1 + \exp(-X_1) + \exp(-X_2)]^{-1}$ &
$.5 \max\{0, X_1+X_2+X_3\} + .5 \max\{0, X_4 + X_5\}$ &
$\max\{0, X_1+X_2+X_3\} - \max\{0, X_4+X_5\}$ \\
$\mathbf{E}$ &
$[1 + \exp(3X_2 + 3 X_3)]^{-1}$ &
$5 \max\{0, X_1 + X_2\}$ &
$2 \max\{(X_1>.1), (X_2 > .1)\} - 1$ \\
$\mathbf{F}$ &
$[1 + \exp(3X_2 + 3 X_3)]^{-1}$ &
$5 \max\{0, X_1 + X_2\}$ &
$X_1 + \log[1 + \exp(X_2)]$ \\
\bottomrule
\end{tabular}}
\end{table*}
These functions correspond to the following settings:
\begin{itemize}
\item \textbf{Setup A}: complicated $e(x)$ and $b(x)$, but a relatively simple $\tau(x)$.
\item \textbf{Setup B}: randomized trial with a complex $b(x)$ and $\tau(x)$.
\item \textbf{Setup C}: complex $e(x)$ and $b(x)$, but a constant $\tau(x)$.
\item \textbf{Setup D}: response under treatment is independent of the response under control, and hence there is no statistical benefit of jointly learning the two responses.
\item \textbf{Setup E}: large $b(x)$ resulting in substantial confounding, discontinuous $\tau(x)$.
\item \textbf{Setup F}: similar to setup $E$, but with a smooth $\tau(x)$.
\end{itemize}

Our experiments are implemented using the \emph{econml}~\citep{econml} package in Python, and we use the FLAML package~\citep{wang2021flaml} to implement all AutoML-based learners.
Each FLAML search has a 60-second time budget.
The searches used by the S- and T-Learners and for nuisance-model selection use five-fold cross-validation.
The final-stage FLAML regressors for the R- and DR-Learners and the FLAML regressors supplied to the X-Learner use ten-fold cross-validation.
The \emph{econml} R- and DR-Learners use ten-fold cross-fitting, while the true-nuisance R-Learner uses a single fold because its nuisance functions are supplied rather than estimated.
For each of the 54 configurations, we generate 100 datasets and report the median test MSE for CATE over the 100 repetitions in Tables~\ref{tab:CATE_full} and~\ref{tab:CATE_full_D-F}.
\begin{table*}[tp]
\centering
\caption{Median test MSE of the CATE estimators over $100$ repetitions under Setups A--C.
  The \textbf{smallest} test MSE is highlighted in boldface and the \underline{second smallest} test MSE is underlined.}
\label{tab:CATE_full}
\resizebox{0.9\textwidth}{!}{
\begin{tabular}{ccc|cc|cc|cc|cc|c|c}
\toprule
\multirow{2}{*}{Setup} & \multirow{2}{*}{N} & \multirow{2}{*}{$\sigma^2$} & \multicolumn{2}{c|}{DR} & \multicolumn{2}{c|}{S} & \multicolumn{2}{c|}{T} & \multicolumn{2}{c|}{X} & R & Oracle \\
\cmidrule{4-13}
 & & & TabPFN & AutoML & TabPFN & AutoML & TabPFN & AutoML & TabPFN & AutoML & AutoML & AutoML\\
\midrule
\multirow{9}{*}{A} & 500 & 0.5 & 0.0403 & 0.0906 & \underline{0.0392} & 0.0428 & 0.0695 & 0.2057 & \textbf{0.0342} & 0.0773 & 0.0770 & 0.0659 \\
 &  & 1 & \textbf{0.0486} & 0.1543 & 0.0528 & 0.0650 & 0.1103 & 0.3178 & \underline{0.0502} & 0.1214 & 0.1116 & 0.1069 \\
 &  & 2 & \underline{0.0611} & 0.2706 & \textbf{0.0610} & 0.0825 & 0.1702 & 0.4556 & 0.0692 & 0.1693 & 0.2060 & 0.1964 \\
 & 1000 & 0.5 & \underline{0.0221} & 0.0494 & 0.0264 & 0.0317 & 0.0551 & 0.1477 & \textbf{0.0190} & 0.0382 & 0.0406 & 0.0381 \\
 &  & 1 & \underline{0.0366} & 0.0699 & 0.0367 & 0.0421 & 0.0725 & 0.2046 & \textbf{0.0317} & 0.0724 & 0.0621 & 0.0604 \\
 &  & 2 & 0.0495 & 0.1238 & \textbf{0.0480} & 0.0562 & 0.1081 & 0.2859 & \underline{0.0494} & 0.1151 & 0.1087 & 0.0889 \\
 & 2000 & 0.5 & \underline{0.0098} & 0.0241 & 0.0146 & 0.0219 & 0.0357 & 0.0941 & \textbf{0.0090} & 0.0244 & 0.0201 & 0.0218 \\
 &  & 1 & \underline{0.0208} & 0.0385 & 0.0245 & 0.0302 & 0.0517 & 0.1375 & \textbf{0.0192} & 0.0399 & 0.0383 & 0.0355 \\
 &  & 2 & \underline{0.0331} & 0.0676 & 0.0378 & 0.0403 & 0.0698 & 0.1913 & \textbf{0.0321} & 0.0645 & 0.0626 & 0.0582 \\
\midrule
\multirow{9}{*}{B} & 500 & 0.5 & \underline{0.0261} & 0.0763 & 0.0355 & 0.0607 & 0.0350 & 0.0929 & \textbf{0.0233} & 0.0492 & 0.0770 & 0.0533 \\
 &  & 1 & \underline{0.0513} & 0.1278 & 0.0599 & 0.0952 & 0.0556 & 0.1397 & \textbf{0.0456} & 0.0856 & 0.1104 & 0.1008 \\
 &  & 2 & \underline{0.0847} & 0.2168 & 0.0896 & 0.1384 & 0.0848 & 0.2119 & \textbf{0.0788} & 0.1220 & 0.1833 & 0.1656 \\
 & 1000 & 0.5 & \underline{0.0130} & 0.0406 & 0.0214 & 0.0421 & 0.0260 & 0.0621 & \textbf{0.0112} & 0.0295 & 0.0366 & 0.0332 \\
 &  & 1 & \underline{0.0256} & 0.0707 & 0.0357 & 0.0614 & 0.0350 & 0.1021 & \textbf{0.0230} & 0.0475 & 0.0622 & 0.0588 \\
 &  & 2 & 0.0537 & 0.1036 & 0.0637 & 0.0858 & \underline{0.0528} & 0.1474 & \textbf{0.0405} & 0.0814 & 0.1008 & 0.0910 \\
 & 2000 & 0.5 & \underline{0.0070} & 0.0216 & 0.0130 & 0.0260 & 0.0174 & 0.0430 & \textbf{0.0065} & 0.0173 & 0.0212 & 0.0195 \\
 &  & 1 & \underline{0.0146} & 0.0328 & 0.0199 & 0.0369 & 0.0241 & 0.0627 & \textbf{0.0117} & 0.0286 & 0.0296 & 0.0310 \\
 &  & 2 & \underline{0.0251} & 0.0553 & 0.0335 & 0.0558 & 0.0365 & 0.0876 & \textbf{0.0213} & 0.0429 & 0.0595 & 0.0541 \\
\midrule
\multirow{9}{*}{C} & 500 & 0.5 & \textbf{0.0033} & 0.0316 & 0.0061 & 0.0640 & 0.0273 & 0.0898 & \underline{0.0053} & 0.0320 & 0.0287 & 0.0239 \\
 &  & 1 & \textbf{0.0066} & 0.0725 & \underline{0.0099} & 0.0832 & 0.0579 & 0.1640 & 0.0125 & 0.0531 & 0.0610 & 0.0565 \\
 &  & 2 & \textbf{0.0135} & 0.1374 & \underline{0.0179} & 0.0998 & 0.1052 & 0.2750 & 0.0256 & 0.0945 & 0.1139 & 0.1037 \\
 & 1000 & 0.5 & \textbf{0.0024} & 0.0159 & 0.0048 & 0.0234 & 0.0173 & 0.0548 & \underline{0.0033} & 0.0146 & 0.0151 & 0.0143 \\
 &  & 1 & \textbf{0.0048} & 0.0337 & 0.0079 & 0.0647 & 0.0330 & 0.0896 & \underline{0.0071} & 0.0343 & 0.0286 & 0.0291 \\
 &  & 2 & \textbf{0.0092} & 0.0652 & \underline{0.0135} & 0.0809 & 0.0689 & 0.1588 & 0.0168 & 0.0675 & 0.0606 & 0.0585 \\
 & 2000 & 0.5 & \textbf{0.0007} & 0.0084 & 0.0028 & 0.0104 & 0.0094 & 0.0382 & \underline{0.0013} & 0.0089 & 0.0087 & 0.0073 \\
 &  & 1 & \textbf{0.0014} & 0.0149 & 0.0043 & 0.0277 & 0.0175 & 0.0562 & \underline{0.0025} & 0.0146 & 0.0134 & 0.0142 \\
 &  & 2 & \textbf{0.0028} & 0.0275 & 0.0071 & 0.0708 & 0.0353 & 0.0791 & \underline{0.0052} & 0.0275 & 0.0271 & 0.0298 \\
\bottomrule
\end{tabular}}
\end{table*}

\begin{table*}[tp]
\centering
\caption{Median test MSE of the CATE estimators over $100$ repetitions under Setups D--F.
  The \textbf{smallest} test MSE is highlighted in boldface and the \underline{second smallest} test MSE is underlined.}
\label{tab:CATE_full_D-F}
\resizebox{0.9\textwidth}{!}{
\begin{tabular}{ccc|cc|cc|cc|cc|c|c}
\toprule
\multirow{2}{*}{Setup} & \multirow{2}{*}{N} & \multirow{2}{*}{$\sigma^2$} & \multicolumn{2}{c|}{DR} & \multicolumn{2}{c|}{S} & \multicolumn{2}{c|}{T} & \multicolumn{2}{c|}{X} & R & Oracle \\
\cmidrule{4-13}
 & & & TabPFN & AutoML & TabPFN & AutoML & TabPFN & AutoML & TabPFN & AutoML & AutoML & AutoML\\
\midrule
\multirow{9}{*}{D} & 500 & 0.5 & 0.0867 & 0.1196 & 0.1255 & 0.1127 & \textbf{0.0629} & 0.1038 & \underline{0.0739} & 0.0928 & 0.1152 & 0.1158 \\
 &  & 1 & 0.1162 & 0.1636 & 0.1367 & 0.1355 & \textbf{0.0986} & 0.1693 & \underline{0.1066} & 0.1223 & 0.1496 & 0.1440 \\
 &  & 2 & 0.1366 & 0.2343 & 0.1387 & 0.1407 & \textbf{0.1230} & 0.2558 & \underline{0.1306} & 0.1615 & 0.2334 & 0.2322 \\
 & 1000 & 0.5 & 0.0515 & 0.0781 & 0.0715 & 0.0653 & \textbf{0.0305} & 0.0616 & \underline{0.0399} & 0.0583 & 0.0738 & 0.0730 \\
 &  & 1 & 0.0803 & 0.1116 & 0.1211 & 0.0972 & \textbf{0.0570} & 0.1048 & \underline{0.0706} & 0.0856 & 0.1105 & 0.1110 \\
 &  & 2 & 0.1102 & 0.1643 & 0.1340 & 0.1339 & \textbf{0.0933} & 0.1603 & \underline{0.1042} & 0.1142 & 0.1515 & 0.1505 \\
 & 2000 & 0.5 & 0.0316 & 0.0545 & 0.0297 & 0.0408 & \textbf{0.0172} & 0.0374 & \underline{0.0249} & 0.0418 & 0.0556 & 0.0531 \\
 &  & 1 & 0.0550 & 0.0758 & 0.0643 & 0.0632 & \textbf{0.0333} & 0.0641 & \underline{0.0432} & 0.0620 & 0.0774 & 0.0793 \\
 &  & 2 & 0.0853 & 0.1163 & 0.1102 & 0.1004 & \textbf{0.0642} & 0.1077 & \underline{0.0763} & 0.0882 & 0.1077 & 0.1100 \\
\midrule
\multirow{9}{*}{E} & 500 & 0.5 & \textbf{0.0904} & 0.1645 & 0.1753 & 0.2411 & 0.1617 & 0.2679 & \underline{0.0968} & 0.1414 & 0.1618 & 0.1160 \\
 &  & 1 & \underline{0.1962} & 0.3025 & 0.2394 & 0.3715 & 0.2365 & 0.4154 & \textbf{0.1857} & 0.2345 & 0.2553 & 0.2382 \\
 &  & 2 & 0.3802 & 0.5092 & 0.3367 & 0.5225 & \underline{0.3315} & 0.6063 & \textbf{0.3187} & 0.4018 & 0.4578 & 0.4405 \\
 & 1000 & 0.5 & \textbf{0.0429} & 0.0759 & 0.1211 & 0.1520 & 0.1018 & 0.1640 & \underline{0.0468} & 0.0767 & 0.0659 & 0.0553 \\
 &  & 1 & \textbf{0.0897} & 0.1313 & 0.1705 & 0.2276 & 0.1609 & 0.2475 & \underline{0.0933} & 0.1151 & 0.1212 & 0.1052 \\
 &  & 2 & \underline{0.1916} & 0.2758 & 0.2337 & 0.3374 & 0.2434 & 0.3649 & \textbf{0.1818} & 0.2200 & 0.2368 & 0.2450 \\
 & 2000 & 0.5 & \underline{0.0230} & 0.0294 & 0.0765 & 0.1011 & 0.0556 & 0.1018 & \textbf{0.0210} & 0.0360 & 0.0235 & 0.0285 \\
 &  & 1 & \underline{0.0495} & 0.0600 & 0.1251 & 0.1533 & 0.0946 & 0.1609 & \textbf{0.0413} & 0.0642 & 0.0562 & 0.0555 \\
 &  & 2 & 0.1113 & 0.1348 & 0.1846 & 0.2359 & 0.1541 & 0.2453 & \textbf{0.0868} & 0.1219 & 0.1002 & \underline{0.0975} \\
\midrule
\multirow{9}{*}{F} & 500 & 0.5 & 0.0334 & 0.0945 & \textbf{0.0253} & 0.1147 & 0.0496 & 0.1559 & \underline{0.0256} & 0.0639 & 0.0960 & 0.0793 \\
 &  & 1 & 0.0682 & 0.1706 & \textbf{0.0378} & 0.1685 & 0.0916 & 0.2627 & \underline{0.0454} & 0.0982 & 0.1538 & 0.1152 \\
 &  & 2 & 0.0996 & 0.2333 & \textbf{0.0656} & 0.2518 & 0.1645 & 0.4423 & \underline{0.0891} & 0.1725 & 0.2226 & 0.1859 \\
 & 1000 & 0.5 & \underline{0.0160} & 0.0467 & 0.0161 & 0.0656 & 0.0299 & 0.0846 & \textbf{0.0132} & 0.0347 & 0.0529 & 0.0422 \\
 &  & 1 & 0.0271 & 0.0878 & \textbf{0.0249} & 0.0975 & 0.0544 & 0.1525 & \underline{0.0258} & 0.0560 & 0.0934 & 0.0807 \\
 &  & 2 & 0.0642 & 0.1471 & \textbf{0.0374} & 0.1500 & 0.0983 & 0.2488 & \underline{0.0494} & 0.1012 & 0.1295 & 0.1441 \\
 & 2000 & 0.5 & 0.0104 & 0.0267 & \underline{0.0099} & 0.0390 & 0.0169 & 0.0456 & \textbf{0.0073} & 0.0212 & 0.0333 & 0.0252 \\
 &  & 1 & 0.0190 & 0.0455 & \underline{0.0150} & 0.0581 & 0.0311 & 0.0729 & \textbf{0.0148} & 0.0326 & 0.0494 & 0.0461 \\
 &  & 2 & 0.0373 & 0.0917 & \textbf{0.0256} & 0.0914 & 0.0557 & 0.1327 & \underline{0.0277} & 0.0532 & 0.0879 & 0.0720 \\
\bottomrule
\end{tabular}}
\end{table*}

\subsection{Real data comparison}
\label{app:causal_acic}
\textbf{Data.}
The Infant Health and Development Program (IHDP)~\citep{brooks1992effects} dataset contains $n=4302$ subjects with $8$ covariates capturing demographic, prenatal, and birth-related characteristics, a binary treatment indicating participation in a specialist early-childhood intervention program, and continuous outcomes representing cognitive test scores measured at age $3$.
We use a collection of semi-synthetic versions of the dataset that were created for the ACIC 2017 Causal Inference Benchmark challenge described in~\citet{hahn2019atlantic}, where treatment assignments and potential outcomes are simulated conditional on the real-world covariates.
This design ensures access to the ground-truth treatment effects, allowing quantitative comparison across estimators via the MSE of the CATE.
We accessed the data via Richard Hahn's personal website: \url{https://math.la.asu.edu/~prhahn/}.

Each semi-synthetic dataset is generated according to one of 24 data-generating processes (DGPs) that were chosen to systematically vary structural and noise properties to probe robustness of causal estimators.
Specifically, the DGPs are organized into three broad settings for the error structure:
\begin{itemize}[leftmargin=*]
\item \textbf{IID:} Errors are generated independently across individuals, corresponding to the standard assumption of homogeneous, uncorrelated residuals.
\item \textbf{Non-additive:} Treatment effects and outcomes are generated via nonlinear functions and interaction terms, inducing complex, non-additive response surfaces.
\item \textbf{Group Corr:} The error terms include intra-group correlation components to mimic unobserved group-level dependencies (e.g., hospitals or study sites), introducing residual correlation beyond observed covariates and thereby violating the IID assumption.
\end{itemize}
A fourth error type, heteroskedastic case, was later discovered by \citet{hahn2019atlantic} to contain inconsistencies and hence omitted from the benchmark comparison.

Within each setting, eight distinct DGPs are defined to explore combinations of three factors that affect causal inference difficulty: the effect magnitude $\xi$ (large vs.\ small), the noise level $\eta$ (low vs.\ high), and the selection strength $\kappa$ (weak vs.\ strong confounding).
Overall, these yield $3$ settings $\times$ $8$ scenarios $\times$ $250$ replications $=6{,}000$ semi-synthetic datasets derived from the IHDP covariates.

\textbf{Evaluation.}
We compare four TabPFN-based meta-learners (S-Learner, T-Learner, X-Learner, and DR-Learner) with the 22 estimators reported in the benchmark challenge.
For a fair comparison, we only include methods with available results across all $6{,}000$ datasets, resulting in 16 baseline approaches.
For all TabPFN-based methods, we use the TabPFN classifier to estimate the propensity score model and the TabPFN regressor to fit the outcome model, as well as any auxiliary models with continuous responses.
The evaluation protocol follows that of \citet{hahn2019atlantic}, and we estimate the RMSE of the CATEs based on 250 repetitions under each scenario.
The results for the baseline approaches were taken directly from Richard Hahn's personal website and were not recalculated.

\begin{figure*}[tp]
    \centering
    \includegraphics[height=0.17\linewidth]{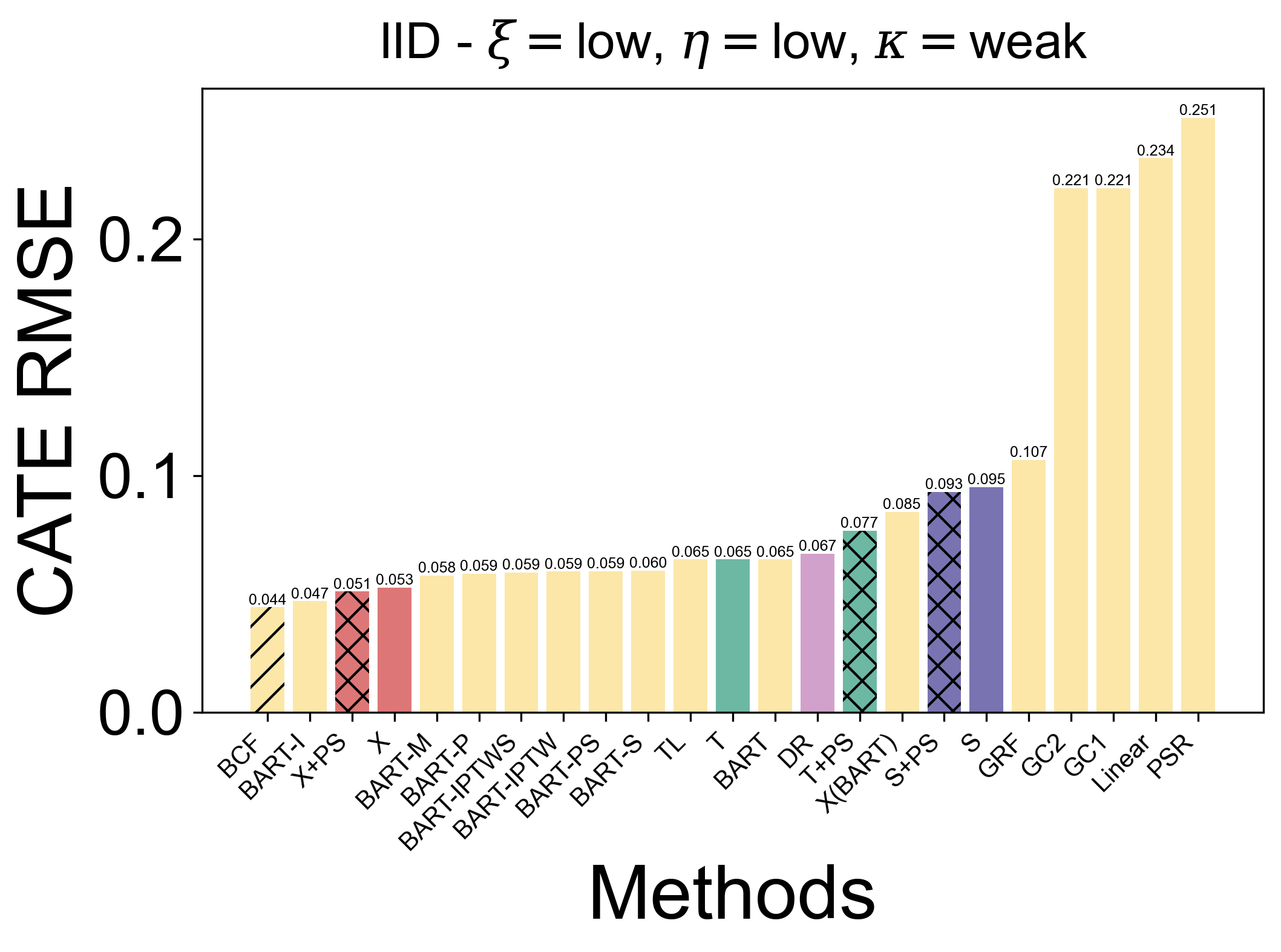}
    \includegraphics[height=0.17\linewidth]{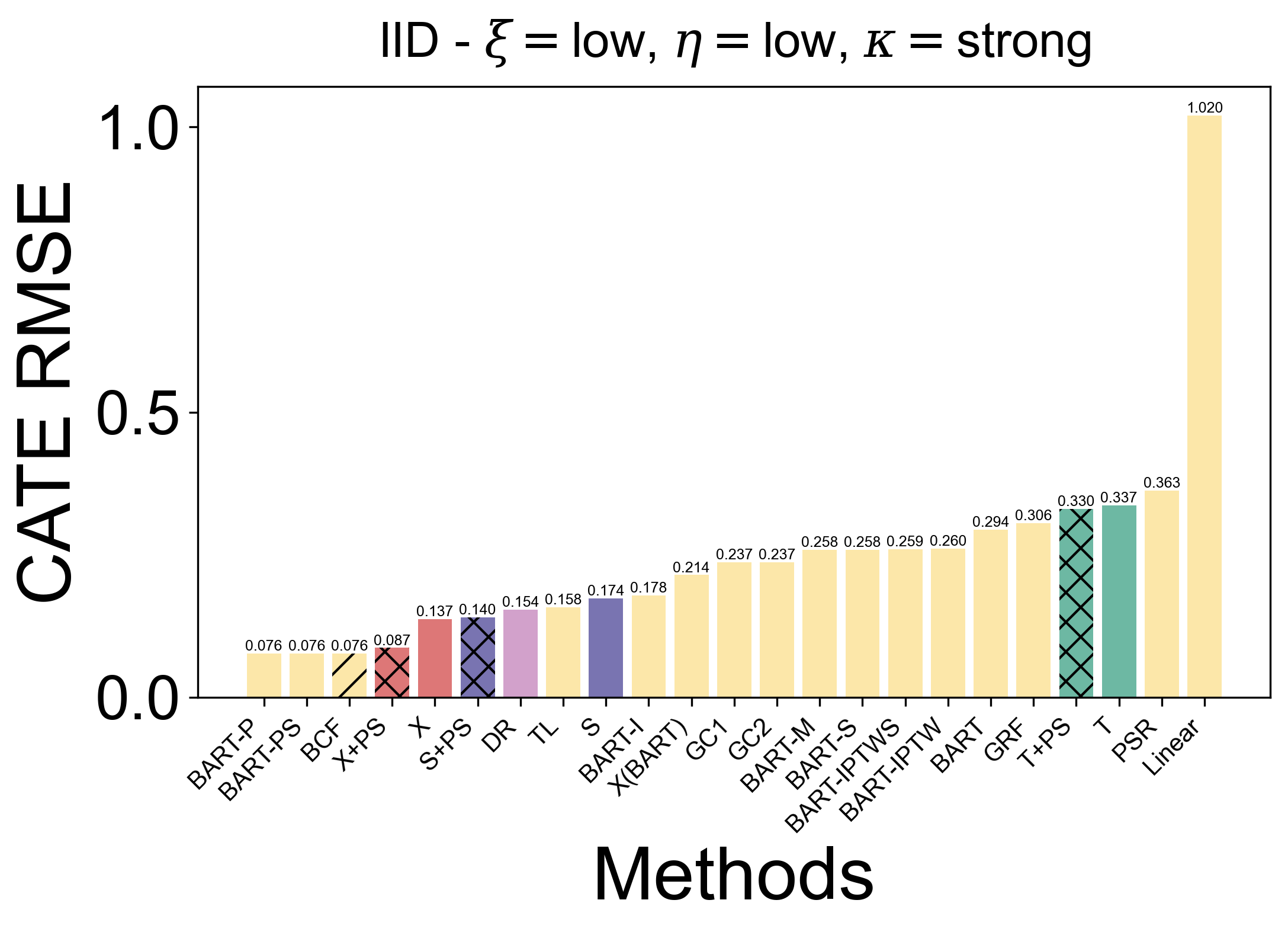}
    \includegraphics[height=0.17\linewidth]{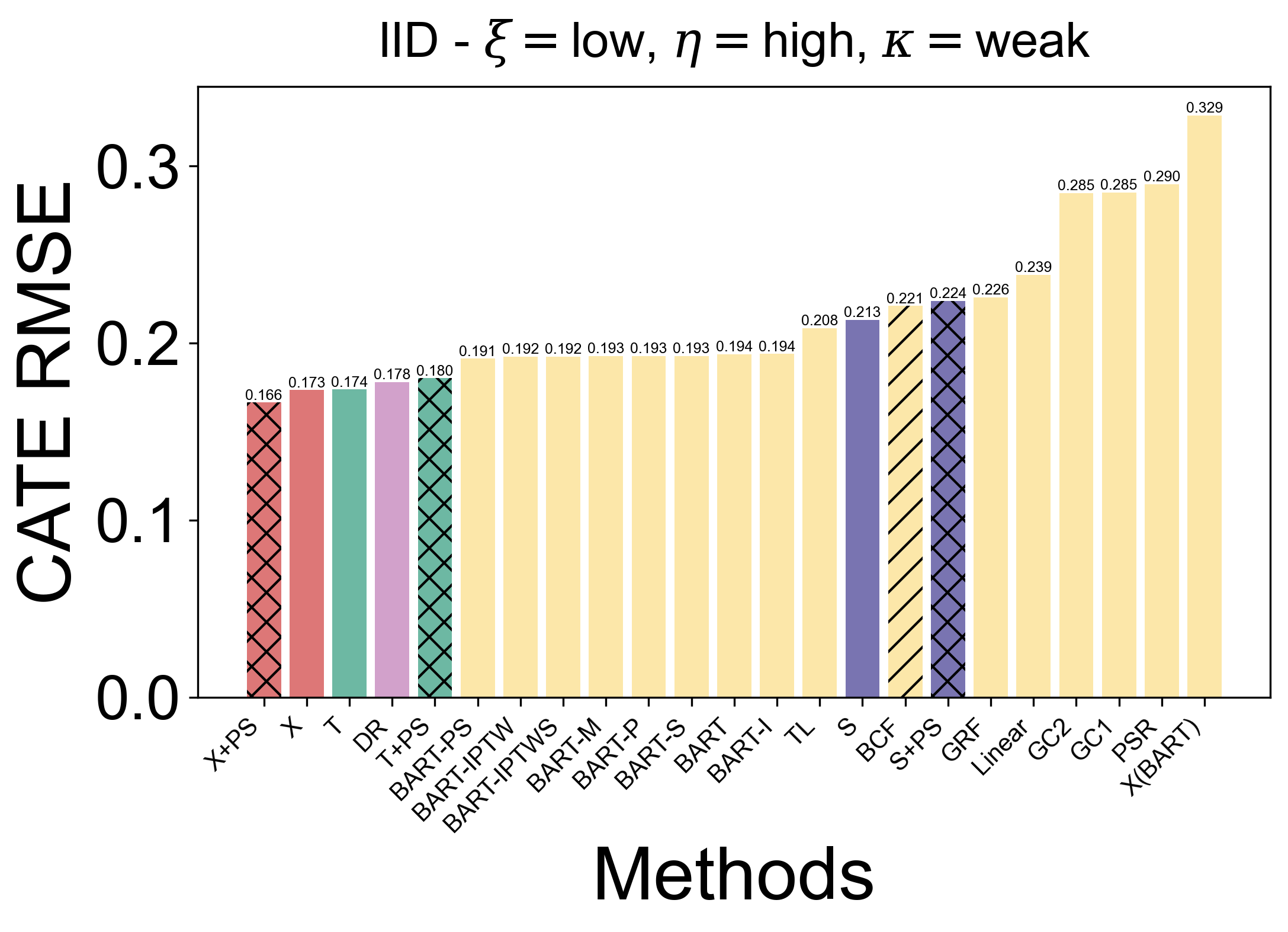}
    \includegraphics[height=0.17\linewidth]{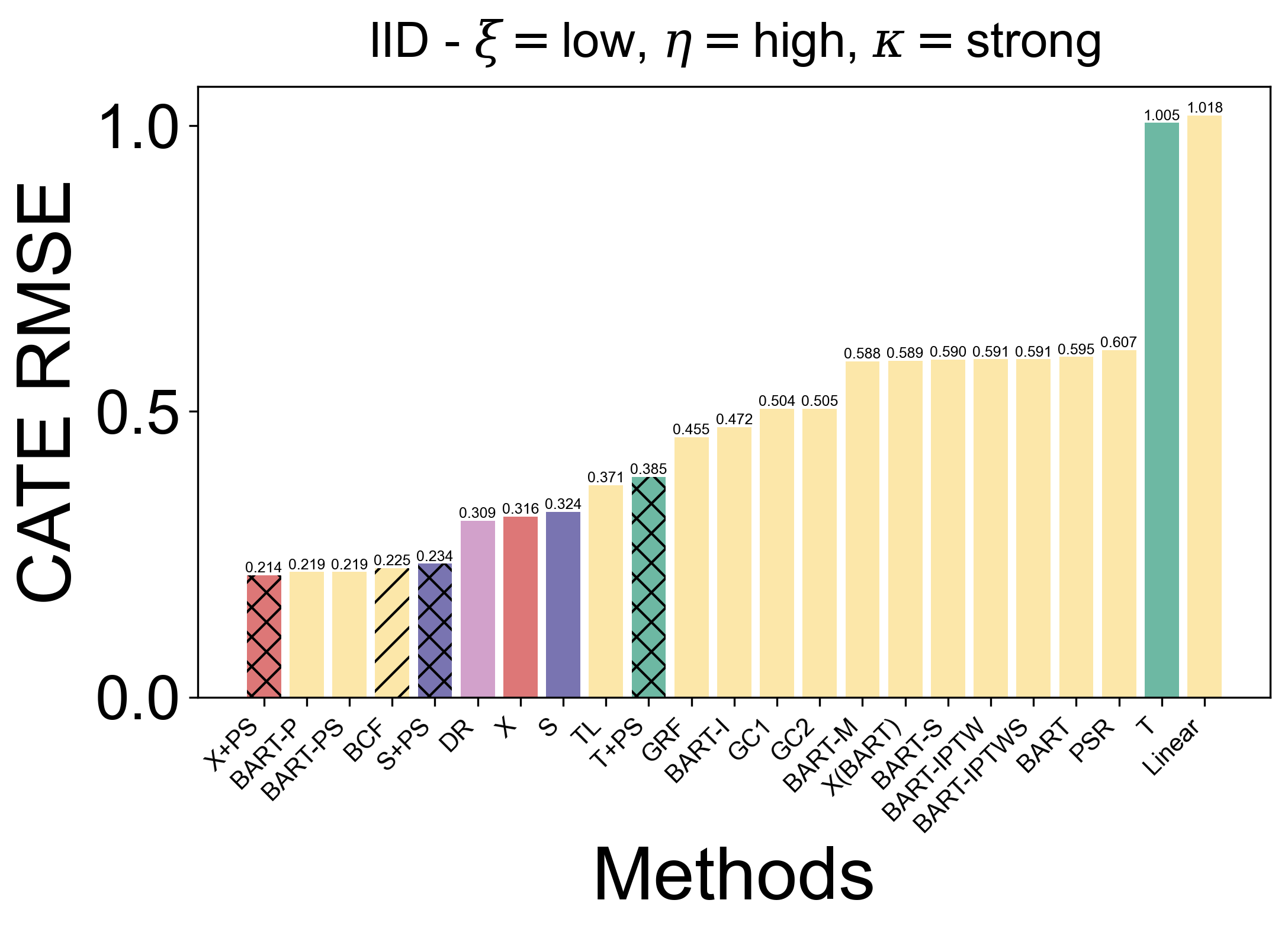}\\
    \includegraphics[height=0.17\linewidth]{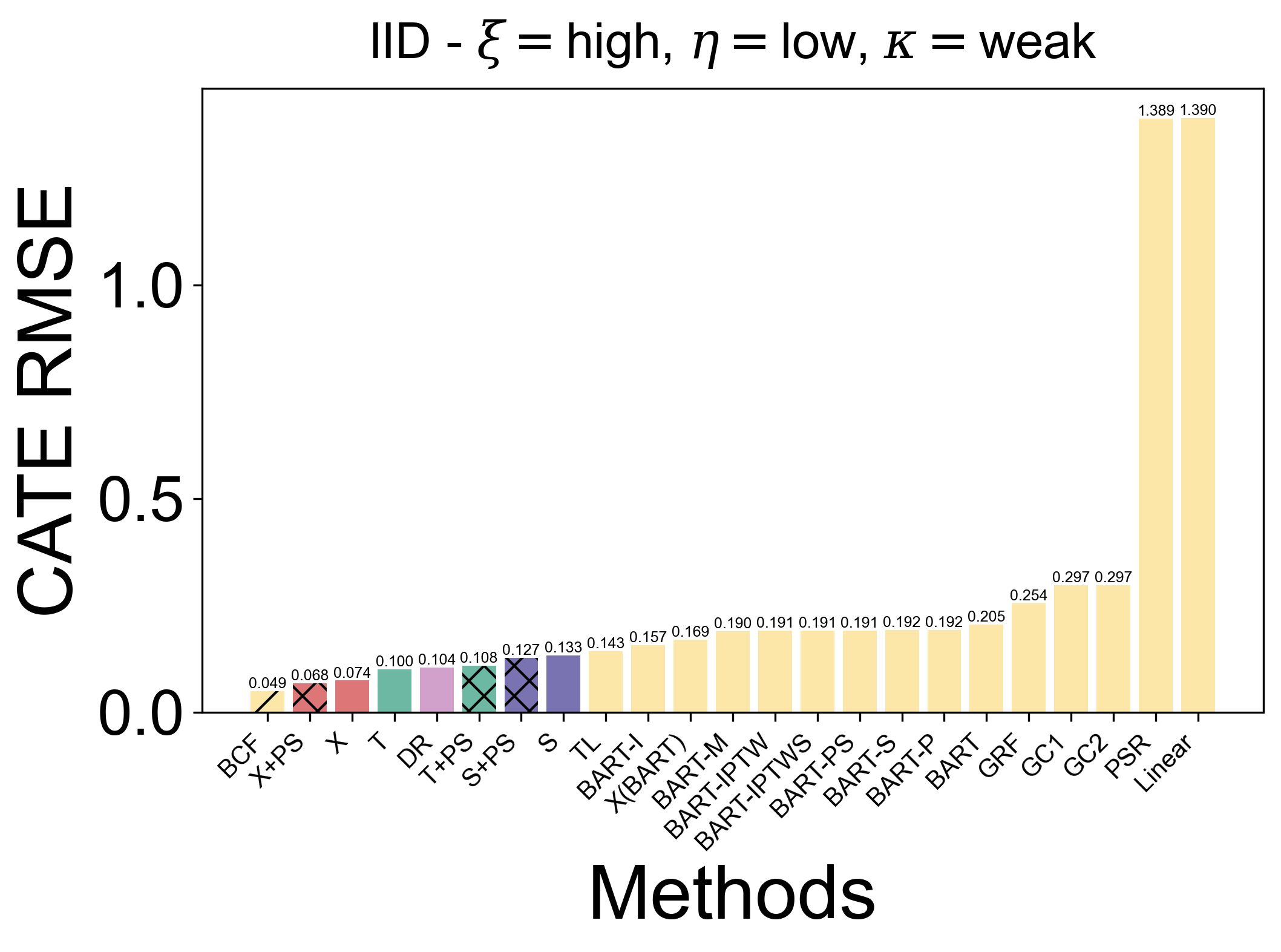}
    \includegraphics[height=0.17\linewidth]{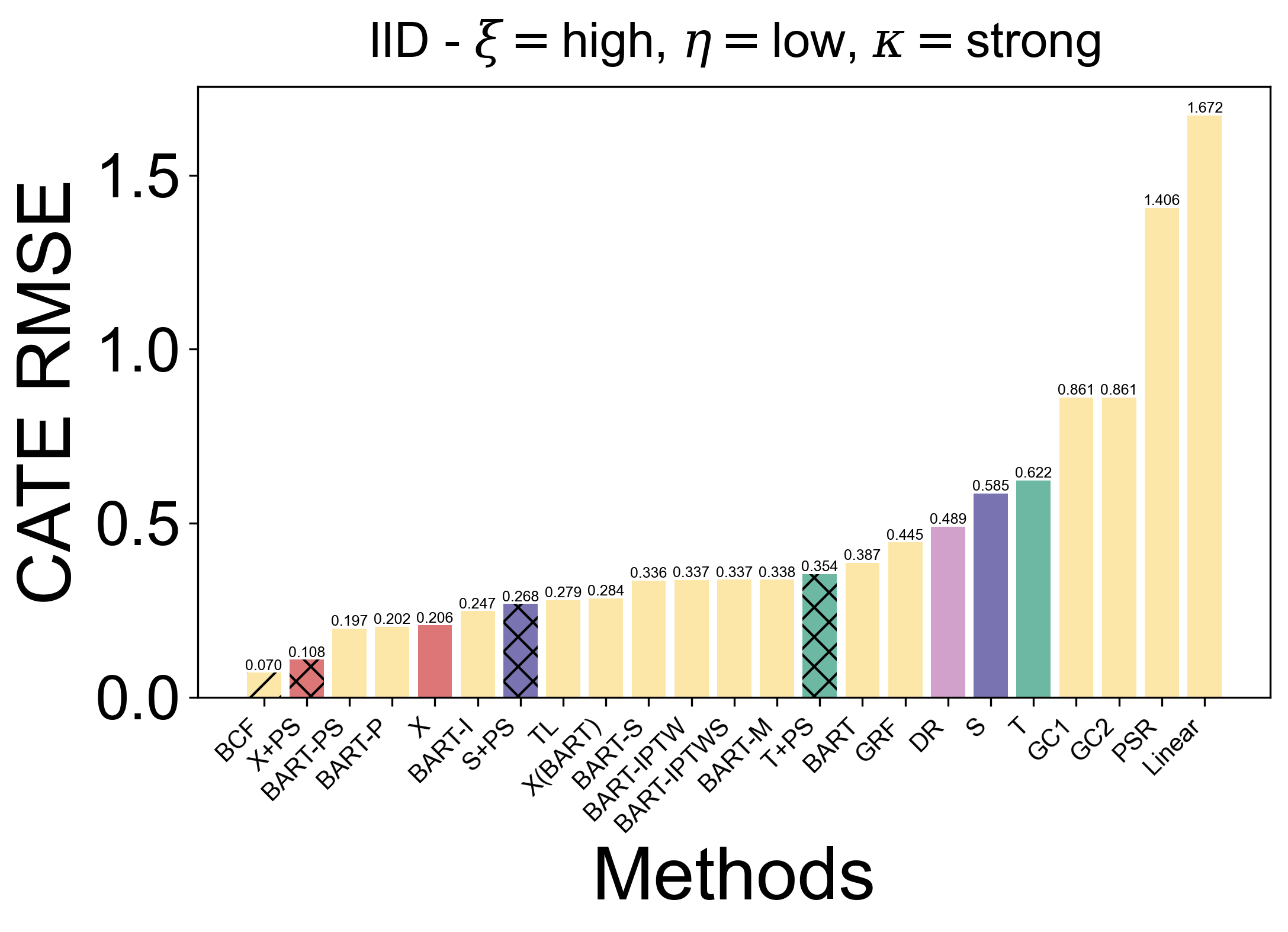}
    \includegraphics[height=0.17\linewidth]{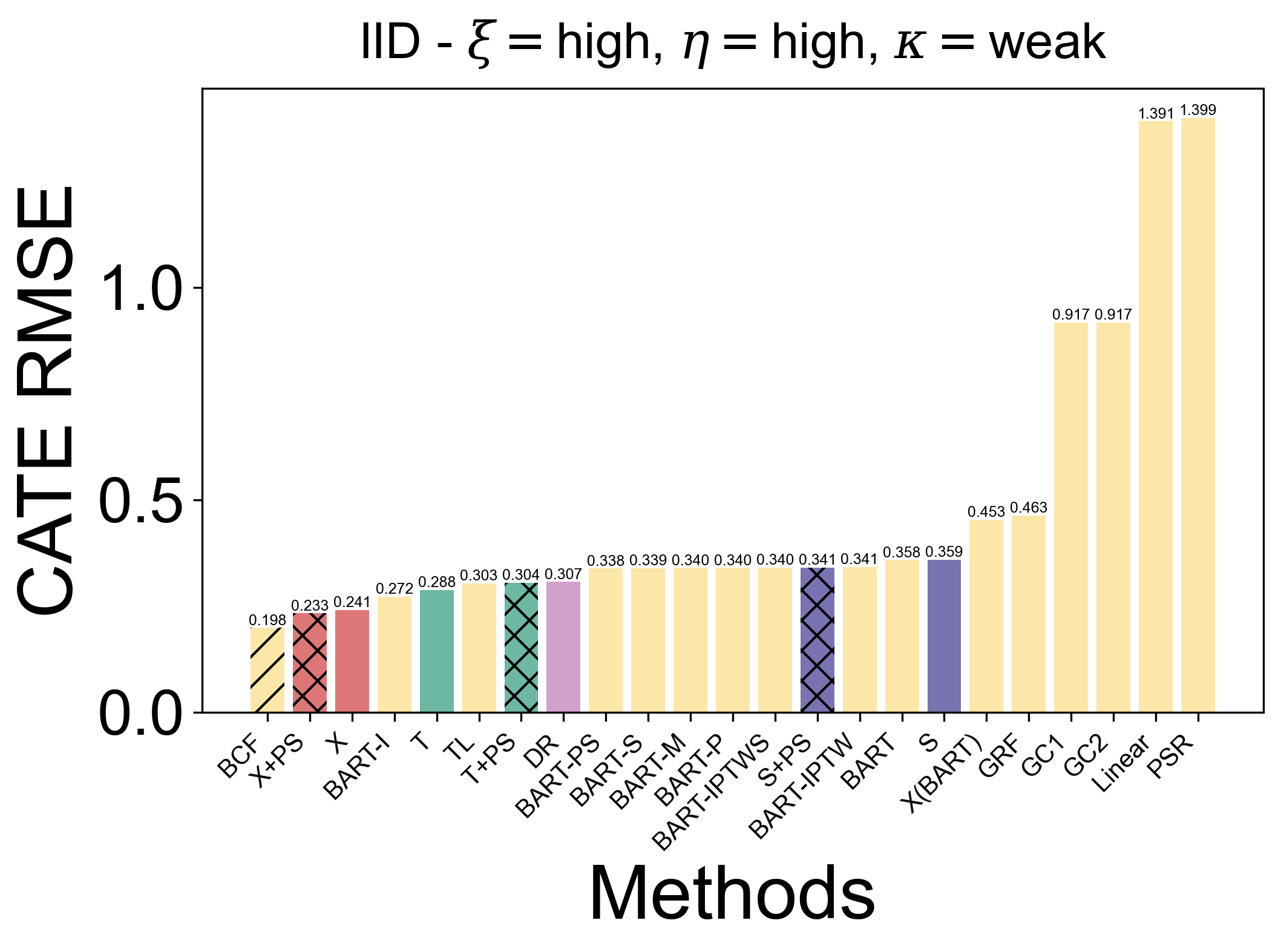}
    \includegraphics[height=0.17\linewidth]{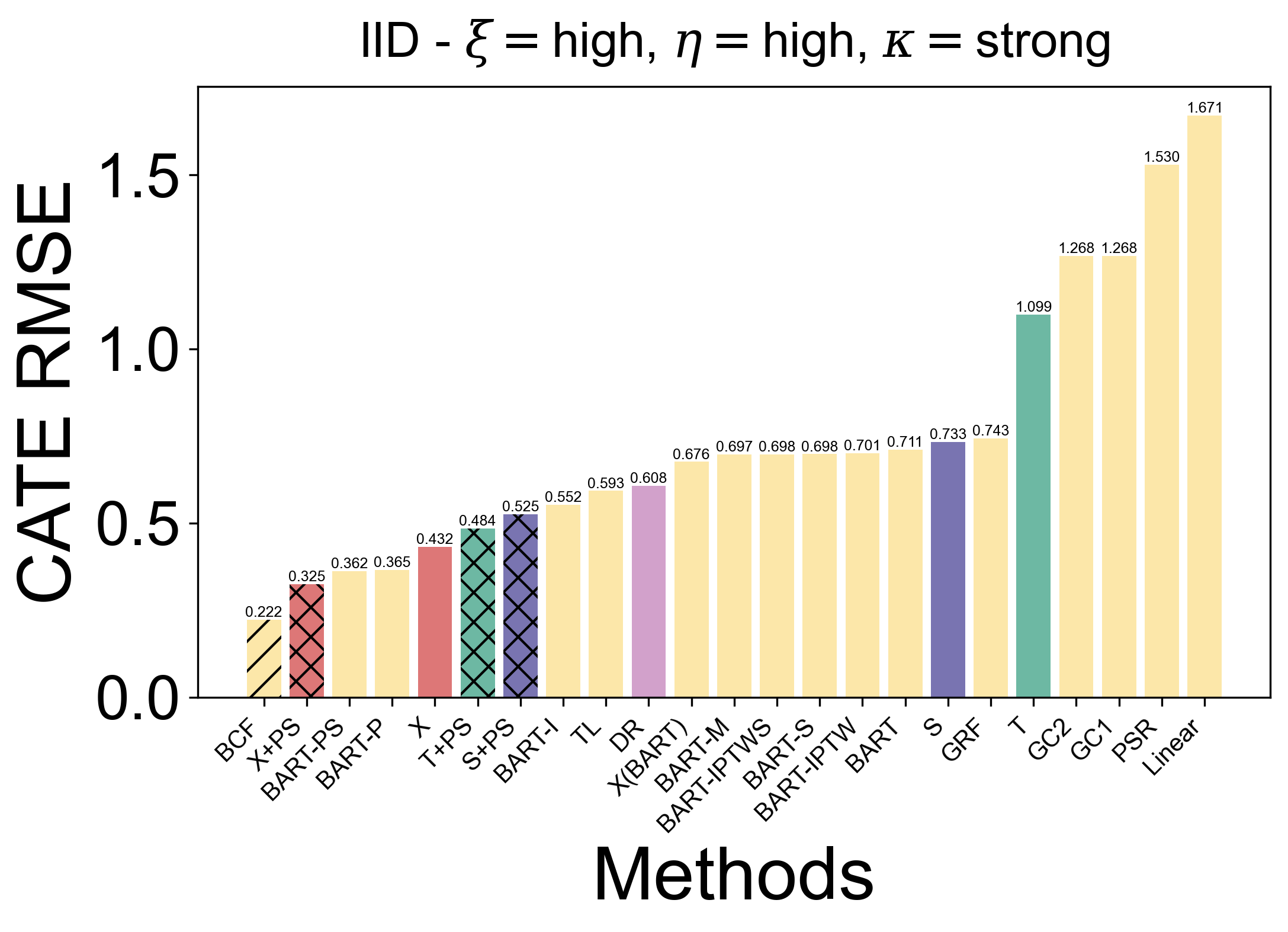}\\
    \includegraphics[height=0.17\linewidth]{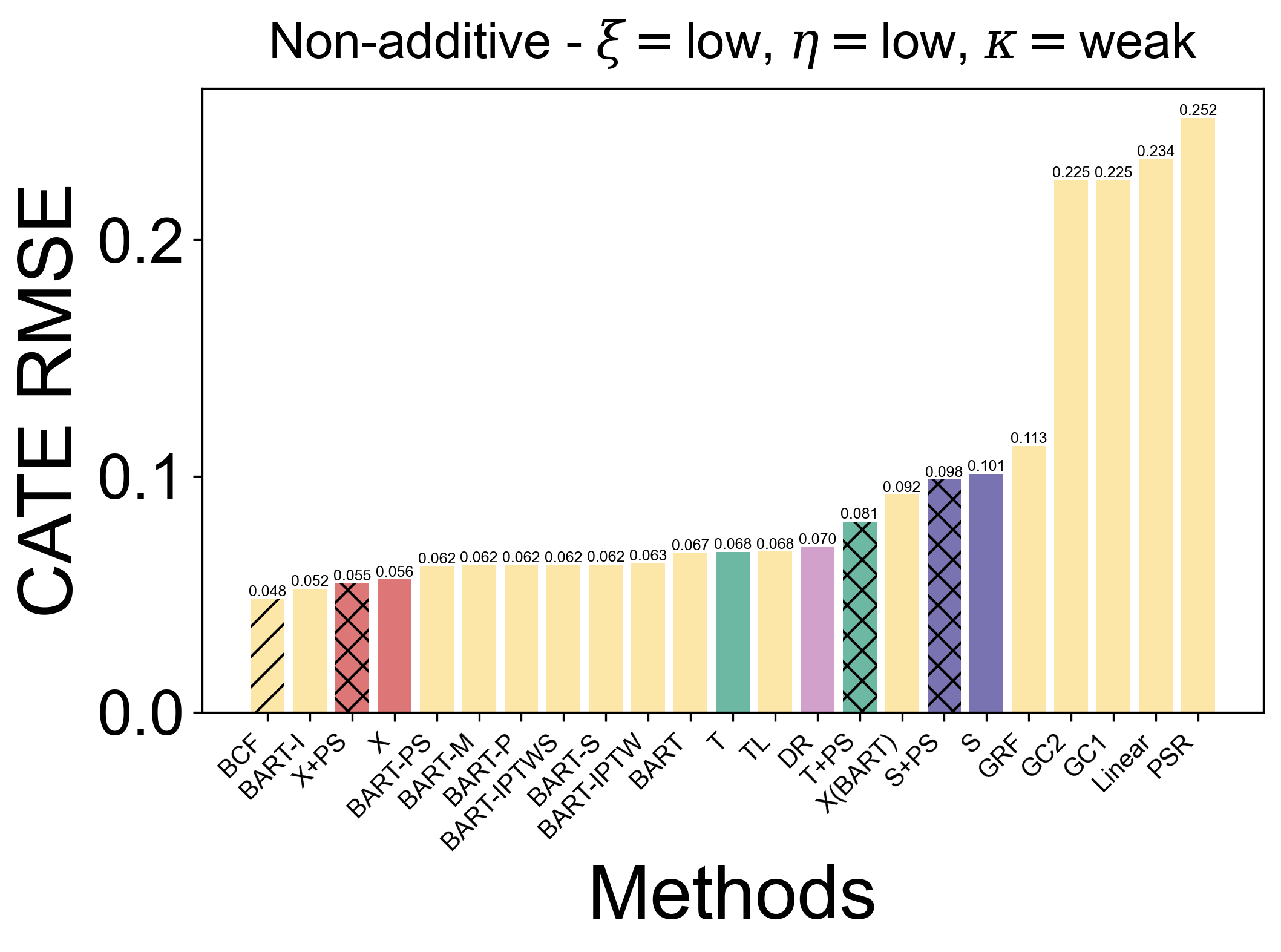}
    \includegraphics[height=0.17\linewidth]{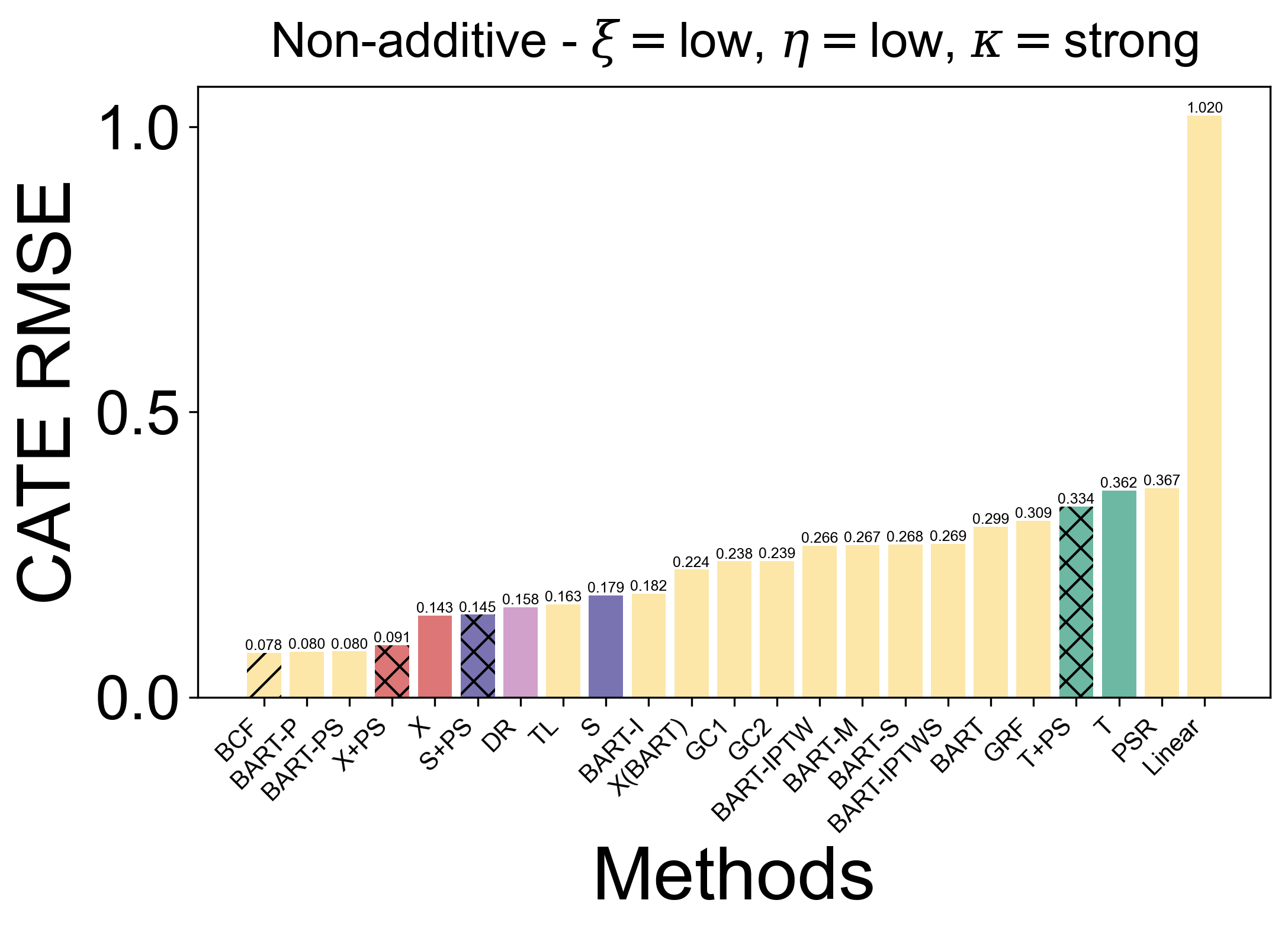}
    \includegraphics[height=0.17\linewidth]{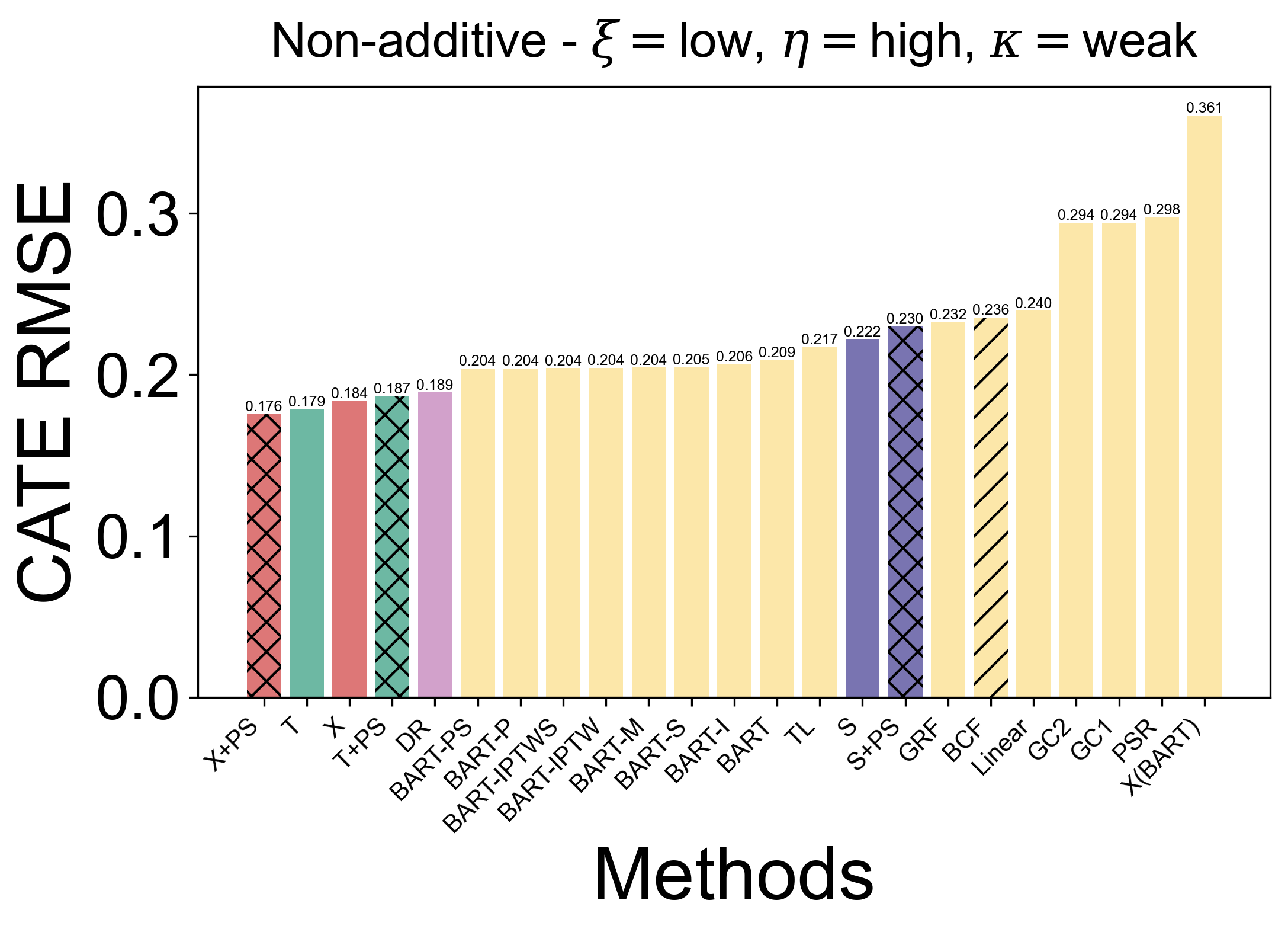}
    \includegraphics[height=0.17\linewidth]{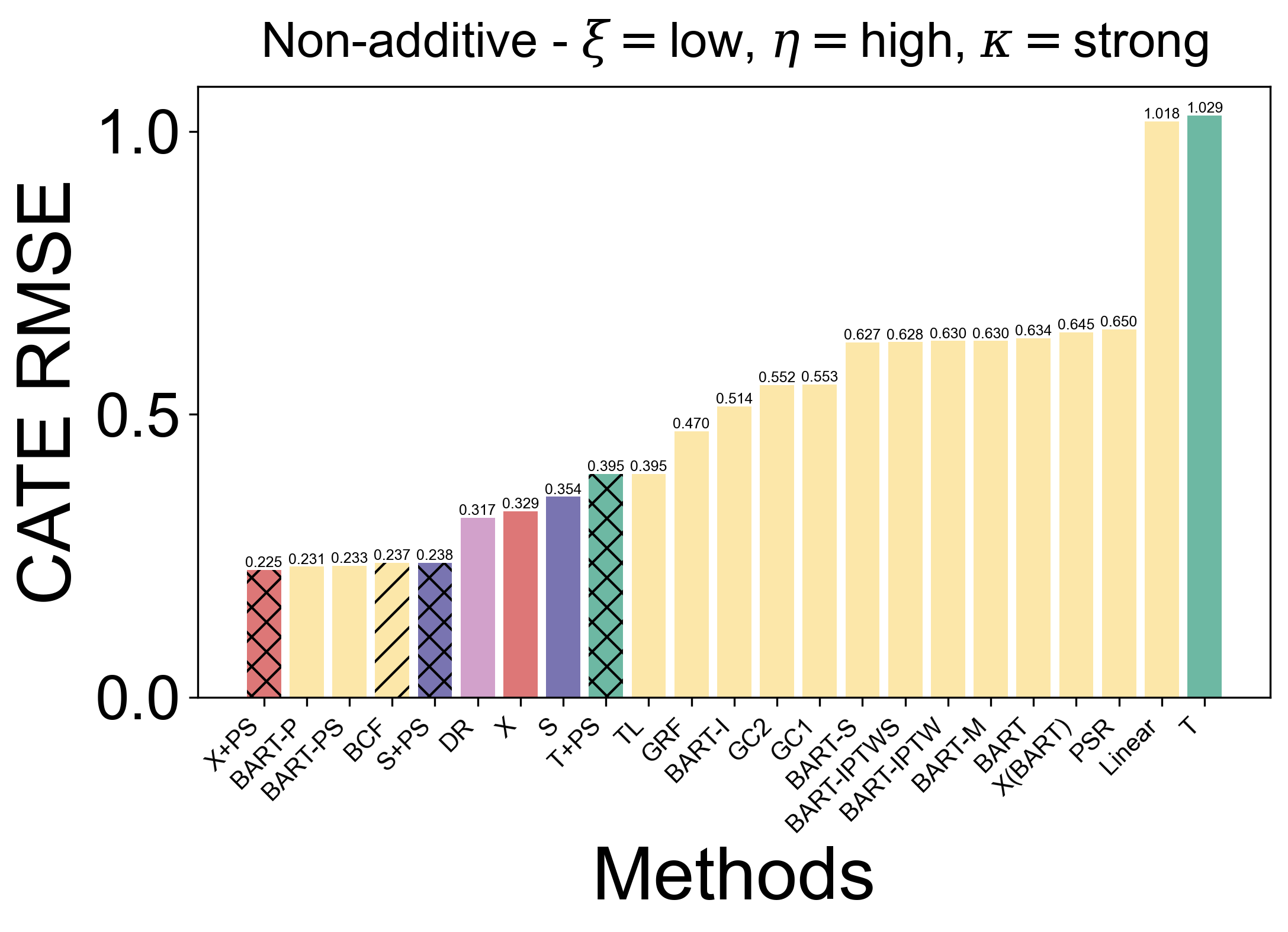}\\
    \includegraphics[height=0.17\linewidth]{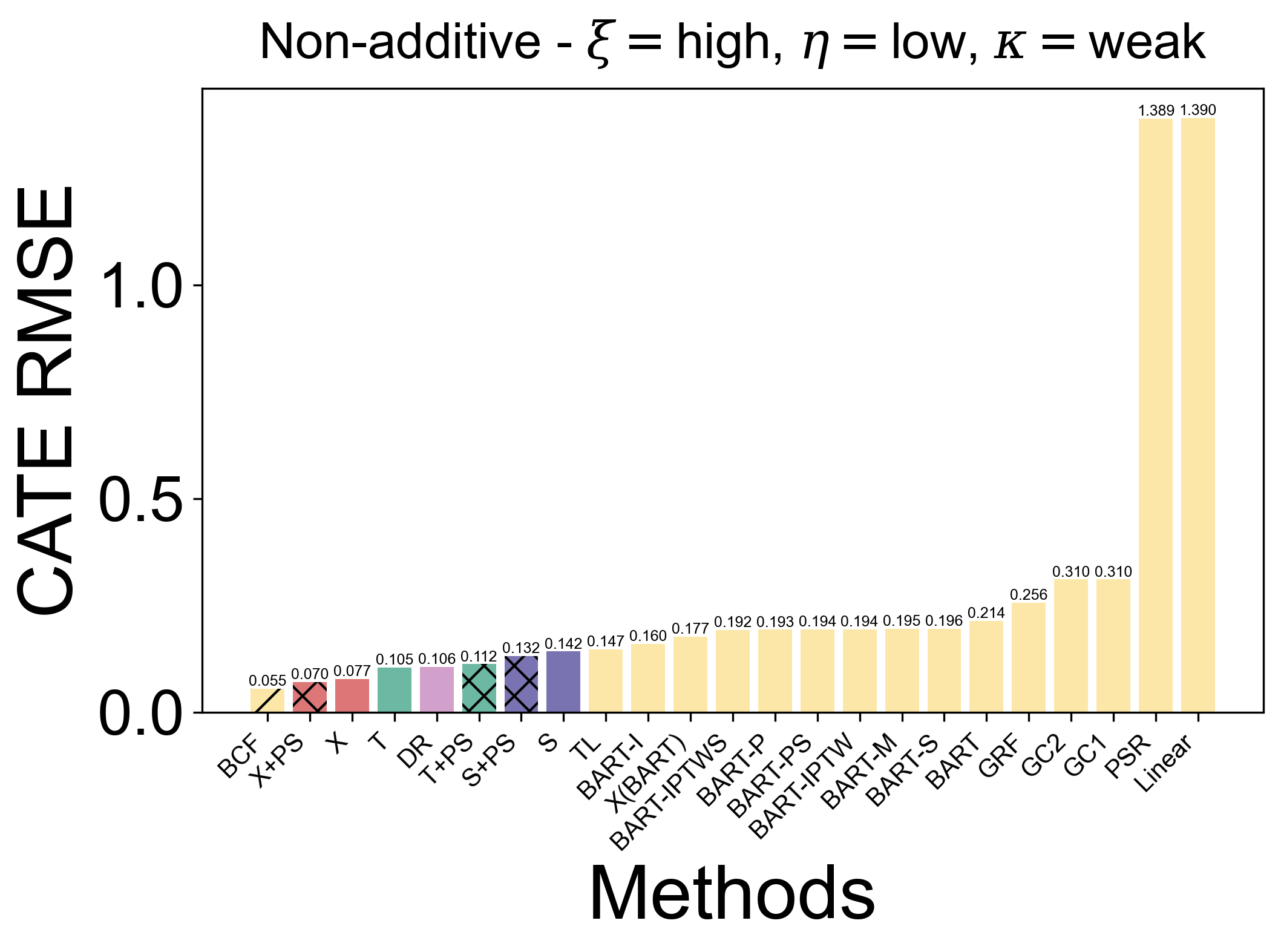}
    \includegraphics[height=0.17\linewidth]{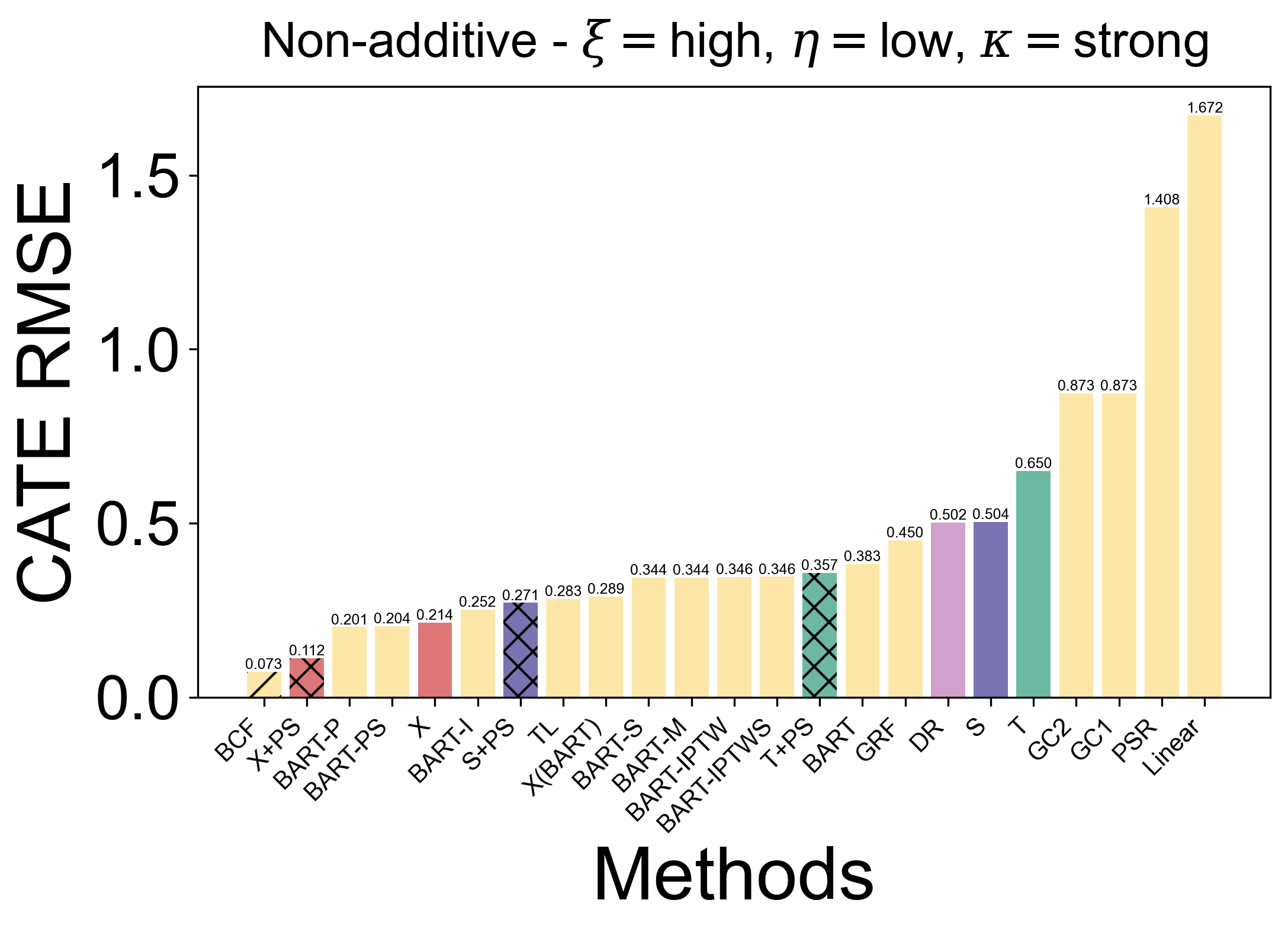}
    \includegraphics[height=0.17\linewidth]{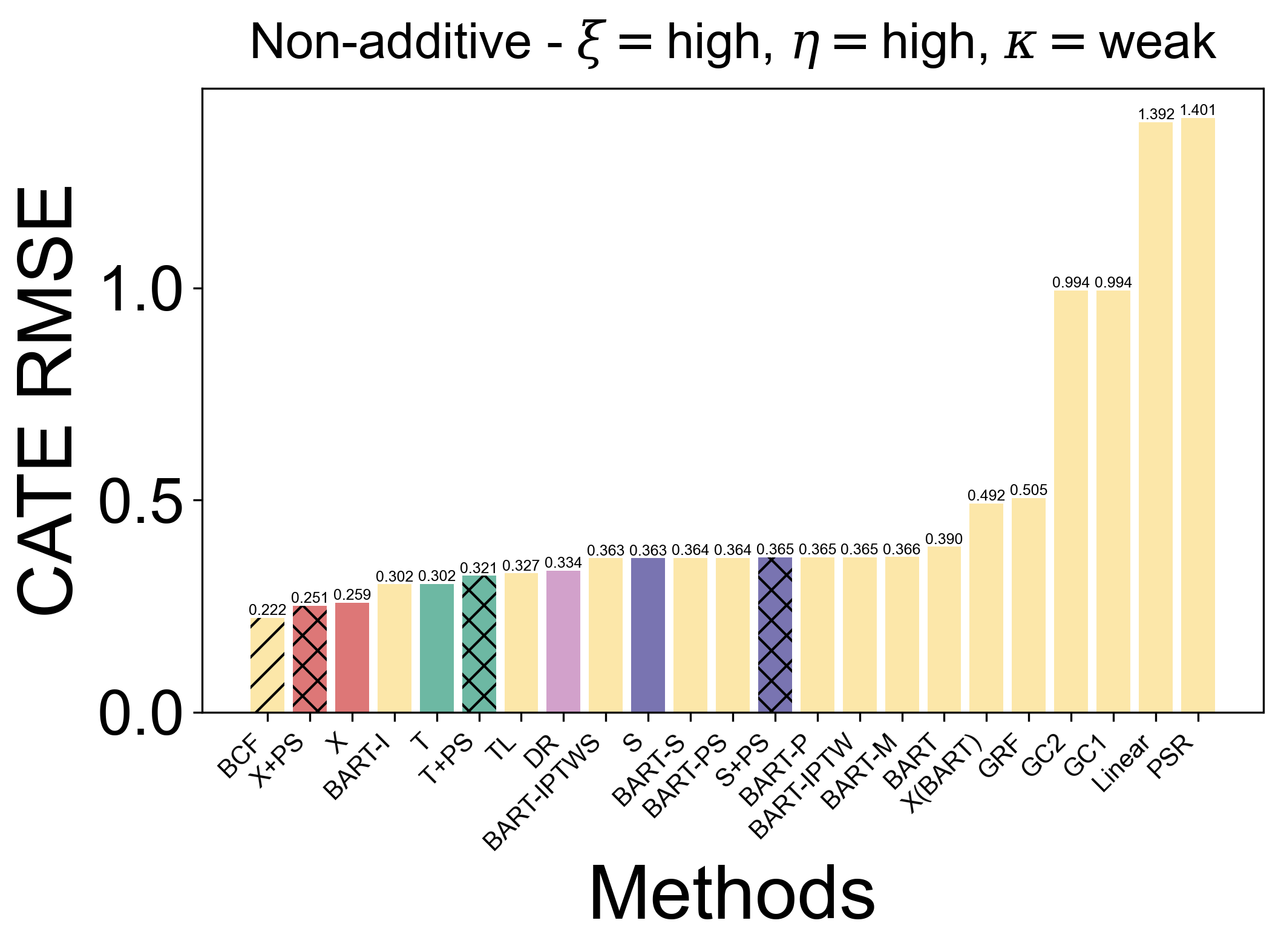}
    \includegraphics[height=0.17\linewidth]{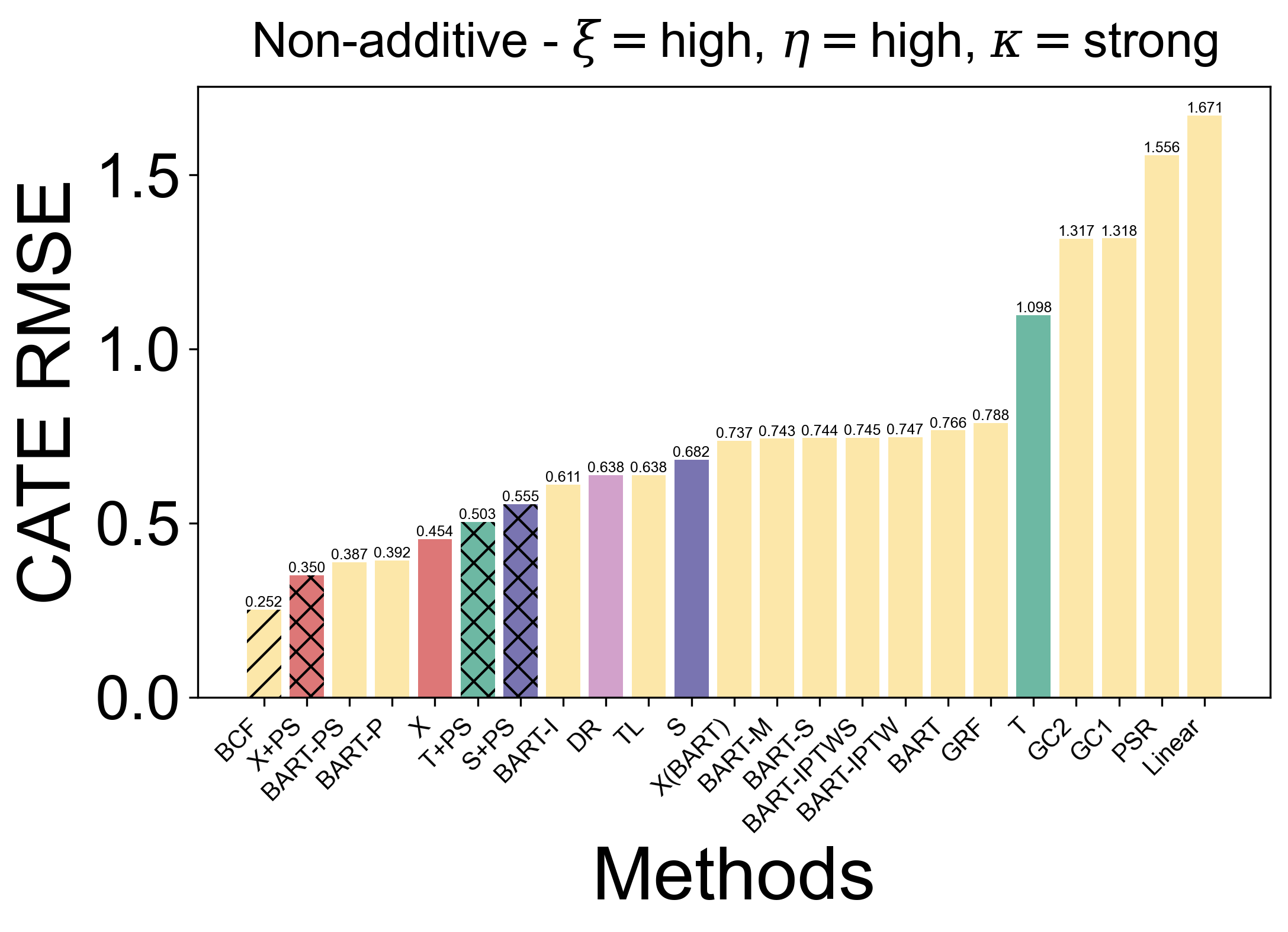}\\
    \includegraphics[height=0.17\linewidth]{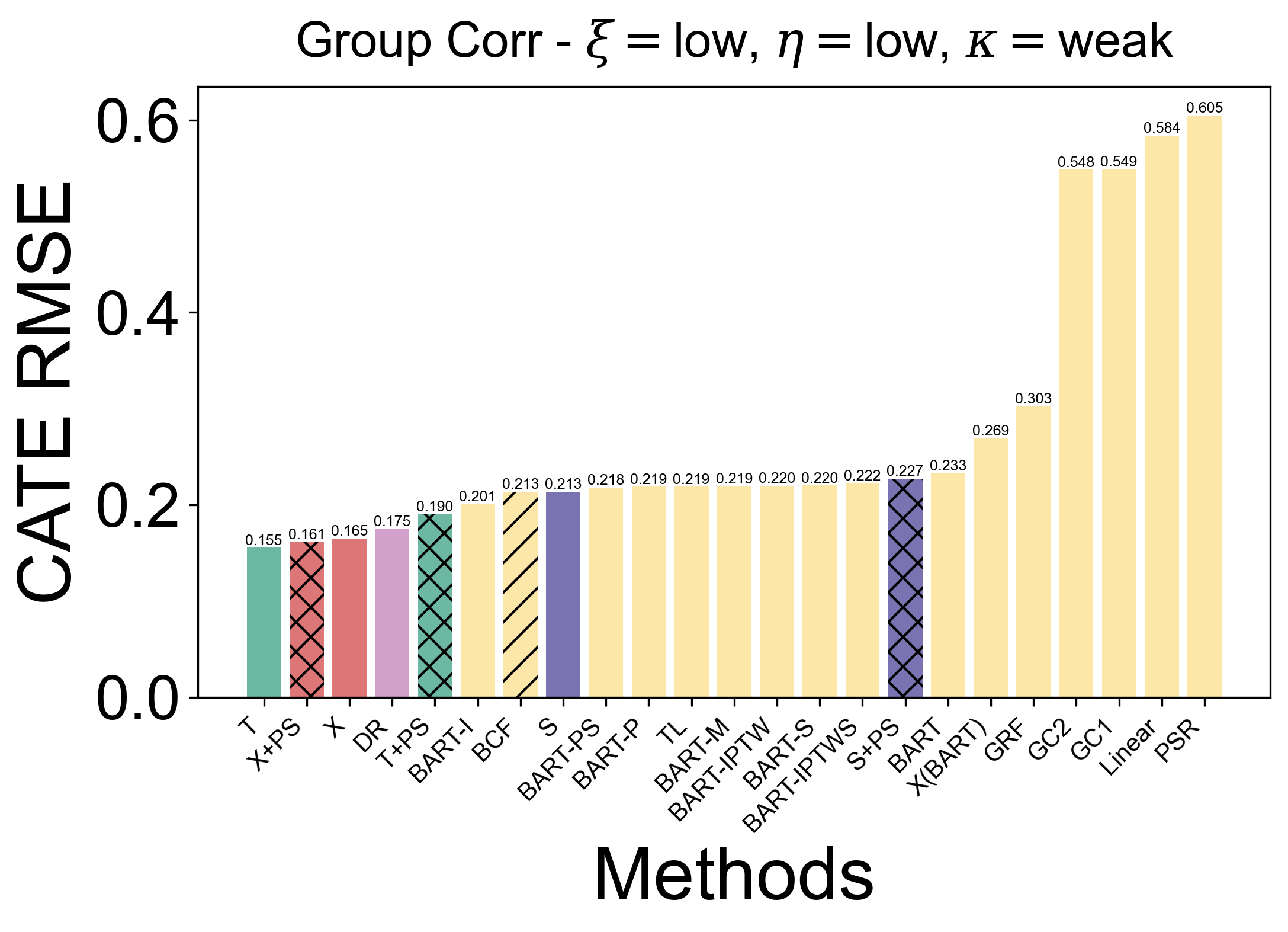}
    \includegraphics[height=0.17\linewidth]{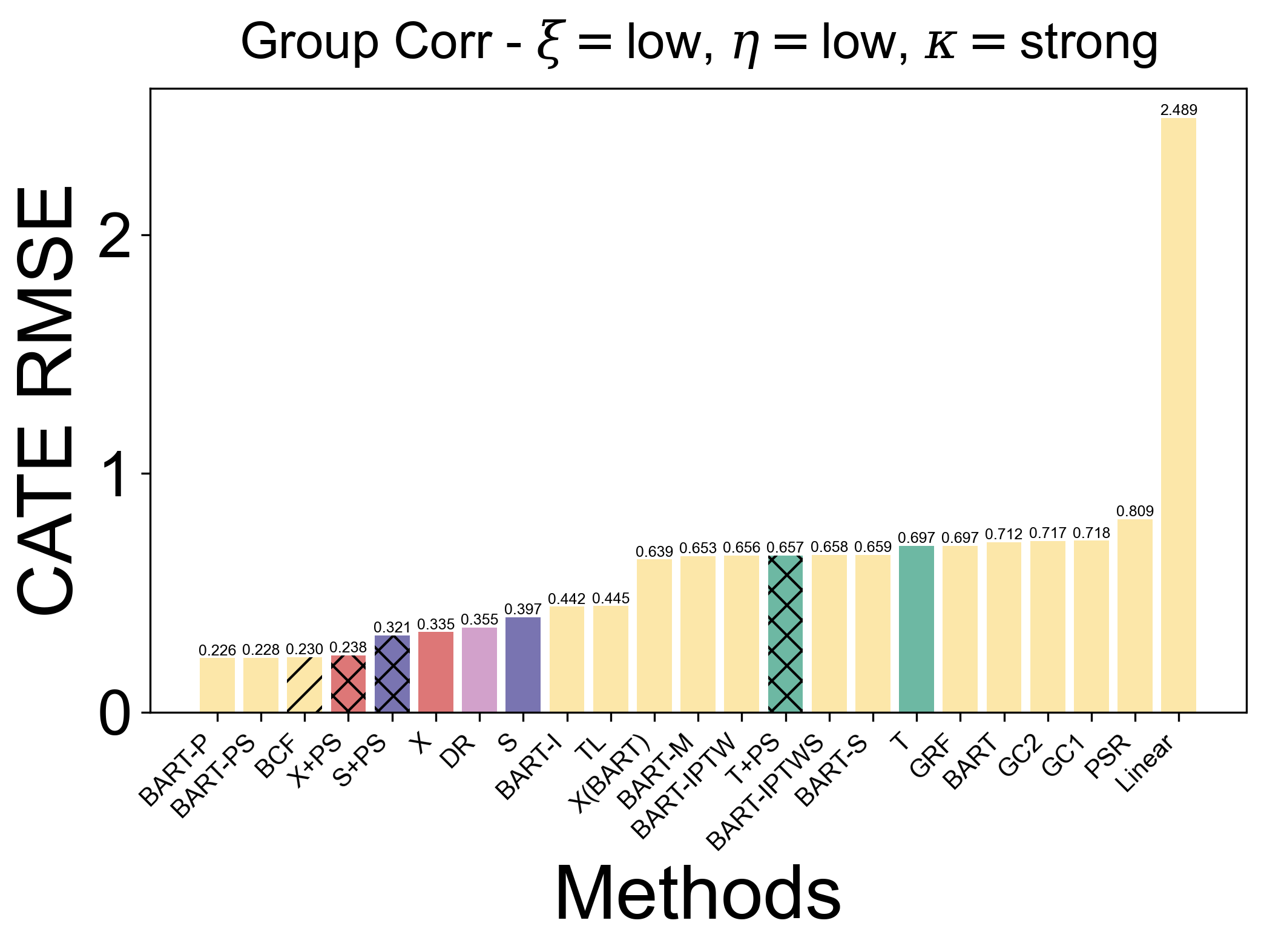}
    \includegraphics[height=0.17\linewidth]{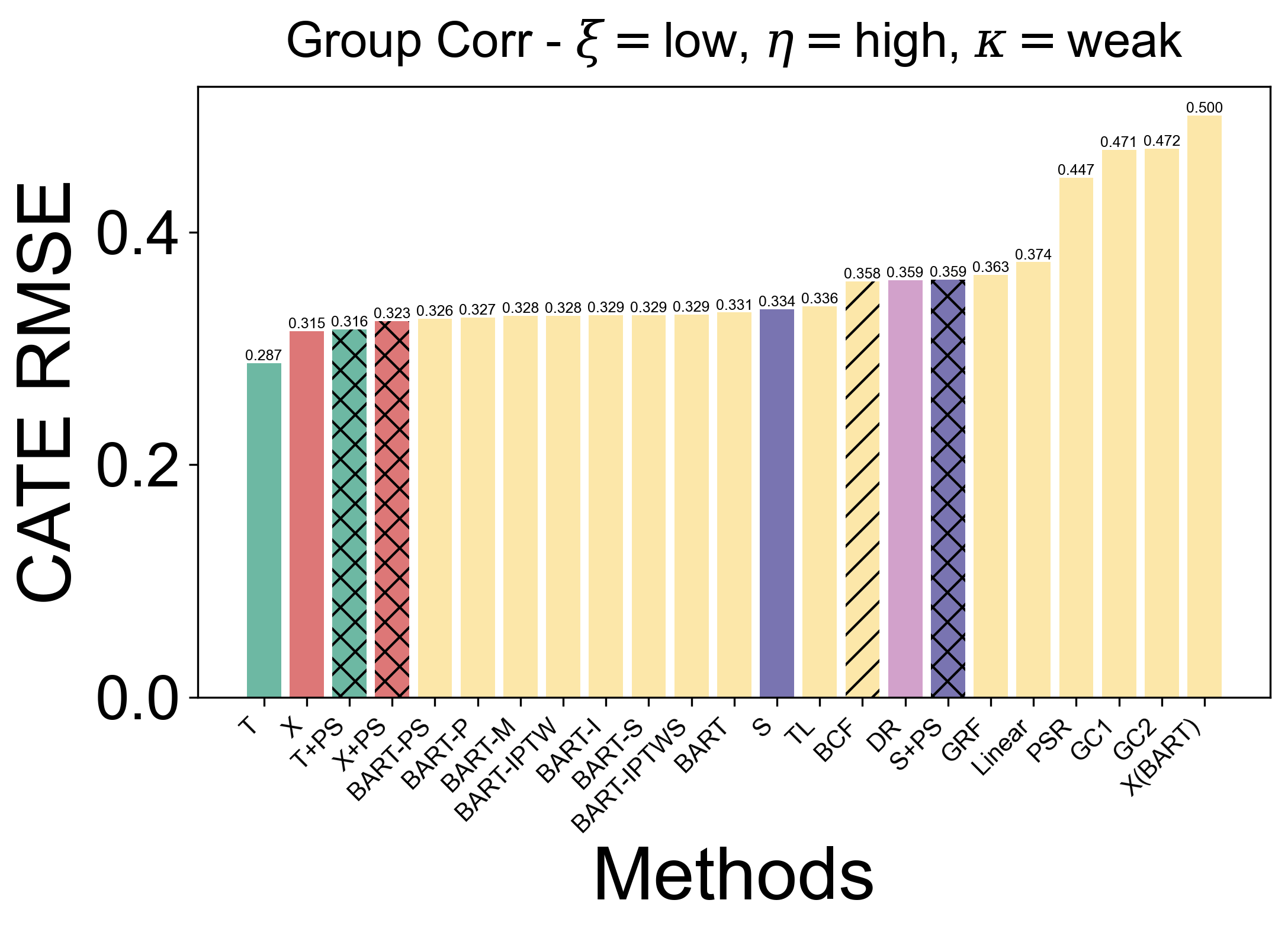}
    \includegraphics[height=0.17\linewidth]{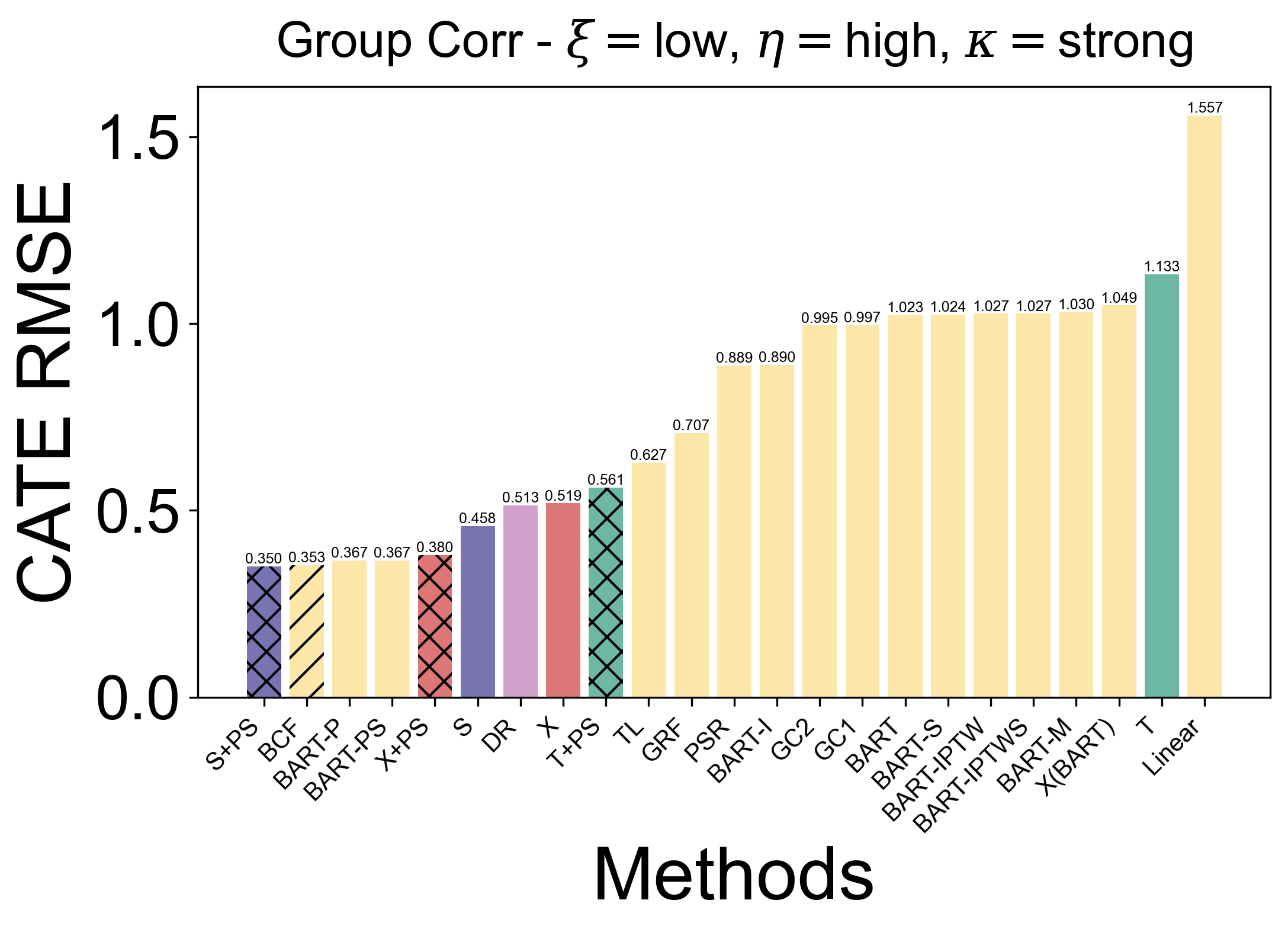}\\
    \includegraphics[height=0.17\linewidth]{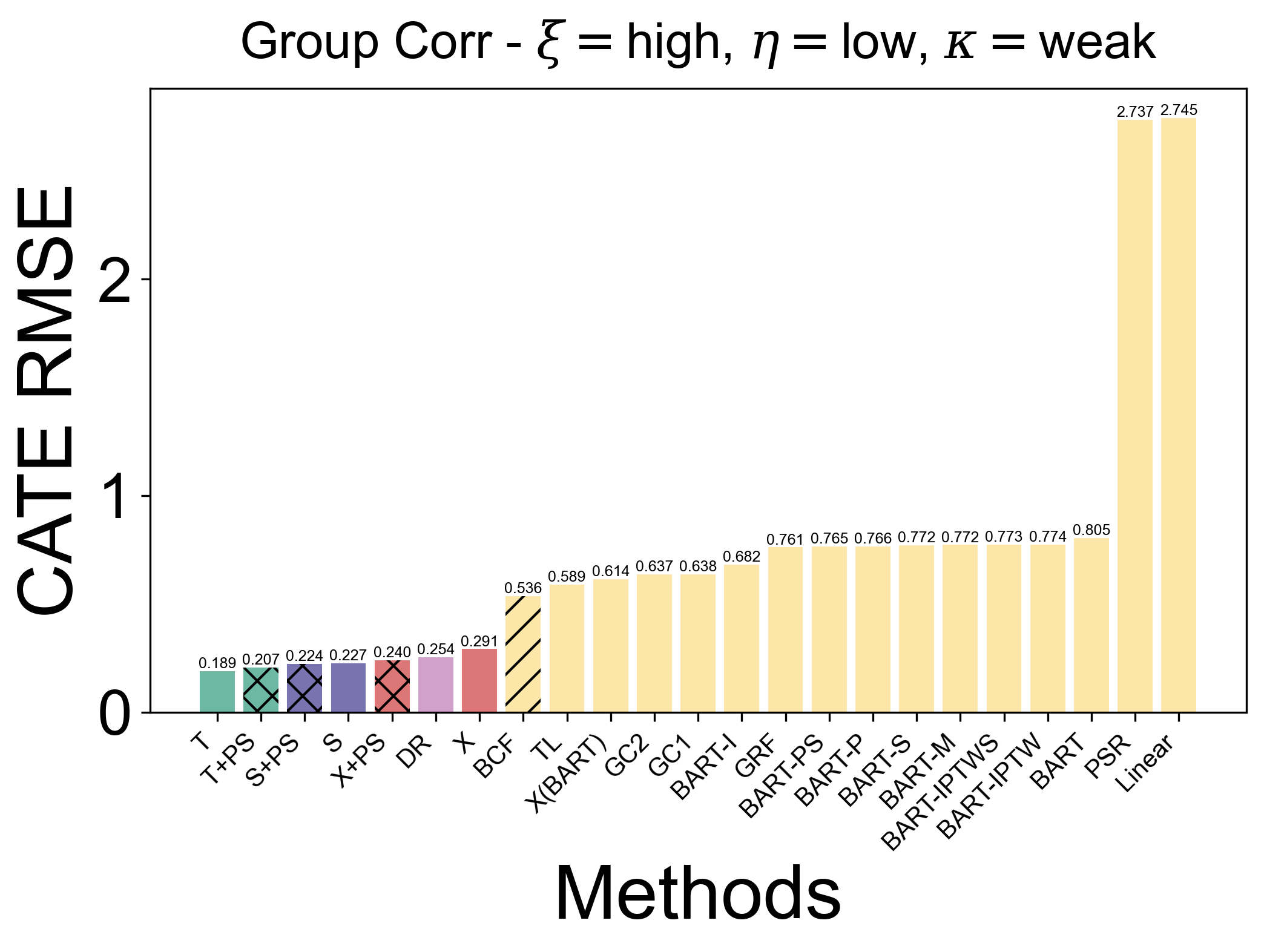}
    \includegraphics[height=0.17\linewidth]{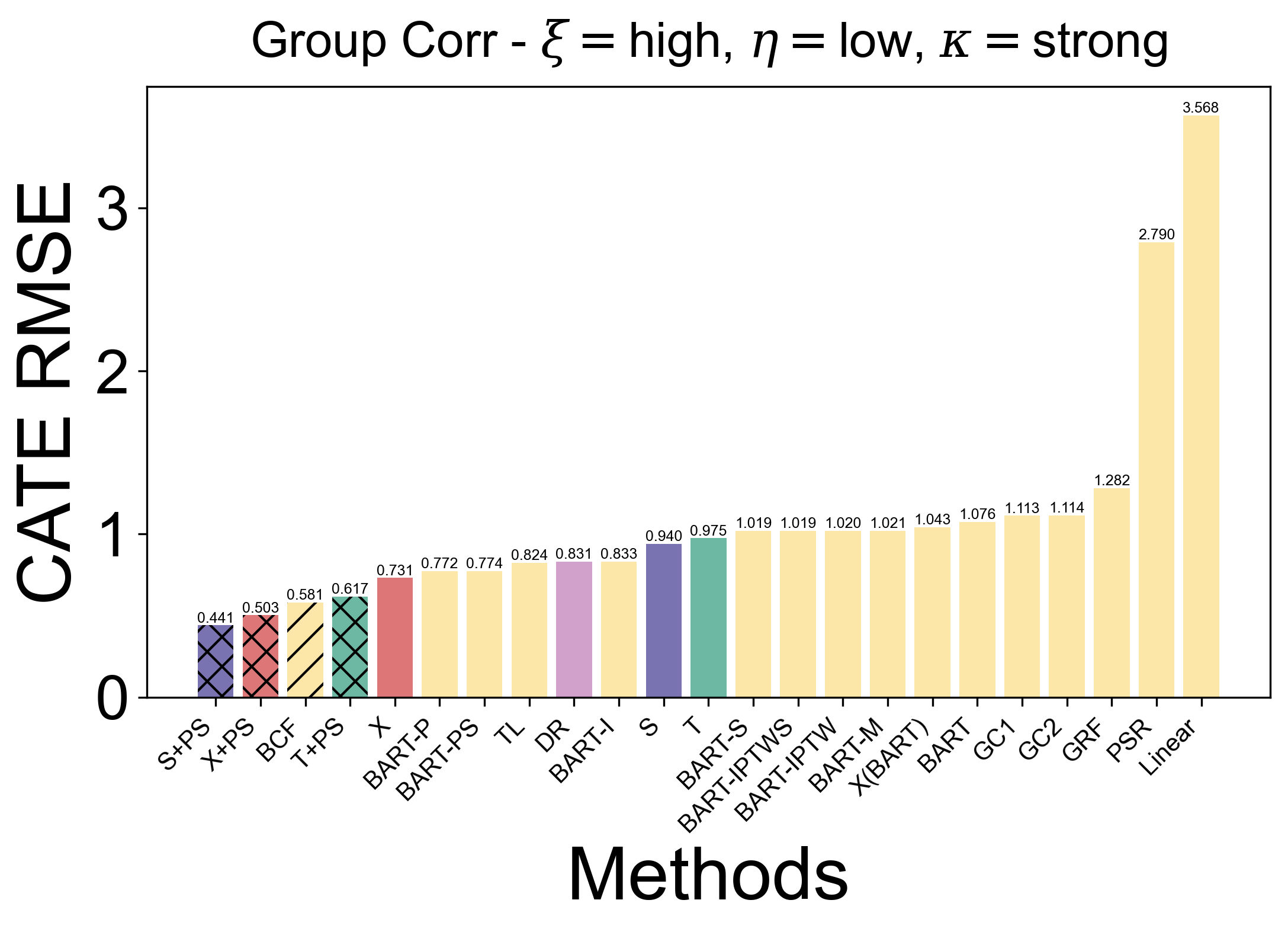}
    \includegraphics[height=0.17\linewidth]{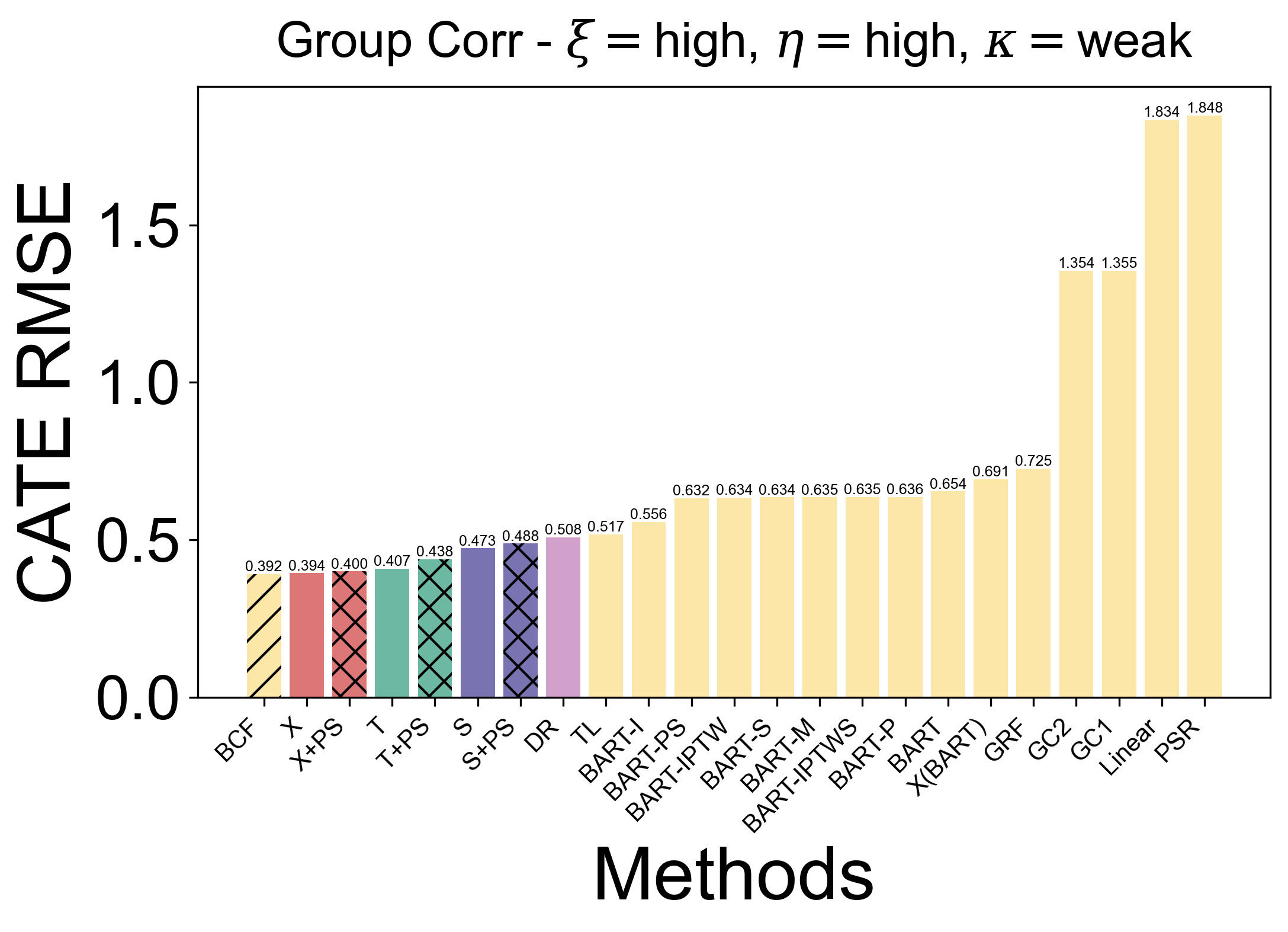}
    \includegraphics[height=0.17\linewidth]{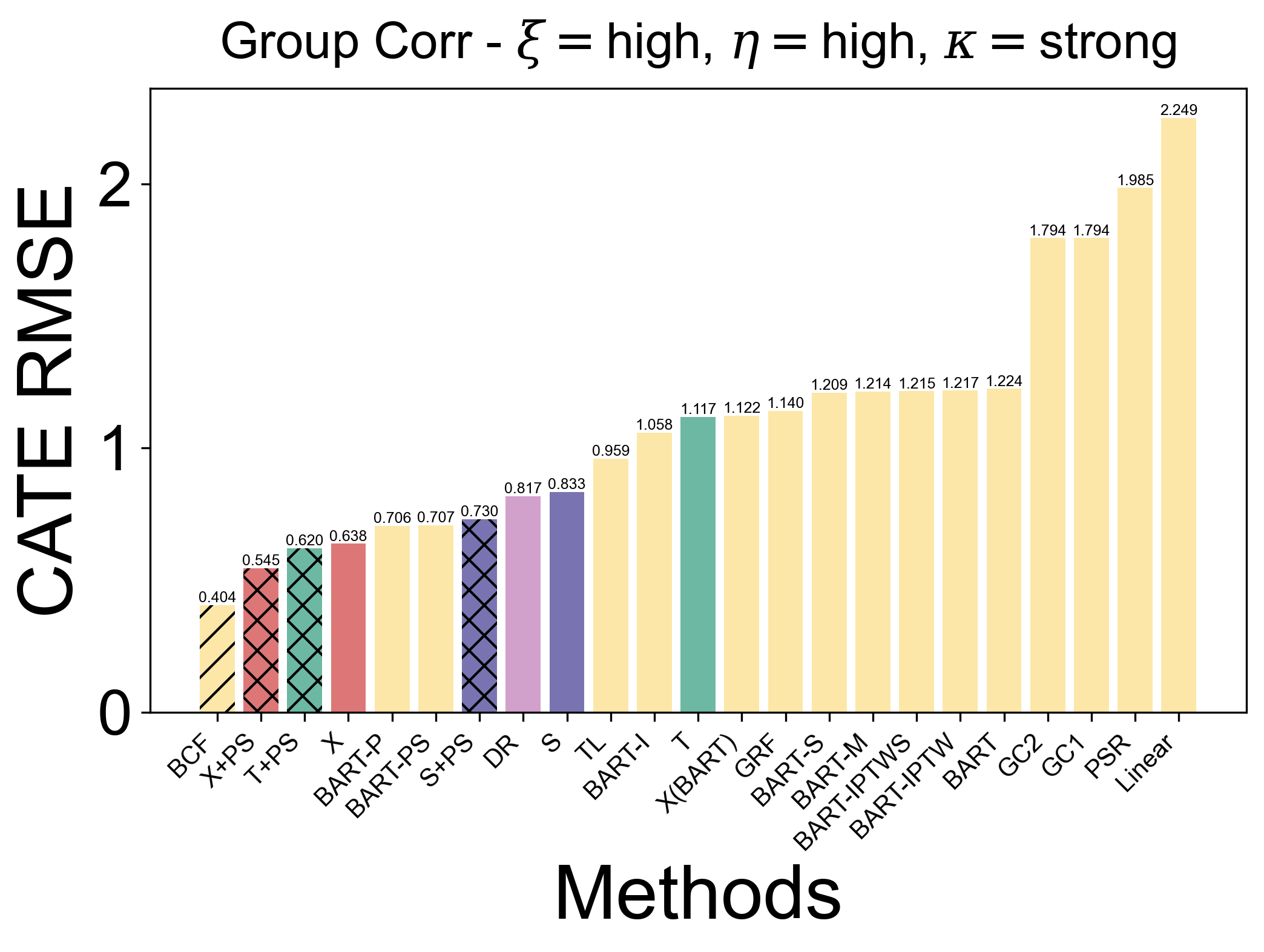}
    \caption{Performance comparison of TabPFN-based CATE estimators against state-of-the-art methods on the ACIC 2017 benchmark.
      Results are shown across panels organized by error type, effect magnitude ($\xi$), noise level ($\eta$), and selection strength ($\kappa$).
      Colored bars represent meta-learners built on TabPFN: \textcolor[HTML]{403990}{S-Learner}, \textcolor[HTML]{2F9B7D}{T-Learner}, \textcolor[HTML]{CF3D3E}{X-Learner}, \textcolor[HTML]{BF7AB5}{DR-Learner}, \textcolor[HTML]{7A6DB0}{S+PS-Learner}, \textcolor[HTML]{6FBFAA}{T+PS-Learner}, and \textcolor[HTML]{E27F7F}{X+PS-Learner}.
      The \textcolor[HTML]{FBDD85}{BCF} is shown for reference as the shaded yellow bar.}
    \label{fig:causal-acic-each-scenario}
\end{figure*}
The comparison of different methods across the 24 scenarios is presented in Figure~\ref{fig:causal-acic-each-scenario}.
TabPFN-based methods, especially the X+PS and X learners, show strong performance in almost all scenarios.

\section{Experimental details for study III}
\label{app:covariate-shift}

\subsection{Additional simulation results}

\begin{figure}[htbp]
    \centering
    \includegraphics[width=0.6\linewidth]{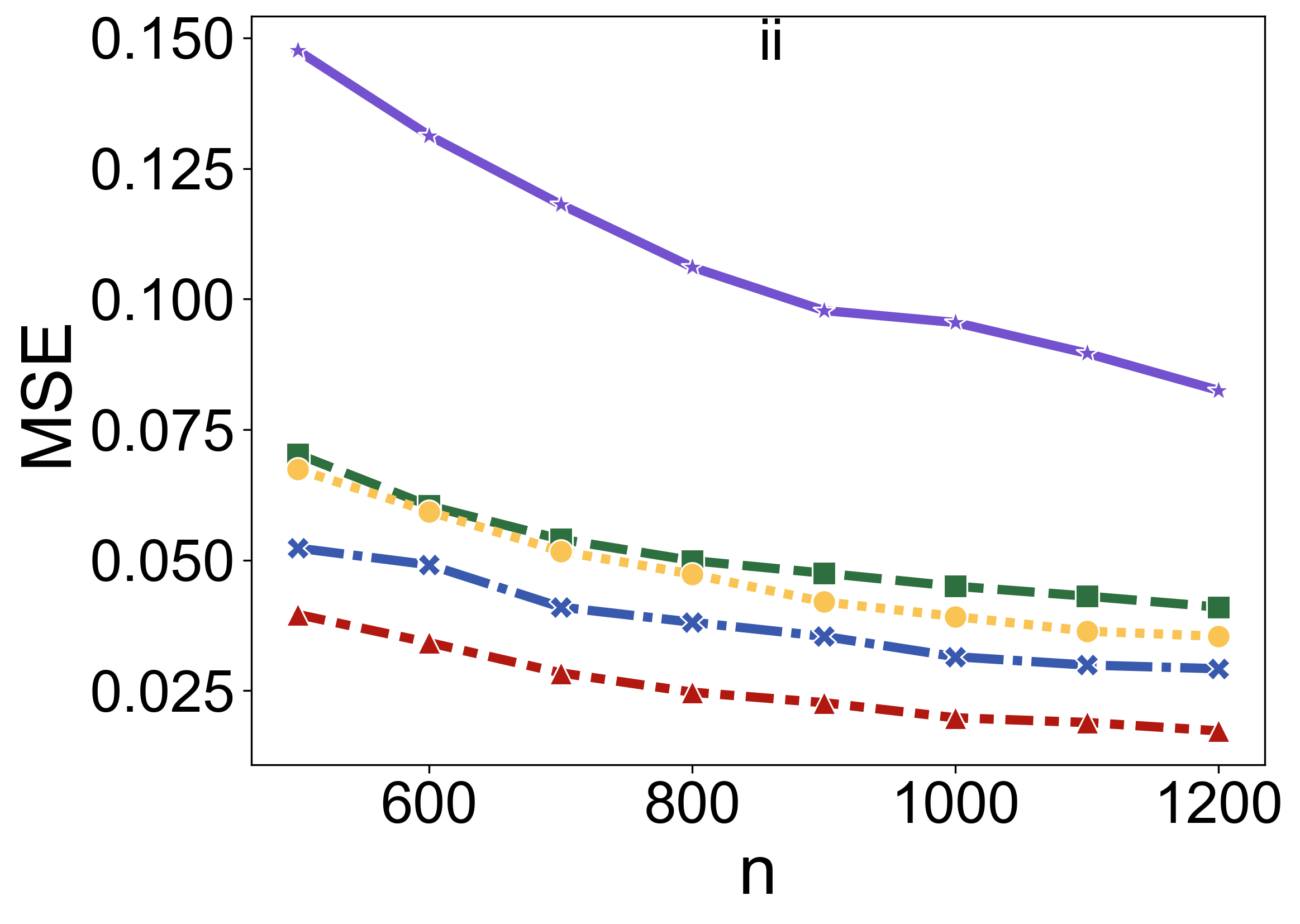}
    \caption{Covariate shift experiment under setting~(ii).}
    \label{fig:covariate-shift-setting2}
\end{figure}

Figure~\ref{fig:covariate-shift-setting2} shows the Setting~(ii) results, where the methods maintain their relative performance ordering observed in Setting~(i).

\subsection{Real data comparison}
\label{app:realdata_covshift}

\textbf{Datasets and preprocessing.}
We use two UCI regression datasets.
Airfoil Self-Noise contains 1,503 observations with five covariates: frequency, angle of attack, chord length, free-stream velocity, and suction-side displacement thickness.
The response is sound pressure level.
Concrete Compressive Strength contains 1,030 observations with eight covariates: cement, blast furnace slag, fly ash, water, superplasticizer, coarse aggregate, fine aggregate, and age.
The response is concrete compressive strength.

For each dataset, we randomly allocate $70\%$ of the observations to source training and $30\%$ to a test pool.
Each raw covariate is standardized using its mean and standard deviation in the source training set.
The same transformation is applied to the test pool.
No logarithmic transformations are applied.
Throughout this subsection, $x$ denotes the standardized covariate vector.

\textbf{Covariate shift.}
We resample the test pool with replacement, preserving its size, with sampling probabilities proportional to
\[
    \omega(x)=\exp(x^{\top}\beta).
\]
For Airfoil, the two shift vectors are
\[
    \beta_{1}=(-1,1,-1,1,-1)^{\top},
    \qquad
    \beta_{2}=(-1,0,0,0,1)^{\top}.
\]
For Concrete, they are
\[
    \beta_{1}=(-1,1,-1,1,-1,1,-1,1)^{\top},
    \qquad
    \beta_{2}=(-1,0,0,0,0,0,0,1)^{\top}.
\]
Each setting is evaluated over ten seeded repetitions.
Within each dataset, the two shift settings share the same source train--test splits, with split seeds 42 through 51.
Target resampling uses a separate deterministic seed for each dataset, shift setting, and repetition.
All methods are evaluated on exactly the same resampled target observations within a repetition.

\textbf{Methods and tuning.}
We compare TabPFN, naive XGBoost, importance-weighted XGBoost (IW), and radial basis function kernel ridge regression (KRR) with pseudo-labeling (PL).
TabPFN uses the original TabPFNv2 regression checkpoint with eight ensemble estimators and is fitted to the full source training set.

The naive and IW XGBoost models are tuned separately using five-fold cross-validation on the source training data.
Both use the same 32-candidate grid:
\[
\begin{aligned}
    \texttt{n\_estimators} &\in \{200,400\}, &
    \texttt{max\_depth} &\in \{3,5\},\\
    \texttt{learning\_rate} &\in \{0.05,0.1\}, &
    \texttt{subsample} &\in \{0.8,1.0\},\\
    \texttt{colsample\_bytree} &\in \{0.8,1.0\}. &&
\end{aligned}
\]
Naive XGBoost minimizes unweighted training loss and selects hyperparameters by ordinary out-of-fold MSE.
IW weights source training observations proportionally to $\exp(x^{\top}\beta)$.
Training weights are normalized to have mean one within each cross-validation training fold and in the final full-source fit; the weight ratios are preserved without clipping.
IW hyperparameters minimize the pooled, self-normalized importance-weighted out-of-fold squared error
\[
    \widehat R_{\mathrm{IW,CV}}
    =
    \frac{
        \sum_{i\in\mathcal D_s}
        \omega(x_i)
        \bigl(y_i-\widehat f^{(-k(i))}(x_i)\bigr)^2
    }{
        \sum_{i\in\mathcal D_s}\omega(x_i)
    },
\]
where $k(i)$ is the validation fold containing observation $i$, and $\widehat f^{(-k(i))}$ is fitted using the other folds.
Target test responses are not used to train or tune either XGBoost model.

For KRR, we retain the source-splitting, pseudo-labeling, and regularization grid of the real-data implementation, using an RBF kernel for the multivariate covariates.
We additionally report a \emph{KRR test-label oracle}, which selects its regularization parameter using the target test responses and is evaluated on those same observations.
This is an optimistic, test-informed benchmark, distinct from the synthetic-data oracle that uses the true conditional mean.

\textbf{Evaluation and uncertainty.}
For every method, prediction MSE is the ordinary, unweighted average of squared prediction errors on the resampled target test set.
Thus, all methods are evaluated using the same target-risk metric.
Table~\ref{tab:covariate-shift-real} reports the mean MSE over the ten repetitions.
The reported standard error is $\operatorname{SE} = s_{\mathrm{MSE}}/\sqrt{10}$, where $s_{\mathrm{MSE}}$ is the sample standard deviation of the ten repetition-level MSEs.
Because the repetitions reuse observations through overlapping splits, these standard errors summarize variation across splitting and resampling repetitions; they are not population-level confidence intervals.

\section{Comparison with Gaussian process regression}
\label{app:extrapolation}
To study the adaptiveness of TabPFN, we evaluate the interpolation and extrapolation behavior of TabPFN trained on data sampled from predefined functions, comparing its performance to Gaussian process regression (GPR) as a baseline.
Experiments are conducted in both one-dimensional (1D) and two-dimensional (2D) settings.

\subsection{Training data generation}
\textbf{1D case}.
The training set of size $n$ consists of features $x_i$ placed at equally spaced points over $[-1, 1]$.
Responses are generated as:
\[
y_i = f(x_i) + \epsilon_i, \quad \epsilon_i \stackrel{\text{i.i.d.}}{\sim} \mathcal{N}(0, 0.1^2),
\]
where $f$ is one of the following:
\begin{itemize}
    \item[(i)] Linear function: $f(x) = x $,
    \item[(ii)] Quadratic function: $f(x) = x^2$,
    \item[(iii)] Step function: $f(x) = \mathbbm{1}(x\geq 0)$,
    \item[(iv)] The piecewise-linear function joining $(-1,0)$, $(-0.6,1)$, $(0.2,-0.5)$, $(0.6,0.7)$, and $(1,0)$.
\end{itemize}
TabPFN is trained on this small dataset and prediction is made over the extended interval $[-4.5, 4.5]$ to assess both interpolation ($x \in [-1, 1]$) and extrapolation ($x \notin [-1, 1]$) performance.

\textbf{2D case}.
The training features form an equally spaced $n\times n$ mesh grid over $[0,1]^2$, resulting in $n^2$ observations.
We examine $n=11$ and $n=31$.
The responses are generated as:
\[
y_i = f(x_{i1}, x_{i2}) + \epsilon_i, \quad \epsilon_i \stackrel{\text{i.i.d.}}{\sim} \gN(0, 0.05^2),
\]
where $f$ is analogously defined as in the 1D case:
\begin{itemize}
\item[(i)] Linear function: $f(x_1, x_2) = x_1 + x_2 $,
\item[(ii)] Quadratic function: $f(x_1, x_2) = (x_1-0.5)^2 +(x_2-0.5)^2$,
\item[(iii)] Step function:
\[
f(x_1,x_2) =
\begin{cases}
0, & \text{if } x_1 < 0.5 \text{ and } x_2<0.5,\\
1, & \text{otherwise}.
\end{cases}
\]
\item[(iv)] A piecewise-bilinear function obtained by bilinear interpolation through a fixed $4\times4$ grid of function values generated from a uniform distribution using seed $1$.
\end{itemize}
TabPFN is trained on this small dataset and prediction is made over a uniform mesh grid on $[-0.5,1.5]^2$ to assess both interpolation ($x \in [0,1]^2$) and extrapolation ($x \notin [0,1]^2$) performance.

\subsection{Baseline for comparison: Gaussian process regression}
For both 1D and 2D cases, we compare TabPFN with Gaussian process regression (GPR).
For Gaussian process regression, we consider $5$ kernels: constant kernel $\times$ RBF kernel, constant kernel $\times$ Matern kernel, constant kernel $\times$ RationalQuadratic kernel, constant kernel $\times$ ExpSineSquared kernel, constant kernel $\times$ RBF kernel $+$ constant kernel $\times$ Matern kernel.
The measurement-error standard deviation is selected from the grid $[0.05, 0.1, 0.15, 0.2]$.
We maximize the marginal log-likelihood to select the kernel, its parameters, and the measurement-error standard deviation.
All GPR predictions and predictive intervals are computed using this selected fit.
The GPR is implemented using the \emph{GaussianProcessRegressor} in \texttt{scikit-learn} package in Python.

\subsection{Interpolation and extrapolation}
\begin{figure*}[htbp]
    \centering
    \includegraphics[width=0.24\linewidth]{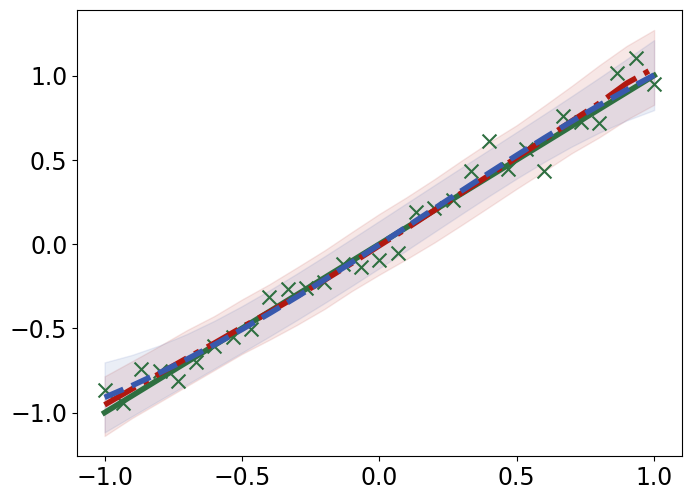}
    \includegraphics[width=0.24\linewidth]{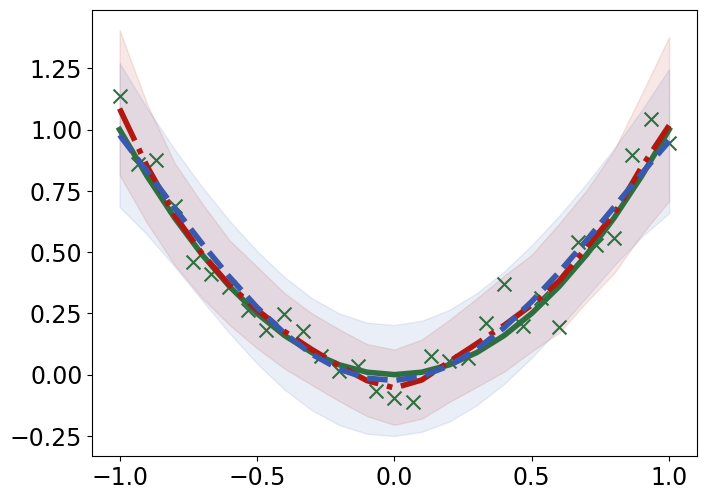}
    \includegraphics[width=0.24\linewidth]{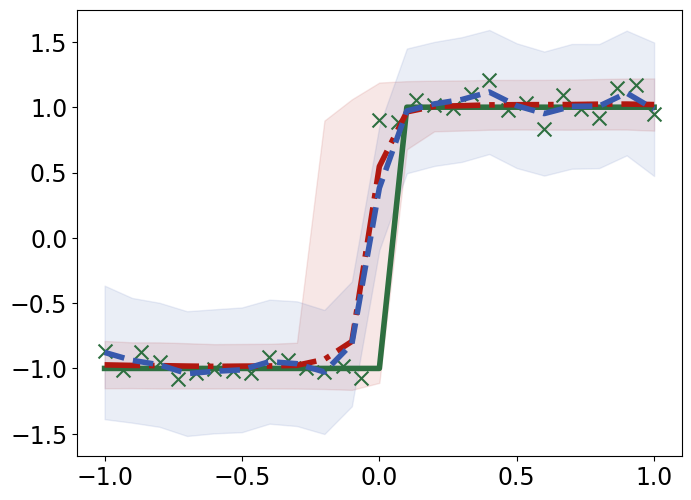}
    \includegraphics[width=0.24\linewidth]{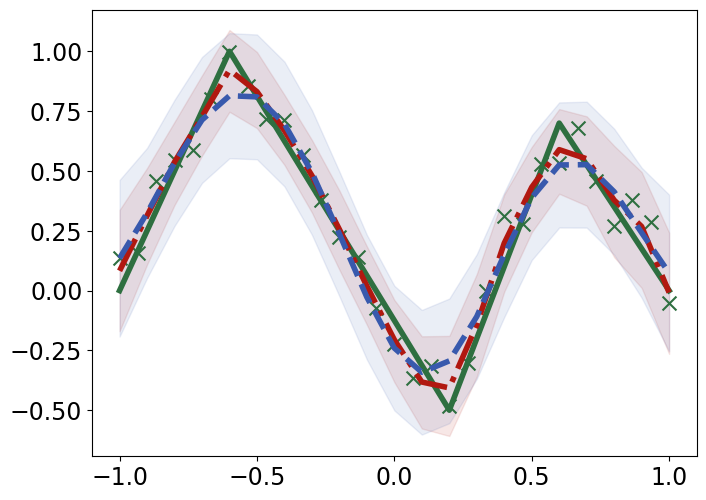}
    \includegraphics[width=0.24\linewidth]{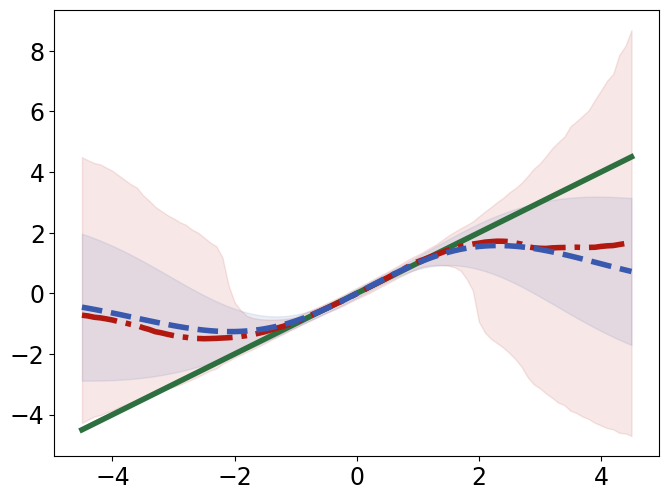}
    \includegraphics[width=0.24\linewidth]{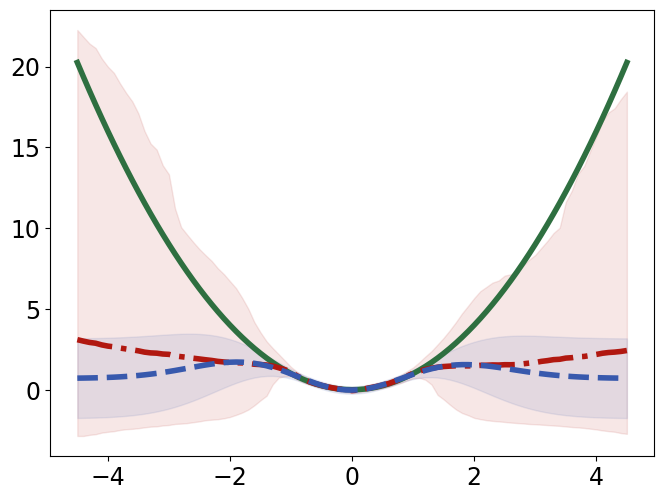}
    \includegraphics[width=0.24\linewidth]{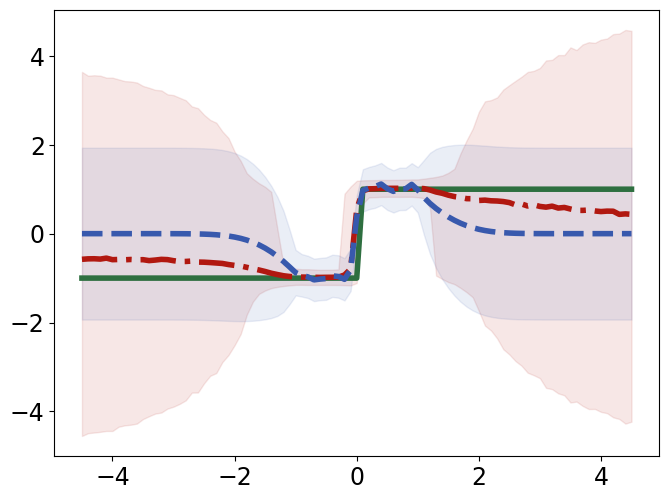}
    \includegraphics[width=0.24\linewidth]{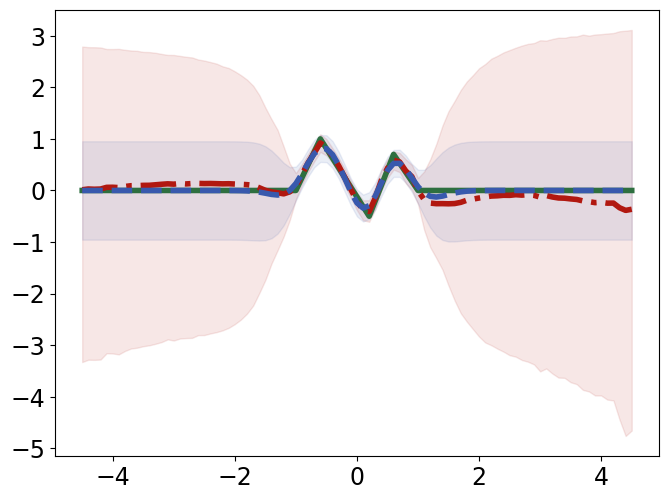}
    \caption{Interpolation (top) and extrapolation (bottom) performance of TabPFN (red) and Gaussian process regression (GPR; blue) regression on synthetic 1D functions, compared to ground truth (green).
      Columns correspond to different true functions: (i) linear, (ii) quadratic, (iii) step, and (iv) piecewise-linear.
      TabPFN shows the predicted mean (dash dotted line) with 95\% prediction intervals (2.5\%–97.5\% quantiles; shaded region), while GPR displays the mean $\pm$ 1.96 SD (dashed lines).
      SD is the standard-deviation of the predictive distribution.}
    \label{fig:extrapolation}
\end{figure*}

\begin{figure*}[ht]
    \centering
    \includegraphics[width=0.24\linewidth]{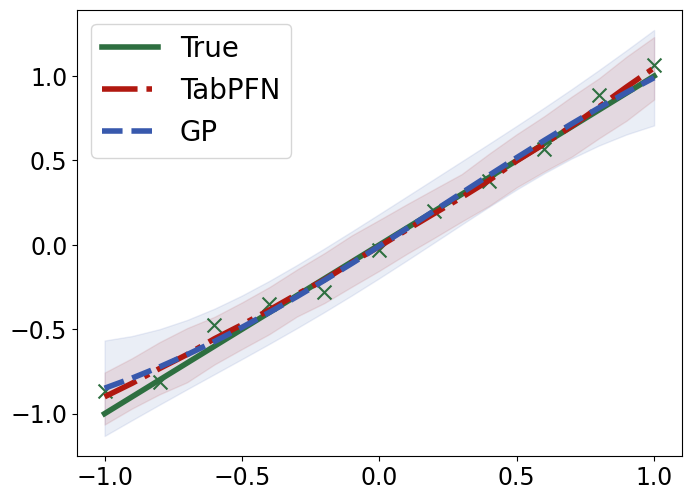}
    \includegraphics[width=0.24\linewidth]{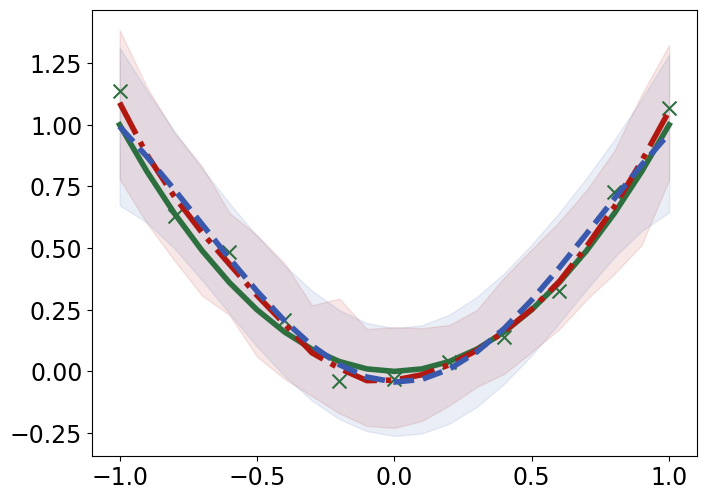}
    \includegraphics[width=0.24\linewidth]{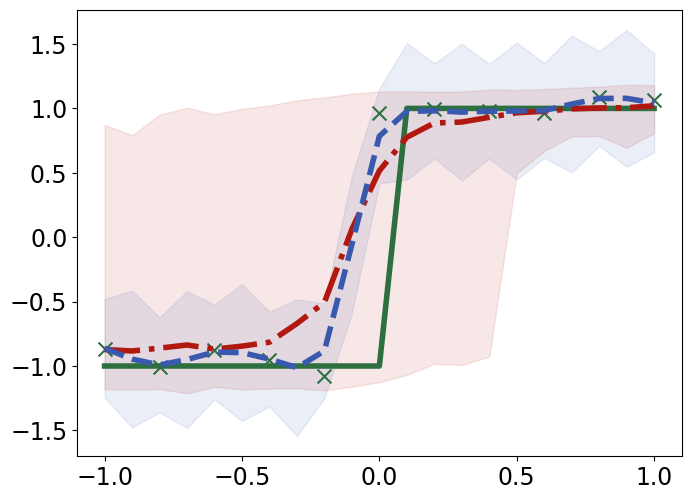}
    \includegraphics[width=0.24\linewidth]{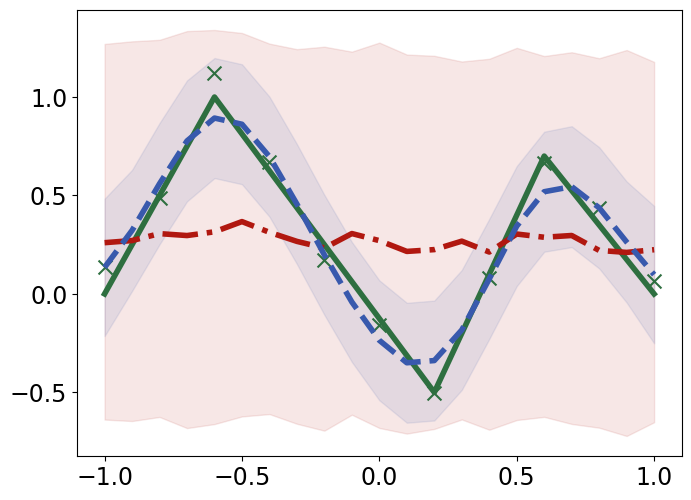}
    \includegraphics[width=0.24\linewidth]{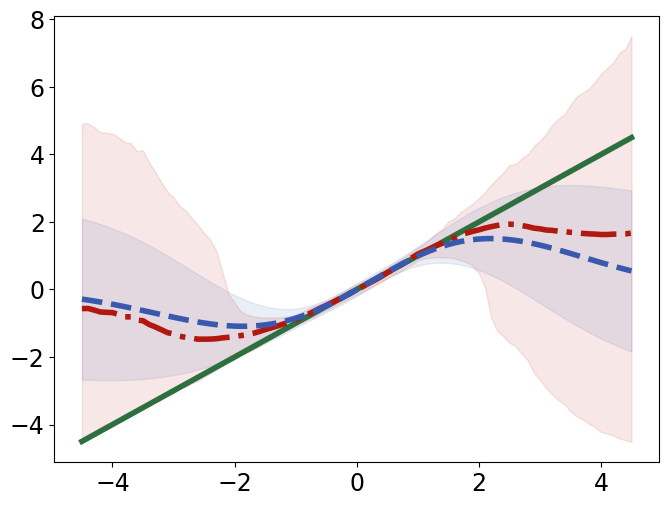}
    \includegraphics[width=0.24\linewidth]{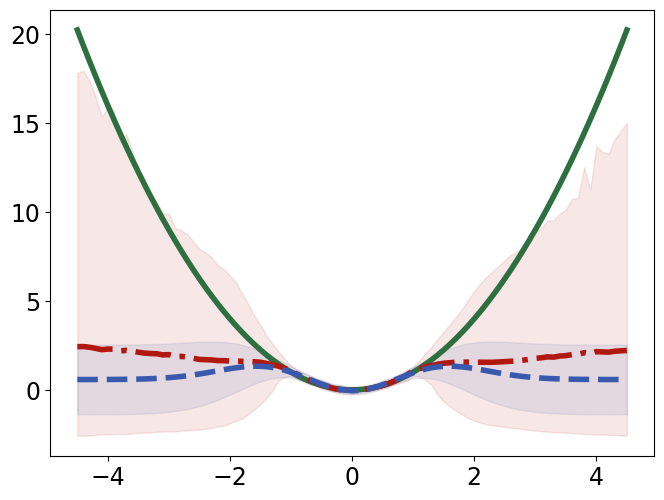}
    \includegraphics[width=0.24\linewidth]{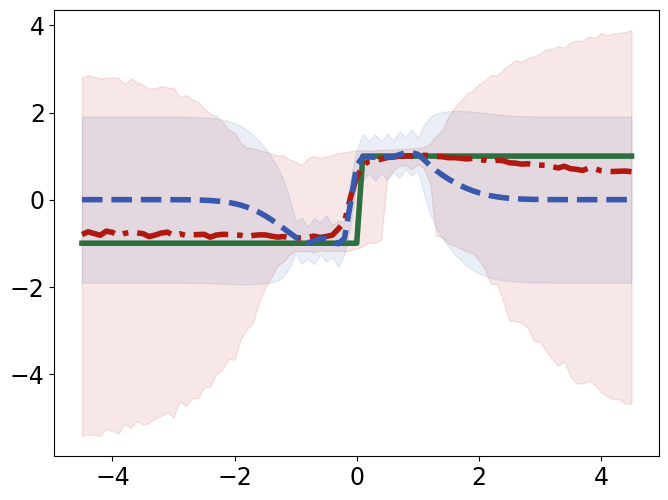}
    \includegraphics[width=0.24\linewidth]{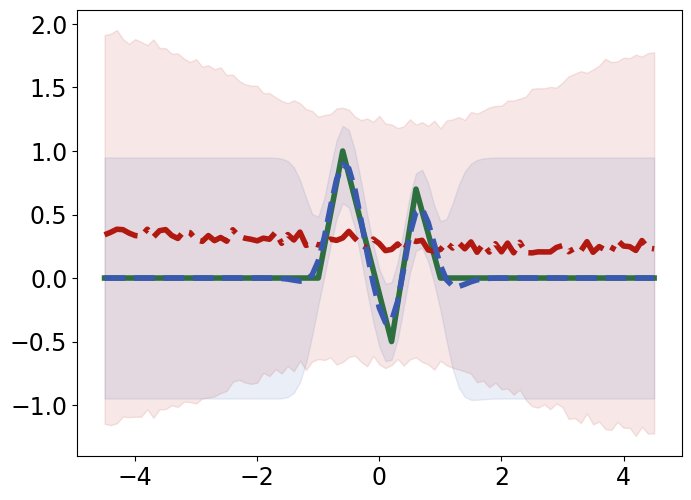}
    \caption{Interpolation (top) and extrapolation (bottom) performance of TabPFN (red) and Gaussian process regression (GPR; blue) regression on synthetic 1D functions, compared to ground truth (green) when $n=11$.
      Columns correspond to different true functions: (i) linear, (ii) quadratic, (iii) step, and (iv) piecewise-linear.
      TabPFN shows the predicted mean (dash dotted line) with 95\% prediction intervals (2.5\%–97.5\% quantiles; shaded red region), while GPR displays the mean $\pm$ 1.96 SD (shaded blue region).
      SD is the standard-deviation of the predictive distribution.}
    \label{fig:extrapolation_1d_n11}
\end{figure*}

\begin{figure*}[ht]
    \centering
    \includegraphics[width=0.24\linewidth]{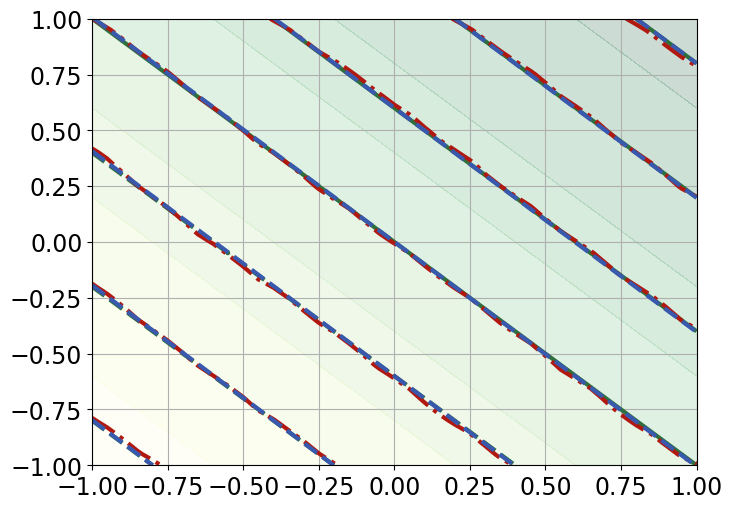}
    \includegraphics[width=0.24\linewidth]{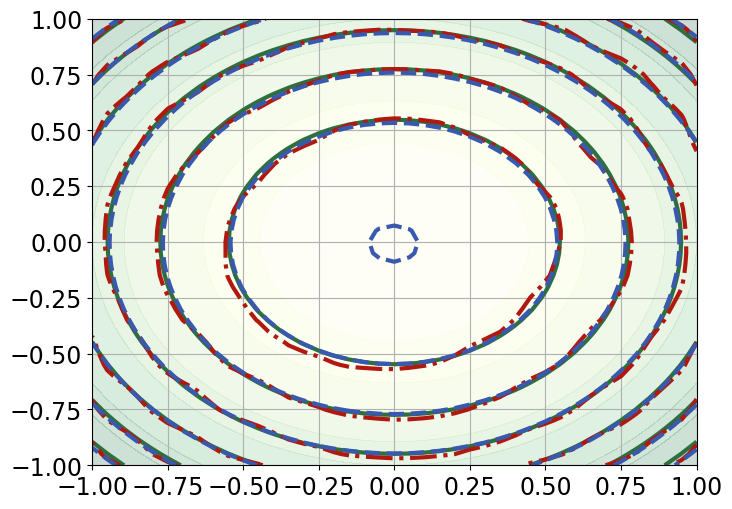}
    \includegraphics[width=0.24\linewidth]{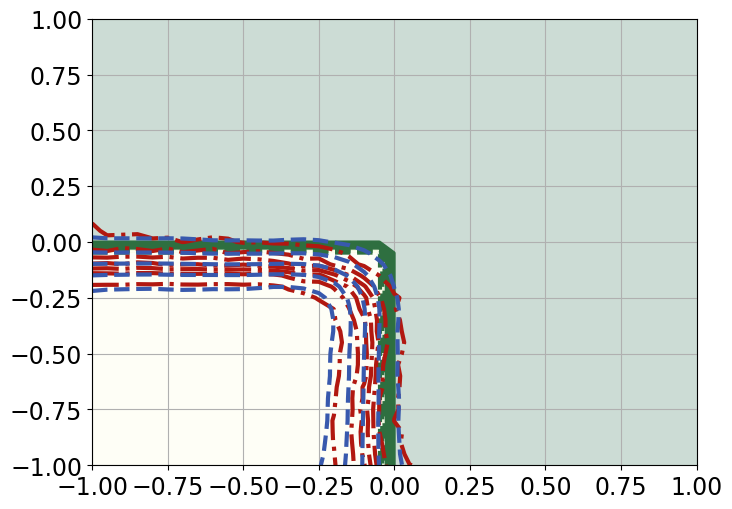}
    \includegraphics[width=0.24\linewidth]{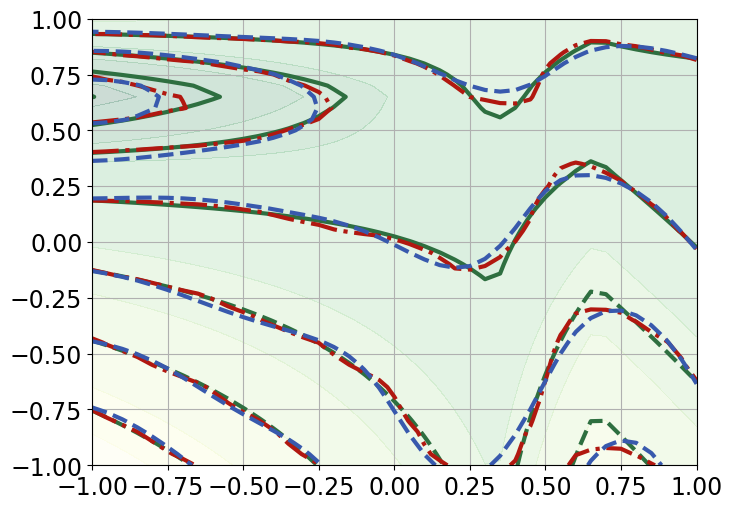}
    \includegraphics[width=0.24\linewidth]{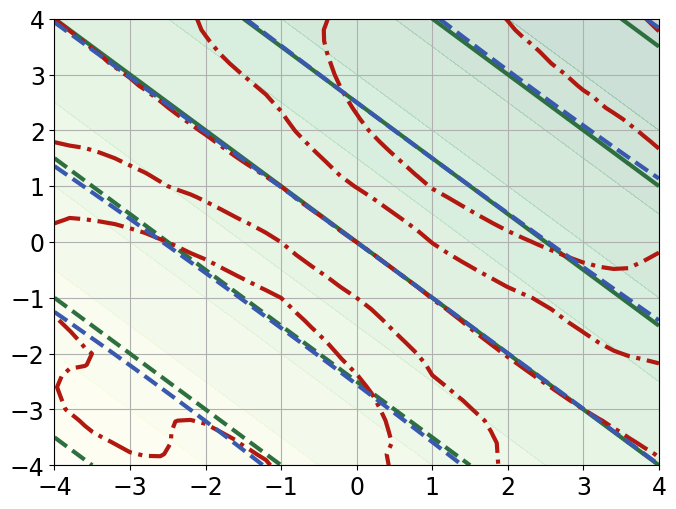}
    \includegraphics[width=0.24\linewidth]{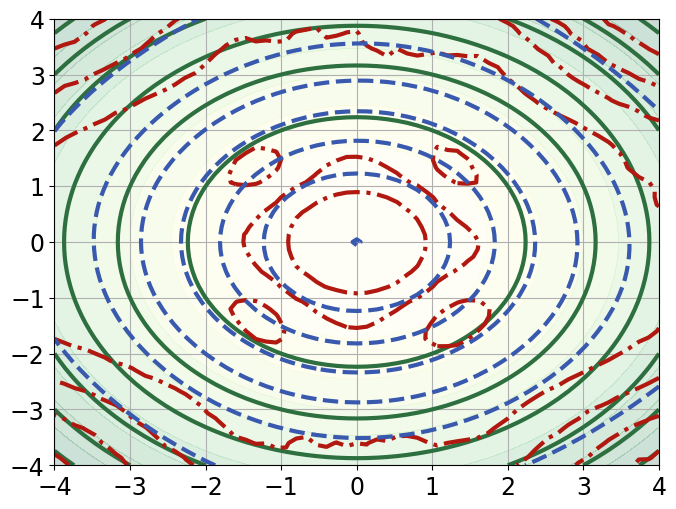}
    \includegraphics[width=0.24\linewidth]{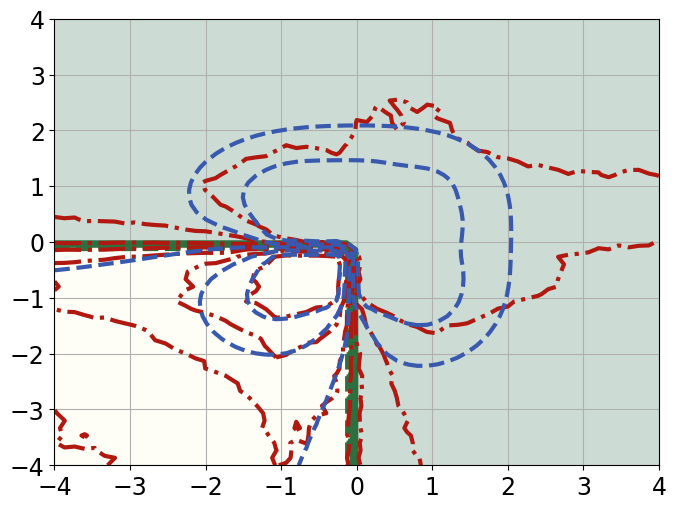}
    \includegraphics[width=0.24\linewidth]{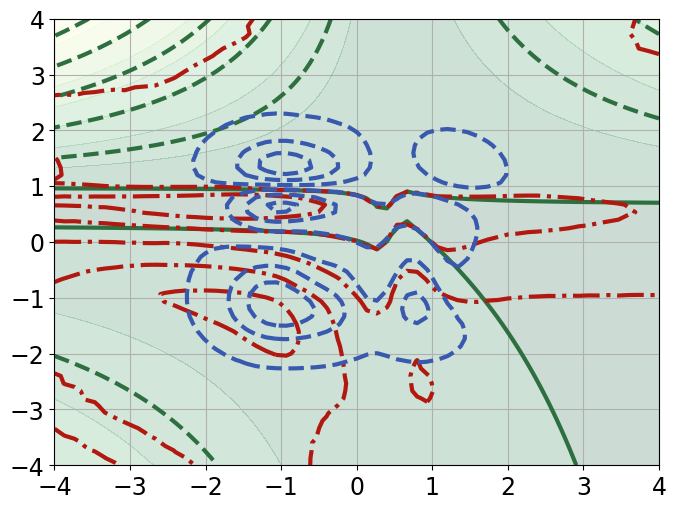}\\
    \includegraphics[width=0.24\linewidth]{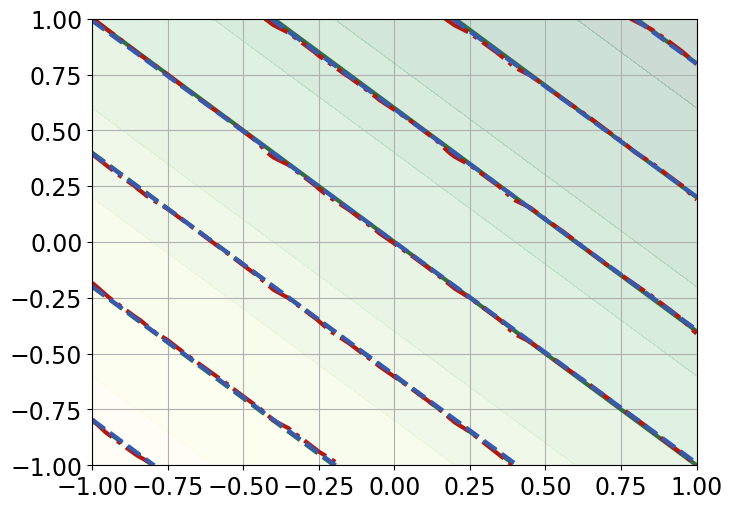}
    \includegraphics[width=0.24\linewidth]{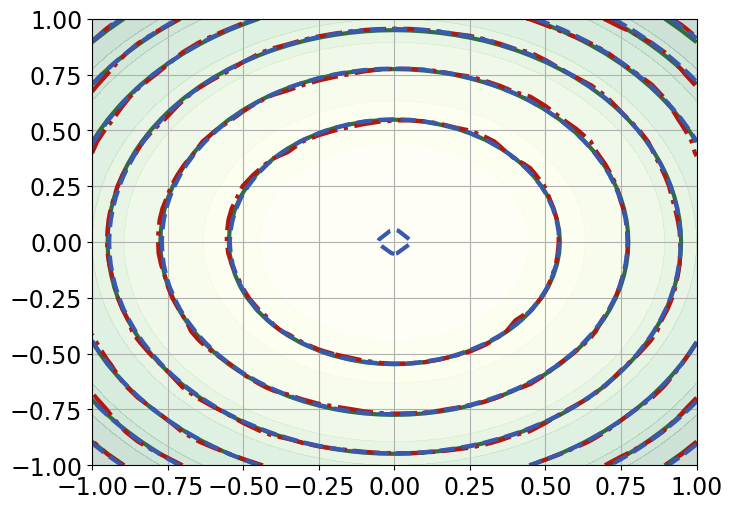}
    \includegraphics[width=0.24\linewidth]{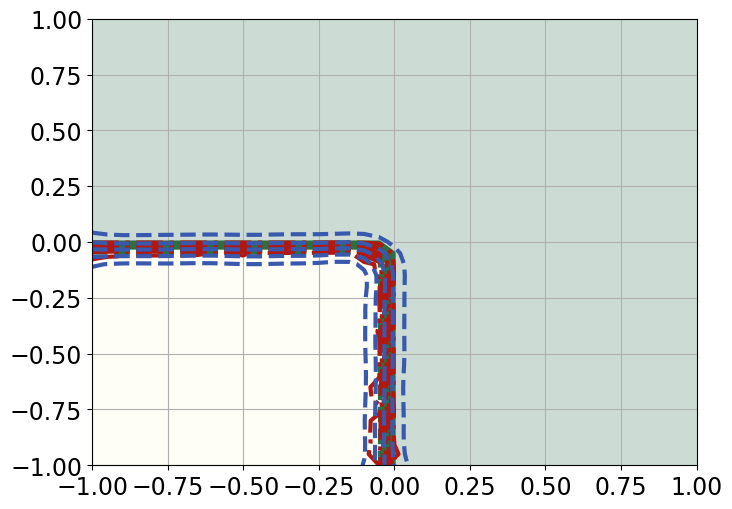}
    \includegraphics[width=0.24\linewidth]{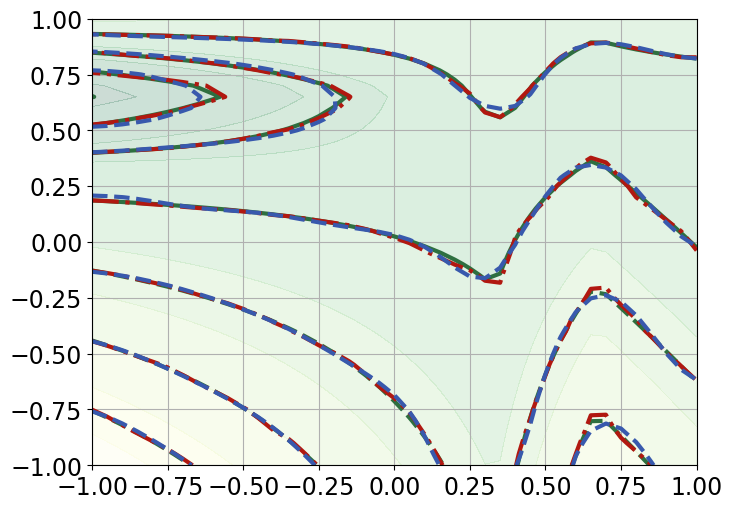}
    \includegraphics[width=0.24\linewidth]{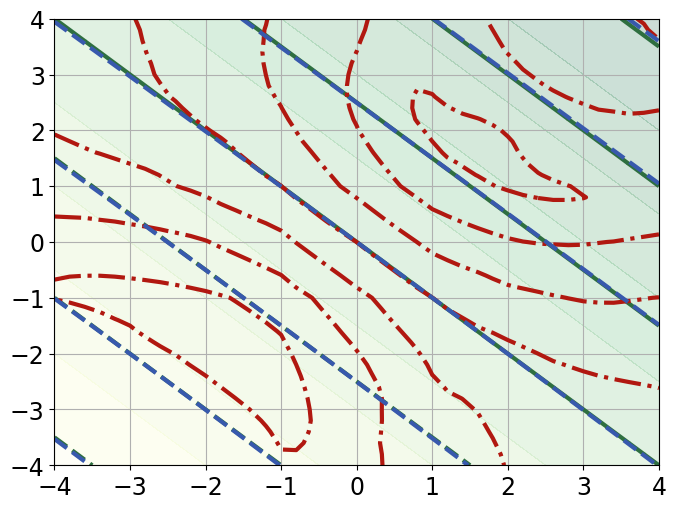}
    \includegraphics[width=0.24\linewidth]{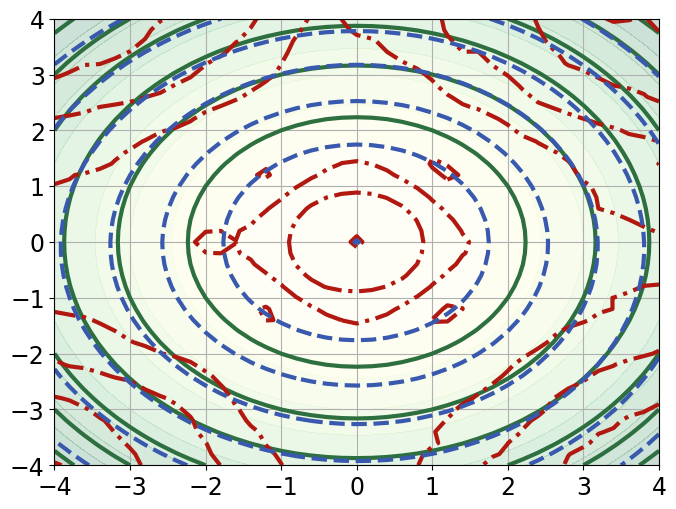}
    \includegraphics[width=0.24\linewidth]{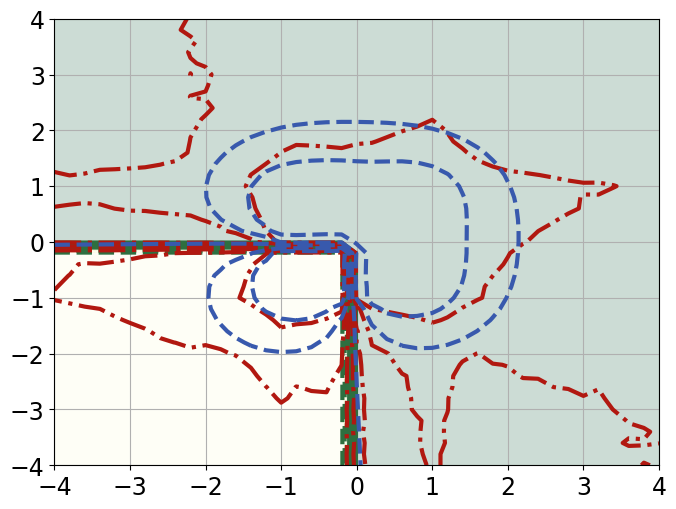}
    \includegraphics[width=0.24\linewidth]{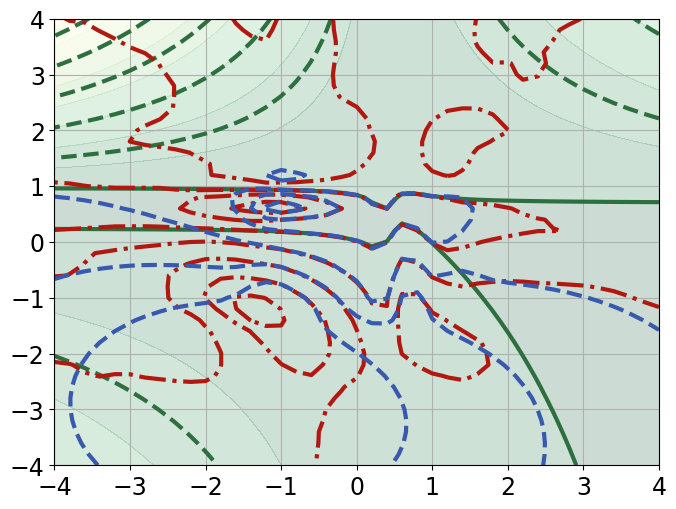}
    \caption{Performance comparison of TabPFN (red) and Gaussian Process (GP; blue) regression on synthetic 2D functions, evaluated against ground truth (green) using contour plots.
      Columns represent different true functions: (i) linear, (ii) quadratic, (iii) step, and (iv) piecewise-linear.
      Rows 1 \& 2: Interpolation and extrapolation performance with a training set size of $n=11^2$.
      Rows 3 \& 4: Interpolation and extrapolation performance with a larger training set $n=31^2$.}
    \label{fig:extrapolation_2d}
\end{figure*}

The results for $n=31$ training data points are shown in Figure \ref{fig:extrapolation}.
The results when $n=11$ under 1D case are visualized in Figure~\ref{fig:extrapolation_1d_n11}.
The results under 2D case are visualized in Figure~\ref{fig:extrapolation_2d}.

We observe that in all cases, TabPFN fits a smoothly-varying curve that broadly resembles the GPR curve within the interpolation region.
The notable differences are:
\begin{itemize}
    \item Unlike GPR, TabPFN seems not to suffer from the Gibbs phenomenon (fluctuations at the point of discontinuity of a step function).
    \item TabPFN extrapolates at roughly the same level of the last observed response, whereas GPR quickly asymptotes to zero (the GPR prior mean).
    \item TabPFN is more conservative than GPR at fitting a quickly-varying function.
    GPR is able to roughly approximate the piecewise-linear function with $n=11$, whereas TabPFN still outputs a constant model.
    \item TabPFN has a more diffuse conditional predictive distribution $p(y|x)$ compared to GPR, especially away from the interpolation region.
\end{itemize}

\subsection{Uncertainty quantification}
\label{app:UQ}
\subsubsection{Synthetic datasets}
To assess the uncertainty quantification capability of TabPFN in comparison with GPR, we evaluate the predictive intervals of both methods in terms of their coverage ratio and interval length.
For each predefined function, we generate $100$ independent datasets following the data generation procedure described above.
For each dataset, the model is trained on the corresponding training set and evaluated on a dense grid with a step size of $0.1$ over the range $[-4.5, 4.5]$, which encompasses both the interpolation region ($[-1, 1]$) and the extrapolation region ($[-4.5, 4.5] \setminus [-1, 1]$).
At each grid point, we record the predictive mean and the associated $95\%$ prediction interval produced by the model.
For TabPFN, the $95\%$ prediction interval is obtained from the empirical $2.5\%$ and $97.5\%$ quantiles of the posterior predictive distribution.
The coverage ratio is computed as the proportion of independently generated noisy responses that fall within the predicted intervals, averaged over all test points and repetitions.
Similarly, the interval length is computed as the mean width of the predictive intervals, also averaged across all test points and repetitions.
This evaluation jointly examines the calibration (via coverage ratio) and sharpness (via interval length) of the uncertainty estimates provided by TabPFN and GPR in both interpolation and extrapolation settings.
The average coverage ratios and interval lengths for interpolation and extrapolation are reported in Table~\ref{tab:interpolation_uq} and Table~\ref{tab:extrapolation_uq}, respectively.
\begin{table*}[tp]
\caption{Average coverage ratio and interval length of TabPFN and Gaussian process regression (GPR) under the interpolation setting ($x \in [-1, 1]$), averaged over 100 repetitions.}
\label{tab:interpolation_uq}
\resizebox{0.8\linewidth}{!}{
\begin{tabular}{lccccccccc}
\toprule
\multirow{2}{*}{$n$}& \multirow{2}{*}{Method}& \multicolumn{2}{c}{Linear} & \multicolumn{2}{c}{Quadratic} & \multicolumn{2}{c}{Step} & \multicolumn{2}{c}{P Linear} \\
\cmidrule{3-10}
& & Coverage      & Length     & Coverage       & Length       & Coverage     & Length    & Coverage       & Length      \\
\midrule
\multirow{2}{*}{11}& TabPFN & 0.9350 & 0.4656 & 0.9525 & 0.7365 & 0.9705 & 1.2301 & 0.9930 & 1.8449 \\
& GPR & 0.7715 & 0.3182 & 0.7235 & 0.3945 & 0.8005 & 0.5933 & 0.7090 & 0.3787 \\
\multirow{2}{*}{31}& TabPFN & 0.9405 & 0.3924 & 0.9420 & 0.4235 & 0.9365 & 0.5501 & 0.9375 & 0.4788 \\
& GPR & 0.8205 & 0.3266 & 0.8590 & 0.3912 & 0.8770 & 0.5190 & 0.9045 & 0.4279 \\
\bottomrule
\end{tabular}}
\end{table*}

\begin{table*}[tp]
\caption{Average coverage ratio and interval length of TabPFN and Gaussian process regression (GPR) under the extrapolation setting ($x \notin [-1, 1]$), averaged over 100 repetitions.}
\label{tab:extrapolation_uq}
\resizebox{0.8\linewidth}{!}{
\begin{tabular}{lccccccccc}
\toprule
\multirow{2}{*}{$n$}& \multirow{2}{*}{Method}& \multicolumn{2}{c}{Linear} & \multicolumn{2}{c}{Quadratic} & \multicolumn{2}{c}{Step} & \multicolumn{2}{c}{P Linear} \\
\cmidrule{3-10}
& & Coverage      & Length     & Coverage       & Length       & Coverage     & Length    & Coverage       & Length      \\
\midrule
\multirow{2}{*}{11}& TabPFN & 0.9276 & 5.1478 & 0.4287 & 8.4742 & 0.9996 & 3.3699 & 1.0000 & 2.4744 \\
& GPR & 0.5599 & 2.3820 & 0.0421 & 0.8183 & 0.9431 & 2.3125 & 0.2983 & 0.4443 \\
\multirow{2}{*}{31}& TabPFN & 0.9827 & 5.9242 & 0.6817 & 11.9492 & 0.9975 & 3.8227 & 0.9993 & 4.9580 \\
& GPR & 0.7192 & 1.9608 & 0.0813 & 1.9430 & 0.9980 & 2.4185 & 0.5327 & 0.8643 \\
\bottomrule
\end{tabular}}
\end{table*}

\textbf{Interpolation performance} (Table \ref{tab:interpolation_uq}).
In the interpolation region ($x \in [-1, 1]$), TabPFN consistently exhibits coverage at around the nominal 95\% level across all function types, indicating that it remains well-calibrated within the training distribution.
In contrast, GPR undercovers for each function, with coverage between $0.71$ and $0.80$ when $n=11$ and between $0.82$ and $0.90$ when $n=31$.
TabPFN generally produces wider intervals than GPR, indicating more conservative uncertainty estimates that achieve coverage closer to the nominal level.

\textbf{Extrapolation performance}(Table \ref{tab:extrapolation_uq}).
In contrast, the extrapolation region ($x \notin [-1, 1]$) reveals a clear divergence between the two models.
TabPFN maintains substantially higher coverage than GPR in nearly all cases.
For smooth functions, GPR's coverage remains below $0.72$ for the linear function and below $0.09$ for the quadratic function, indicating overconfidence and poor uncertainty calibration outside the training range.
TabPFN, however, achieves much higher coverage (0.93–0.98 for linear; 0.43–0.68 for quadratic) and adapts its interval width dynamically to reflect function complexity—producing wider intervals in regions of greater uncertainty (e.g., the quadratic case).
For the step and piecewise-linear functions, TabPFN attains essentially full coverage, whereas GPR coverage ranges from $0.30$ to $1.00$ across the four settings.
Overall, TabPFN exhibits markedly better-calibrated uncertainty estimates than GPR under extrapolation, effectively avoiding the overconfident predictions that GPR commonly produces outside its training support.

\textbf{Summary}.
These results demonstrate that TabPFN provides better-calibrated uncertainty quantification than GPR under both interpolation and extrapolation, while adapting its interval width in regions far from the training data.

\subsubsection{Real datasets}
\textbf{Experimental setup.}
To complement the synthetic calibration analysis, we evaluate the uncertainty quantification of TabPFN on five real regression benchmarks drawn from the same suite used in Section~\ref{sec:real_data_semi}: kin8nm (OpenML ID~189), house\_8L (OpenML ID~218), puma8NH (OpenML ID~225), BNG(stock) (OpenML ID~1200), and bank8FM (OpenML ID~572).
For each dataset we subsample 1{,}250 observations uniformly at random so that Gaussian process fitting remains tractable, split them 80/20 into training and test sets, and repeat this procedure over 20 independent random seeds.
Both features and the response are standardised using training-set statistics so that interval widths and CRPS scores are comparable across datasets.

We compare TabPFN against GPR with the noise variance and kernel selected jointly by maximising the log-marginal likelihood over three kernel families (RBF, Mat\'{e}rn-$3/2$, Mat\'{e}rn-$5/2$) and four candidate measurement-error standard deviations on the grid $[0.05, 0.1, 0.15, 0.2]$.
We assess calibration through three complementary tools:
(i)~a \emph{reliability diagram} (calibration plot) comparing nominal and empirical coverage at six levels ($50\%$--$95\%$);
(ii)~\emph{PIT histograms}, which check whether the probability integral transform values $\hat{u}_i = \hat{F}_i(y_i)$ are uniformly distributed, as required for a calibrated predictive distribution \citep{gneiting2007probabilistic}; and (iii)~a scalar summary table reporting coverage at 95\%, mean interval width at 95\%, CRPS (approximated via quantile scoring on a uniform grid of 19 levels), and the Kolmogorov--Smirnov statistic against the uniform distribution (KS-PIT; smaller is better).
For both methods, we compute PIT values by linearly interpolating the predictive CDF over the estimated quantiles on the probability grid from $0.025$ to $0.975$; observations outside this grid are assigned PIT values of $0$ or $1$, respectively.
All figures aggregate by first averaging each metric over the 20 repetitions within each dataset, then averaging (and computing the standard deviation) across the five datasets.

\begin{figure*}[tp]
  \centering
  \subfloat[Reliability diagram.
    Each point is the mean empirical coverage averaged first over 20 repetitions within each dataset and then over datasets; error bars are $\pm 1$ SD across datasets.
    The dashed line is the ideal diagonal.]{
        \includegraphics[width=0.3\linewidth]{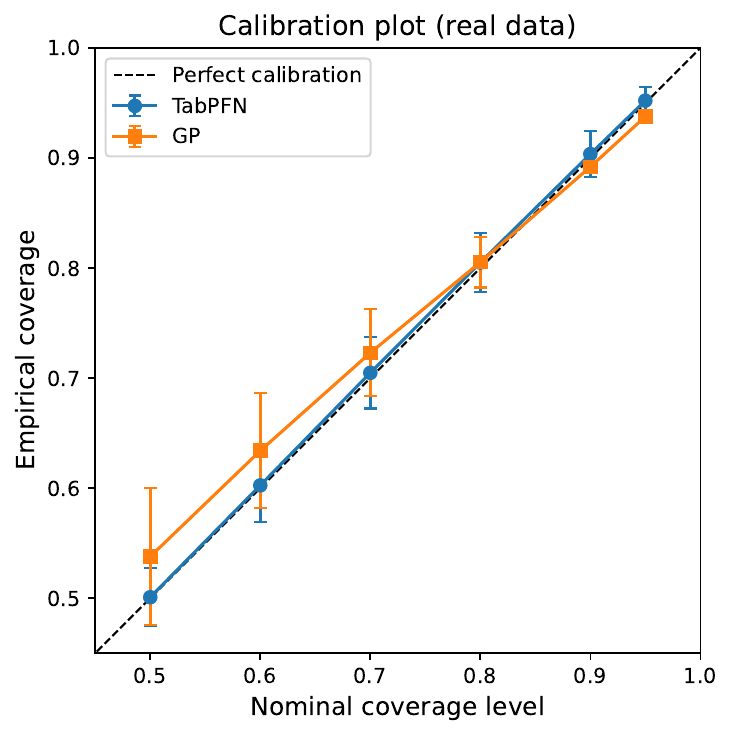}
    }
    \label{fig:realdata-calibration}
  \hfill
    \subfloat[PIT histograms for TabPFN (left) and GPR (right).
      Bars show the mean density histogram averaged first over repetitions and then over datasets.
      The dashed line at density $= 1$ is the $\mathrm{Uniform}(0,1)$ target.]{\includegraphics[width=0.6\linewidth]{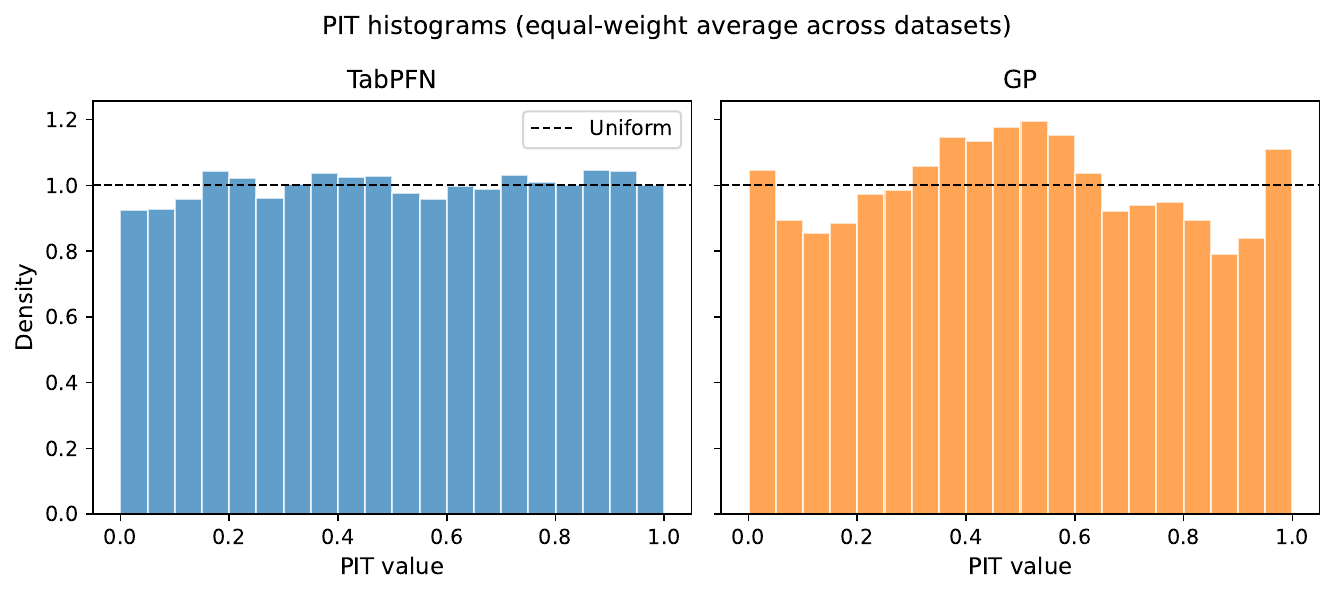}}
    \label{fig:realdata-pit}
  \caption{Calibration diagnostics on five real regression benchmarks.}
  \label{fig:realdata-uq}
\end{figure*}

\textbf{Results.}
Figure~\ref{fig:realdata-uq} and Table~\ref{tab:realdata-uq} summarise the results.

\begin{table*}[tp]
  \centering
  \caption{Real-data UQ evaluation across five regression benchmarks.
    Coverage, width, and CRPS are averaged over 20 repetitions per dataset; KS-PIT is computed from the pooled PIT values across those repetitions.
    Coverage\@95\%: empirical coverage of the nominal $95\%$ prediction interval (target: $0.950$).
    Width\@95\%: mean interval width in standardised units (lower is sharper).
    CRPS: continuous ranked probability score via quantile scoring (lower is better).
    KS-PIT: Kolmogorov--Smirnov statistic of PIT values against $\mathrm{Uniform}(0,1)$ (lower indicates better calibration).}
  \label{tab:realdata-uq}
  \small
  \begin{tabular}{llcccc}
    \toprule
    Dataset & Method & Coverage\@95\% & Width\@95\% & CRPS & KS-PIT \\
    \midrule
    \multirow{2}{*}{kin8nm}
      & TabPFN & 0.968 & 1.618 & 0.207 & 0.030 \\
      & GPR     & 0.943 & 1.439 & 0.208 & 0.032 \\
    \addlinespace
    \multirow{2}{*}{house\_8L}
      & TabPFN & 0.954 & 1.654 & 0.213 & 0.034 \\
      & GPR     & 0.929 & 2.035 & 0.295 & 0.119 \\
    \addlinespace
    \multirow{2}{*}{puma8NH}
      & TabPFN & 0.939 & 1.993 & 0.304 & 0.035 \\
      & GPR     & 0.939 & 2.529 & 0.376 & 0.038 \\
    \addlinespace
    \multirow{2}{*}{BNG(stock)}
      & TabPFN & 0.940 & 1.653 & 0.244 & 0.031 \\
      & GPR     & 0.942 & 1.981 & 0.291 & 0.031 \\
    \addlinespace
    \multirow{2}{*}{bank8FM}
      & TabPFN & 0.960 & 0.648 & 0.094 & 0.021 \\
      & GPR     & 0.936 & 0.773 & 0.113 & 0.040 \\
    \midrule
    \multirow{2}{*}{Mean}
      & TabPFN & 0.952 & 1.513 & 0.212 & 0.031 \\
      & GPR     & 0.938 & 1.751 & 0.257 & 0.052 \\
    \bottomrule
  \end{tabular}
\end{table*}
TabPFN displays dataset-dependent coverage behavior at the $95\%$ nominal level: it \emph{over-covers} on three of the five datasets---kin8nm ($96.8\%$), house\_8L ($95.4\%$), and bank8FM ($96.0\%$)---and \emph{under-covers} on the remaining two---puma8NH ($93.9\%$) and BNG(stock) ($94.0\%$).
The overall mean of $95.2\%$ is close to the nominal level, while the per-dataset variation indicates that TabPFN prediction intervals are neither uniformly conservative nor uniformly miscalibrated.
GPR, by contrast, under-covers on all five datasets (mean $93.8\%$, range $92.9\%$--$94.3\%$), showing a systematic tendency regardless of dataset.

The calibration plot (Figure~\ref{fig:realdata-uq}(a)) confirms this pattern across the full range of nominal levels: TabPFN's averaged coverage curve tracks the diagonal closely.
GPR's curve lies above the diagonal at lower nominal levels but below it at the $90\%$ and $95\%$ levels, indicating that the direction of its calibration error changes across the distribution.
The PIT histograms (Figure~\ref{fig:realdata-uq}(b)) provide further detail.
TabPFN's PIT values are close to uniform (mean KS-PIT $= 0.031$).
GPR's PIT distribution departs more substantially from uniformity (mean KS-PIT $= 0.052$), driven primarily by house\_8L (KS-PIT $= 0.119$).

On sharpness, TabPFN produces narrower $95\%$ intervals than GPR on four of the five datasets (mean width $1.51$ vs.\ $1.75$, a $14\%$ reduction); the exception is kin8nm, where GPR is narrower ($1.44$ vs.\ $1.62$) but also covers less ($94.3\%$ vs.\ $96.8\%$).
TabPFN achieves lower CRPS on all five datasets (mean $0.212$ vs.\ $0.257$), confirming that its predictive distributions are jointly sharper and better calibrated in the aggregate.

\section{Sparse linear regression}
\subsection{Hyper-parameter selection}
The LASSO estimator is implemented using the \emph{Lasso} function in the \texttt{scikit-learn} package, while the MCP and SCAD estimators are implemented using the \texttt{pycasso} package in Python.
All three methods involve a regularization hyperparameter controlling the strength of penalization.
For each method, we evaluate $20$ candidate values of the tuning parameter and select the optimal one based on the prediction error computed on an independent validation set of size $1000$.
For LASSO, the regularization grid is chosen over the interval $[0.001\|X_{\text{train}}^{\top}y_{\text{train}}\|_{\infty}, \|X_{\text{train}}^{\top}y_{\text{train}}\|_{\infty}]$.
For SCAD and MCP, the tuning parameter grids are respectively defined as $[3, 100]$ and $[1.1, 100]$.
All grids are generated on a logarithmic scale with $20$ evenly spaced points within the corresponding intervals.

\subsection{Additional results for sparse linear regression}
\label{app:linear_regression_more}
\begin{figure*}[tp]
    \centering
    \includegraphics[width=0.24\linewidth]{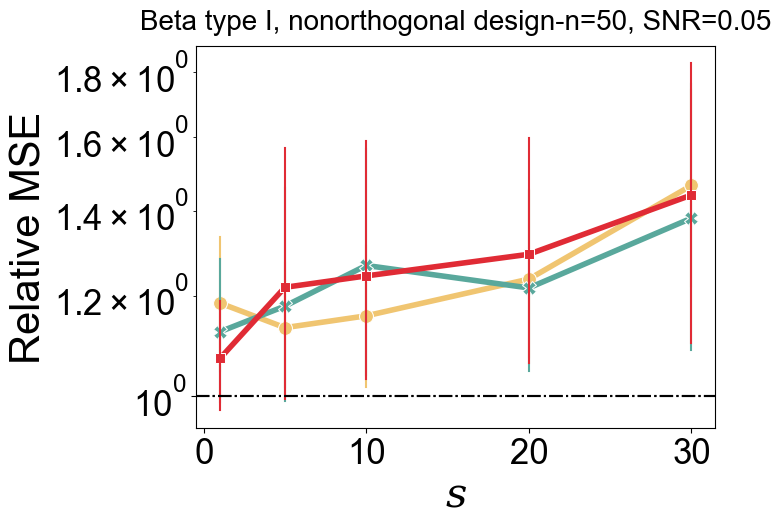}
\includegraphics[width=0.24\linewidth]{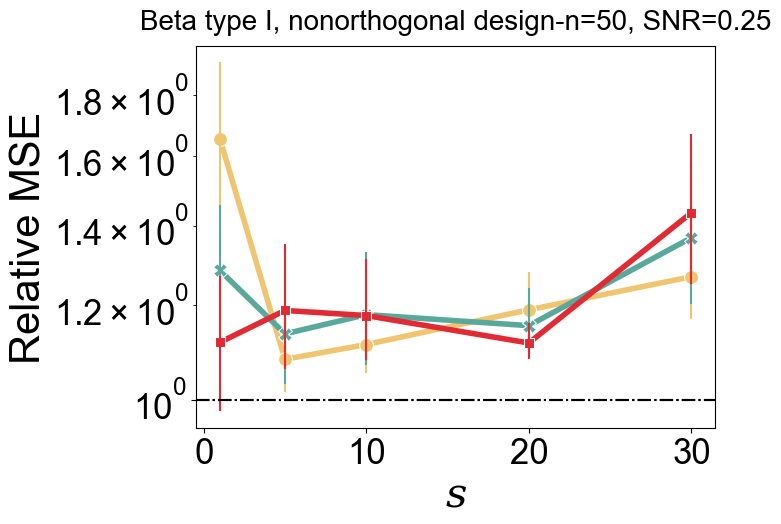}
\includegraphics[width=0.24\linewidth]{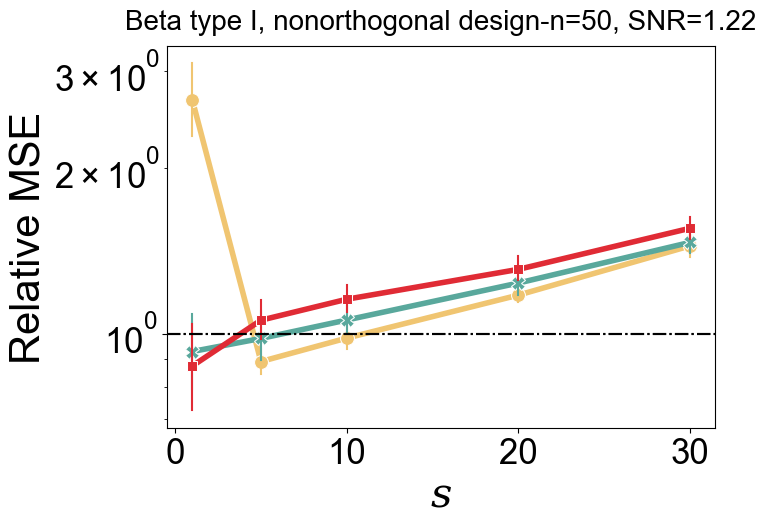}
\includegraphics[width=0.24\linewidth]{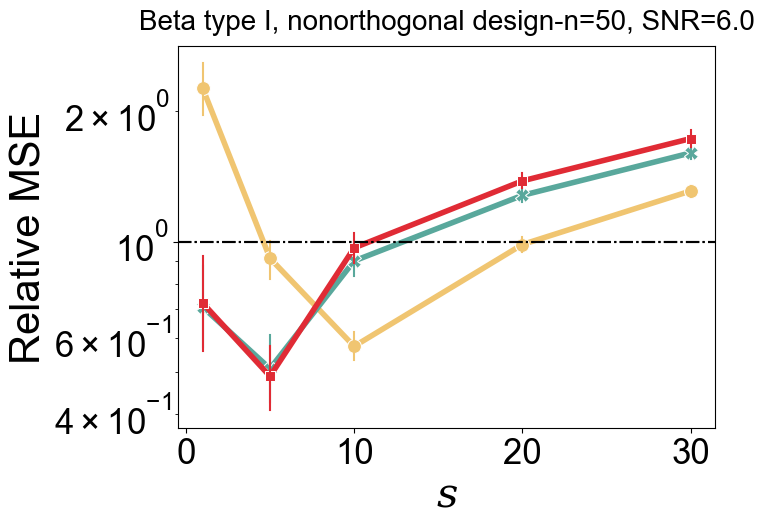}\\
\includegraphics[width=0.24\linewidth]{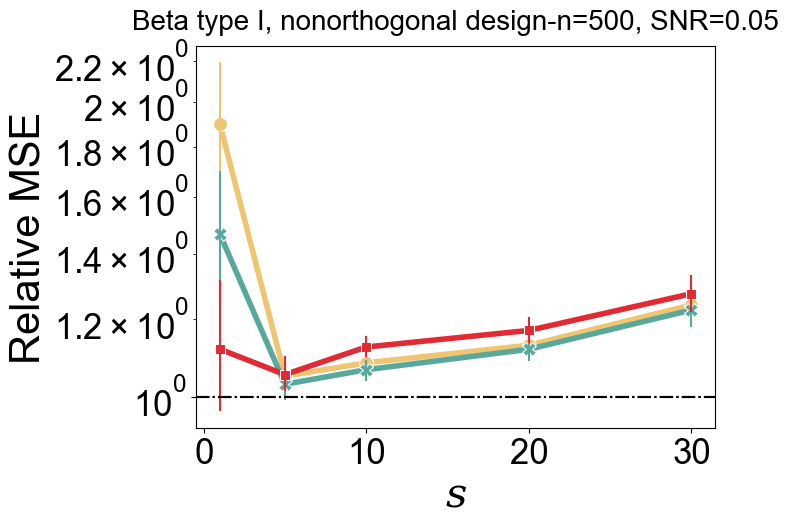}
\includegraphics[width=0.24\linewidth]{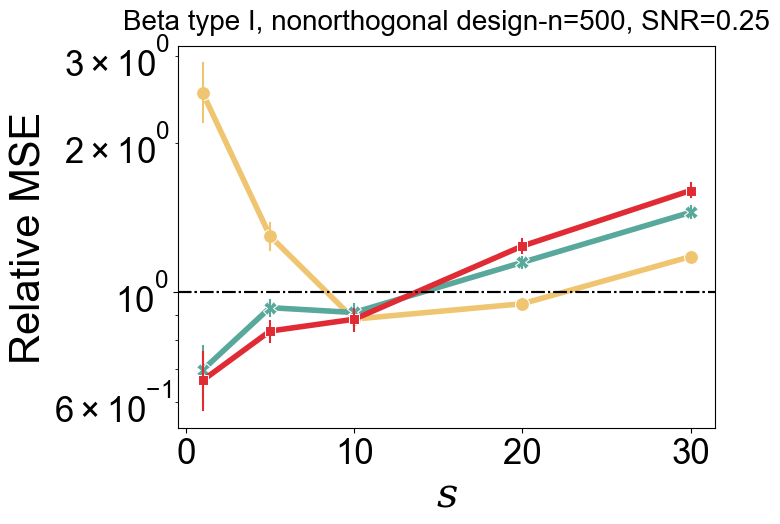}
\includegraphics[width=0.24\linewidth]{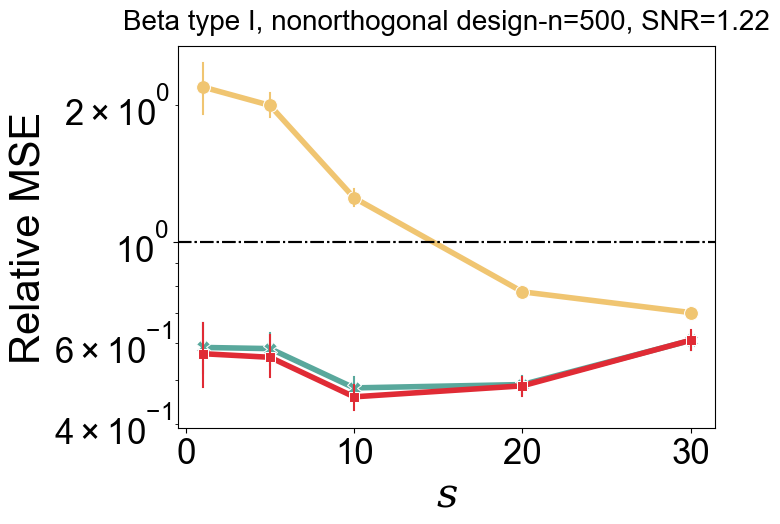}
\includegraphics[width=0.24\linewidth]{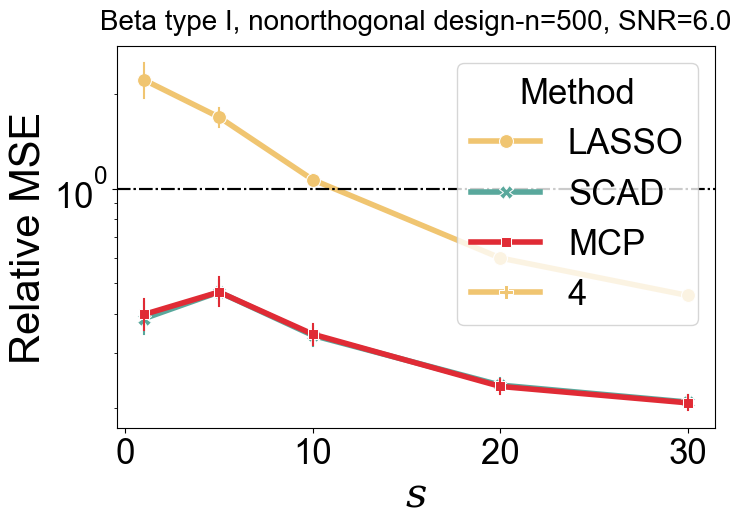}
\caption{Relative test MSE of various methods compared to TabPFN.
  Experiments use beta-type 1 with nonorthogonal design.
  The top and bottom rows display results for $n=50$ and $n=500$ respectively, while the columns (from left to right) correspond to increasing SNR values.}
\label{fig:test_MSE_beta1band}
\end{figure*}

\begin{figure*}[tp]
    \centering
    \includegraphics[width=0.24\linewidth]{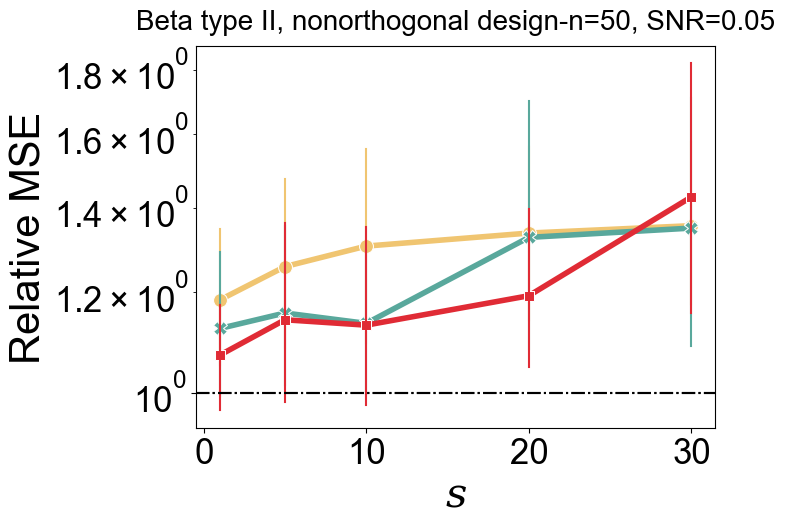}
    \includegraphics[width=0.24\linewidth]{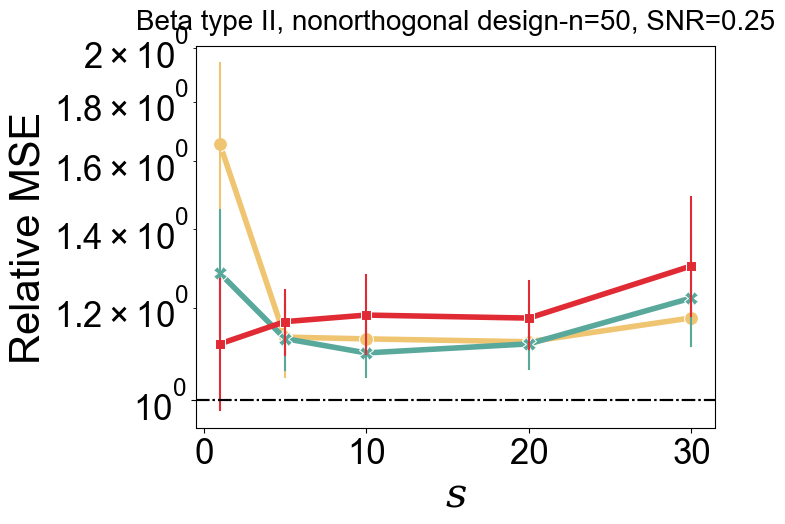}
    \includegraphics[width=0.24\linewidth]{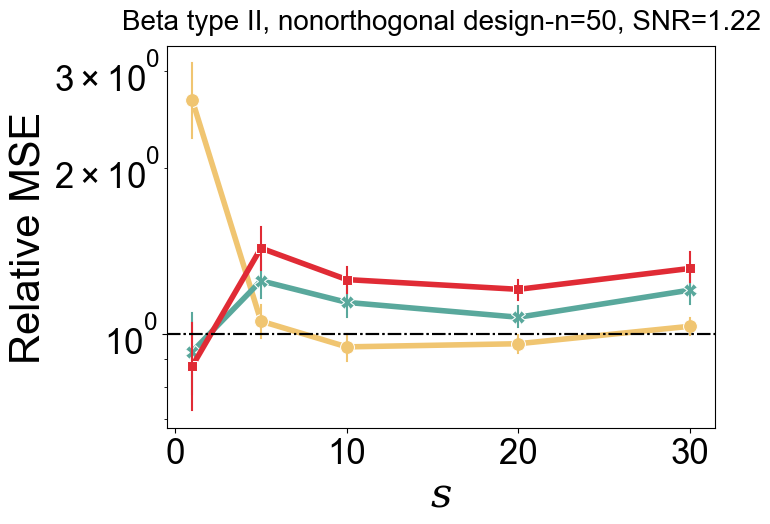}
    \includegraphics[width=0.24\linewidth]{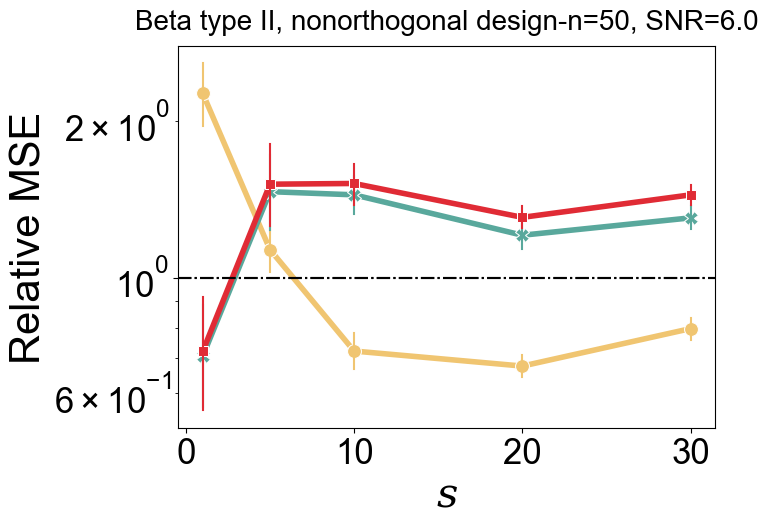}\\
    \includegraphics[width=0.24\linewidth]{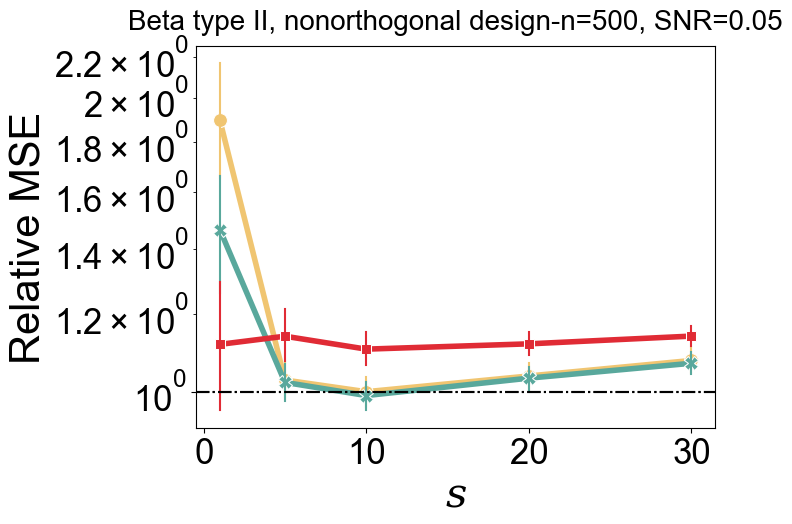}
    \includegraphics[width=0.24\linewidth]{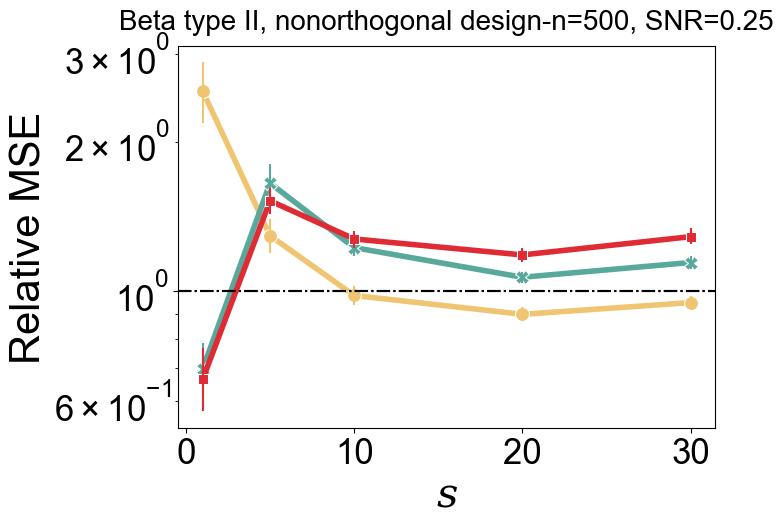}
    \includegraphics[width=0.24\linewidth]{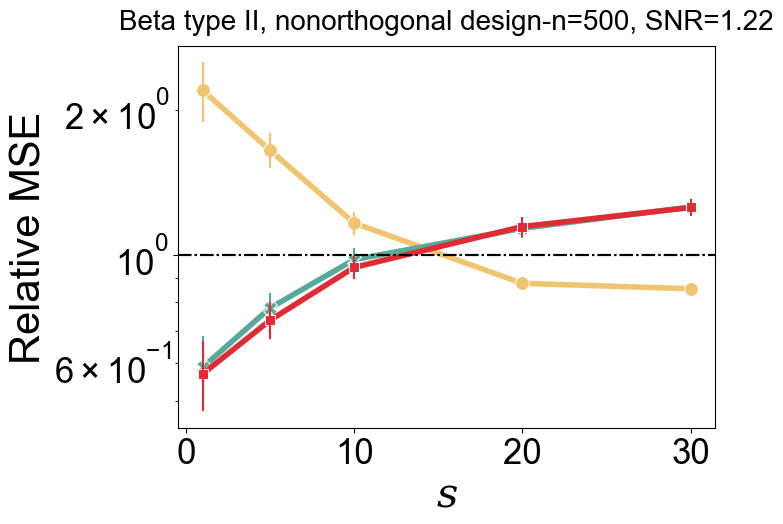}
    \includegraphics[width=0.24\linewidth]{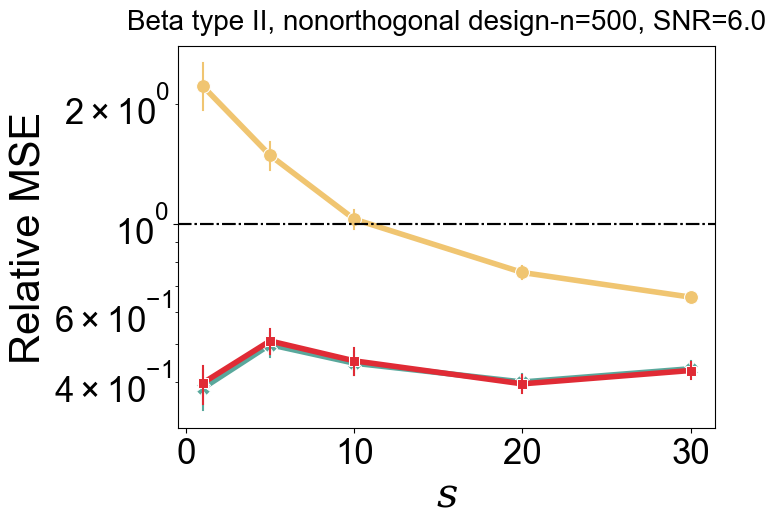}
    \caption{Relative test MSE of various methods compared to TabPFN.
      Experiments use beta-type 2 with nonorthogonal design.
      The top and bottom rows display results for $n=50$ and $n=500$ respectively, while the columns (from left to right) correspond to increasing SNR values.}
    \label{fig:test_MSE_beta2band}
\end{figure*}

The results under non-orthogonal design with beta-type 1 and beta-type 2 are given respectively in Figure~\ref{fig:test_MSE_beta1band} and Figure~\ref{fig:test_MSE_beta2band}.

\subsection{ALE plot for TabPFN}

\begin{figure*}[ht]
\includegraphics[width=0.19\linewidth]{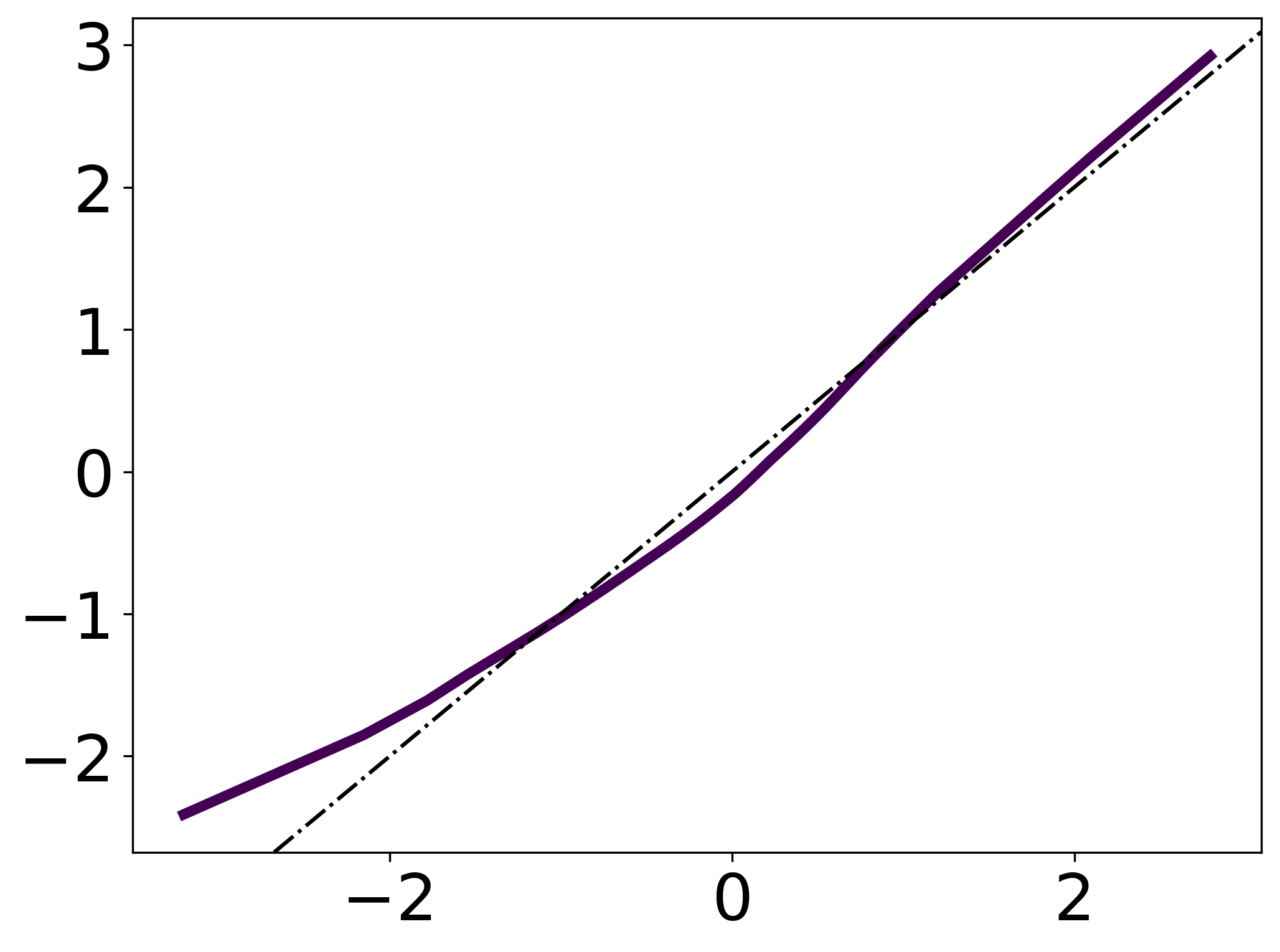}
\includegraphics[width=0.19\linewidth]{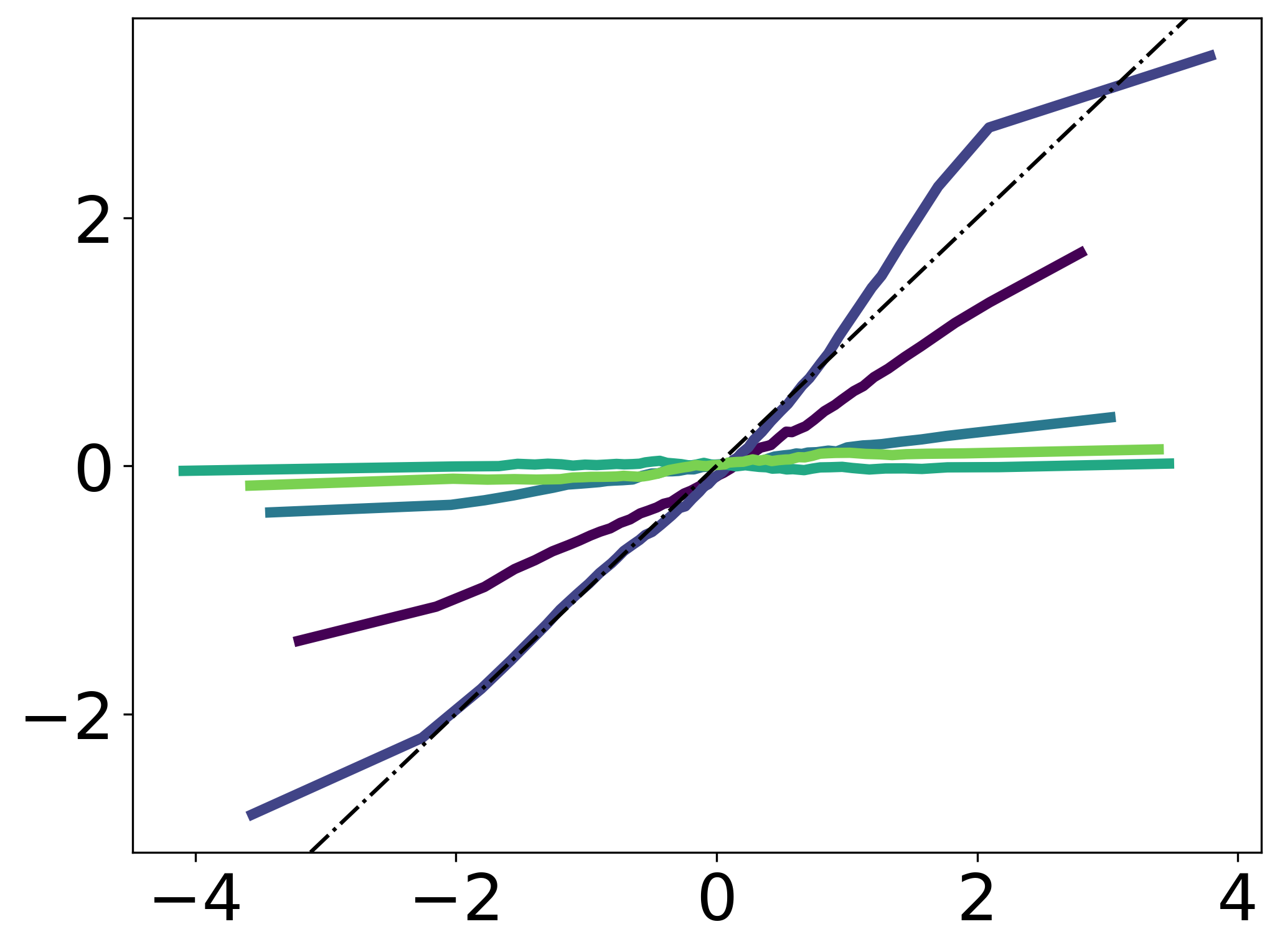}
\includegraphics[width=0.19\linewidth]{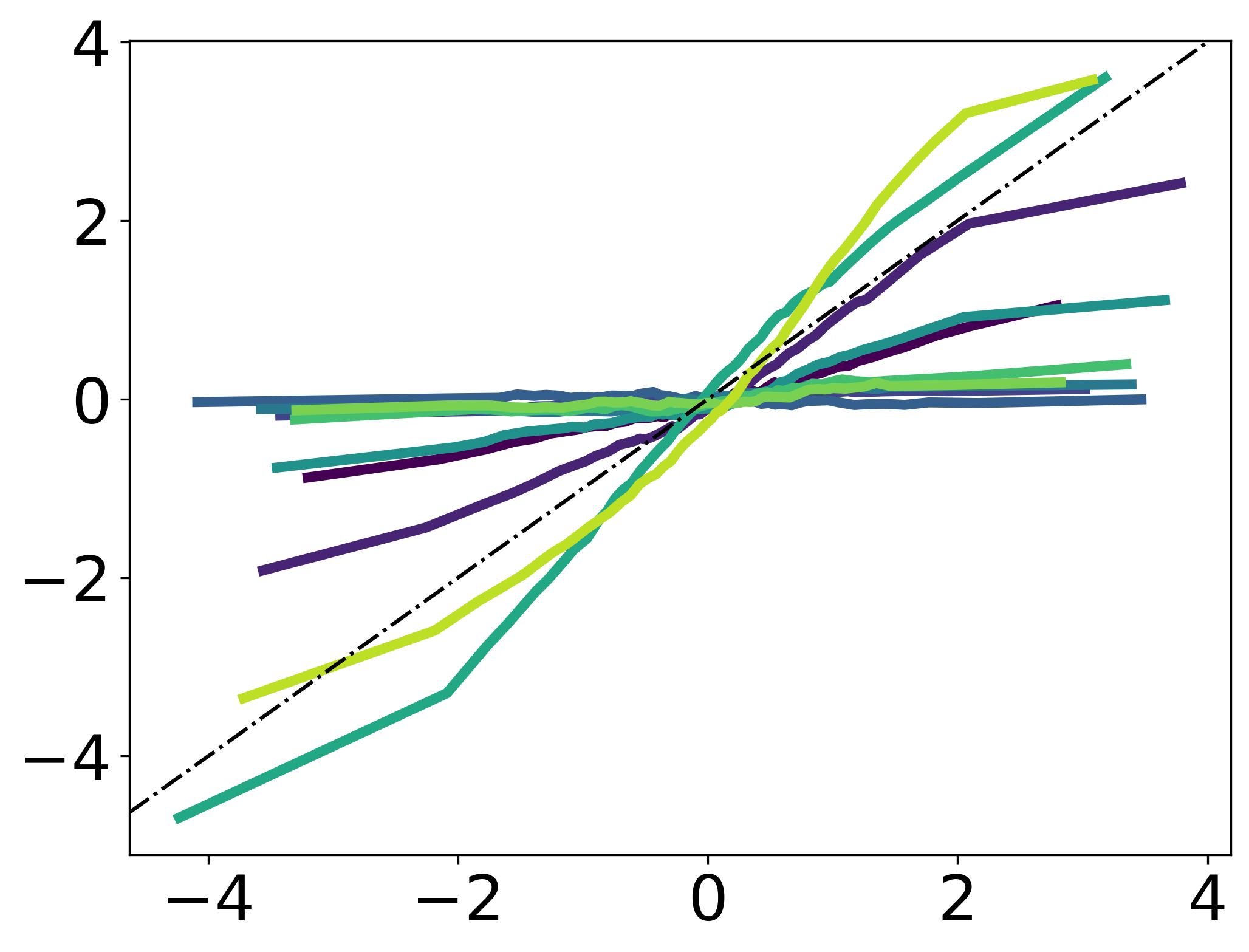}
\includegraphics[width=0.19\linewidth]{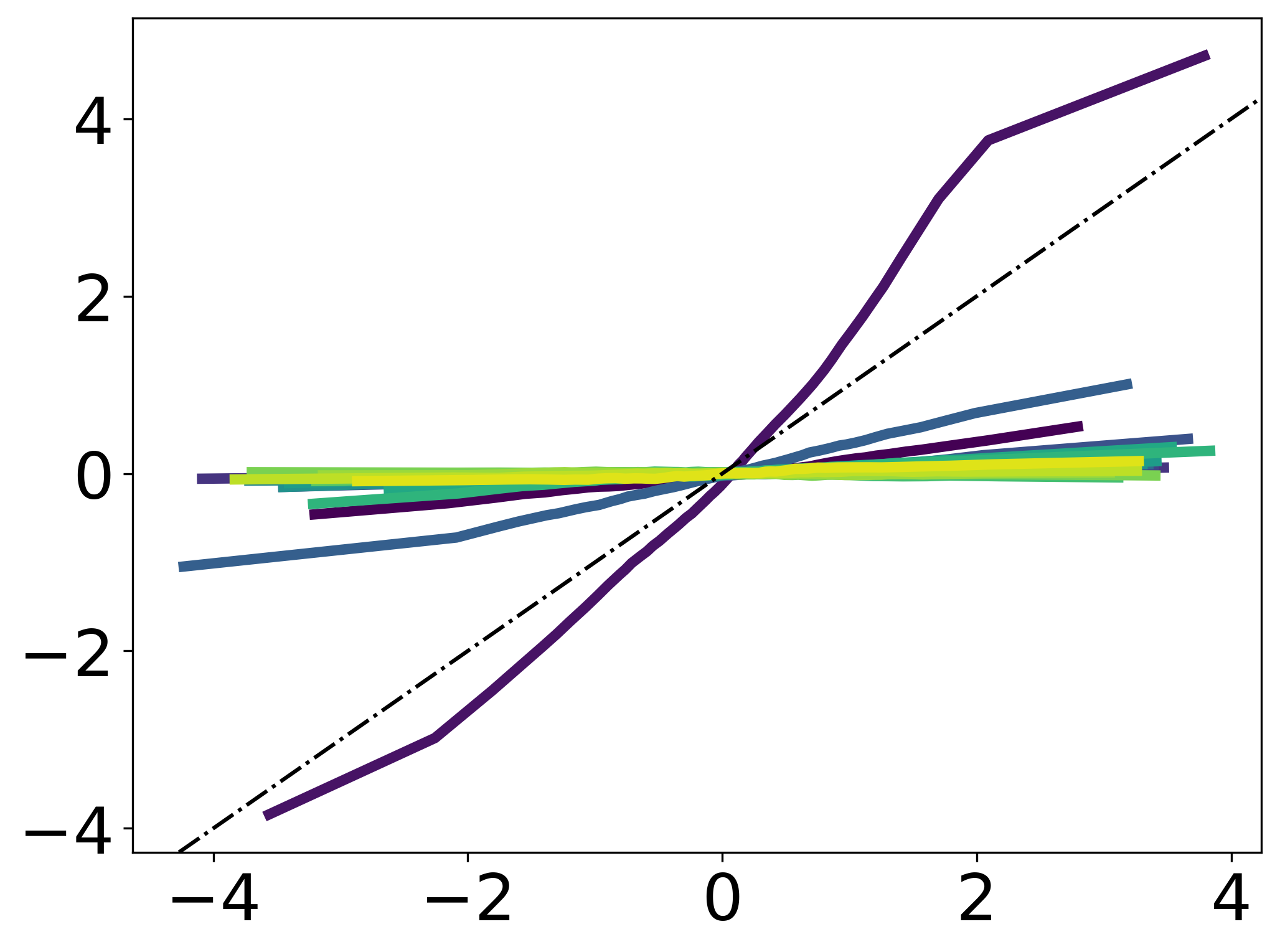}
\includegraphics[width=0.19\linewidth]{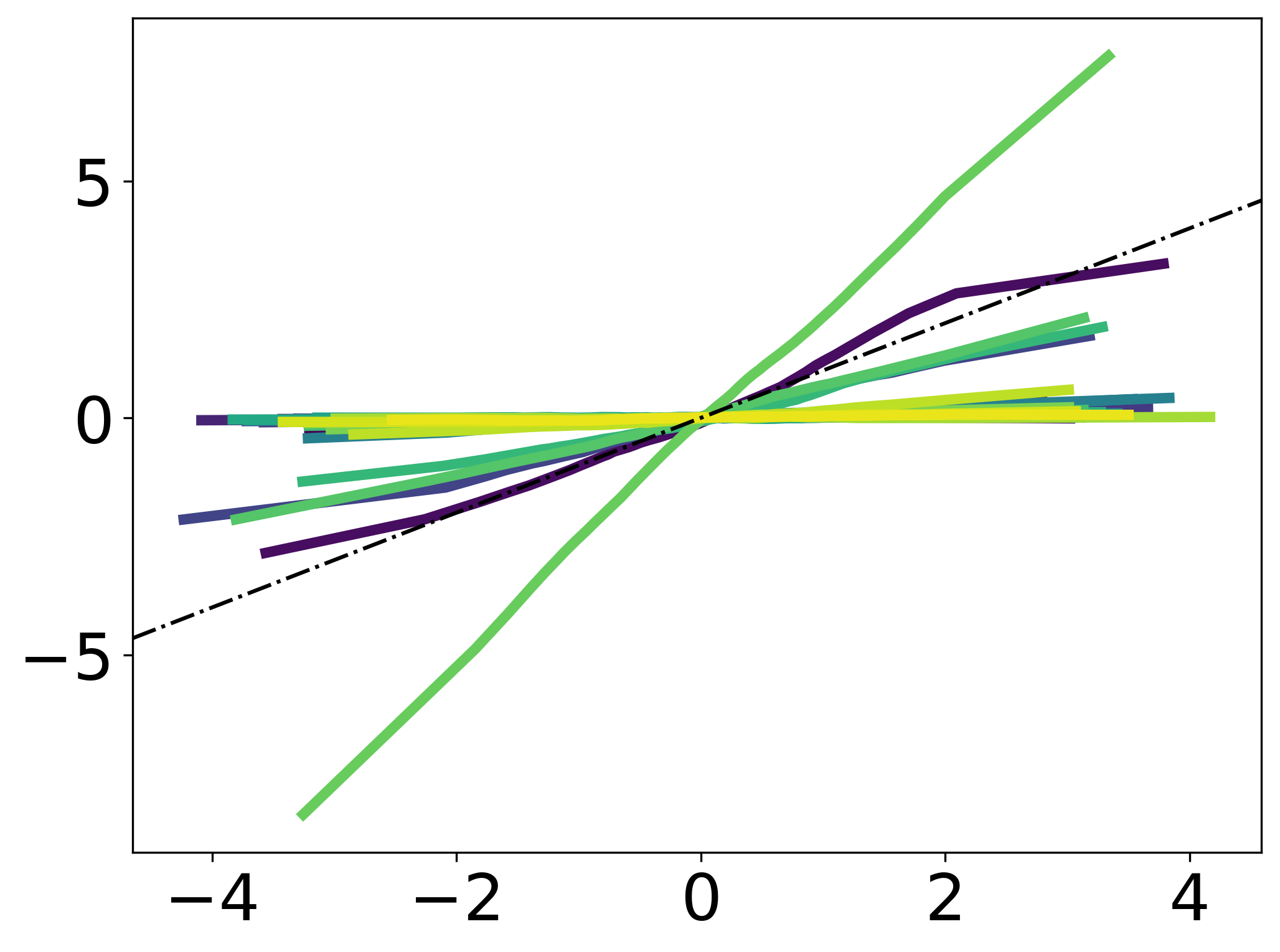}\\
\includegraphics[width=0.19\linewidth]{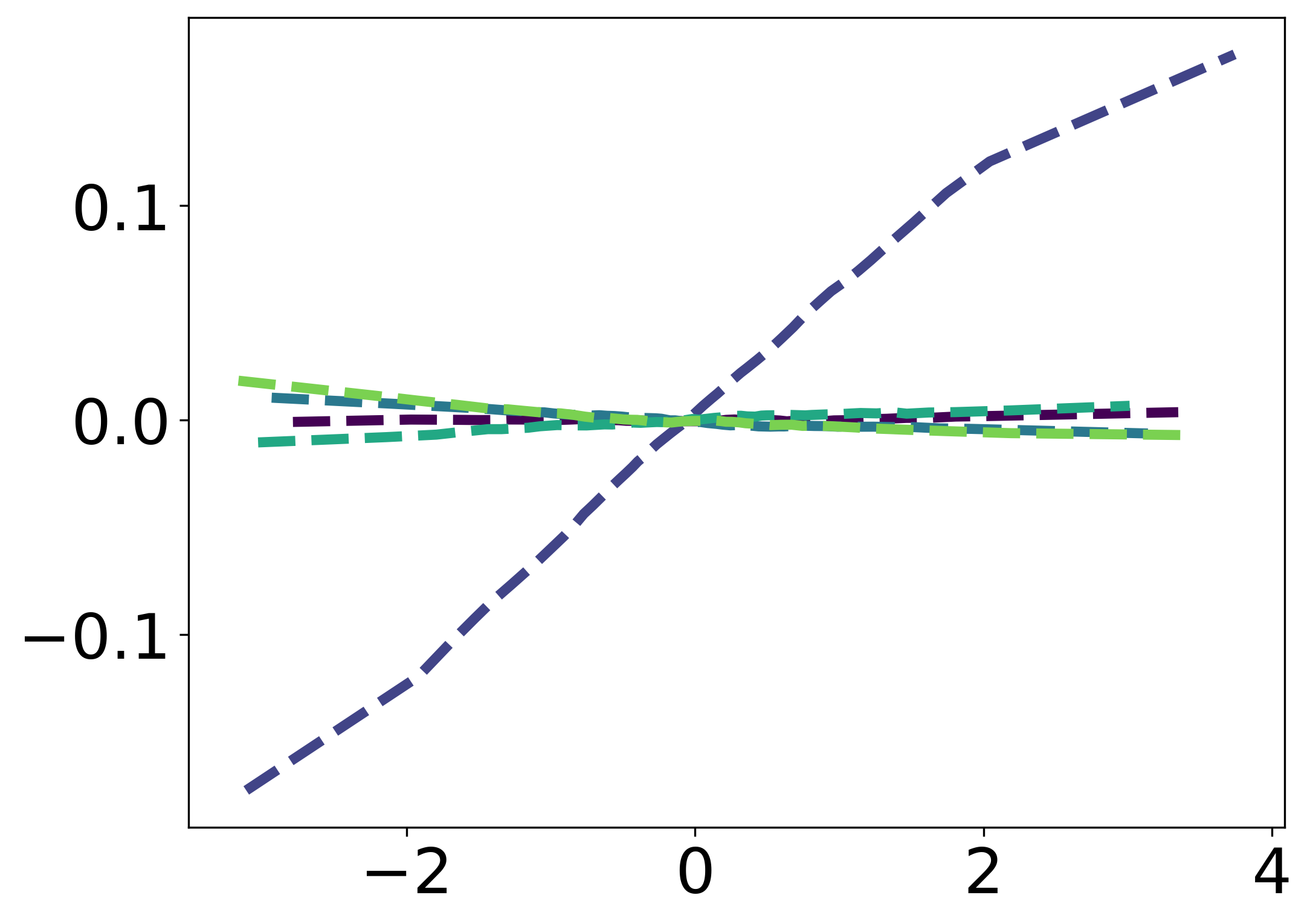}
\includegraphics[width=0.19\linewidth]{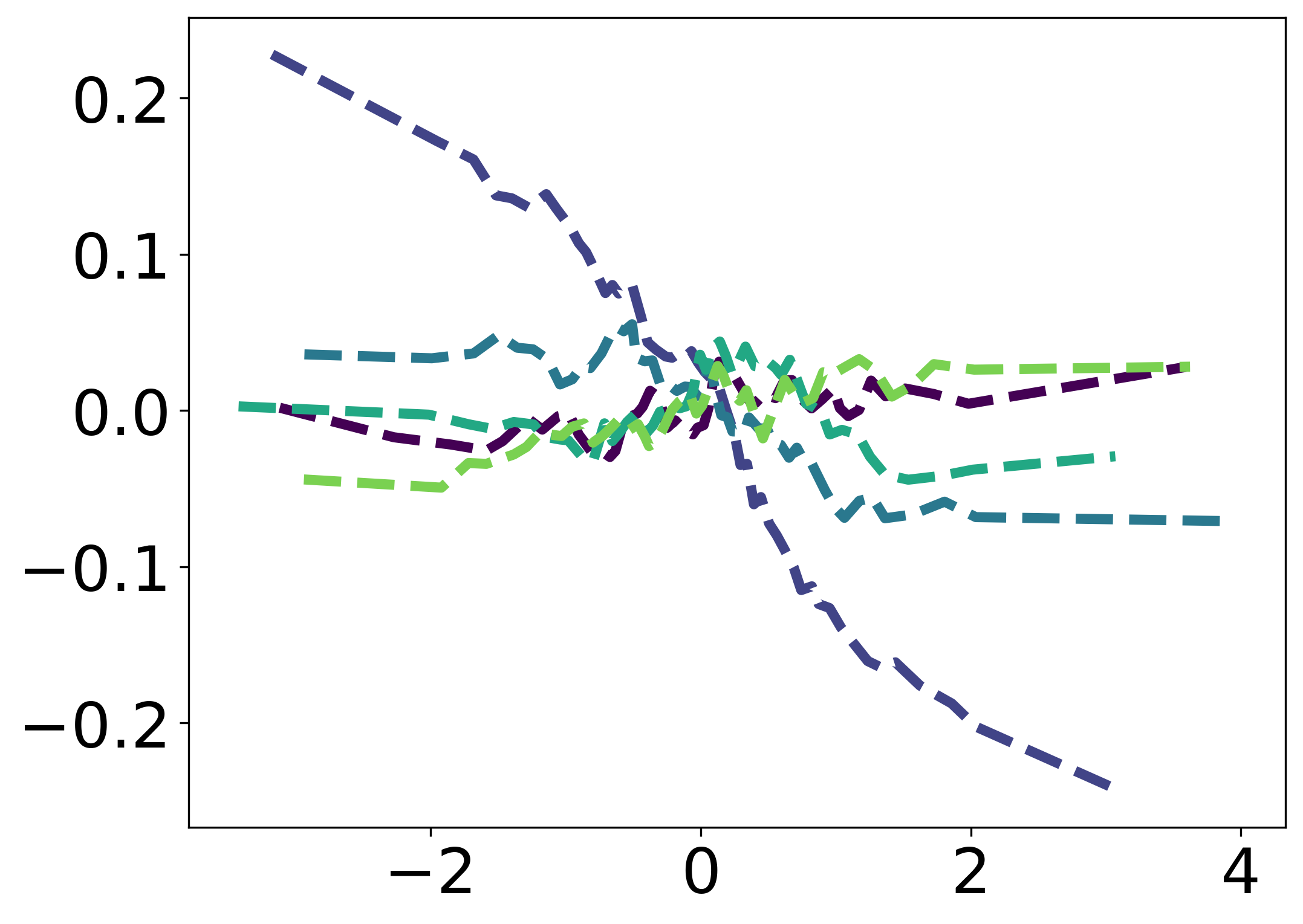}
\includegraphics[width=0.19\linewidth]{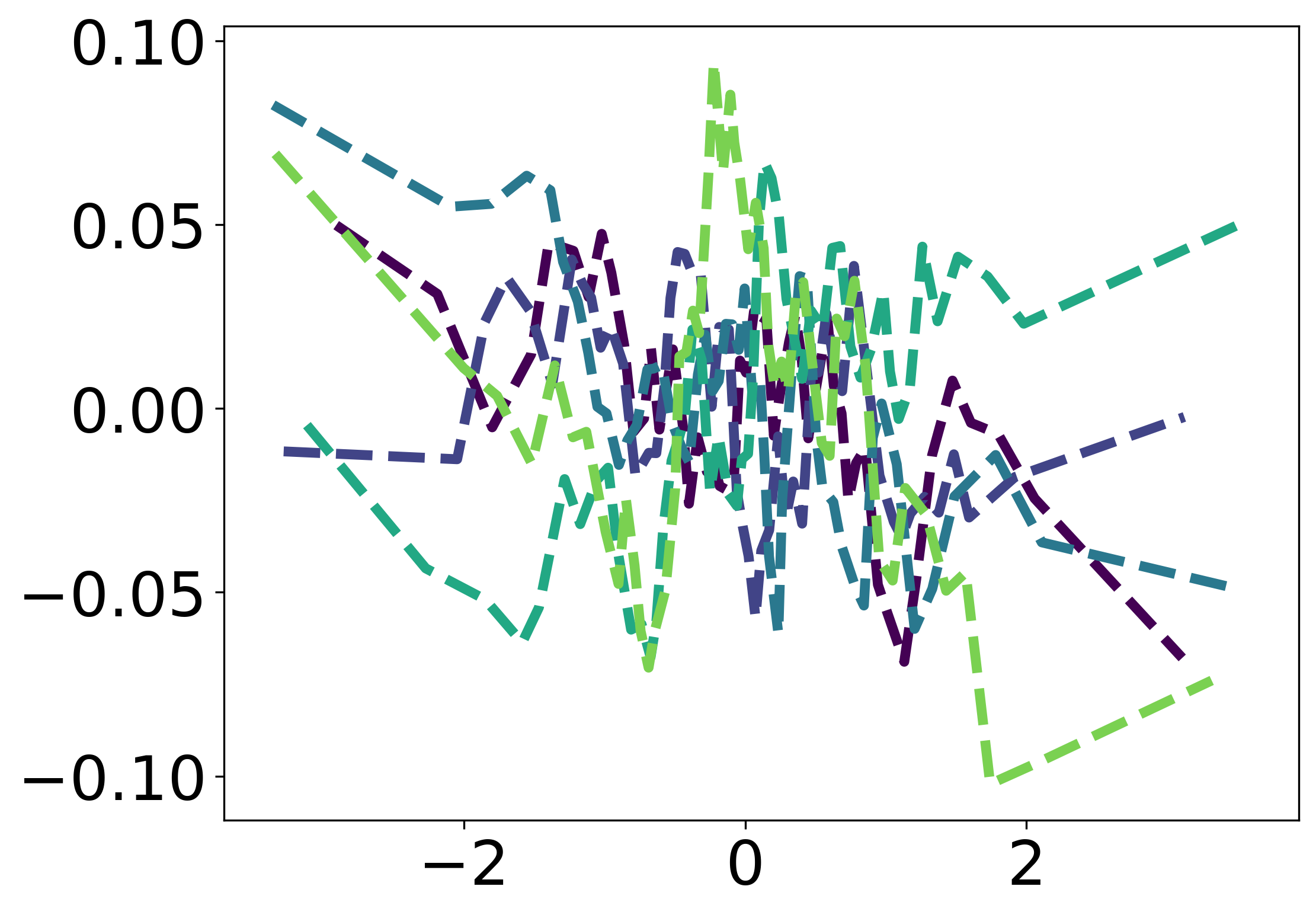}
\includegraphics[width=0.19\linewidth]{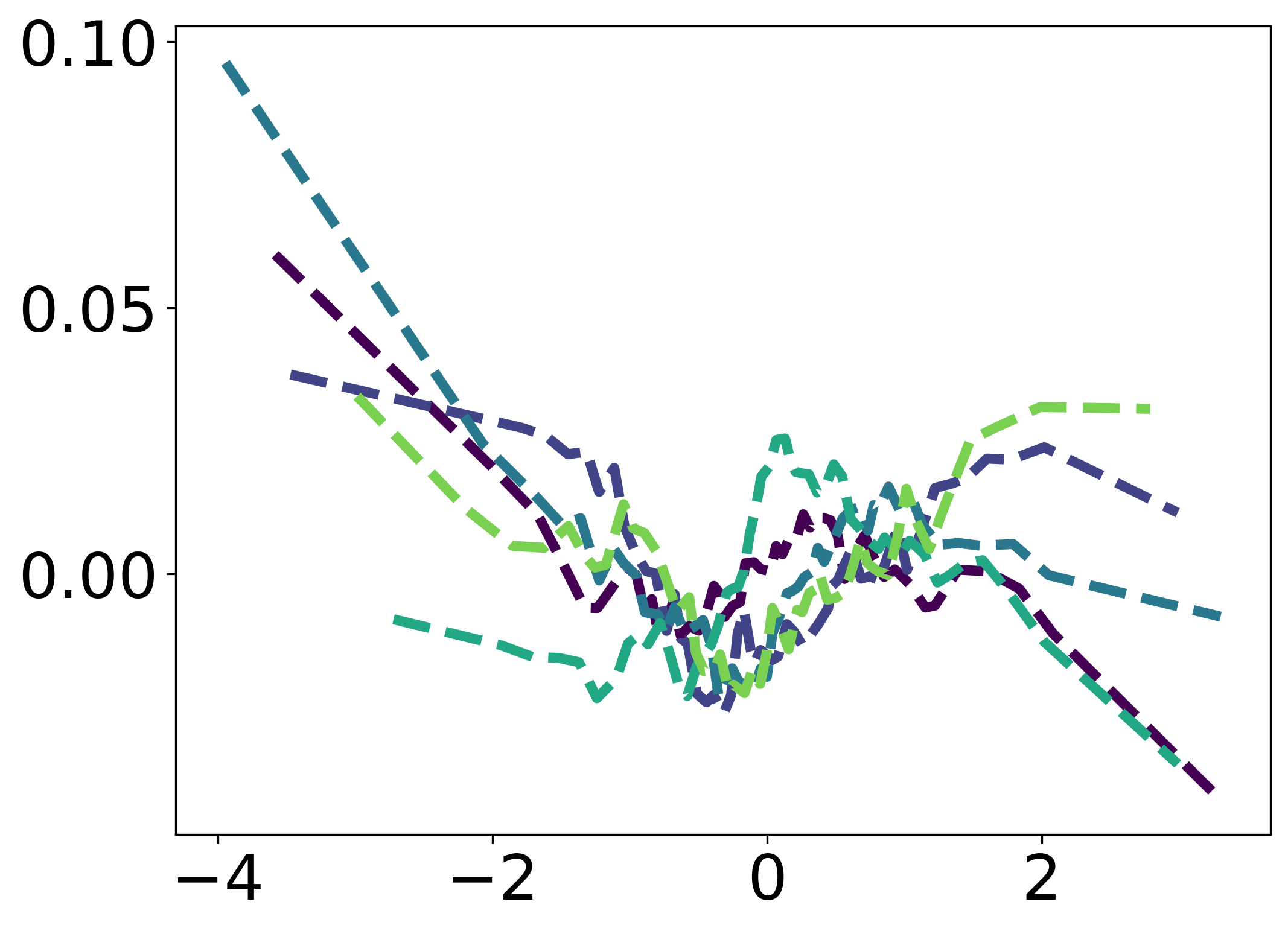}
\includegraphics[width=0.19\linewidth]{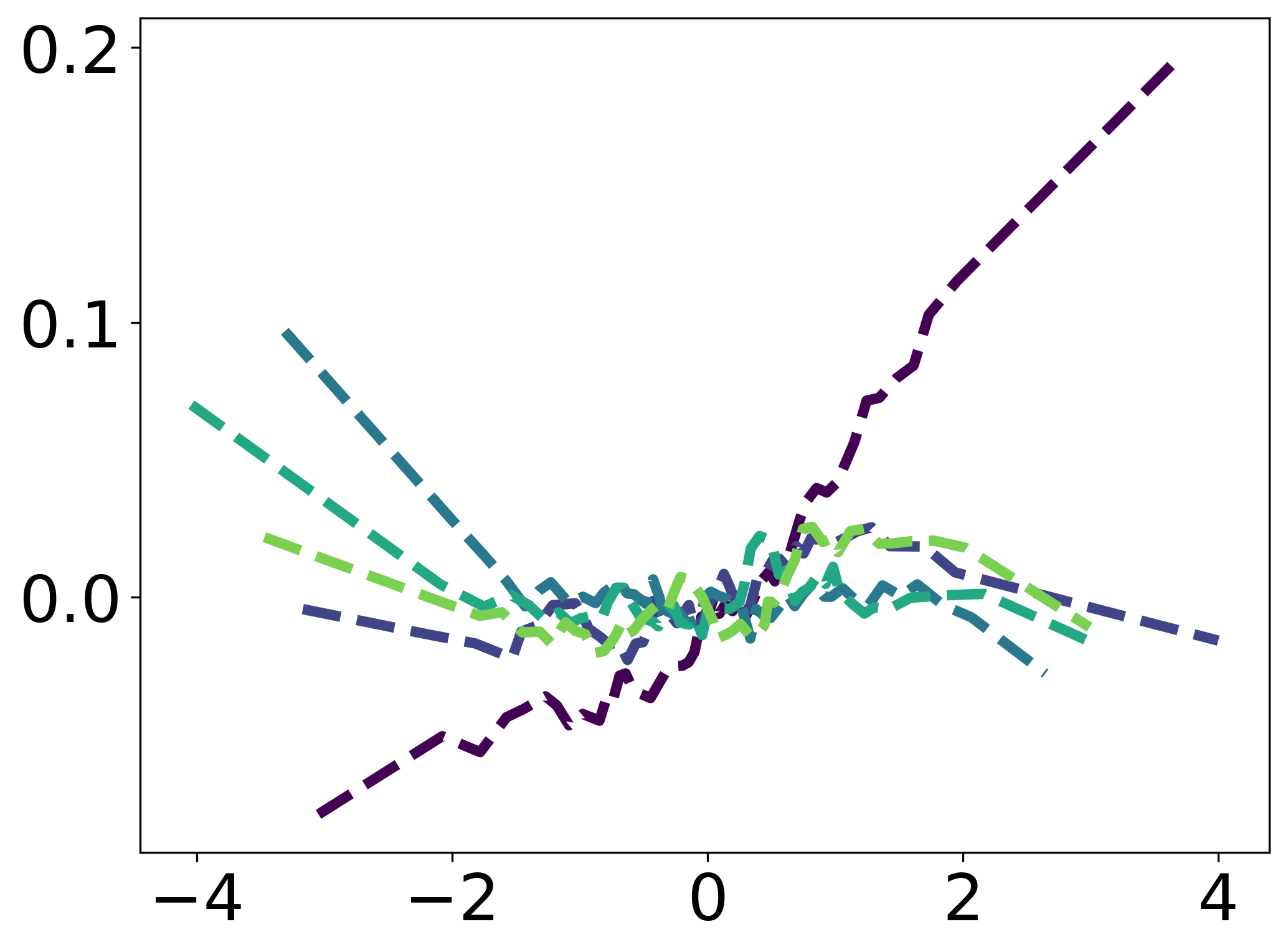}\\
\includegraphics[width=0.19\linewidth]{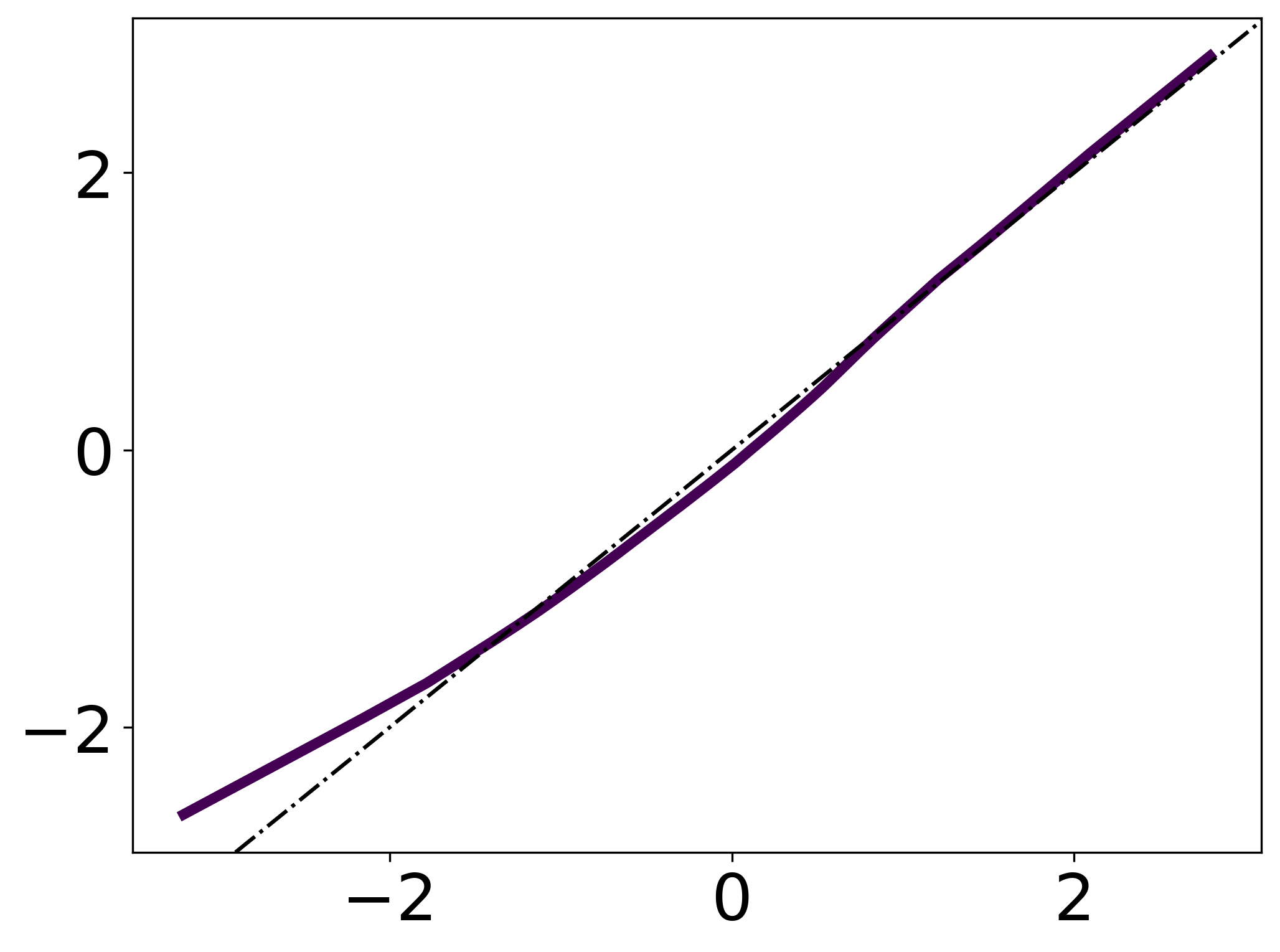}
\includegraphics[width=0.19\linewidth]{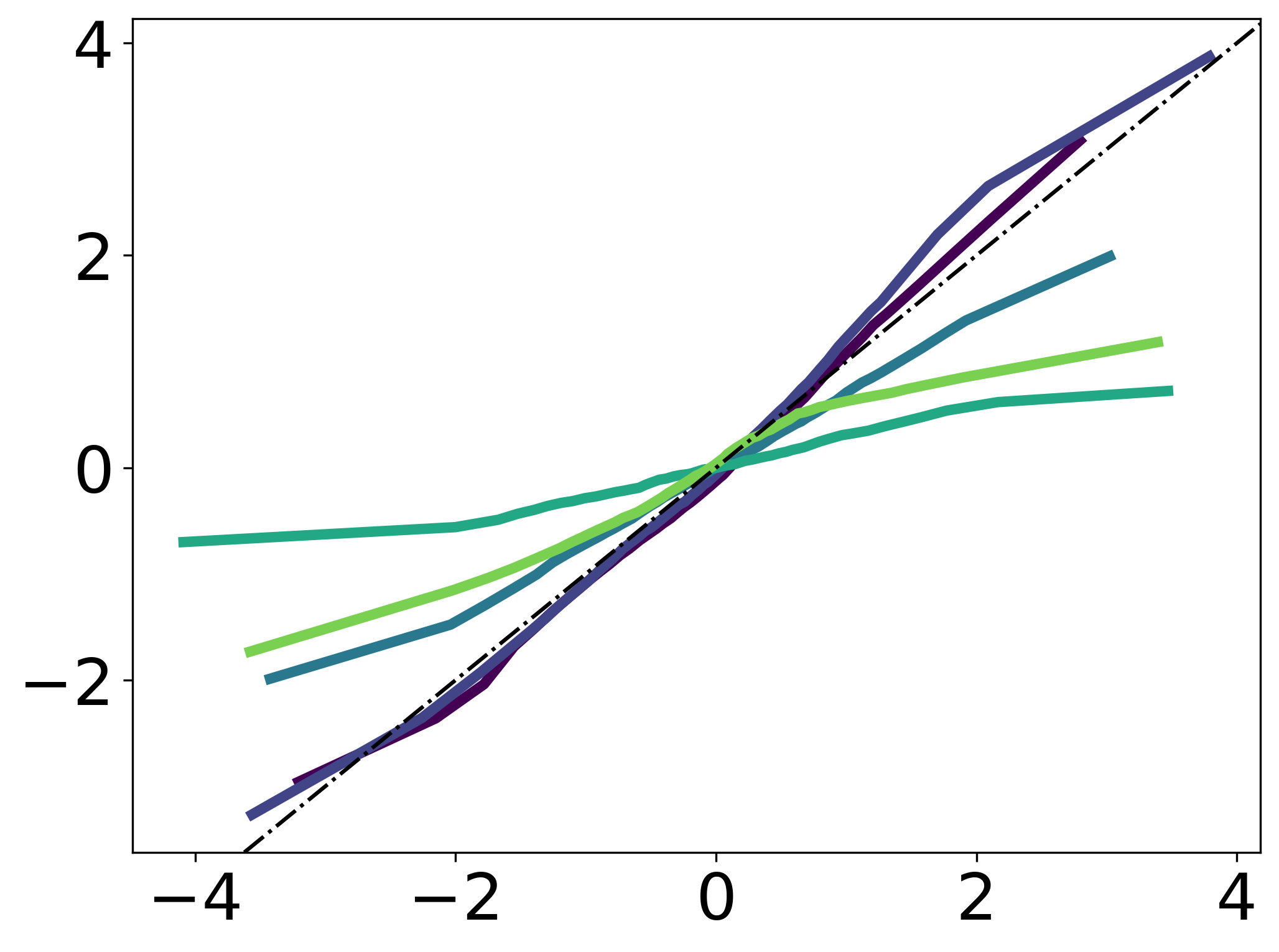}
\includegraphics[width=0.19\linewidth]{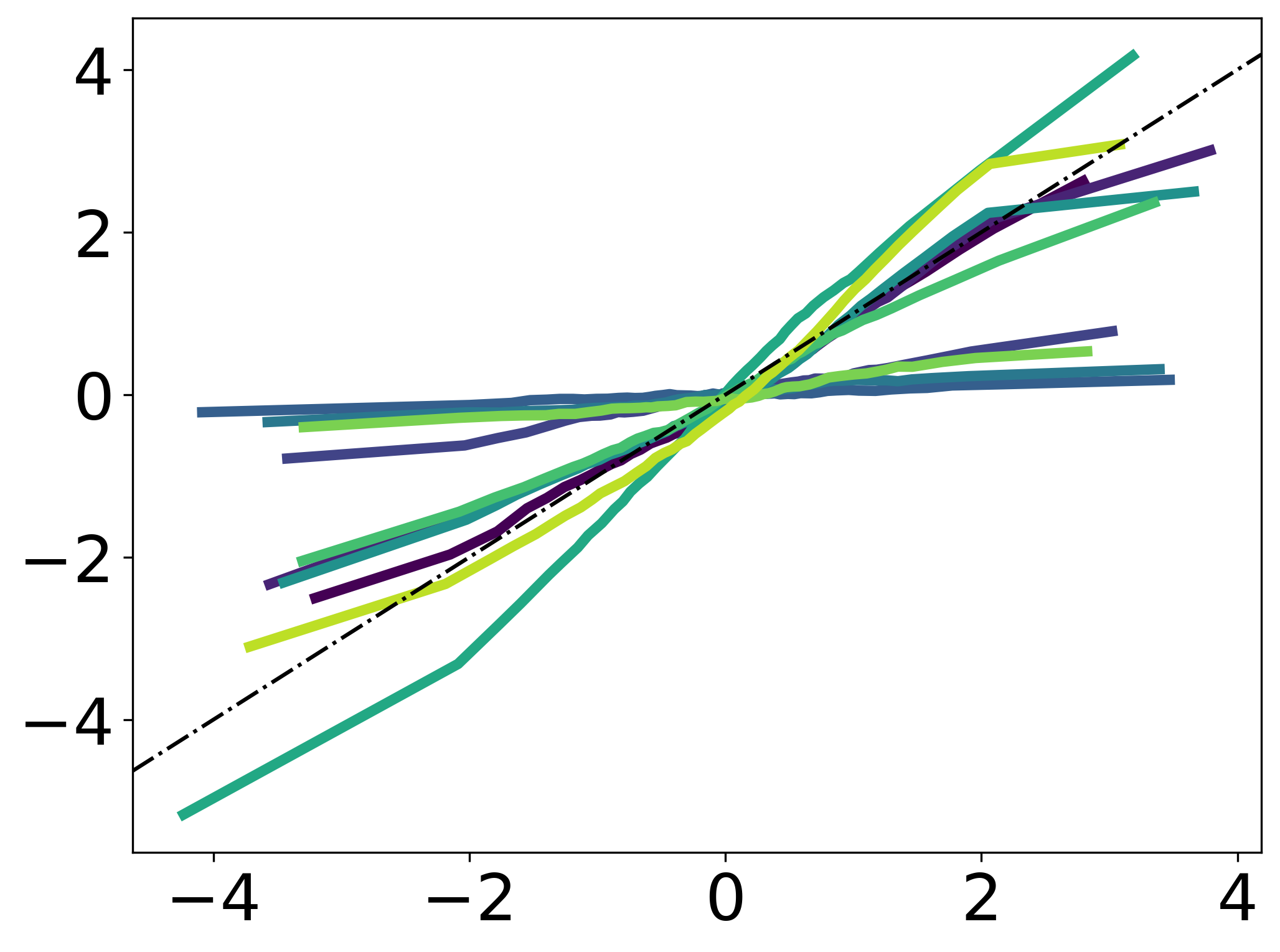}
\includegraphics[width=0.19\linewidth]{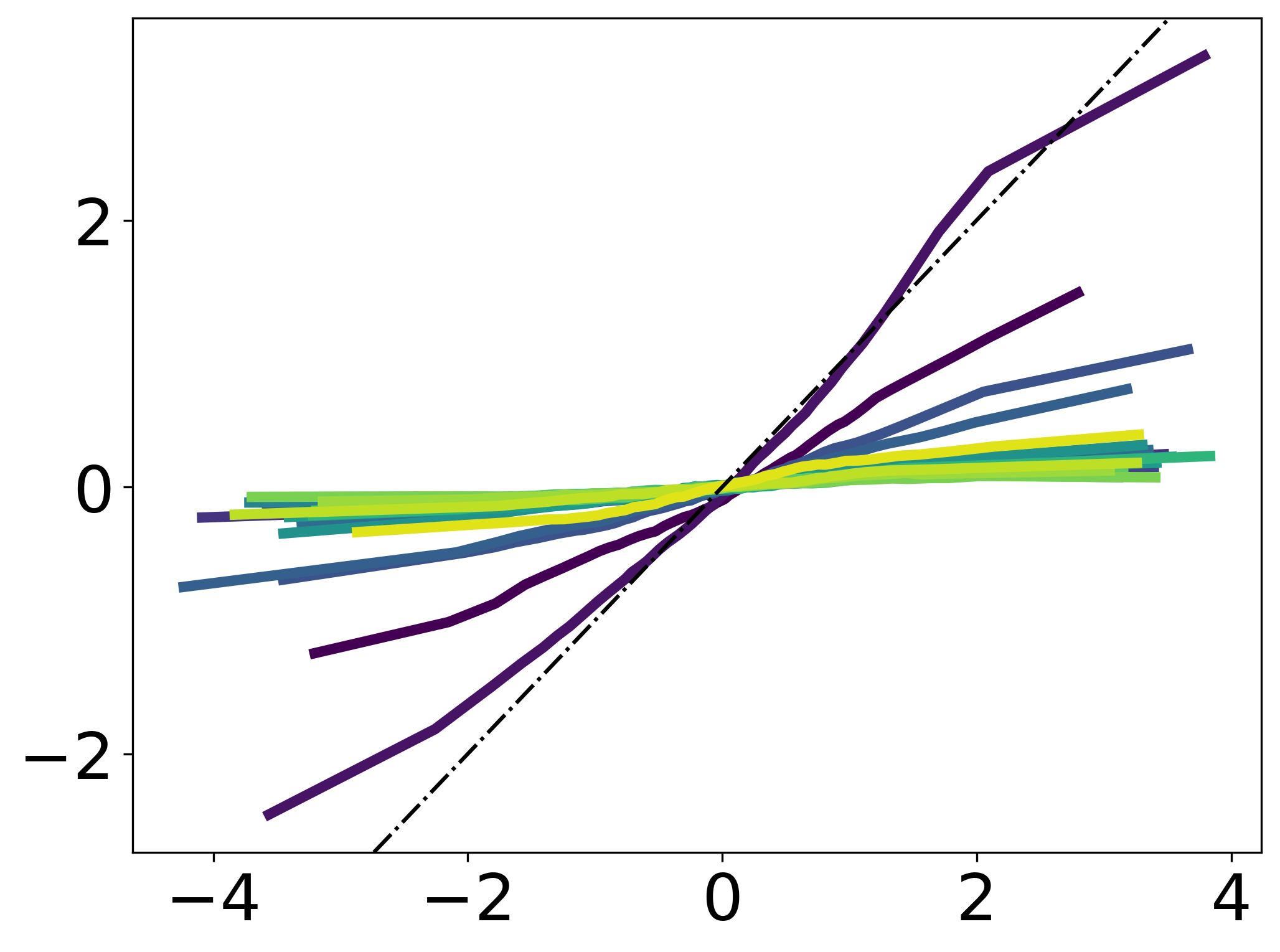}
\includegraphics[width=0.19\linewidth]{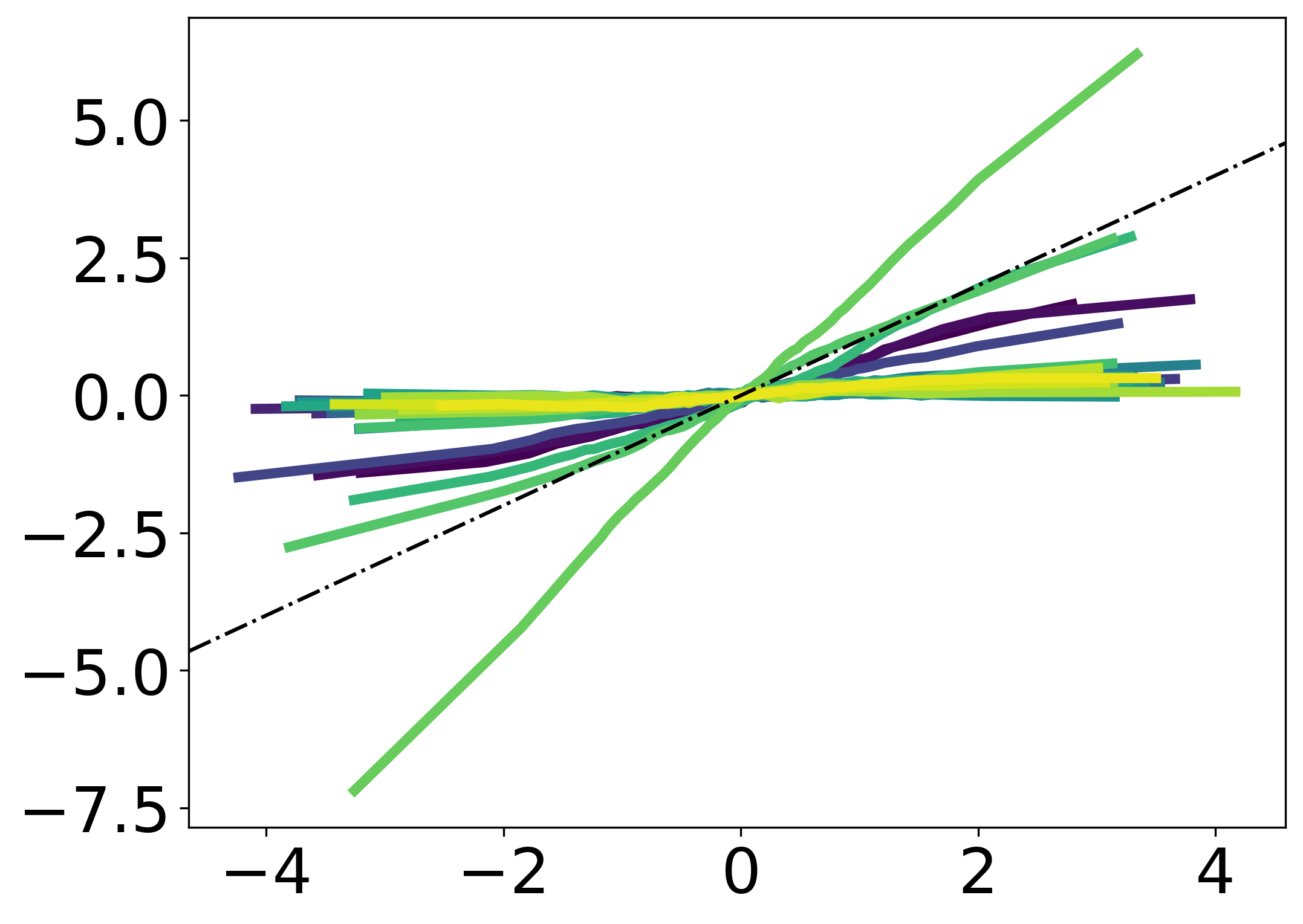}\\
\includegraphics[width=0.19\linewidth]{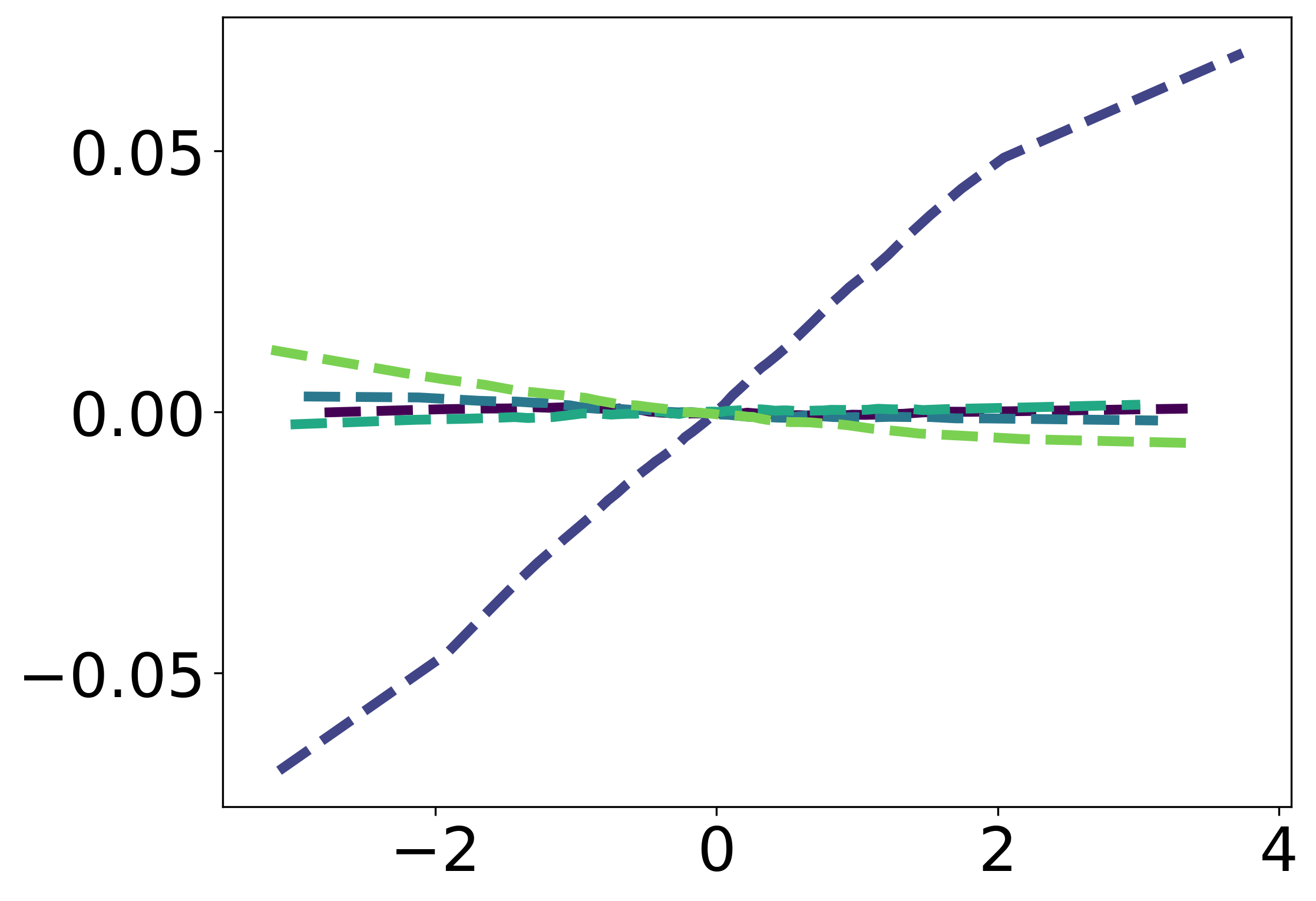}
\includegraphics[width=0.19\linewidth]{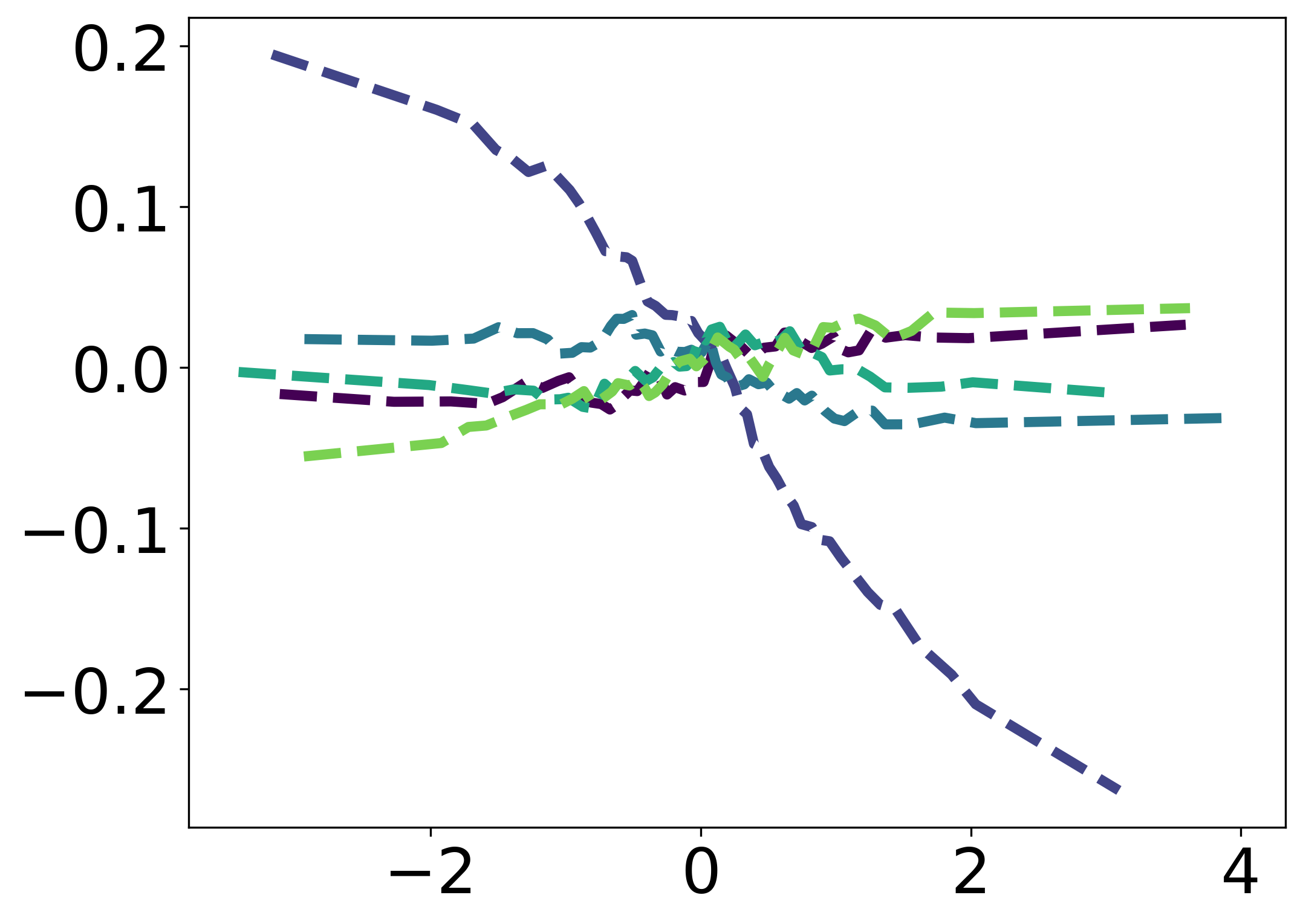}
\includegraphics[width=0.19\linewidth]{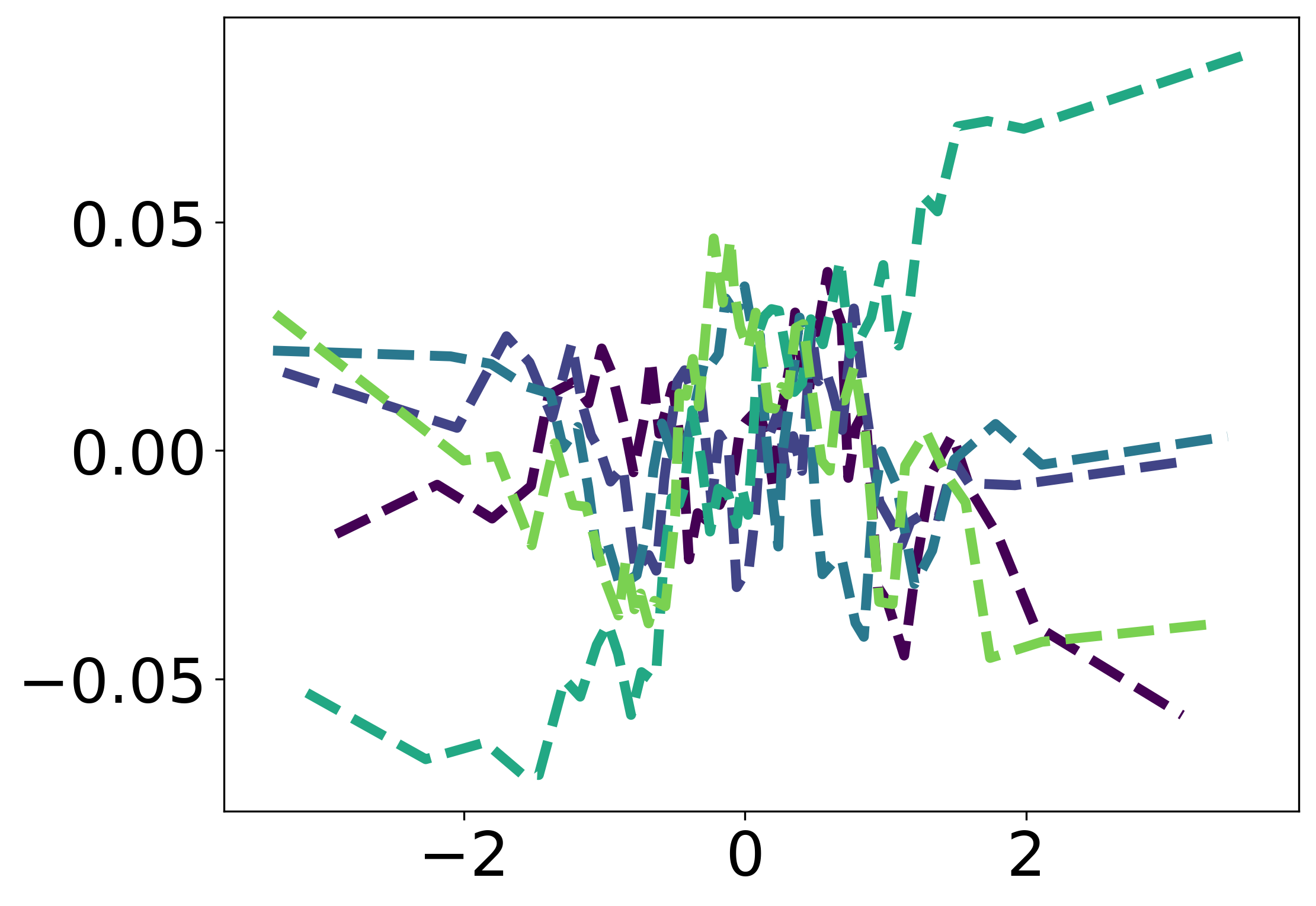}
\includegraphics[width=0.19\linewidth]{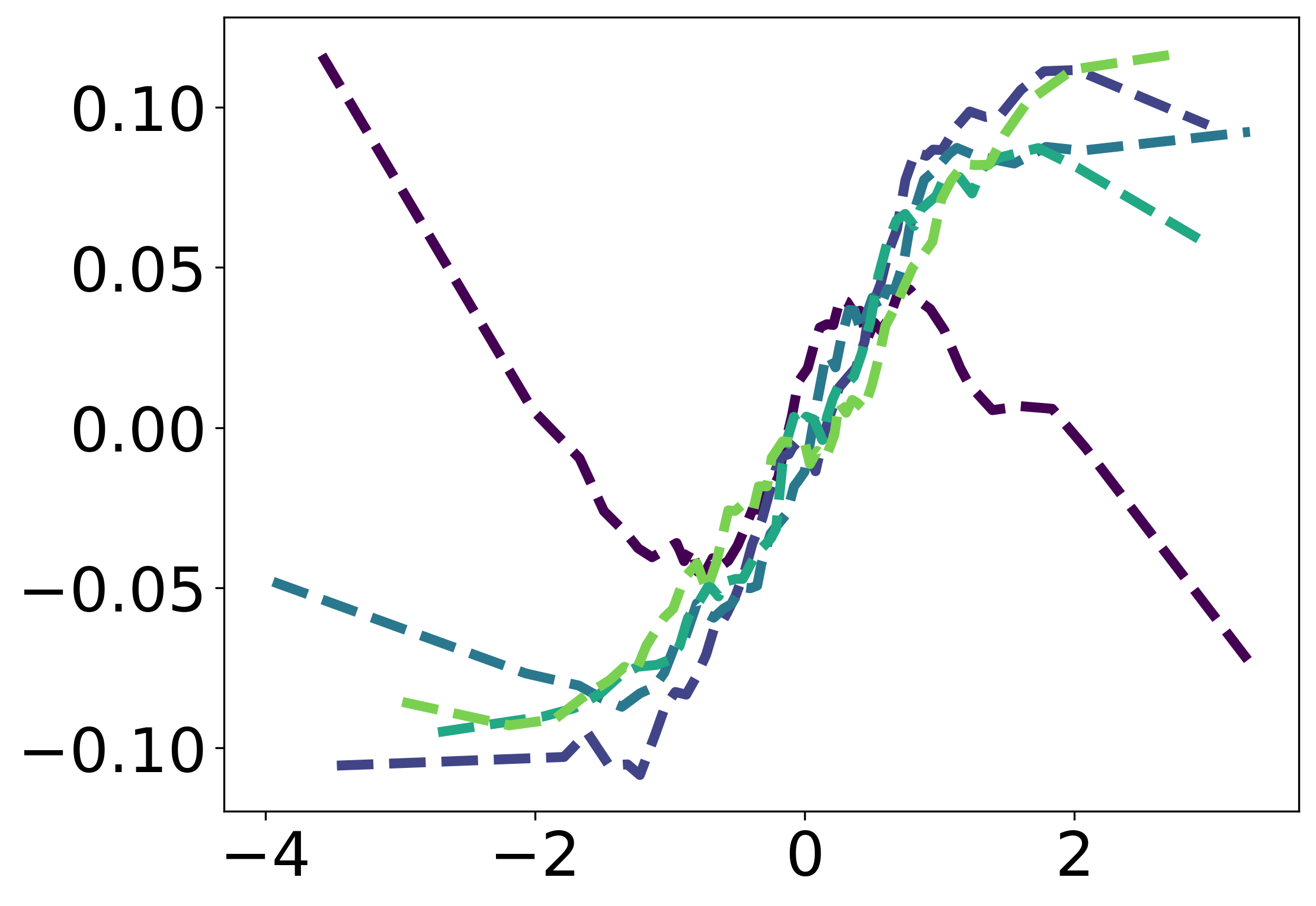}
\includegraphics[width=0.19\linewidth]{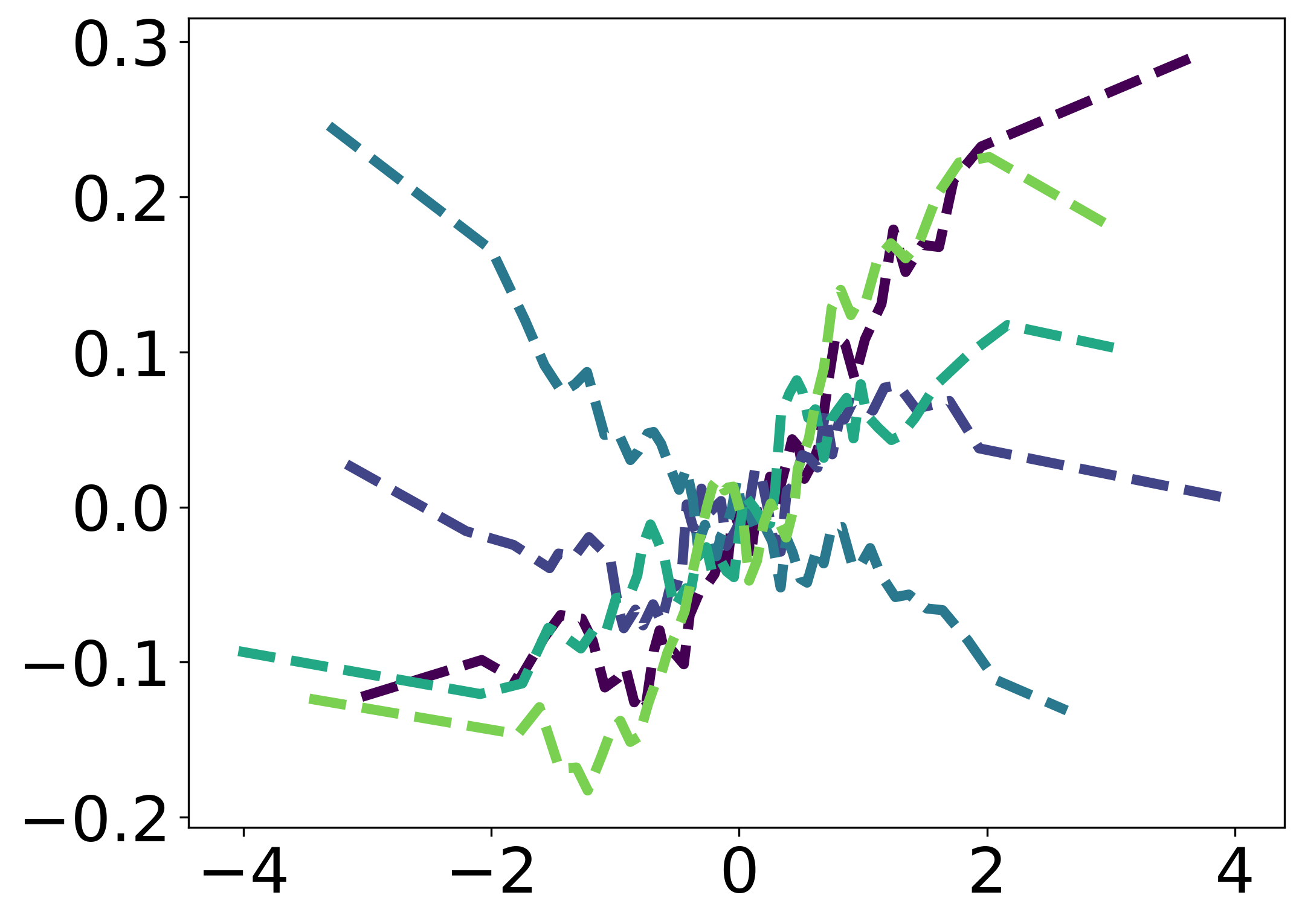}\\
\hrule
\includegraphics[width=0.19\linewidth]{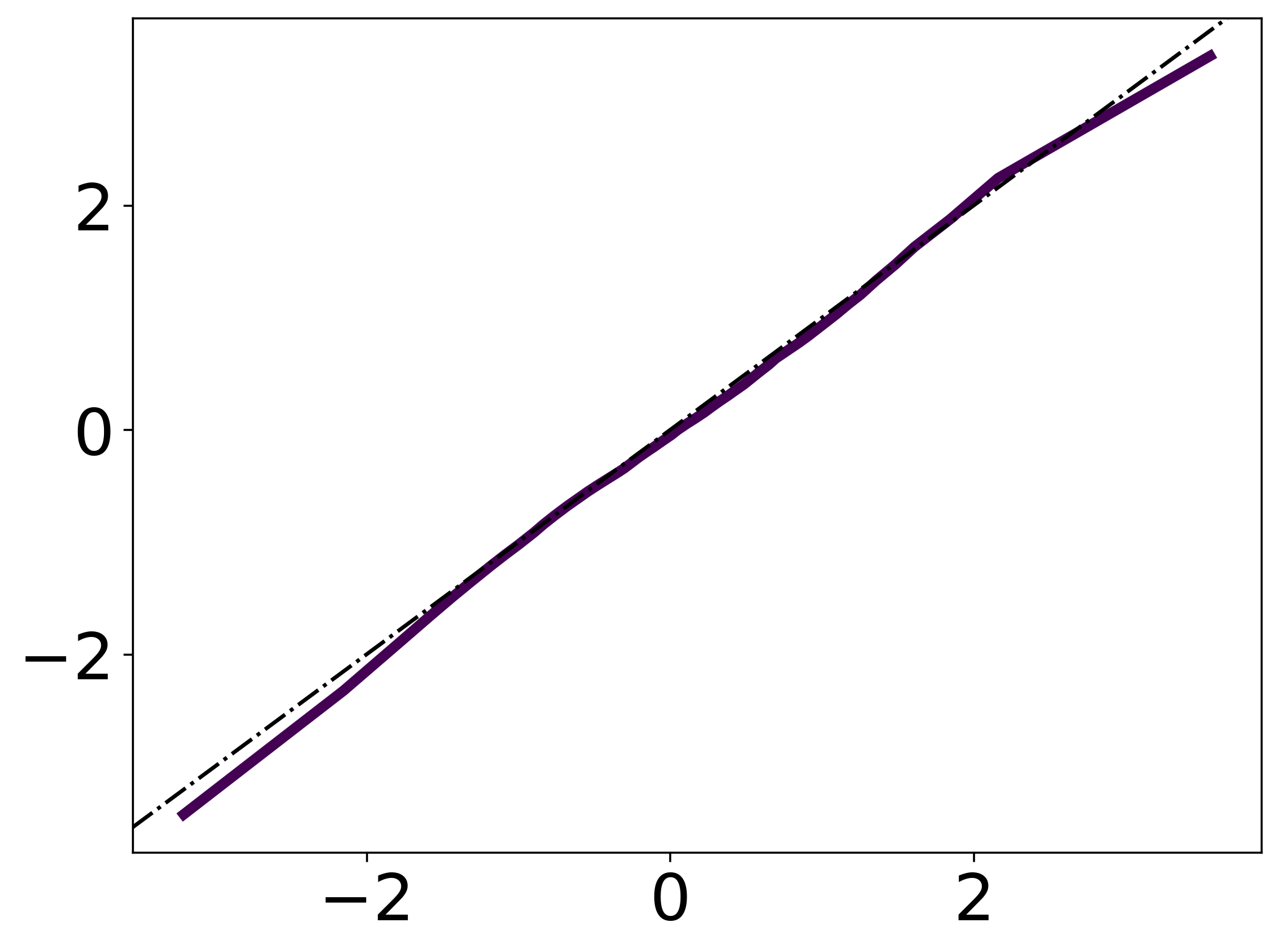}
\includegraphics[width=0.19\linewidth]{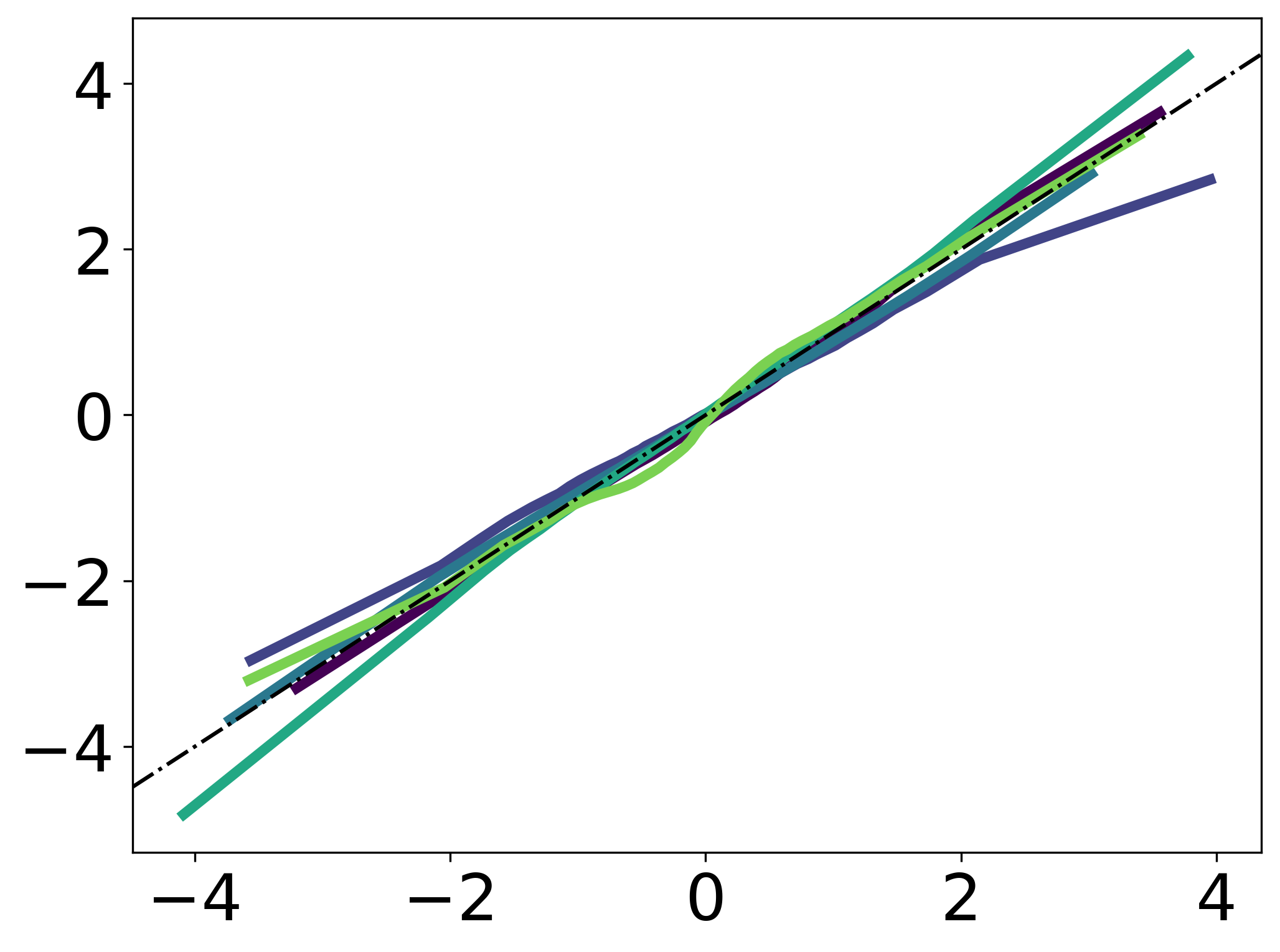}
\includegraphics[width=0.19\linewidth]{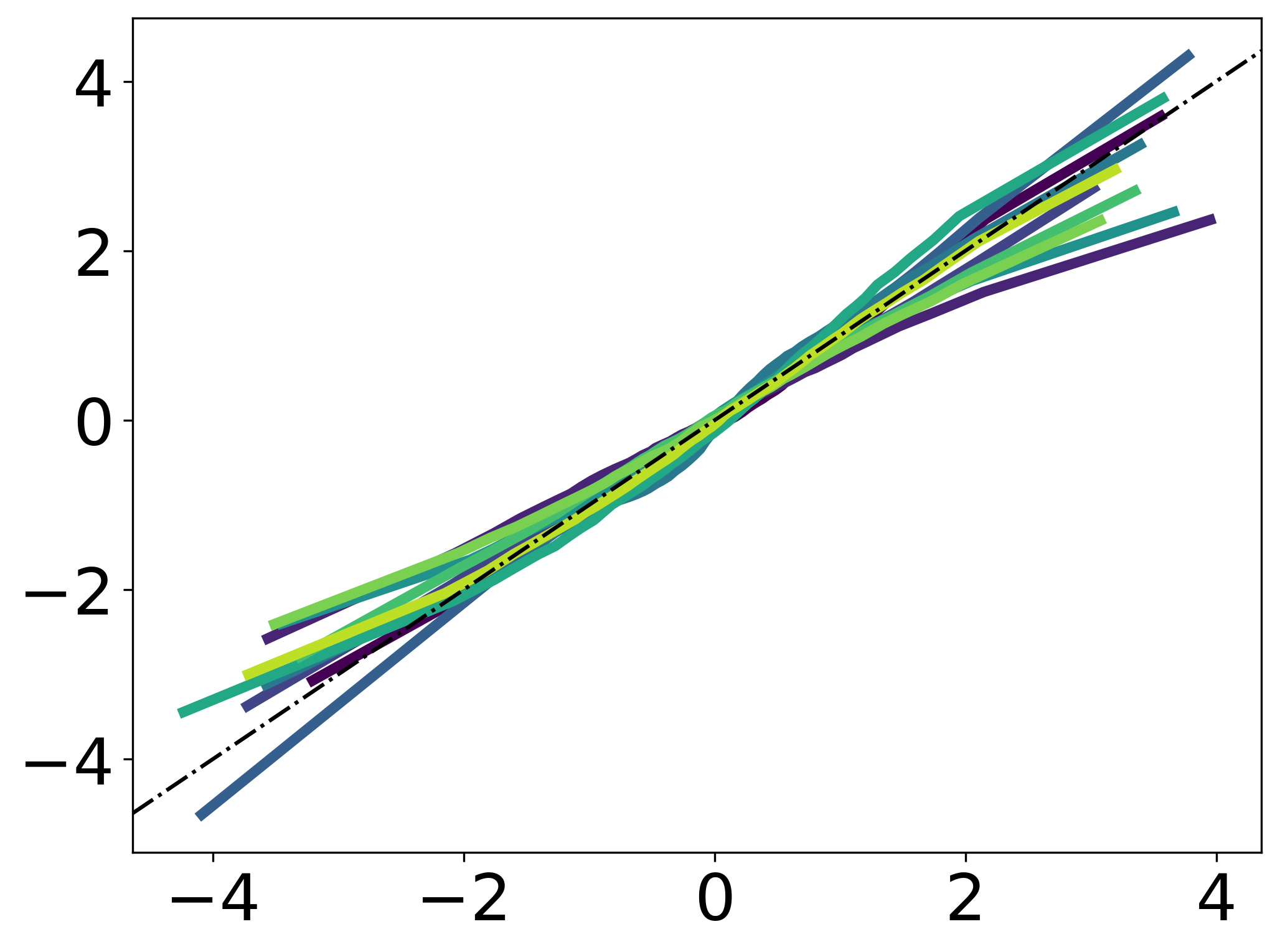}
\includegraphics[width=0.19\linewidth]{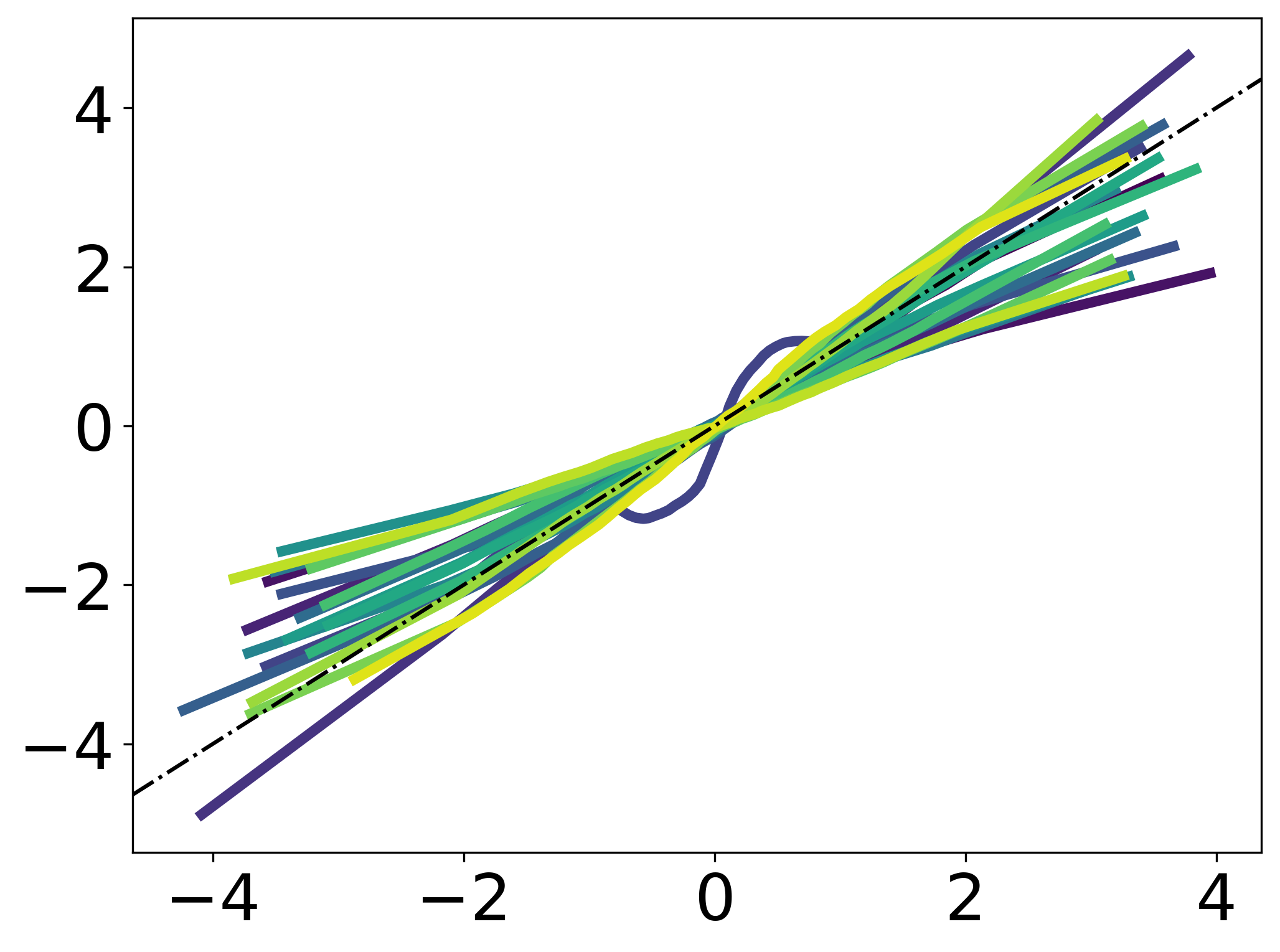}
\includegraphics[width=0.19\linewidth]{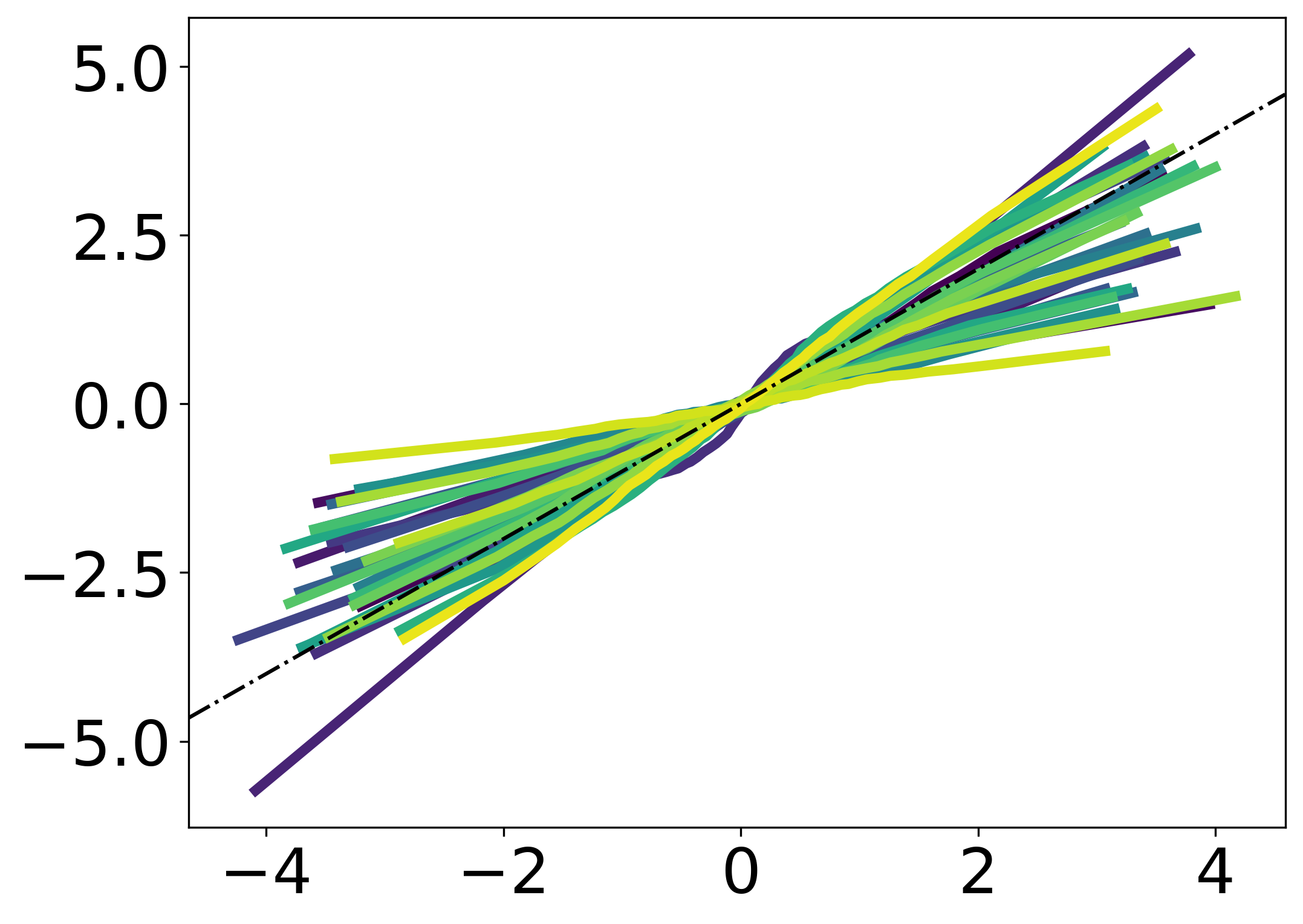}\\
\includegraphics[width=0.19\linewidth]{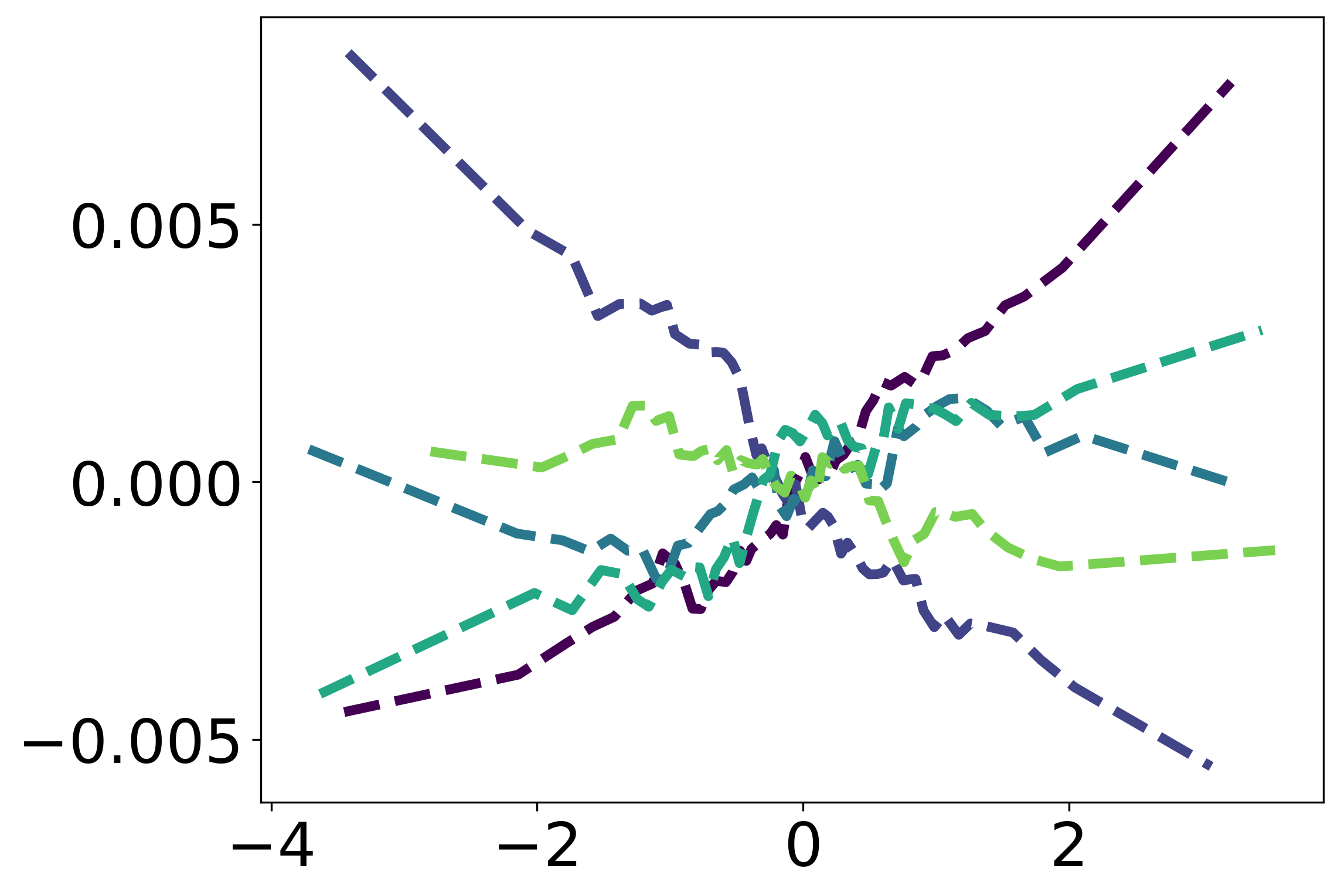}
\includegraphics[width=0.19\linewidth]{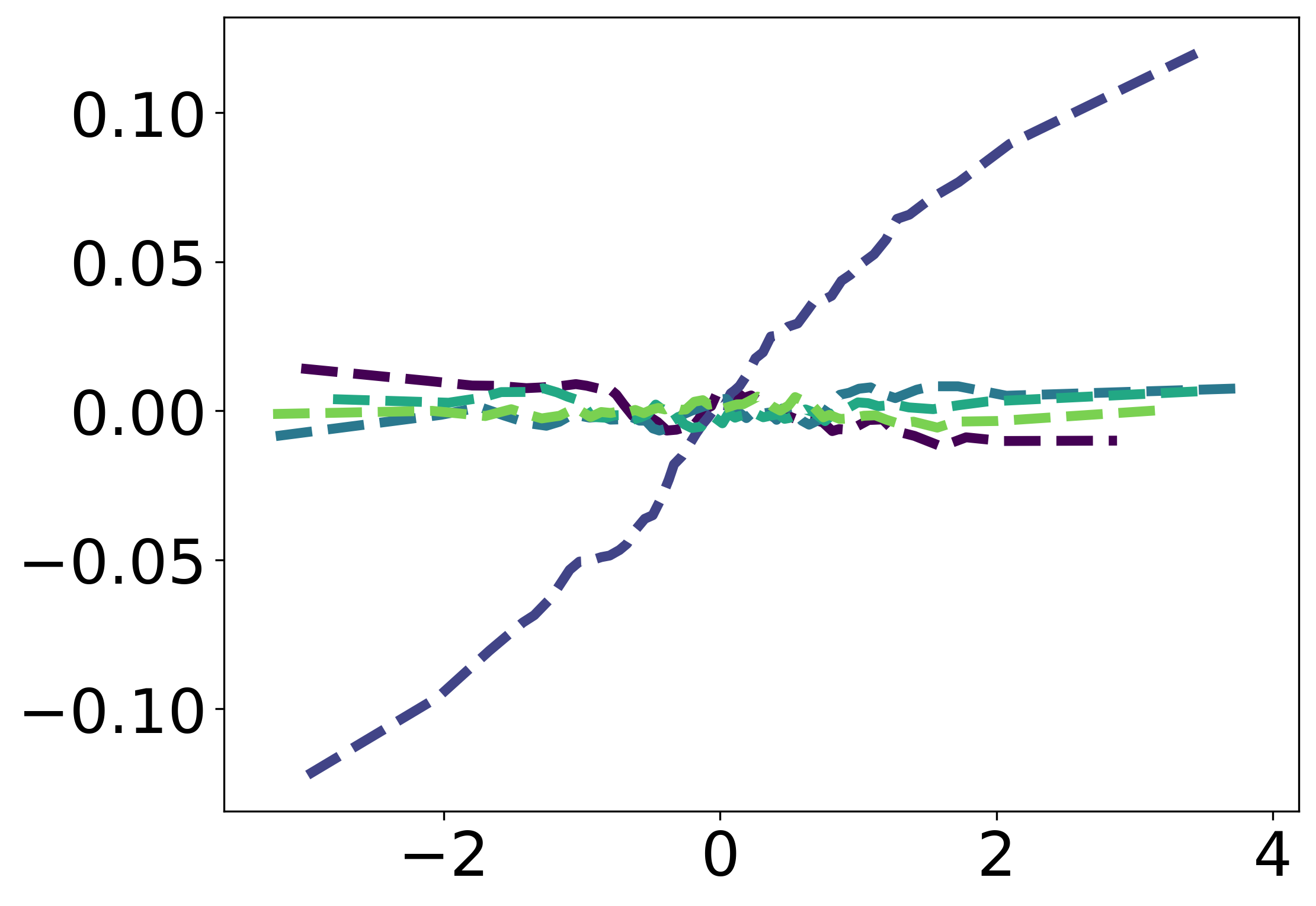}
\includegraphics[width=0.19\linewidth]{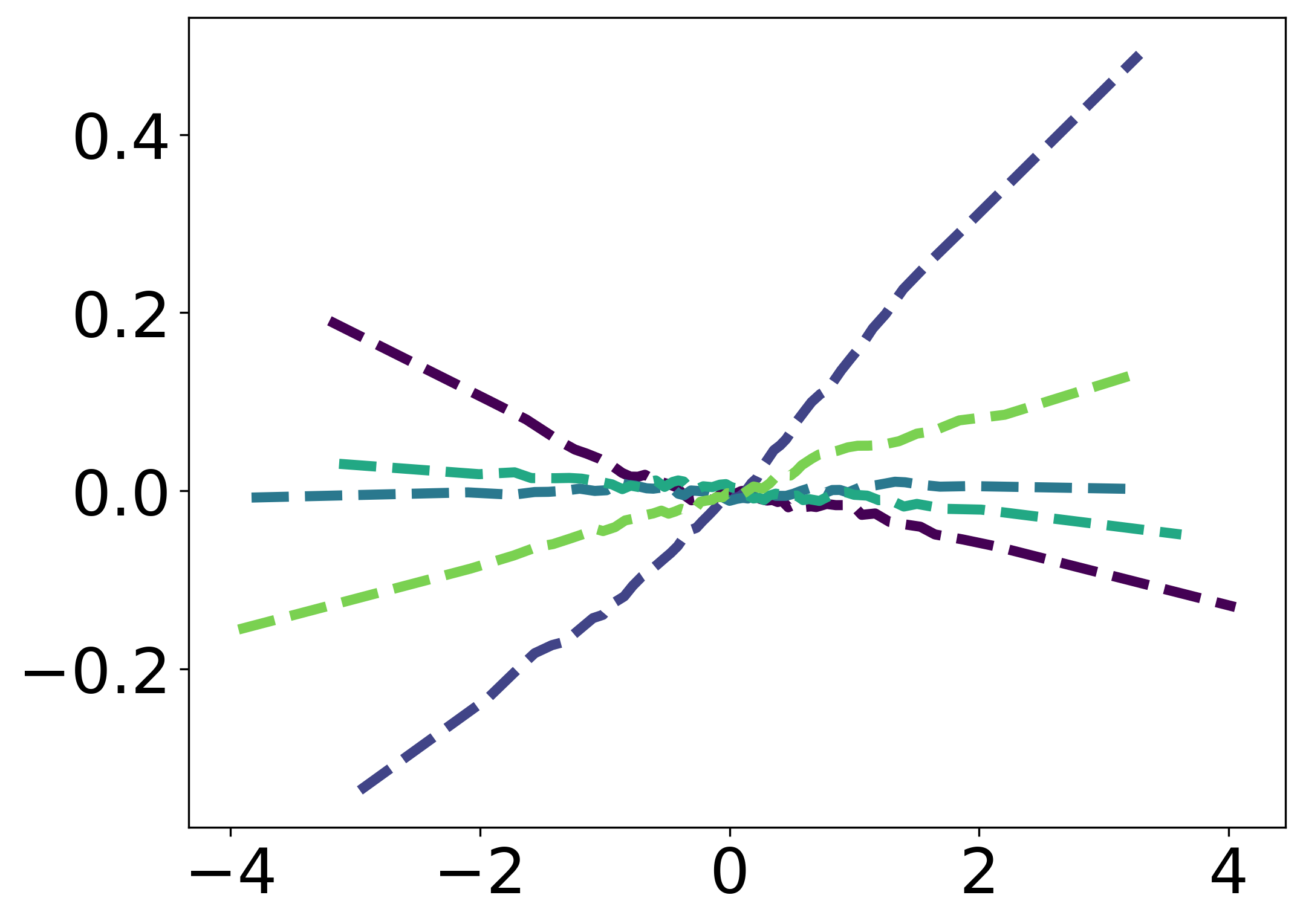}
\includegraphics[width=0.19\linewidth]{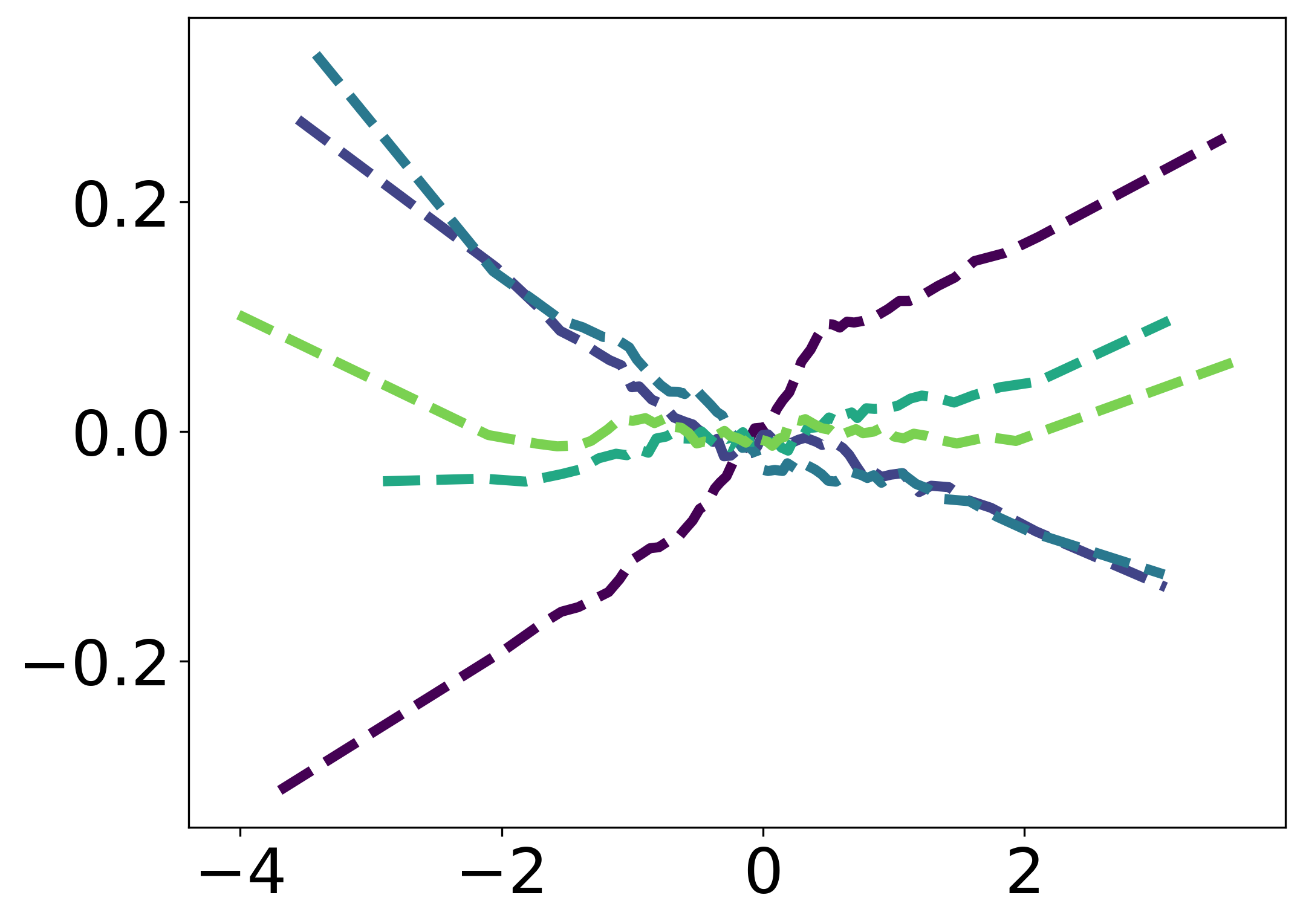}
\includegraphics[width=0.19\linewidth]{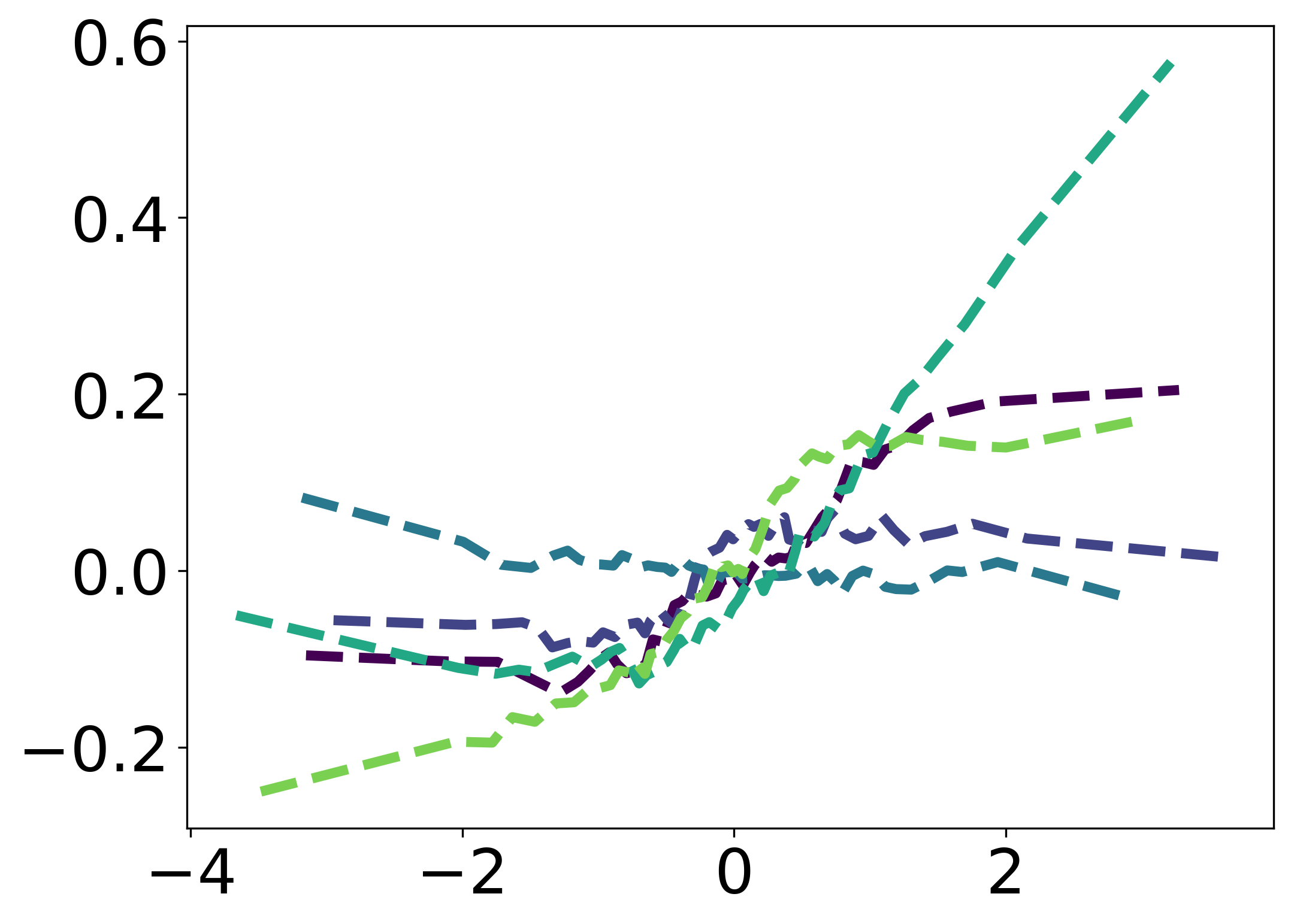}\\
\includegraphics[width=0.19\linewidth]{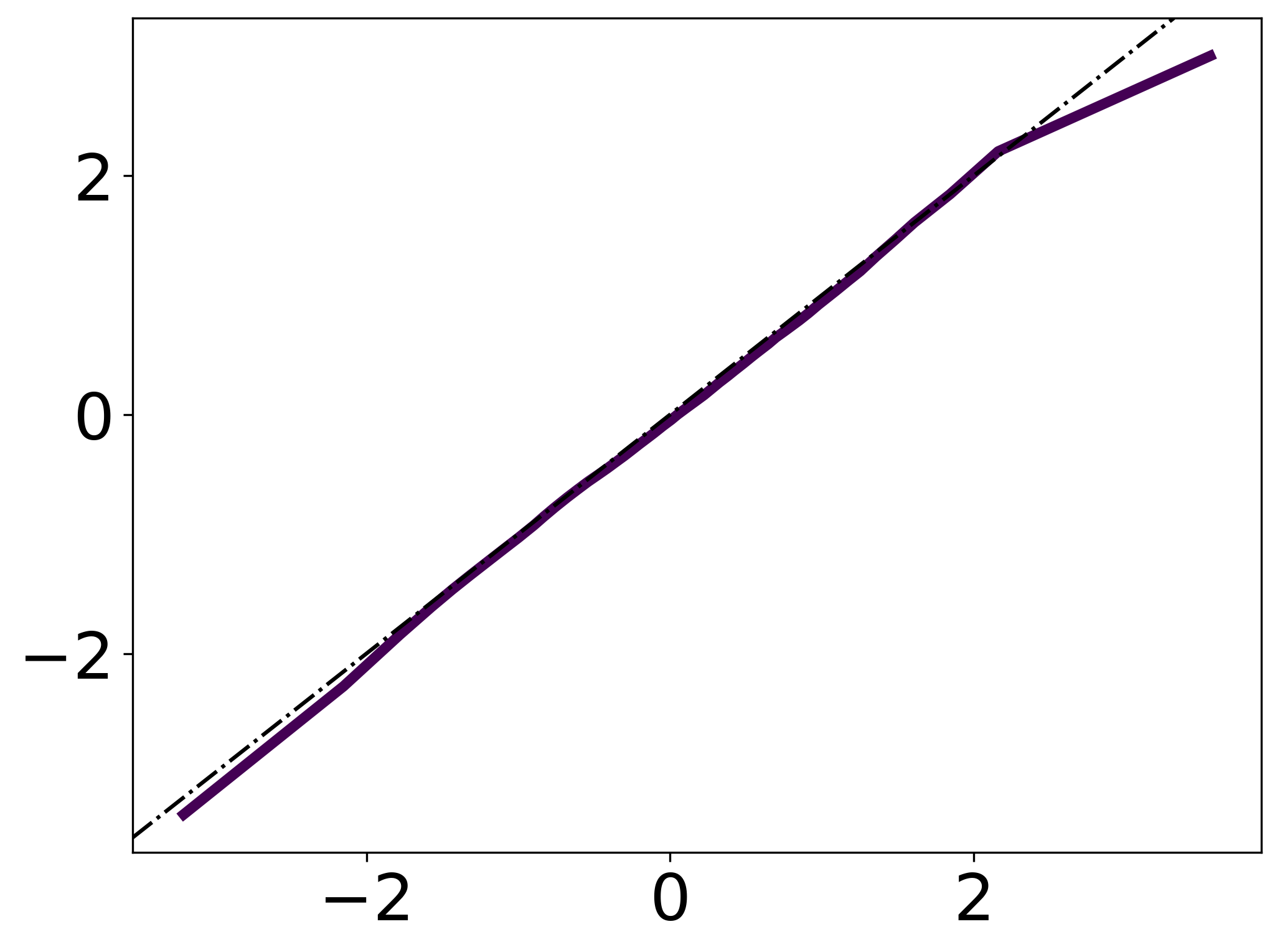}
\includegraphics[width=0.19\linewidth]{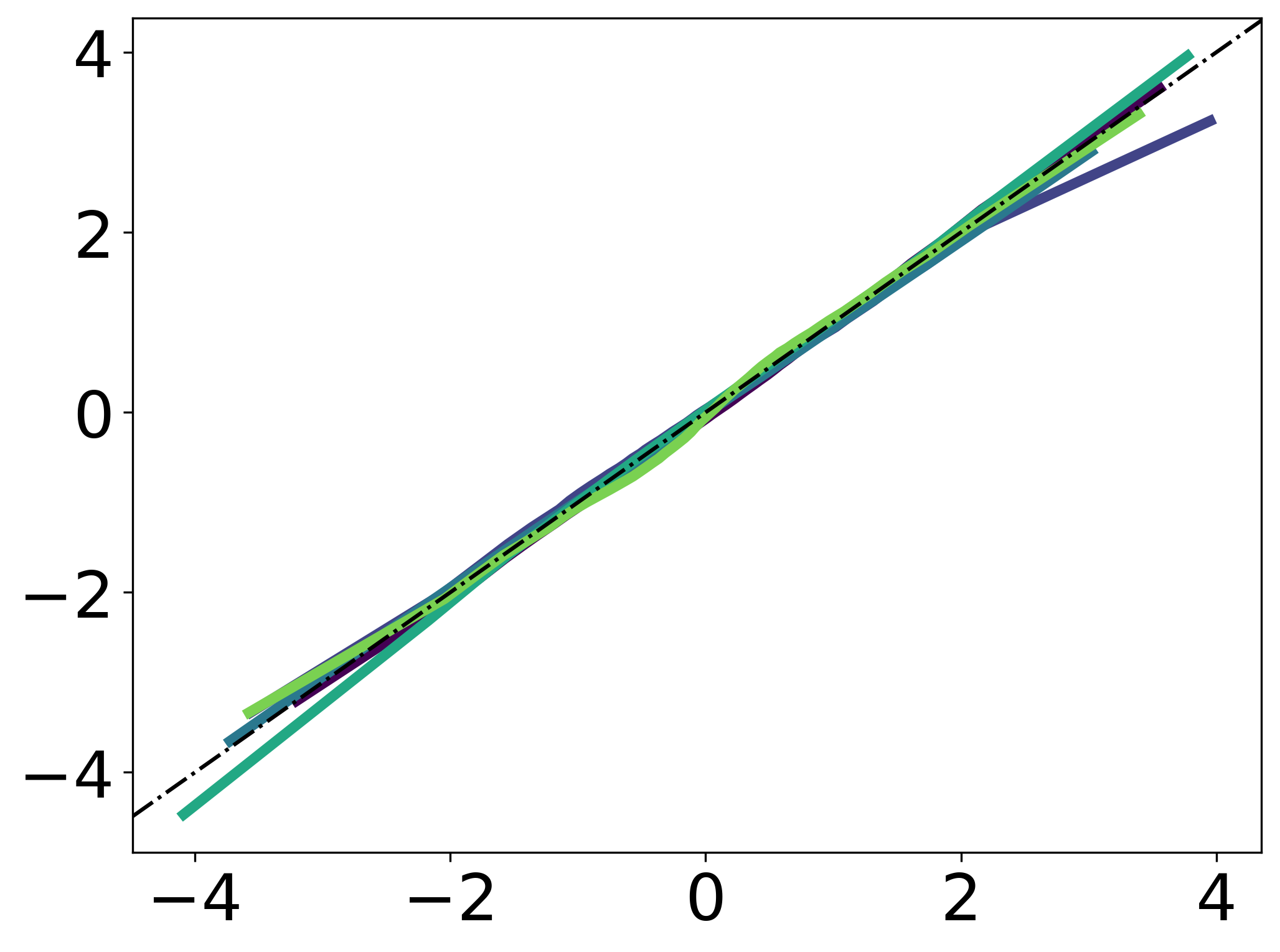}
\includegraphics[width=0.19\linewidth]{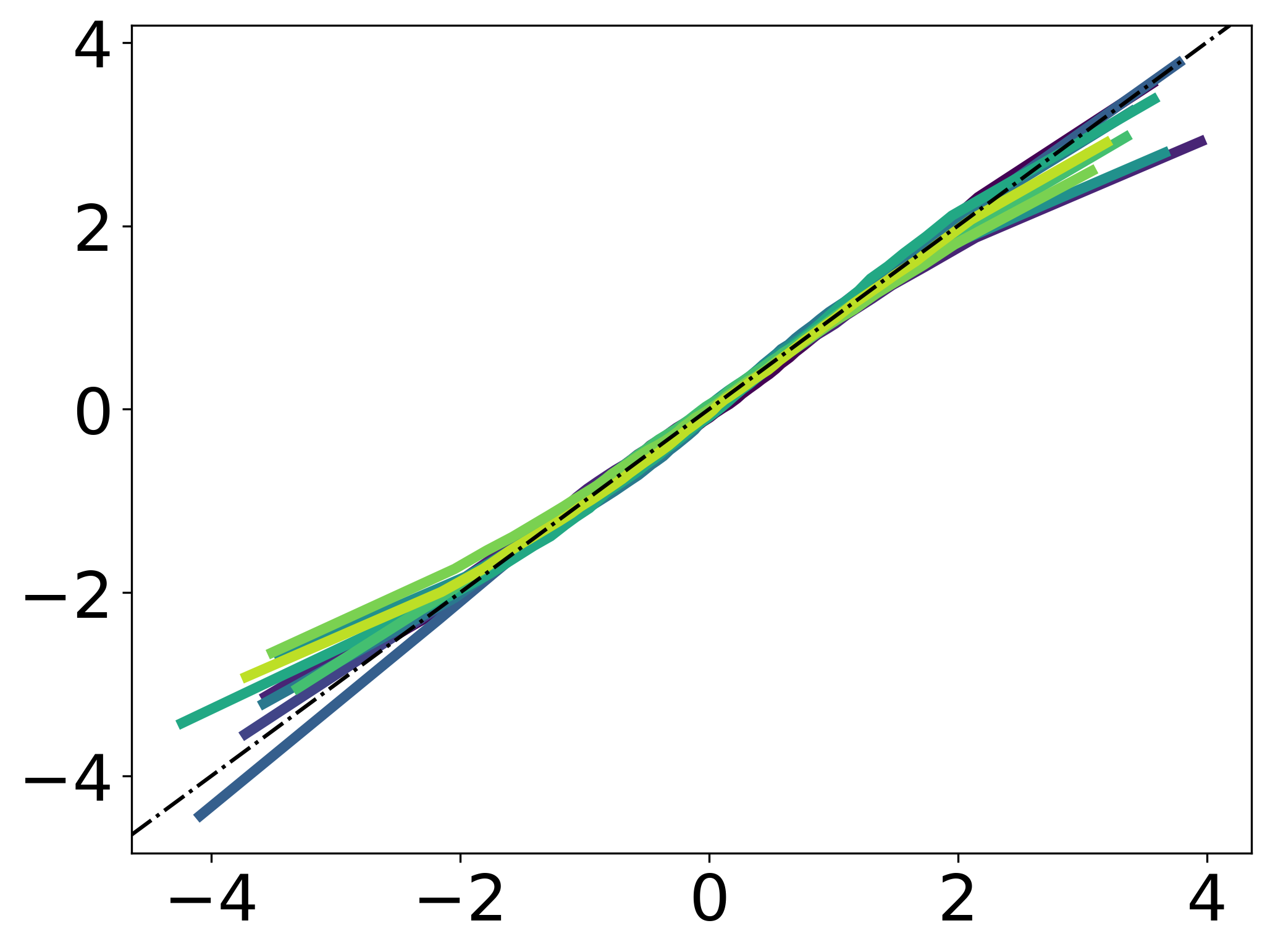}
\includegraphics[width=0.19\linewidth]{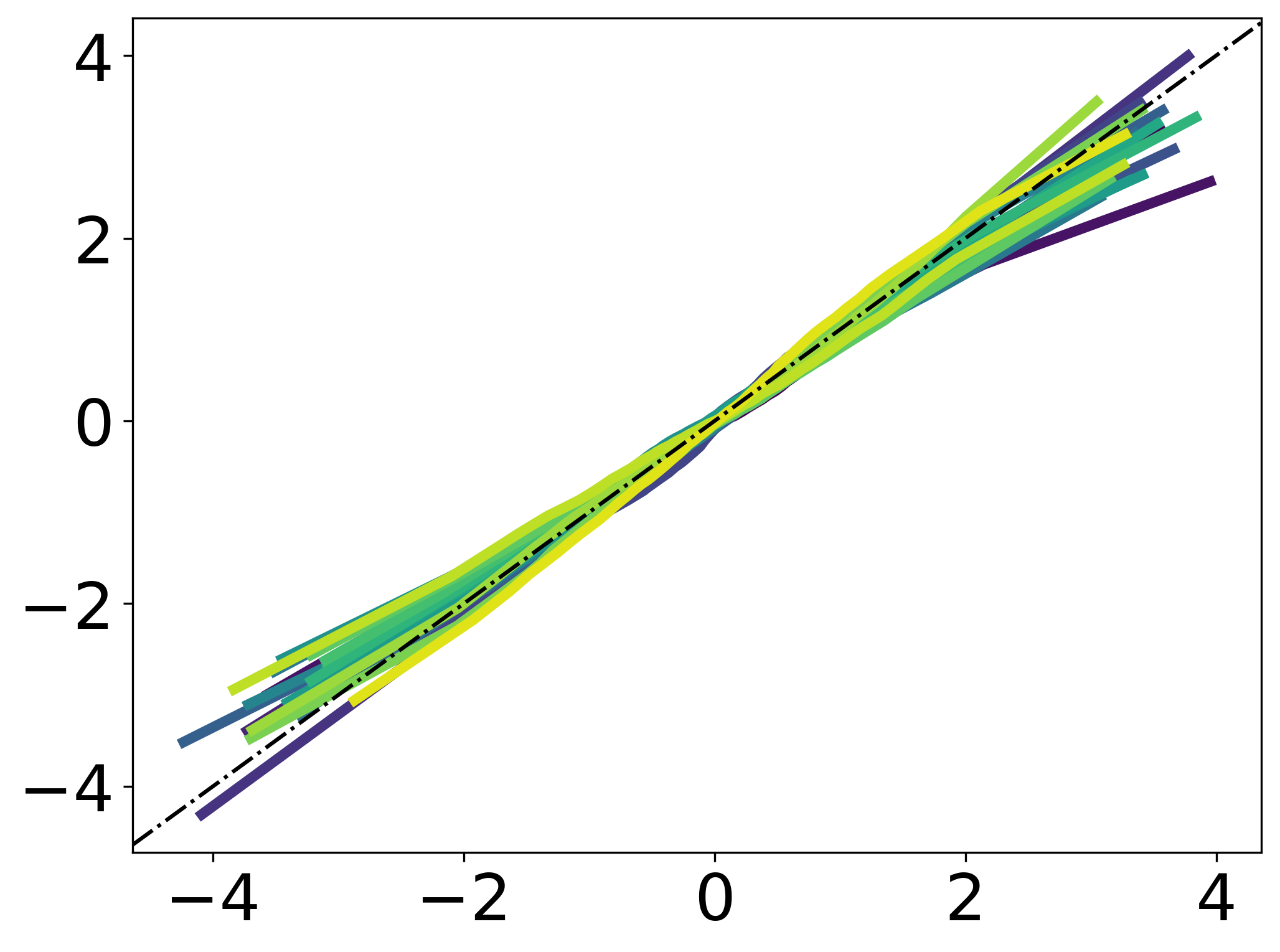}
\includegraphics[width=0.19\linewidth]{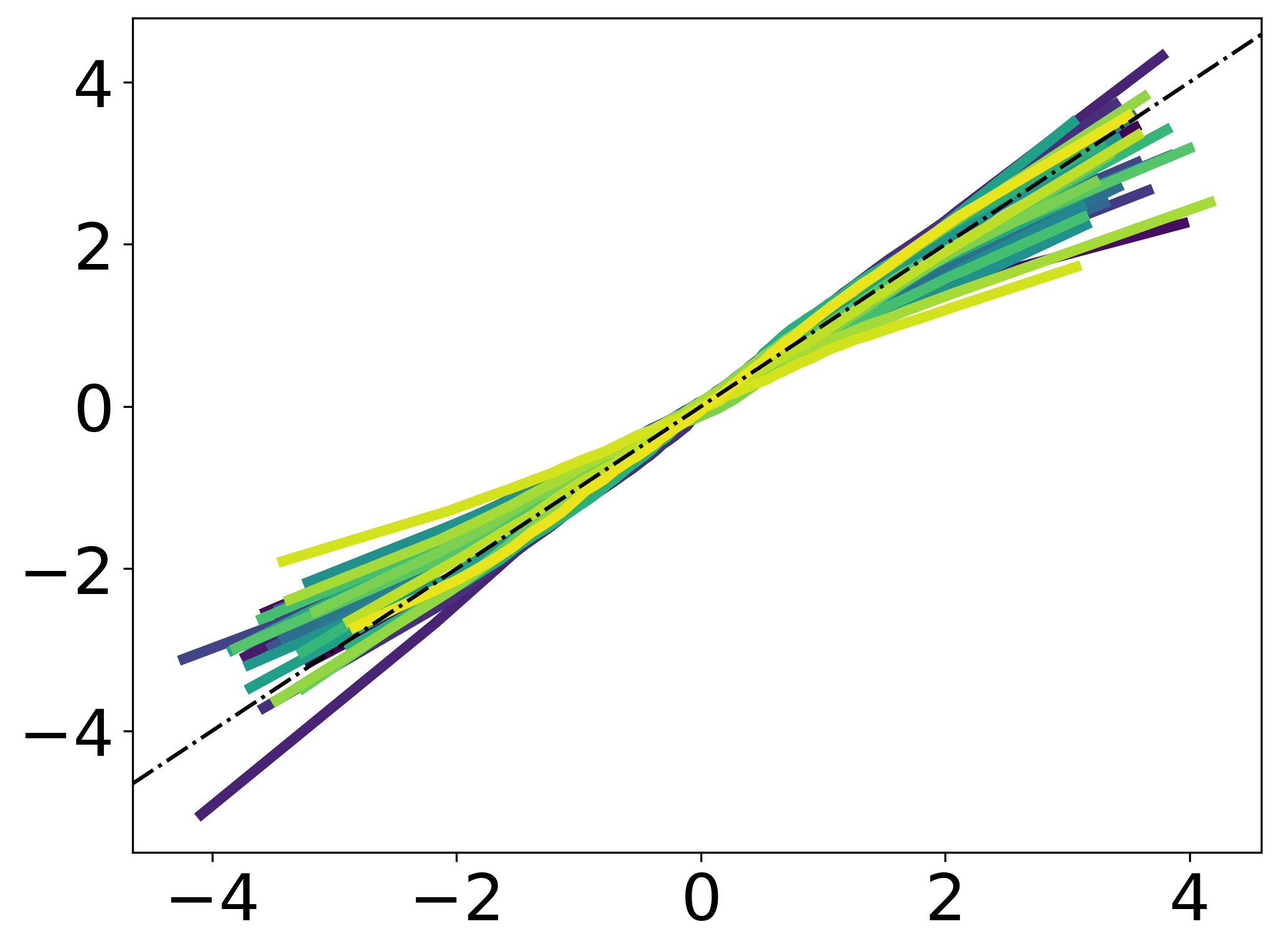}\\
\includegraphics[width=0.19\linewidth]{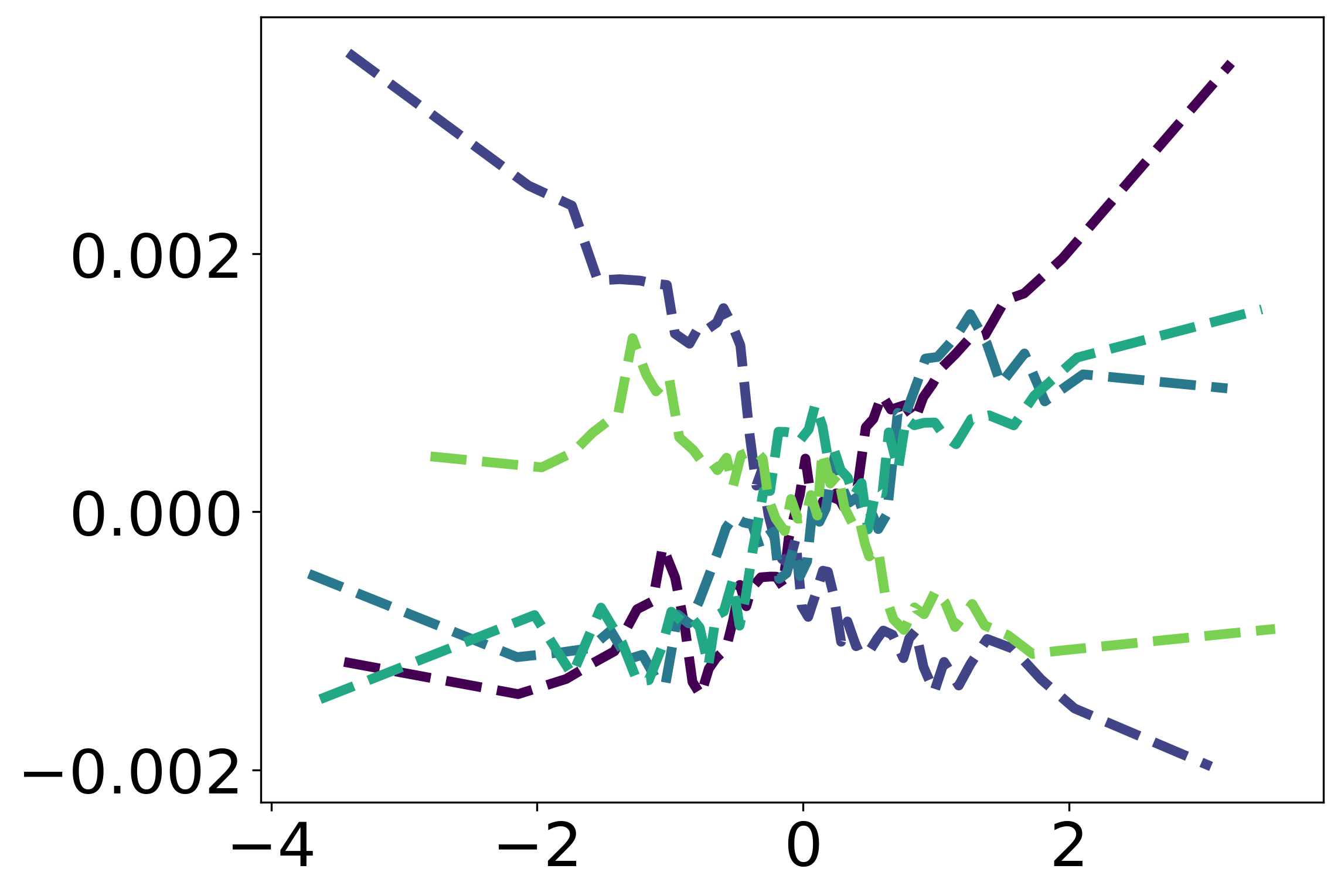}
\includegraphics[width=0.19\linewidth]{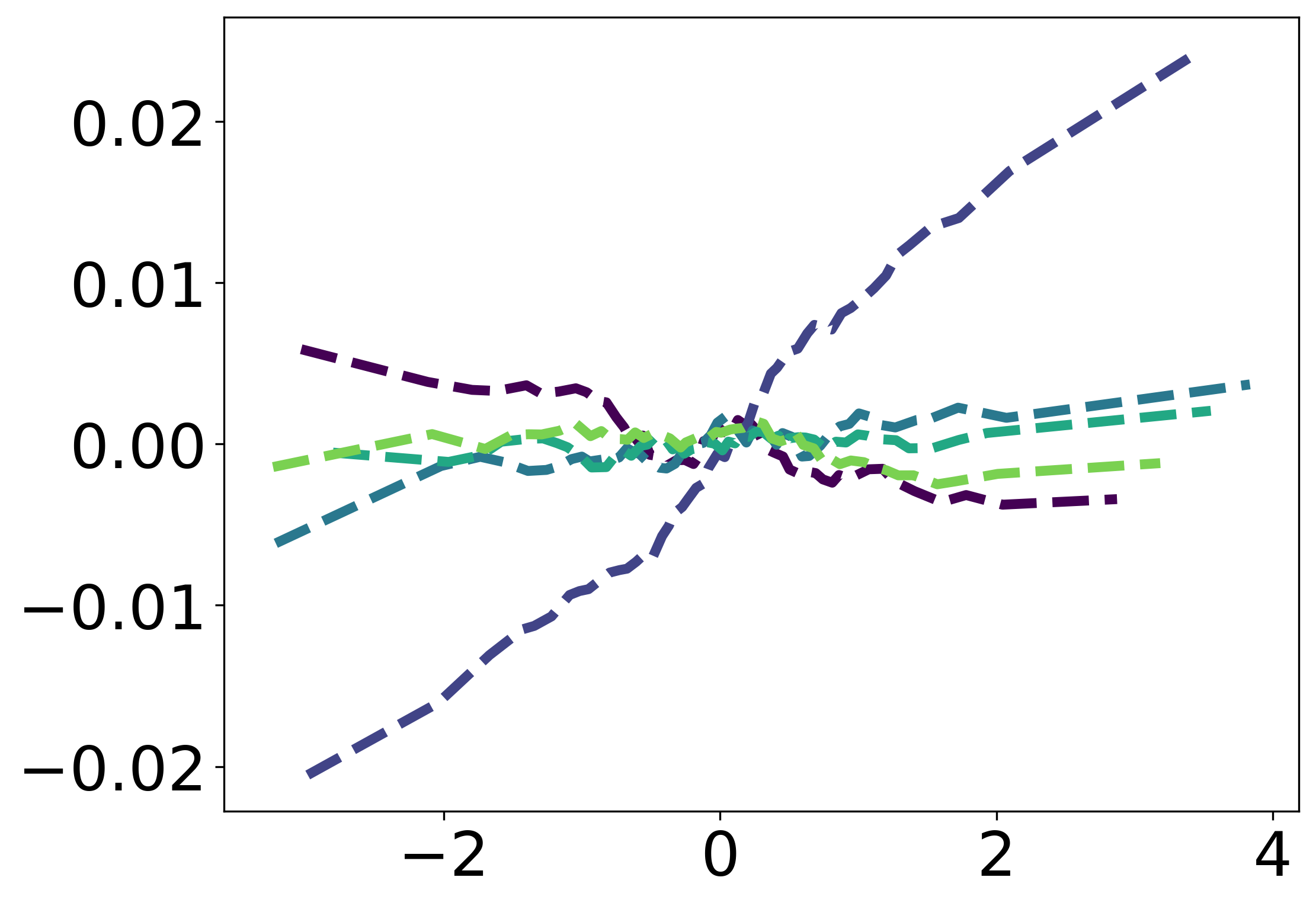}
\includegraphics[width=0.19\linewidth]{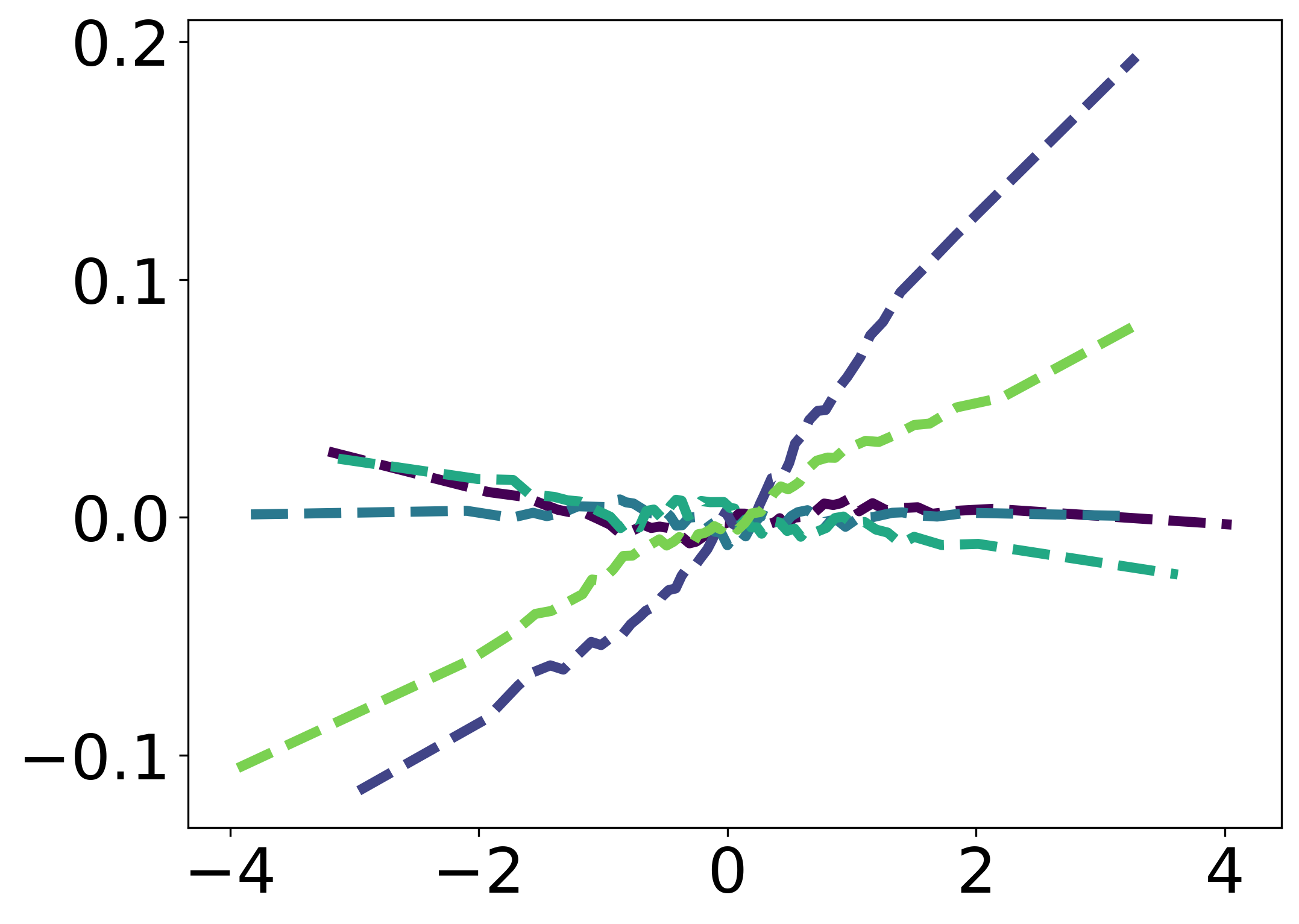}
\includegraphics[width=0.19\linewidth]{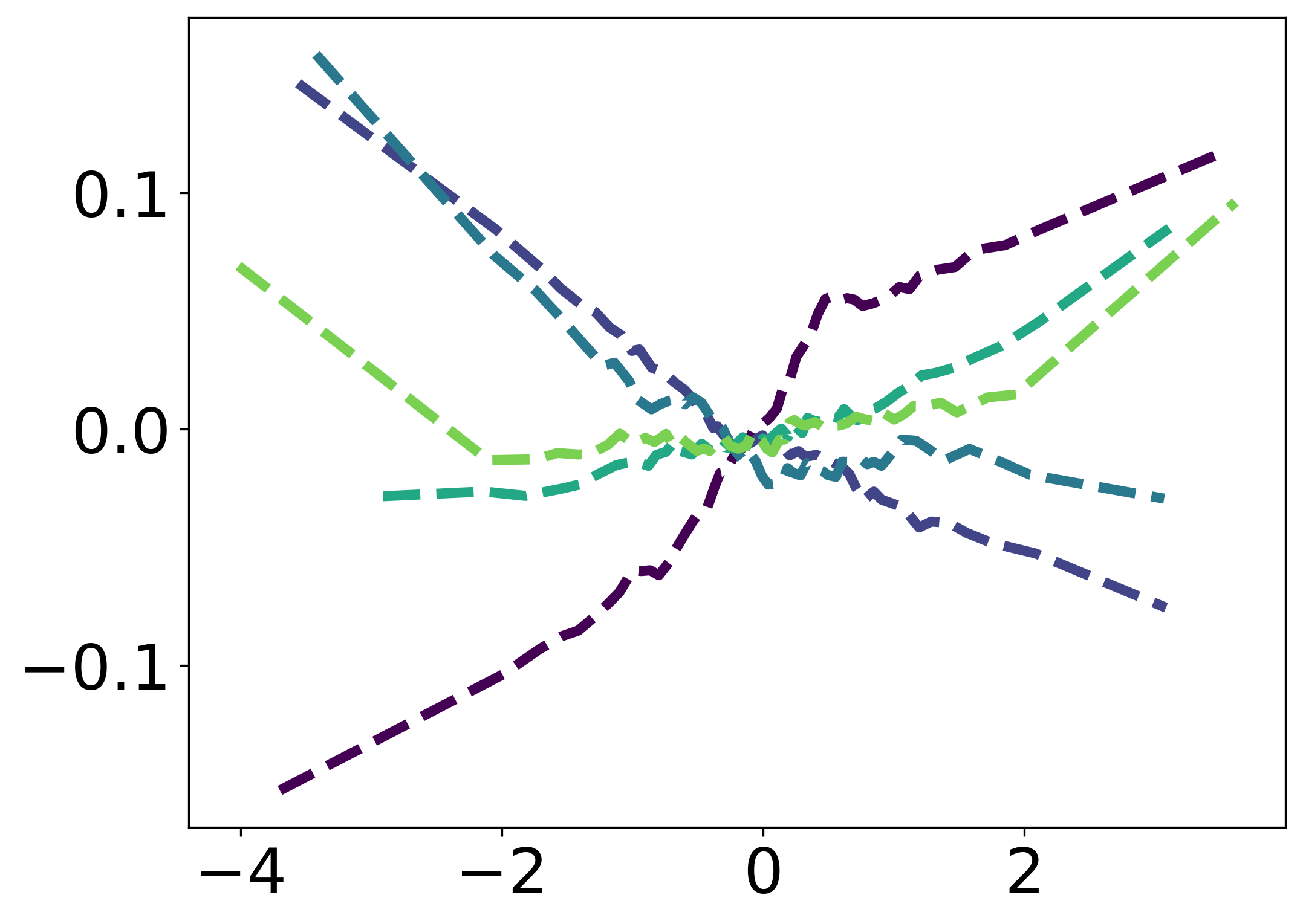}
\includegraphics[width=0.19\linewidth]{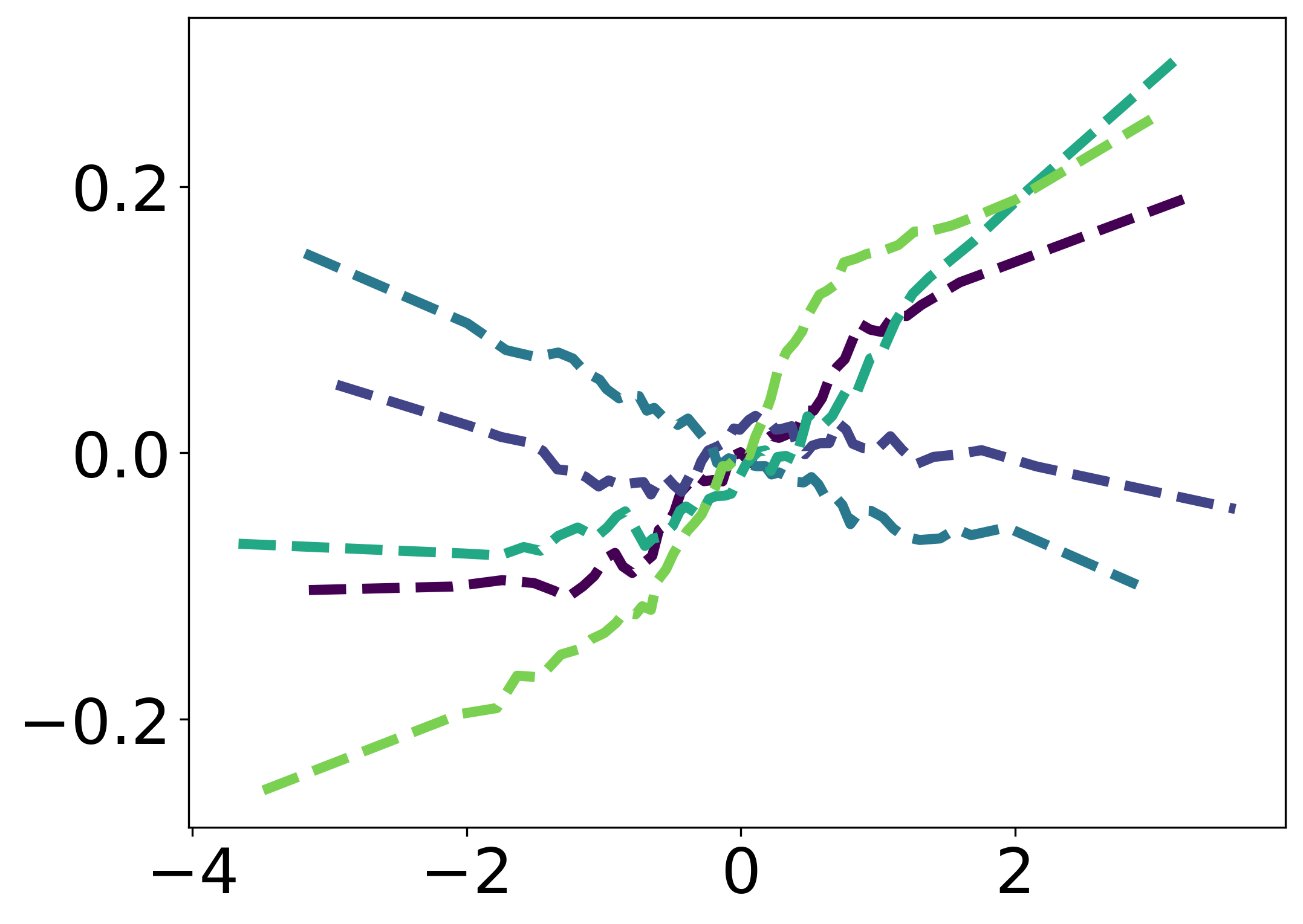}
\caption{Accumulated Local Effects plot for TabPFN under sparse linear regression when $p=100$ under orthogonal design.
  Columns (left to right) correspond to increasing sparsity levels (1, 5, 10, 20, 30).
  Rows 1--4 show results when training set is of size $n=50$.
  First row: Relevant feature; second row: randomly sampled $5$ irrelevant features (SNR = 1.22).
  Third row: Relevant feature; fourth row: randomly sampled $5$ irrelevant features (SNR = 6).
  Rows 5--8 show the same results when training set is of size $n=500$.}
\label{fig:ALE}
\end{figure*}

The ALE plots of TabPFN under sparse linear regression are given in this section in Figure~\ref{fig:ALE}.

\section{Implementation Details for LDA and kNN}
\label{app:noisy_label}
When $X|Y=0 \sim \gN(\mu_0, \Sigma)$ and $X|Y=1 \sim \gN(\mu_1, \Sigma)$, with $\pi = \mathbb{P}(Y=1)$, the optimal Bayes classifier takes the form:
\[
C^{\text{Bayes}}(x) =
\begin{cases}
1, & \begin{aligned}[t]
& \log\left(\frac{\pi}{1-\pi}\right) \\
&\quad {}+ \left(x-\frac{\mu_0+\mu_1}{2}\right)^{\top}\Sigma^{-1}(\mu_1-\mu_0)\geq 0,
\end{aligned}\\
0, & \text{otherwise}.
\end{cases}
\]
The LDA classifier replaces population parameters with their sample estimates.
Letting $\tilde{Y}_i = Y_i$ for clean training data, we compute:
\[
\begin{split}
\hat\pi &= \frac{1}{n}\sum_{i=1}^n \mathbbm{1}(\tilde{Y}_i=1), \\
\hat\mu_r &= \frac{\sum_{i=1}^n X_i \mathbbm{1}(\tilde{Y}_i=r)}{\sum_{i=1}^n \mathbbm{1}(\tilde{Y}_i=r)}, \quad r \in \{0,1\} \\
\hat\Sigma &= \frac{1}{n-2}\sum_{r=0}^1 \sum_{i:\tilde{Y}_i=r} (X_i-\hat\mu_r)(X_i-\hat\mu_r)^\top,
\end{split}
\]
yielding the LDA classifier:
\[
C^{\text{LDA}}(x) =
\begin{cases}
1, & \begin{aligned}[t]
& \log\left(\frac{\hat\pi}{1-\hat\pi}\right) \\
&\quad {}+ \left(x-\frac{\hat\mu_0+\hat\mu_1}{2}\right)^{\top}\hat\Sigma^{-1}(\hat\mu_1-\hat\mu_0)\geq 0,
\end{aligned}\\
0, & \text{otherwise}.
\end{cases}
\]

For $k$-nearest neighbors ($k$NN) classification, we select the optimal number of neighbors $k$ via $5$-fold cross-validation.
The search grid consists of $10$ equally spaced integers between $\lfloor n^{1/4} \rfloor$ and $\lfloor n^{3/4} \rfloor$, with the value maximizing cross-validation accuracy chosen for each experimental replication.